%% file: ms.tex
\RequirePackage{fix-cm}
\documentclass[twocolumn]{svjour3}          
\smartqed  
\usepackage{arydshln}
\usepackage{natbib}
\setcitestyle{authoryear,open={(},close={)}}
\usepackage{graphicx}
\usepackage{url}
\usepackage{tabularx}
\usepackage{times}
\usepackage{amsmath}
\usepackage{epsfig}
\usepackage{graphicx}
\usepackage{amsmath}
\usepackage{amssymb}
\usepackage{cuted}
\usepackage{multirow}
\usepackage{float}
\usepackage{mathtools}
\usepackage{color}
\usepackage{balance}
\usepackage{epstopdf}
\usepackage{subfig}
\usepackage{bbm}
\usepackage[normalem]{ulem}
\useunder{\uline}{\ul}{}
\usepackage[linesnumbered,ruled,vlined,onelanguage]{algorithm2e}
\usepackage[table,xcdraw]{xcolor}

\definecolor{newcolor}{rgb}{0,0,1}
\usepackage{hyperref}
\hypersetup{
    colorlinks=true,
    linkcolor=newcolor,
    filecolor=magenta,      
    urlcolor=blue,
}

\begin{document}

\title{DGSAC: Density Guided Sampling and Consensus
}
\subtitle{A Data Driven Unified Pipeline for Automatic Robust Multiple Structure Recovery}

\titlerunning{DGSAC: Density Guided Sampling and Consensus}        

\author{Lokender Tiwari        \and
        Saket Anand 
}


\institute{Lokender Tiwari \at
              IIIT-Delhi, India \\
              \email{lokendert@iiitd.ac.in}           
           \and
           Saket Anand \at
              IIIT-Delhi, India 
}

\date{Received: date / Accepted: date}

\maketitle

\input{abstract}

\input{new_comm}
\input{intro}

\input{related_work}
\input{dgsac_overview}

\input{building_blocks}
\input{KDGS_sec}

\input{KRD_Frac}

\input{GO_model}

\input{point-to-model_assignment}
\input{experiments}

\input{analysis_discussion}

\bibliography{dgsac}

\end{document}

%% file: abstract.tex
\begin{abstract}

Robust multiple model fitting plays a crucial role in many computer vision applications. Unlike single model fitting problems, the multi-model fitting has additional challenges. The \textit{unknown number of models} and the \textit{inlier noise scale} are the two most important of them, which are in general provided by the user using ground-truth or some other auxiliary information. Mode seeking/ clustering-based approaches crucially depend on the quality of model hypotheses generated. While preference analysis based guided sampling approaches have shown remarkable performance, they operate in a time budget framework, and the user provides the time as a \textit{reasonable guess}. In this paper, we deviate from the mode seeking and time budget framework. We propose a concept called \textit{Kernel Residual Density} (KRD) and apply it to various components of a multiple-model fitting pipeline. The Kernel Residual Density act as a key differentiator between inliers and outliers.  We use KRD to guide and automatically stop the sampling process. The sampling process stops after generating a set of hypotheses that can explain all the data points. An \textit{explanation score} is maintained for each data point, which is updated \textit{on-the-fly}. We propose two model selection algorithms, an \textit{optimal} quadratic program based, and a \textit{greedy}. Unlike mode seeking approaches, our model selection algorithms seek to find one representative hypothesis for each genuine structure present in the data.  We evaluate our method (called DGSAC) on a wide variety of tasks like \textit{planar segmentation}, \textit{motion segmentation}, \textit{vanishing point estimation}, \textit{plane fitting to 3D point cloud}, \textit{line}, and \textit{circle fitting}, which shows the effectiveness of our method and its unified nature.

\keywords{Kernel Residual Density \and Multi-Model Fitting \and Guided Sampling \and Two-View Estimation \and Vanishing Point}
\end{abstract}

%% file: new_comm.tex
\def\eg{\emph{e.g.}} \def\Eg{\emph{E.g}}
\def\ie{\emph{i.e.}} \def\Ie{\emph{I.e.}}
\def\cf{\emph{c.f.}} \def\Cf{\emph{C.f.}}
\def\etc{\emph{etc}} \def\vs{\emph{vs}}
\def\wrt{w.r.t.} \def\dof{d.o.f.}
\def\etal{\emph{et al.}}
\DeclarePairedDelimiter\ceil{\lceil}{\rceil}
\DeclarePairedDelimiter\floor{\lfloor}{\rfloor}

\newcolumntype{?}{!{\vrule width 1.5pt}}
\newcommand{\bH}{\textbf{H}}
\newcommand{\bhi}{\textbf{h}_i}
\newcommand{\bsi}{\textbf{s}_i}
\newcommand{\bS}{\textbf{S}}
\newcommand{\bqi}{\textbf{q}_i}
\newcommand{\bpj}{\textbf{p}_j}
\newcommand{\bR}{\textbf{R}}
\newcommand{\bri}{\textbf{r}_i}
\newcommand{\brj}{\textbf{r}^j}
\newcommand{\xj}{x_j}
\newcommand{\tw}{\text{w}}
\newcommand{\ti}{\text{i}}
\newcommand{\mcalM}{\mathcal{M}}
\newcommand{\blj}{\textbf{l}_j}
\newcommand{\jth}{$j^{th}$}
\newcommand{\ith}{$i^{th}$}
\newcommand{\kth}{$k^{th}$}
\newcommand{\fhat}{\hat{f}}
\newcommand{\sighat}{\hat{\sigma}}

%% file: intro.tex
\section{Introduction}
Robust multiple model fitting plays a crucial role in many computer vision applications. Unlike single model fitting problems, the multi-model fitting has additional challenges, the \textit{unknown number of models} and the \textit{inlier noise scale} are two most important of them, which are in general provided by the user using ground-truth or some other auxiliary information. \textit{Mode seeking/ clustering} based approaches crucially depends on the quality of model hypotheses generated. While preference analysis based guided sampling approaches have shown remarkable performance, they operate in a time budget framework, and the user provides the time as a \textit{reasonable guess}. In this paper, we deviate from the mode seeking and time budget framework. We propose a concept called \textit{Kernel Residual Density} (KRD) and apply it to various components of a multiple-model fitting pipeline. The Kernel Robust model fitting is the task of recovering an underlying geometric structure present in the noisy data contaminated with outliers. In computer vision, frequently occurred geometric structures ( or models) are planar homographies, fundamental matrices, vanishing points, lines, circles. These geometric models are the backbone of many applications including, motion segmentation, 3D reconstruction, visual tracking, image-based 3D modeling. 
Accurate estimation of these model parameters facilitates an interpretable and straightforward representation of scene geometry, rigid body motion or camera pose, and can aid scene understanding. Often, low-level techniques are employed to obtain features/observations/data points. These techniques are oblivious to the underlying scene geometry, typically produce a significant number of outliers, i.e., data points that do not adhere to any genuine model instances. 
We consider a multiple model fitting scenario, where the given data has multiple instances of a geometric model, corrupted by \textit{noise} and the \textit{outliers}. The data points/observations that follow a particular model instance are inliers to that instance and act as a \textit{pseudo-outliers} to other instances. The data points that do not follow any model instances are \textit{gross-outliers}. It is highly likely that each model instance may have different inlier noise scales. Assume that the three crucial information, i.e., classification of data points into inliers-outliers, the inlier noise scale(s), and the number of model instances are \textit{unknown}. The goal of multiple model-fitting approaches is to recover all genuine model instances and their corresponding inliers. Once the inliers obtained, the model parameters can be estimated using standard techniques.

\begin{figure*}[h!]
    \centering
    \includegraphics[width=17cm]{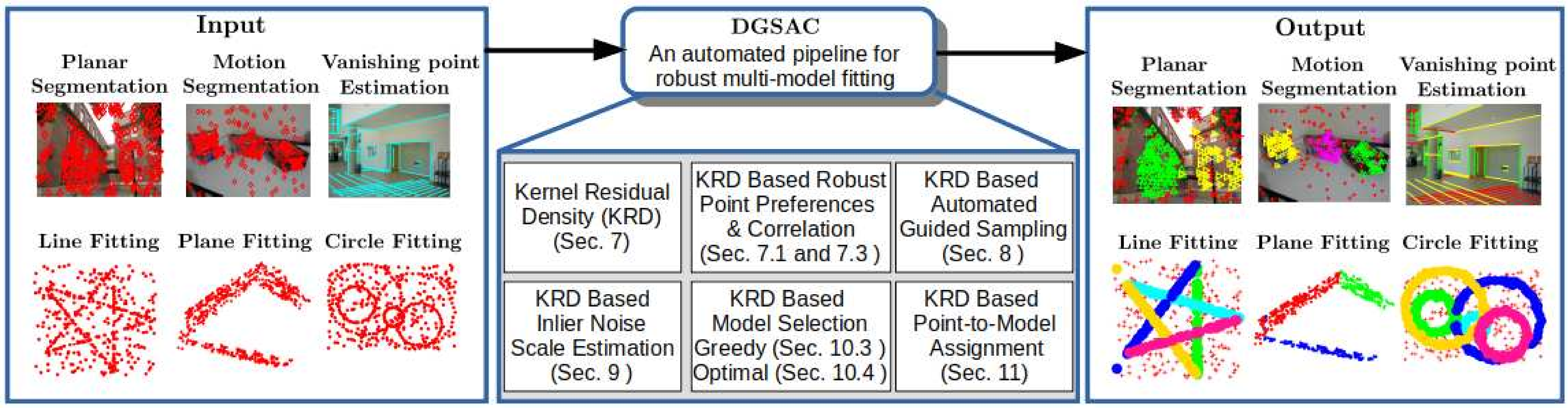}
    \caption{DGSAC Overview}
    \label{fig:introfig}
\end{figure*}

The single model fitting problem has been largely studied in the context of consensus maximization and often solved using heuristic methods like Random Sample Consensus \cite{fischler81} (RANSAC) and its variants or optimal global method like  \cite{chin2015efficient}. However, the performance of these approaches critically relies on the correct estimate of the inlier noise scale (or equivalently, the fraction of inliers), which is usually provided by a user. For single model fitting, this process
can be automated by applying scale estimation methods  \cite{wang12,wang13,magri2015scale} or by adopting a marginalized over scale approach \cite{barath2019magsac}. However, in the case of the multi-model fitting, the problem becomes challenging due to additional challenges. Many proposed methods  \cite{pham2014random,chin2012accelerated,wong2013simultaneous,lai2016efficient,lai2017unified,magri14tl,magri15rpa,magri16set} like assume, the \emph{number of models} or \emph{inlier scale} or \emph{both} are known a priori, which introduces user dependency.  
In this paper, we achieve automatic robust multi-model fitting by exploiting the concept of \emph{kernel residual density}, which is a key differentiator between inliers and outliers. Inliers yield a small residual value and form a dense cluster around the regression surface, whereas outliers (or pseudo-outliers) can have arbitrarily large residuals and a much lower density. Geometrically, we can say that inliers are densely packed around the regression surface, while outliers are spread sparsely in the residual space. The Density Guided Sampling and Consensus (DGSAC) is an automatic pipeline for robust multi-model fitting. We present an illustration of the components of the DGSAC pipeline and its possible applications in Fig. \ref{fig:introfig}. Our specific contributions are summarized below:

\begin{itemize}
    \item \textit{Kernel Residual Density (KRD)}: We present a Kernel Version of Residual Density proposed in \cite{tiwari2018dgsac} and its use in various components of multi-model fitting pipeline.
    \item \textit{Kernel Density Guided Hypothesis Generation}: We present a novel iterative guided sampling approach driven by KRD based point correlations and is shown to generate more relevant model hypotheses from \emph{all} inlier structures. To best our knowledge, we present the first non-time budget guided sampling approach, with density-driven stopping criteria. 
    \item \textit{KRD Based Inlier Noise Scale Estimation}: We present an inlier noise scale estimation approach based on simple yet effective information derived from residual dispersion and kernel residual density. 
    \item \textit{Model Goodness Score}: We propose a model goodness score, derived from kernel residual density and the estimated inlier noise scale to derive objective/cost function for model selection algorithms.
    \item \textit{Greedy/Optimal Model Selection}: We propose two (\textit{greedy} and \textit{optimal}) model selection algorithms. Both use the residual density to uniquely identify a model hypothesis from each genuine structure. The optimal model selection is modeled as a quadratic program. 
    \item \textit{KRD Based Point-to-model assignment}: We propose, Kernel Residual Density based hard inlier assignment to handle \textit{multi-structure inliers} (the inliers that belongs to multiple structures \eg~ In the case of line fitting, the data points at the intersection of two lines can be referred as \textit{multi-structure inliers}. 
    \item \textit{Multi-model fitting}: We combine the above modules to engineer an end-to-end automatic multi-model fitting solution that eliminates the need for user input.  
    \item \textit{Extensive Evaluation}: We extensively evaluate DGSAC on a wide variety of tasks like motion segmentation, planar segmentation, vanishing point estimation/line segment classification, plane fitting to the 3D point cloud, line and circle fitting.  
\end{itemize}

This paper is an extension of our previous work in \cite{tiwari2018dgsac}. The following are the substantially improved additions in this paper: Unlike residual density proposed in \cite{tiwari2016robust}, in this paper, we present an entirely new kernel version of the residual density. An early version of the density guided sampling process is presented in \cite{tiwari2018dgsac}. In this paper, we present a substantially new Kernel Density Guided Sampling (KDGS) process with an extensive evaluation and comparison with other competing methods. 

The remainder of the paper is organized as follows. We categorically discuss recent approaches for guided sampling and full multi-model fitting, and their limitations in Sec. \ref{sec:relted_work}, a formal problem statement is described in Sec. \ref{sec:prob_stat} followed by DGSAC Methodology in Sec. \ref{sec:dgsac_methodo}, the preliminaries are described in Sec. \ref{sec:prelim}. We introduce Kernel Residual Density in Sec. \ref{sec:krd}, guided hypotheses generation in Sec. \ref{sec:kdgs}, inlier noise scale estimation in Sec. \ref{sec:KRD_frac_est}, model selection algorithms in Sec. \ref{sec:GO_model} and point-to-model assignment in Sec. \ref{sec:p2m}. The extensive experimental analysis is shown in Sec. \ref{sec:exps} followed by discussion and conclusion in Sec. \ref{sec:ana_discussion}.

%% file: related_work.tex
\section{Related Work}
\label{sec:relted_work}
The robust multi-model fitting has a vast literature. In this section, we review recent approaches that best fit in the context of this work. Individually, we review approaches into two categories. 1. Approaches for guided sampling for multi-model fitting. 2. Approaches for full robust multi-model fitting.

\subsection{Guided Sampling for Multi-Model Fitting}

\noindent \textbf{Preference Analysis Based}: Most of the guided sampling strategies for multi-model fitting approaches exploit the concept of \textit{preferences} \cite{chin2012accelerated,wong2011dynamic}. Multi-GS \cite{chin2012accelerated} compute \textit{hypothesis preferences} based on \textit{ordered residuals} to derive a conditional distribution and use it for sampling minimal sample sets. DHF \cite{wong2013simultaneous} and ITKSF \cite{wong2013simultaneous} show improvement over Multi-GS \cite{chin2012accelerated}. Both DHF and ITKSF use \textit{hypothesis and data preferences} to simultaneous sample and prune hypotheses. DHF  \cite{wong2013simultaneous} follow a per data point strategy, where its goal is to associate a hypothesis to each data point. These approaches operate in a time budget framework, and there is no such stopping criteria or time budget constraint that ensures at-least one good hypothesis for all genuine structures. \\
\noindent \textbf{Matching Score Based}: Recently, two approaches EGHG \cite{lai2016efficient} and UHG \cite{lai2017unified} has been proposed. These approaches exploit matching scores of a feature matcher to derive the conditional distribution for sampling minimal sample sets. EGHG maintains two sampling loops a global and a local. A set of good hypotheses maintained by a global sampling loop is used by local sampling to impose the epipolar constraints further, to improve the hypothesis generation. UHF first cluster the data points using T-Linkage \cite{magri14tl} to prune out outliers. It uses the matching score of the clustered data points to derive conditional distribution for sampling minimal sample sets. However, the performance of T-Linkage \cite{magri14tl}, depends on the user-specified inlier threshold, which affects the performance of UHG. The concept of translating matching scores into inlier probability is not convincing because false matches may also have a high matching score. Also, this limits the usability of the algorithm to correspondence based applications. Both EGHG \cite{lai2016efficient}  and UHF \cite{lai2017unified}  require user-specified inlier threshold and operate in a time budget framework. A user provides the time as a reasonable guess.

\subsection{Full Multi-Model Fitting Approaches}
Irrespective of the hypothesis generation process, in this section, we categorize recent best performing full end-to-end multi-model fitting approaches based on their model selection strategies and review them. 

\noindent\textbf{Clustering Based}: J-Linkage \cite{jlink} and T-Linkage \cite{magri14tl} are two of widely used clustering-based approaches in this category. J-Linkage \cite{jlink} first represents the data points into their conceptual space and uses an agglomerative clustering called J-linkage to cluster data points belonging in the conceptual space. T-linkage \cite{magri14tl} is a variant of J-Linkage with a continuous conceptual representation of data points. T-Linkage \cite{magri14tl} show improvement over J-Linkage \cite{jlink}. However, the performance of both methods critically depends on the user-specified inlier threshold. \\
\noindent\textbf{Matrix Factorization Based}:
Recently, matrix factorization based approaches like RPA \cite{magri15rpa} or NMU \cite{tepper2016nonnegative} have been proposed for model selection and have shown to outperform clustering-based approaches. RPA assumes the knowledge of the inlier noise scale (albeit for the entire dataset) and the number of structures known a priori. It constructs a data point similarity matrix and decomposes it using symmetric NMF  \cite{symnmf15}. The decomposed matrix, along with the preference matrix, is used for final model selection. NMU outperforms RPA by enforcing an additional constraint of under-approximation. However, it also requires a user-specified noise scale estimate.

\noindent\textbf{Optimization Based}: This category of approaches form model selection as an optimization problem. QP-MF \cite{yu2011global} formulate model selection as quadratic program. RCMSA \cite{pham2014random} mode formulated the multi-model fitting problem in simulated annealing and graph cut framework. It uses data preferences to construct a weighted graph. RansaCov \cite{magri16set} formulate model selection as a set-coverage problem. However, the performance of these approaches critically depends on the number of models and inlier threshold provided by a user.\\

%% file: dgsac_overview.tex
\section{Notations}
A matrix is represented by capital bold \textbf{H}. Its respective \ith and \jth~ row and column are represented by its counterpart small bold as $\textbf{h}_i$ and $\textbf{h}^j$ respectively. The \ith~row and \jth column element of the matrix \textbf{H} is represented by small plain text as $h^j_i$. The sub-vector constituting the first $k$ elements of the row vector $\textbf{h}_i$ is represented by $\textbf{h}_i^{1:k}$. The sub-vector constituting the elements of the row vector $\textbf{h}_i$ whose indices are in the set $w$ is represented by $\textbf{h}_i^{\{w\}}$. The cardinality of a set $w$ is denoted by $|w|$. The mean of the elements of a vector a is denoted by $mean(\textbf{h}_i)$.
\section{Problem Statement}
\label{sec:prob_stat}
We consider a multiple model fitting scenario. Given is the data set $ \mathbf{X}=\{x^j,~j=1,\ldots,n\} $ of $n$ data points originated from $\kappa$ multiple instances of a geometric model. Here, we call $x^j$ \textit{a data point} in the abstract setting, it can be a point correspondence in homography fitting, or a 3D point in plane fitting, a line in vanishing point estimation, a 2D point in line fitting. Let the fraction of inliers of each model instance is denoted by $f_{i}=\frac{\vert\mathcal{I}_i\vert}{n}, ~i=1,...,\kappa  $, where $\mathcal{I}_{i}$ is an index set of inlier points of $i^{th}$ model instance and $ f_{0}=1-\sum_{i=1}^{\kappa}f_{i} $ denotes the fraction of gross outliers. Our goal is to output a set of $\kappa$ tuples $(\textbf{h}_1,\mathcal{I}_1), (\textbf{h}_2,\mathcal{I}_2),...,(\textbf{h}_{\kappa},\mathcal{I}_{\kappa}) $, where $\textbf{h}_i$ and $\mathcal{I}_i$ are the estimated model parameters and inlier set of the $i^{th}$ model instance respectively. \\ \textbf{Note:} We assume the number of models $\kappa$, inlier fractions $f_{i}$'s ( or the respective inlier noise scales $\sigma_i$'s) are \textit{unknown.} However, if provided these parameter can easily be accommodated in the proposed pipeline.

\section{DGSAC Methodology}
\label{sec:dgsac_methodo}
DGSAC begins with generating quality hypotheses using density guided sampling (KDGS) (Sec. \ref{sec:kdgs}). The KDGS aims to generate good model hypotheses for each data point. The sampling process is guided by the conditional distribution derived from density-based point correlation and potential good model hypotheses of each data point (Sec. \ref{sec:kdgs_flow}). KDGS maintains a density-driven \textit{explanation score} for each data point to ensure every point is explained by at least one good hypothesis. Once the KDGS terminates, a model hypothesis is assigned to each data point based on Kernel Residual Density. The density and residual profiles of the hypotheses obtained from KDGS are used to estimate the inlier noise scale (Sec. \ref{sec:KRD_frac_est}), which is then used to compute the goodness score of each model hypothesis. Hypotheses, along with their goodness score, are then feed into either greedy or optimal model selection algorithm (Sec. \ref{sec:GO_model}). On the one hand, the greedy algorithm (Sec. \ref{sec:gms}) starts with selecting a hypothesis with a high goodness score, maintaining a model diversity based on the overlap between estimated inlier sets of already selected and the remaining hypotheses.
On the other hand, optimal model selection (Sec. \ref{sec:oms}) is solved as a quadratic program. While the quadratic program's objective is to maximize the total goodness score, we further impose the penalty (Sec. \ref{sec:dpm}) of selecting similar hypotheses. The final set of hypotheses obtained by greedy or optimal model selection along with their inlier set are then fed into a point-to-model assignment (Sec. \ref{sec:p2m}) to get the final output of the DGSAC pipeline.

%% file: building_blocks.tex
\section{Preliminaries}
\label{sec:prelim}
A model hypothesis $\mathbf{h}_i$ can be obtained by fitting the model equation to a set of data points. Usually, this set contains the minimum number of data points required to estimate the model parameters, such a set is known as \emph{minimal subset} (MSS). For example, if the geometric structure is a line, the cardinality of the minimal subset $ \eta$ is $2$, and $\bhi$ contains the value of slope and the line's intercept. \\
\noindent \textbf{Residual.} Residual of a data point $x^j$ \wrt~a hypothesis $\bhi$ is a measure of disagreement of the data point \wrt~ $\bhi$. It is computed using model specific residual function. Assume that we have generated a total of $m$ model hypotheses $\bH = \{ \bhi \}_{i=1}^{m}$. We compute residual of all data points with respect to the hypothesis $\bhi$ using model specific residual function, which is defined as $\psi(\textbf{h}_i,x_j): \mathbb{R}^d \rightarrow \mathbb{R}_+$ and store in a residual vector $\bri$ as given in equation \eqref{eqn:res_fn}.

\begin{equation}
\label{eqn:res_fn}
\textbf{r}_i=[r_i^1=\psi(\textbf{h}_i,x_1),...,r_i^n=\psi(\textbf{h}_i,x_n)]
\end{equation}

\subsection{Hypothesis/Point preferences}
\label{sec:hyp_pt_pref}
Preference of a hypothesis is the rank ordering of all the data points based on some criteria \eg~residual \cite{chin2012accelerated} or residual density \cite{tiwari2018dgsac} and vice-versa, the preference of a data point is the rank ordering of all model hypotheses. To compute \textit{residual} based \textit{hypothesis preferences}, we sort residuals in the vector $\bri$ in ascending order $r_i^{q_i^{1}} \leq r_i^{q_i^{2}}\leq ,..., \leq r_i^{q_i^{n}}$ to get the sorted residual vector $\boldsymbol{\rho}_i=[r_i^{q_i^{1}}, r_i^{q_i^{2}},..., r_i^{q_i^{n}}]$. The ordered indices of data points $\bqi = [q_i^{1},q_i^{2},...,q_i^{n}]$ encodes the preferences of the hypothesis $\bhi$. Similarly we order $\brj$
in ascending order $r^j_{l_j^{1}} \leq r^j_{l_j^{2}}\leq ,..., \leq r^j_{l_j^{m}}$. The ordered indices of data points $\blj = [l^j_{1},l^j_{2},...,l^j_{m}]$ encodes the preferences of the data point $\xj$. Since, we use \textit{residual} as a ordering criteria, therefore we refer $\bqi$ and $\blj$ as \textit{residual based hypothesis} and \textit{point preferences} respectively.

\input{KRD_pref_figure}
\noindent Preference analysis has been widely used in robust multi-model fitting. Previous approaches  \cite{chin2012accelerated,wong2011dynamic,wong2013simultaneous,yu2011global} uses residual as a criteria to derive point/hypothesis preferences. We demonstrate with an example in Fig. \ref{fig:krd_pref}, a data point's top preference (a hypothesis to which it has the smallest residual) does not always correspond to a good model hypothesis (a model hypothesis that best describes the data point and the underlying model). The residual-based point's preference is conditioned only on the residual of the data point itself, and it ignores what other points in the neighborhood are preferring. We show point's preference derived from the proposed \textit{Kernel Residual Density} (KRD), incorporate local consensus information, and provide robust, meaningful preferences. 

In the next section, we introduce \textit{kernel residual density} (KRD) for computing the KRD points preferences. First We mathematically define the KRD and then discuss its importance. KRD is the most important component of DGSAC, we will use it for computing point correlation (sec. \ref{sec:KRD_point_corr}), guiding hypothesis generation (sec. \ref{sec:kdgs}), inlier noise scale estimation (sec. \ref{sec:KRD_frac_est}) and model selection (sec. \ref{sec:GO_model}).

\section{Kernel Residual Density}
\label{sec:krd}
Previously in  \cite{tiwari2018dgsac}, we have shown that residual density provides crucial information that helps differentiate inliers from outliers. It can be used at various stages of a multiple-model fitting pipeline and is estimated directly from the data without any user input. Here, we introduce a new variant of residual density, which we refer to as \textit{variable bandwidth kernel residual density}. Mathematically, we define kernel residual density (KRD) at a data point in the residual space of the hypothesis $\bhi$ as follows: 
\begin{equation}
\begin{aligned}
d^{q_i^j}_i=\Phi(\bhi,x_{q_i^j})=&\frac{1}{n} \sum_{k=1}^{n} \frac{1}{b^{q_i^j}_{i}} K\Big(\frac{\rho_i^j-\rho_i^k}{b^{q_i^j}_{i}}\Big)\\
=&\frac{1}{n} \sum_{k=1}^{n} \frac{1}{b^{q_i^j}_{i}} K\Big(\frac{r_i^{q_i^j}-r_i^{q_i^k}}{b^{q_i^j}_{i}}\Big)
\end{aligned}
\label{eq:krd_eq}
\end{equation}
Where, $b^{q_i^j}_{i}$ is the variable bandwidth at the data point $x_{q_i^j}$ in the one-dimensional residual space of $\bhi$. Here, $K(.)$ can be any kernel function that must be symmetric around the kernel origin. In this paper, we use Epanechnikov kernel \eqref{eq:epan}. Other kernels like Gaussian, etc., can also be used.
\begin{equation}
\begin{aligned}
K(u) = \frac{3}{4}(1-u^2)\\
\text{s.t}~~~~ |u| \leq 1
\end{aligned}
\label{eq:epan}
\end{equation}
The KRD captures the density of the data points in the residual space.  \\

The critical factor in our KRD formulation is the choice of variable bandwidth. We choose the bandwidth $b^{q_i^j}_{i}$ at the data point $x_{q_i^j}$ in the 1-dimensional residual space of hypothesis $\bhi$ to be $\rho_i^j$. Our choice of variable bandwidth ensures that only one dominating density peak would exist around the regression surface. All the data points with residual smaller than $\rho_i^j$ contribute in the density estimation at data point $x_{q_i^j}$ in 1-D residual space of $\textbf{h}_i$. On the other hand, if we fix the bandwidth, it would give rise to many density peaks due to spurious local structures. This behavior is also shown in  \cite{wang2004robust}.\\
An example of kernel residual density computed for a good and bad hypothesis is shown in Fig. \ref{fig:krd_pref}(d). The corresponding good and bad hypotheses are shown in Fig. \ref{fig:krd_pref}(c). It can be seen from the plots; for a good hypothesis, the kernel residual density is large near the regression surface (inlier region) compared to the outlier region. For a bad hypothesis, KRD is nearly flat throughout except a  small peak near the regression surface. \\
Next, we discuss the advantage of using Kernel Residual Density, but first, we introduce KRD based point preferences.

\subsection{KRD Based Point Preferences}
\label{sec:krd_based_pt_pref}
Unlike residual based point preference in Sec. \ref{sec:hyp_pt_pref} where for each data point we ordered hypotheses in \textit{increasing} order of their residual values, in density based point preferences we rank ordered hypotheses based on \textit{decreasing} order of kernel densities. For each data point $x^j$, we find permutation $\textbf{v}^j = [v^{j}_{1},v^{j}_{2},...,v^{j}_{m}]$ such that $d^j_{v_1^{j}} \geq d^j_{v_2^{j}}\geq ,..., \geq d^j_{v_m^{j}}$. The permutation vector $\textbf{v}^j$ encodes the KRD based point preference of \jth~data point. \\ \textit{Note:} hypotheses are ordered in decreasing order of their density values.

\subsection{Why Kernel Residual Density?}
We demonstrate the benefit of Kernel Residual Density using ea line fitting example in Fig. \ref{fig:krd_pref}. From a pool of hypotheses, we plotted stop-1 preferred hypothesis of 5 different inliers of the ground truth structure. The green hypotheses are the residual-based top-1 preference of \jth~ data point (\ie~ hypothesis with index $l^j_1$), while in magenta are the KRD based top-1 preferred hypothesis (\ie~ hypothesis with index $v^j_1$). We can see only density-based top-1 preferences are the good hypotheses that belong to a genuine geometric structure. This phenomenon is because, the kernel density captures the local consensus information, in the form of data points having similar residuals. Intuitively, density-based preference of a data point $x^j sc$'s, indirectly accounts all the data points lying within the variable bandwidth $b^{q_i^j}_{i}$ of a hypothesis $\textbf{h}_i$ at data point $x^j$.

Another advantage of using kernel density is that it helps differentiate inliers and outliers. In the case of a good hypothesis, its inliers due to their smaller residuals form a dense cluster around the regression surface. In contrast, its outliers due to comparative larger residuals form a nearly flat region away from the regression surface. The above phenomenon can be verified from the kernel residual density profiles of a good and bad hypothesis shown in Fig. \ref{fig:krd_pref}(d) . We can see that the good hypothesis has a higher density around the regression surface and a nearly flat region as we move away from the regression surface. On the contrary, we can see a bad hypothesis has a nearly flat density profile throughout except for a small peak near the regression surface. In the following sections, we show the use of KRD in various components of DGSAC. \\

Next, we describe the density-based point correlation and discuss its advantage over residual-based point correlation.

\subsection{KRD based point correlation}
\label{sec:KRD_point_corr}
We defined pairwise point correlation as the fraction of overlapped top-T point preferences. It is computed by applying the intersection kernel over top-T preferences as shown in Eq. \ref{eqn:den_pcm}, which shows the pairwise point correlation between data point $x^i$ and $x^j$. Here, $\textbf{v}^{i}_{1:T} $ and $\textbf{v}^{j}_{1:T}$ are top-T preference of point $x^i$ and $x^j$ computed as shown in Sec. \ref{sec:krd_based_pt_pref}.

\begin{eqnarray}
c_i^j = \frac{\textbf{v}^{i}_{1:T} \cap \textbf{v}^{j}_{1:T}}{T}
\label{eqn:den_pcm}
\end{eqnarray}
Let the pariwise KRD based point correlation matrix (PCM) is denoted by matrix \textbf{C}, where its elements are computed using Eq. \ref{eqn:den_pcm}.

\input{point_corr_visual}
\subsubsection{Analysis of Residual and KRD Based PC}
Similar to the KRD based point correlation, we can compute residual-based point correlation by replacing the density-based preferences $\textbf{v}^J$ with residual-based preferences $\textbf{l}^j$ (Sec. \ref{sec:hyp_pt_pref}) in the Eq. \ref{eqn:den_pcm}. We use \emph{breadcubechips} example from the AdelaideRMF dataset  \cite{wong2011dynamic}, to show a qualitative comparison between residual and KRD based PCM in Fig. \ref{fig:res_vs_den_PCM}. Ideally, a good point correlation computation strategy is the one, which provides a high correlation between a pair of inliers of the same structure and zero correlation otherwise. We show residual and KRD based point correlation matrix in Fig. \ref{fig:res_vs_den_PCM}(a) and \ref{fig:res_vs_den_PCM}(b) respectively, with brighter pixels indicating higher pairwise correlation. For better illustration, the points are ordered by structure membership, with the top set of rows being the gross outliers. For each three structures \textit{bread}, \textit{cube} and \textit{chips} The percentage of \textit{uncorrelated} outliers (\ie~zeros pairwise correlation value computed using Eq. \ref{eqn:den_pcm})  varying with the iteration of Algo. \ref{kdgs} is plotted in \ref{fig:res_vs_den_PCM}(d), (e) and (f) respectively. At each iteration, the percentage is significantly larger for density-based PC than residual, which indicates that a high percentage of inliers are correlated with the other inliers of the same structure. Since the inliers of a hypothesis have a non-zero correlation with the other inliers and zero correlation with the outliers, any information derived from KRD based PC \eg conditional PMF in Sec. \ref{sec:kdgs} is robust to the outliers. The value of $ T $ is set after thorough empirical validation independently for both the residual and KRD based PCM. In the case of residual-based PCM, the choice of $ T\!=\!\lfloor 0.1m\rfloor$ is selected according to the prior work  \cite{chin2012accelerated,wong2013simultaneous}. In this work (DGSAC), we use \textit{only} KRD based point correlation and the value of $ T $ is set to $T\!=\!5$ for \emph{all} experiments in Sec. \ref{sec:exps}. Since the value of $ T $ is the function of $ m $ for residual based PC, its computation time will increase linearly with $m$, while for density-based PC it will be constant \wrt $m$ because the value of $T\!=\!5$ is fixed. The constant computation time of pairwise density-based correlation makes it more advantageous over the residual-based PC.

%% file: KRD_pref_figure.tex
\begin{figure*}[h!]
	\vspace{-0.5cm}
	\centering
	\subfloat[]{ \includegraphics[width=4.1cm]{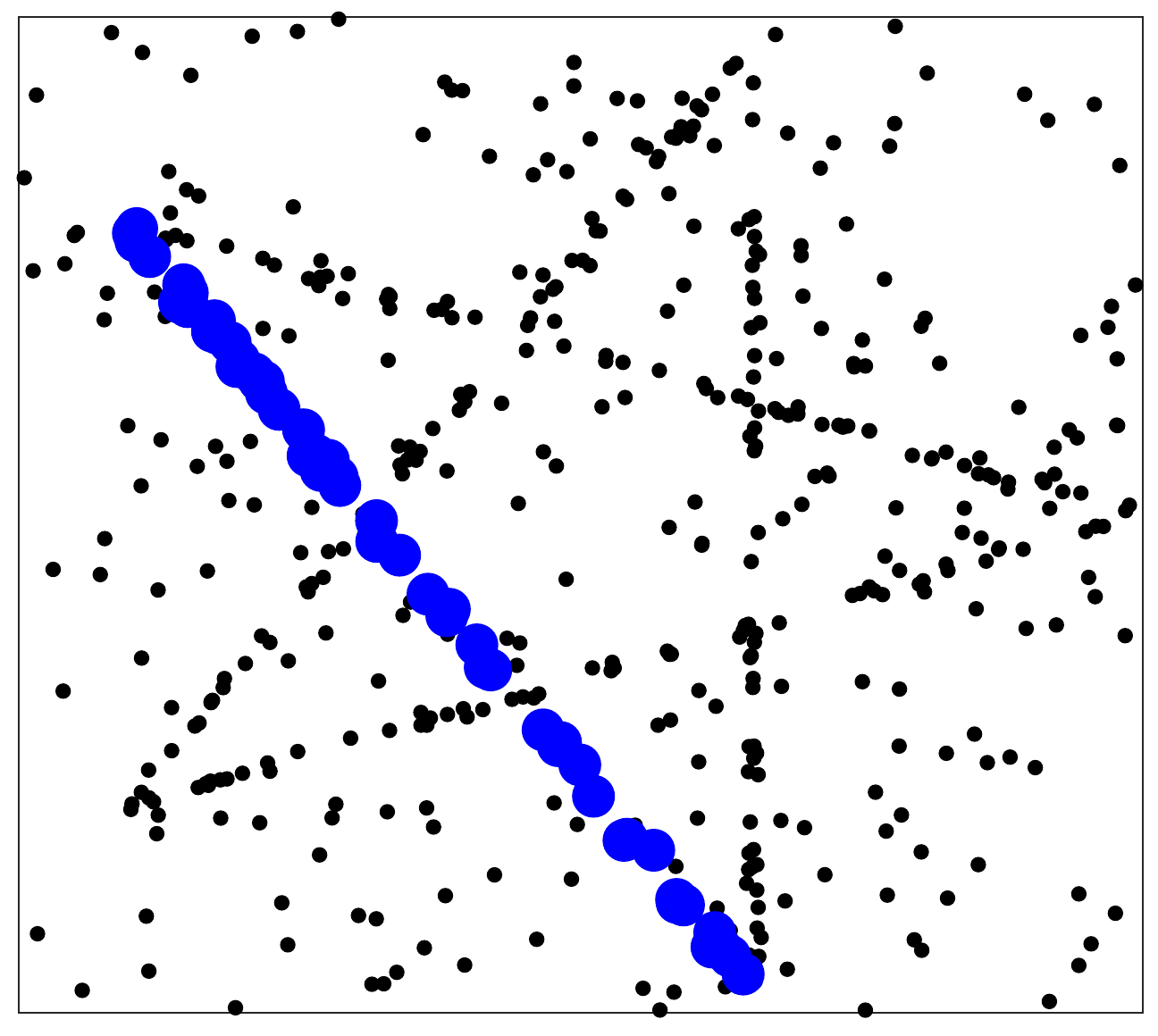}} 
	\subfloat[]{ \includegraphics[width=4.1cm]{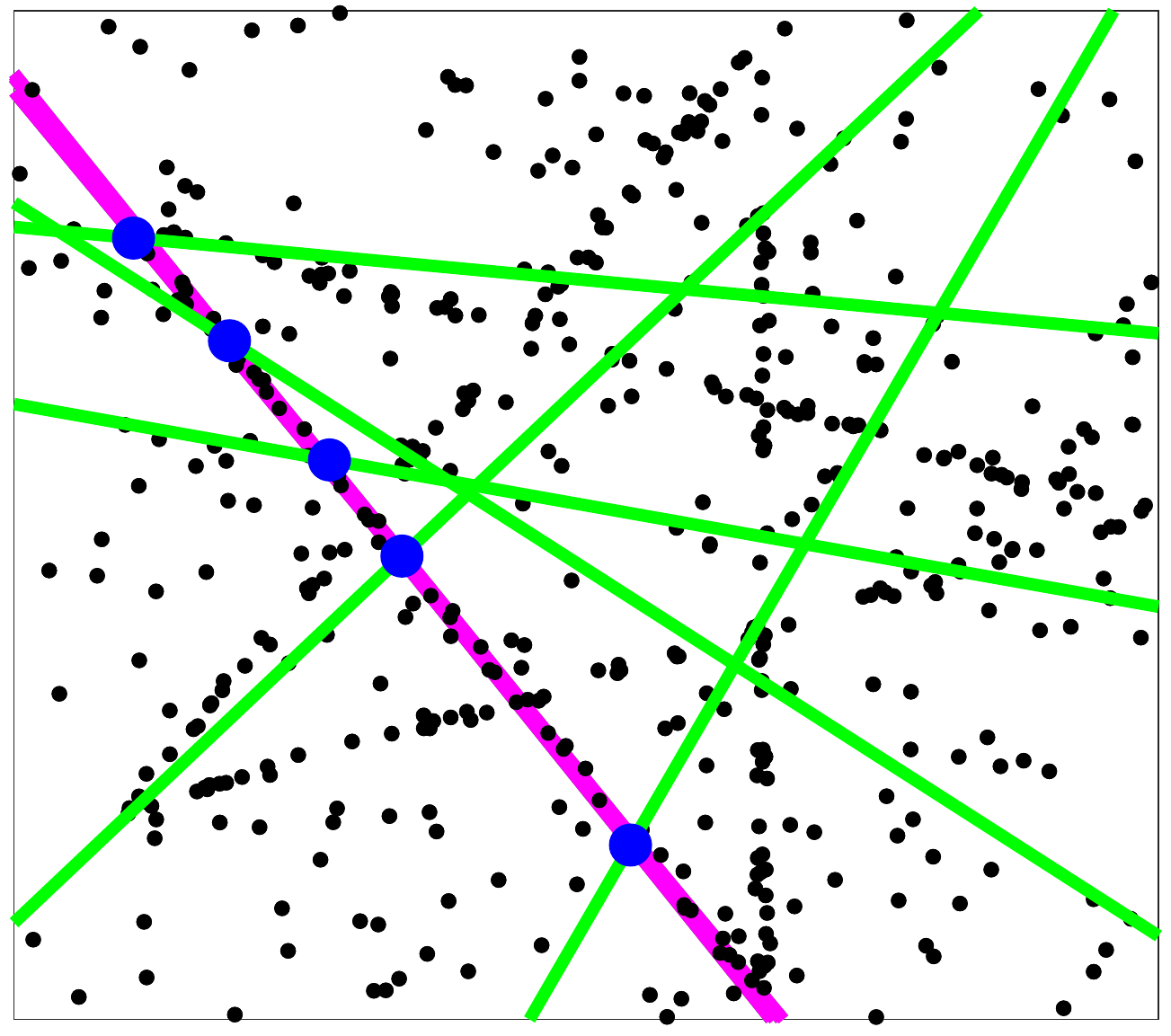}}
		\subfloat[]{ \includegraphics[width=4.1cm]{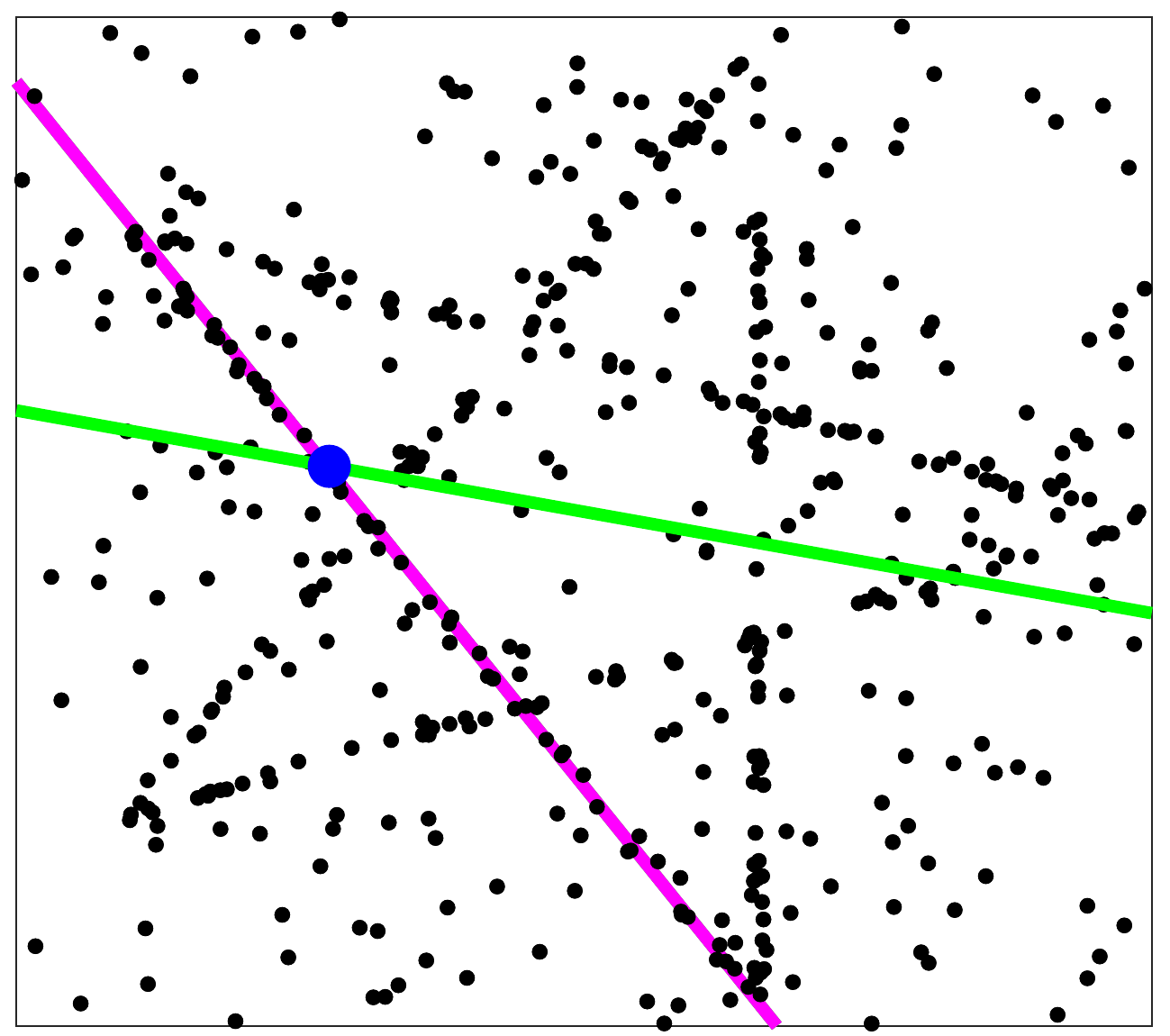}}
			\subfloat[]{ \includegraphics[width=4.7cm]{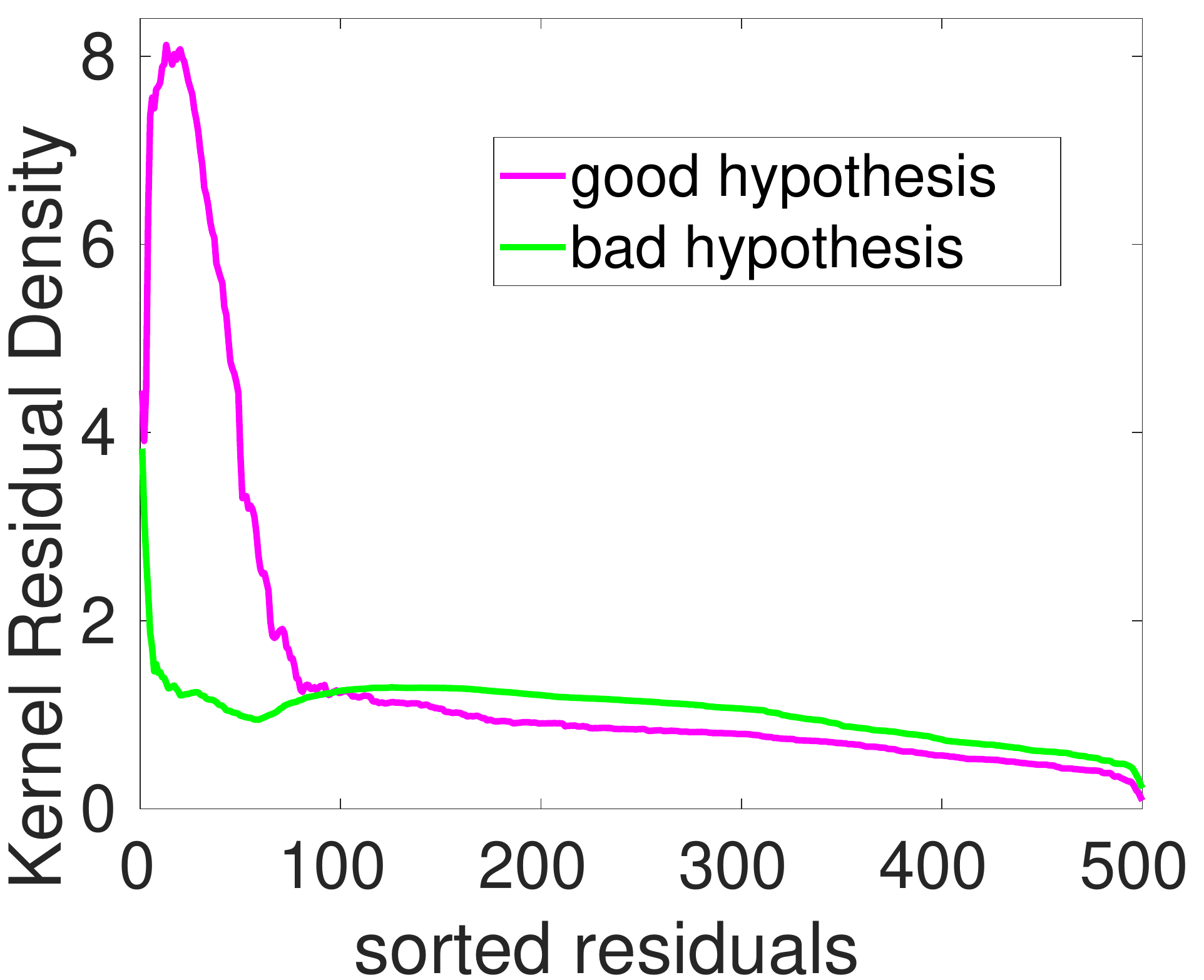}}
	\caption{ (a) Ground-Truth Structure, (b) Density based (magenta) and Residual based (green) top-1 preference of 5 inliers of the ground-truth structure (blue). It can be seen, density based top preference of the data points preferring the hypothesis from the dense structure, while residual based preferences are arbitrary hypotheses to which data points have small residuals only. (c) a good (magenta) and a bad (green) hypothesis of the ground-truth structure in (a), (d). the Kernel Residual Density profile of corresponding good and bad hypothesis shown in (c). It is expected for a good hypothesis to have high density around regression surface (equivalently inliers are densely packed around the regression surface). For a bas hypothesis density would be nearly flat except a very small pear around the regression surface due to small spurious structures.}%
	\label{fig:krd_pref}%
\end{figure*}

%% file: point_corr_visual.tex
\begin{figure}
\begin{tabular}{ccc}
\includegraphics[width=3.7cm,height=3.7cm]{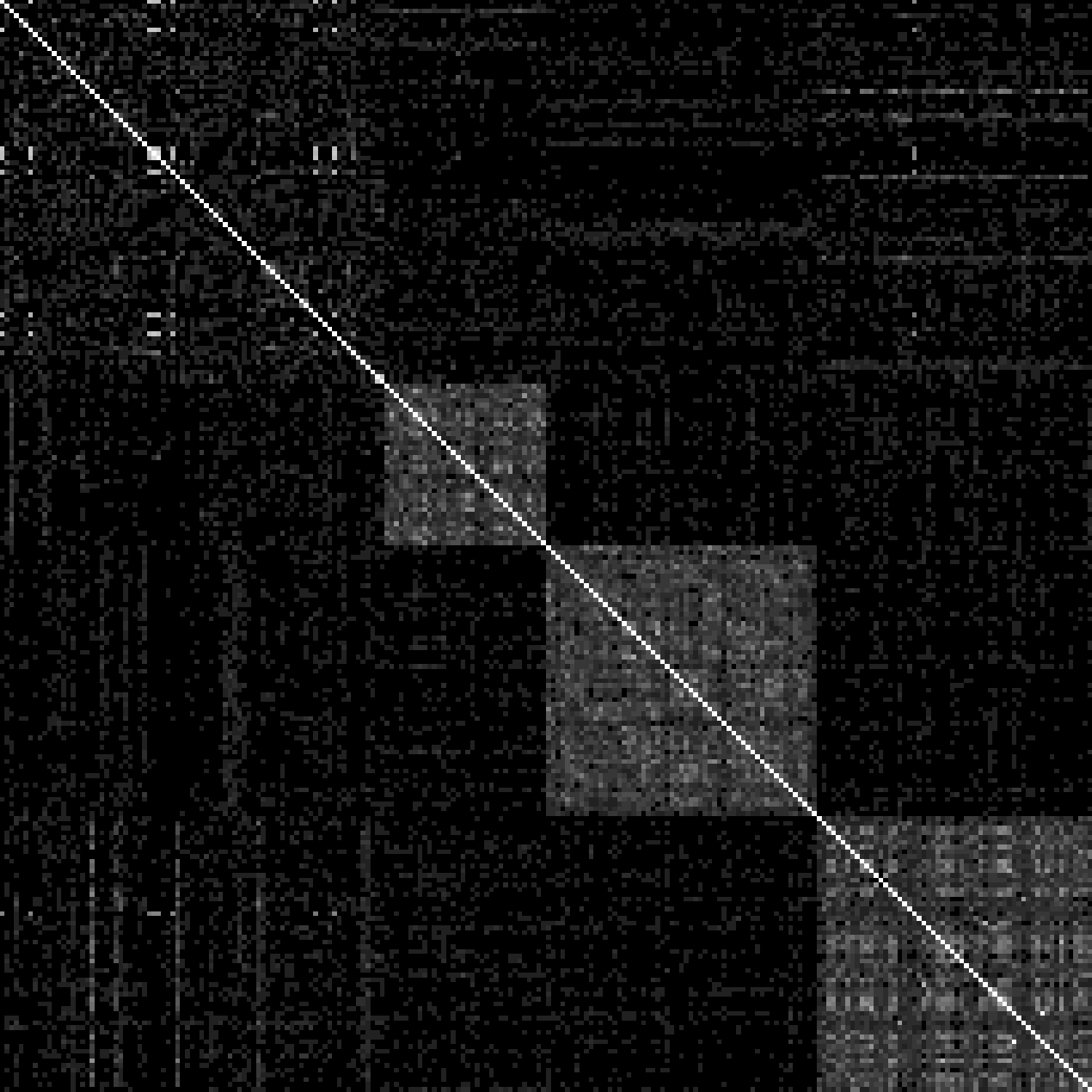}
& \includegraphics[width=3.7cm,height=3.7cm]{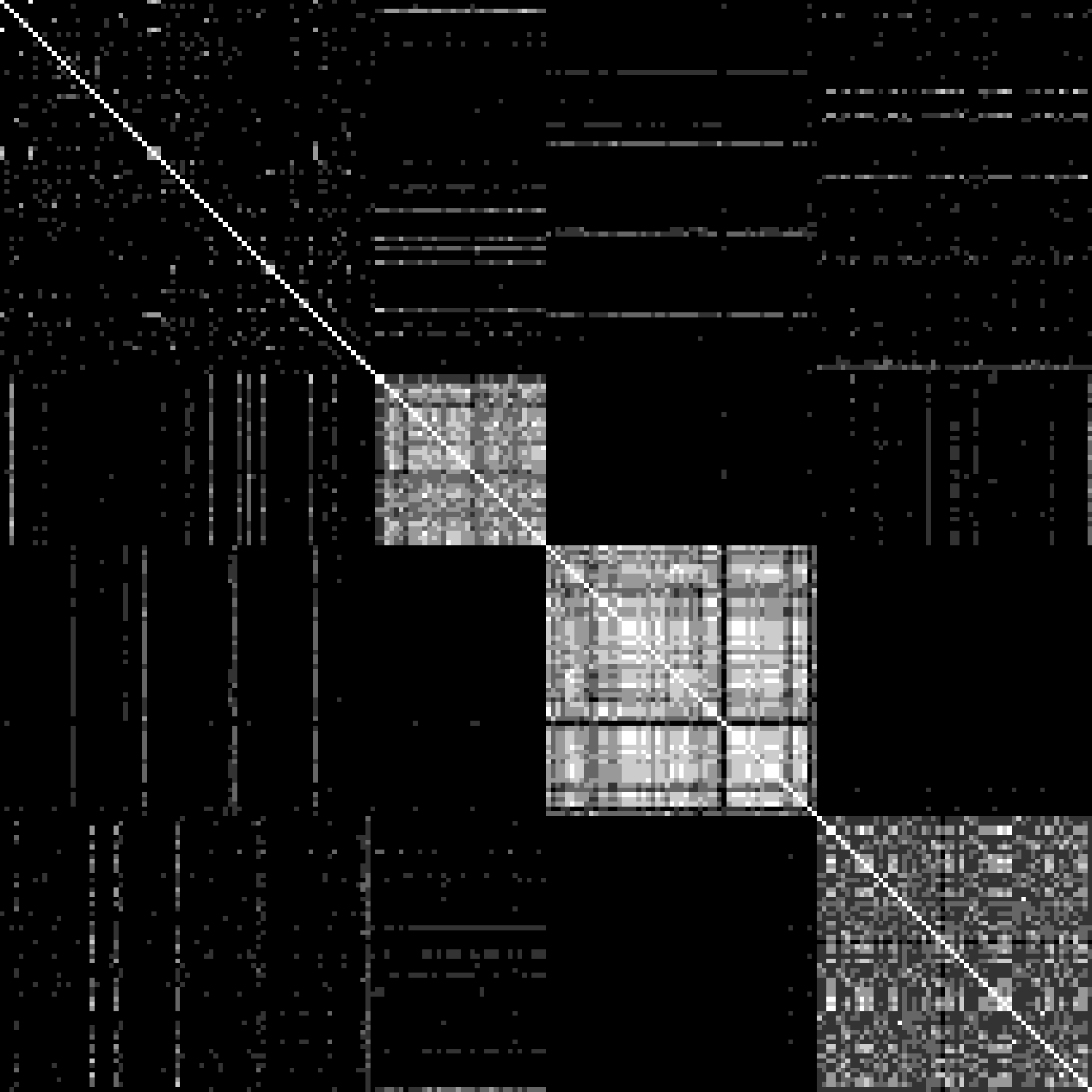} &
 \\
 (a) Residual Based PC & (b) Density Based PC
 \\
 & & \\
  \hspace{-1.1cm} \includegraphics[width=2.6cm]{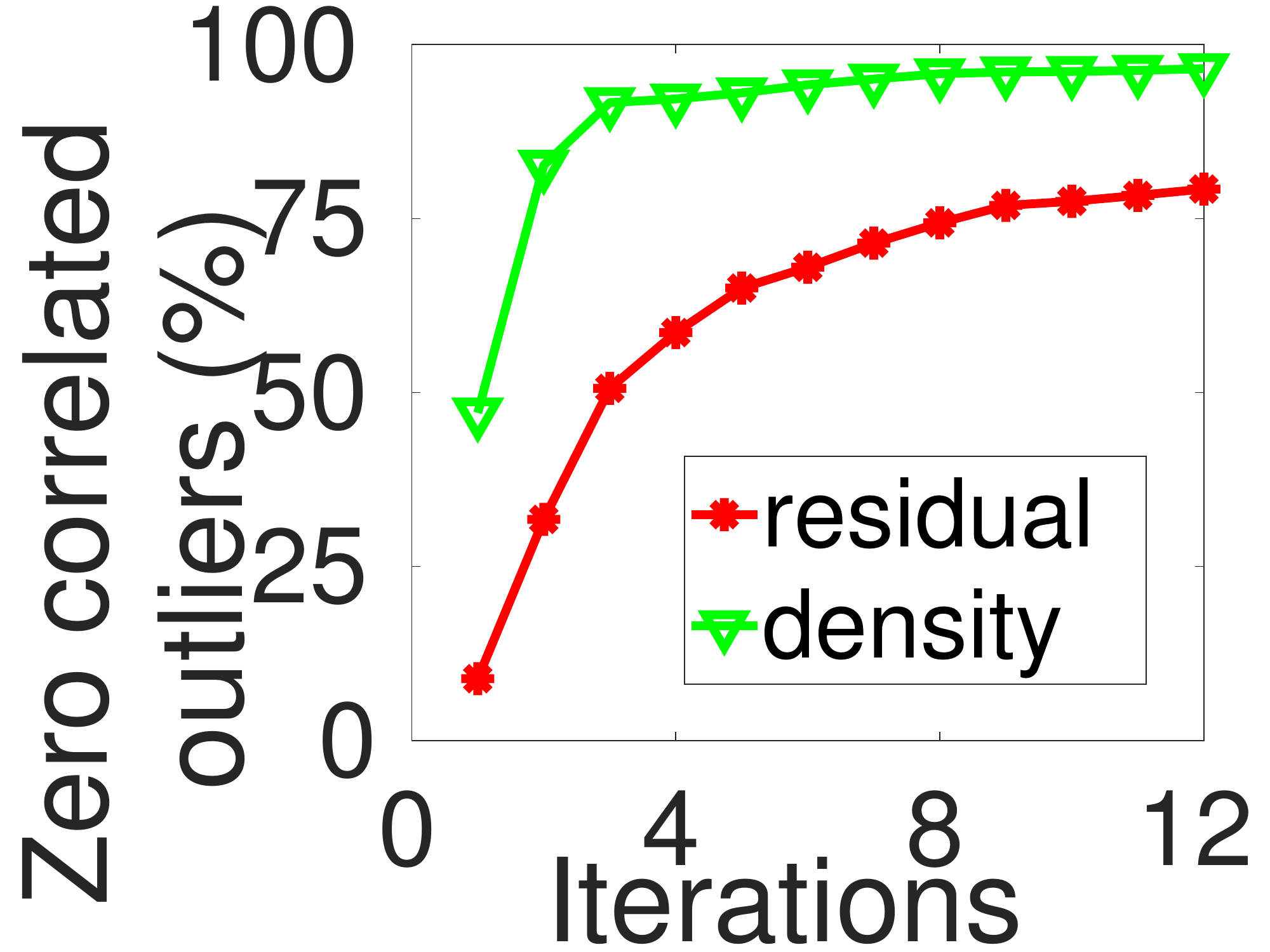}        &    \hspace{-4.0cm}\includegraphics[width=2.6cm]{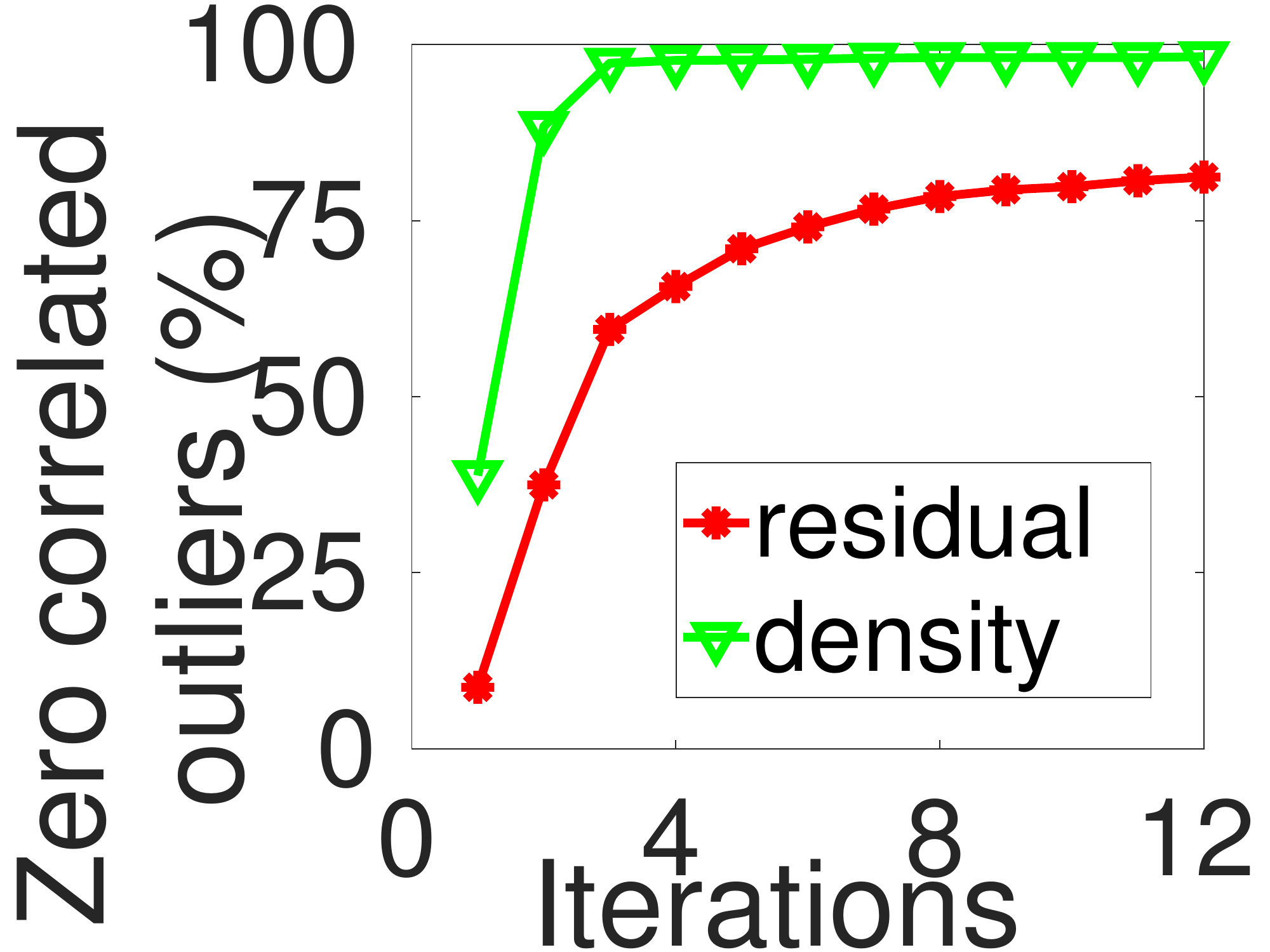}       &   \hspace{-3.3cm}\includegraphics[width=2.6cm]{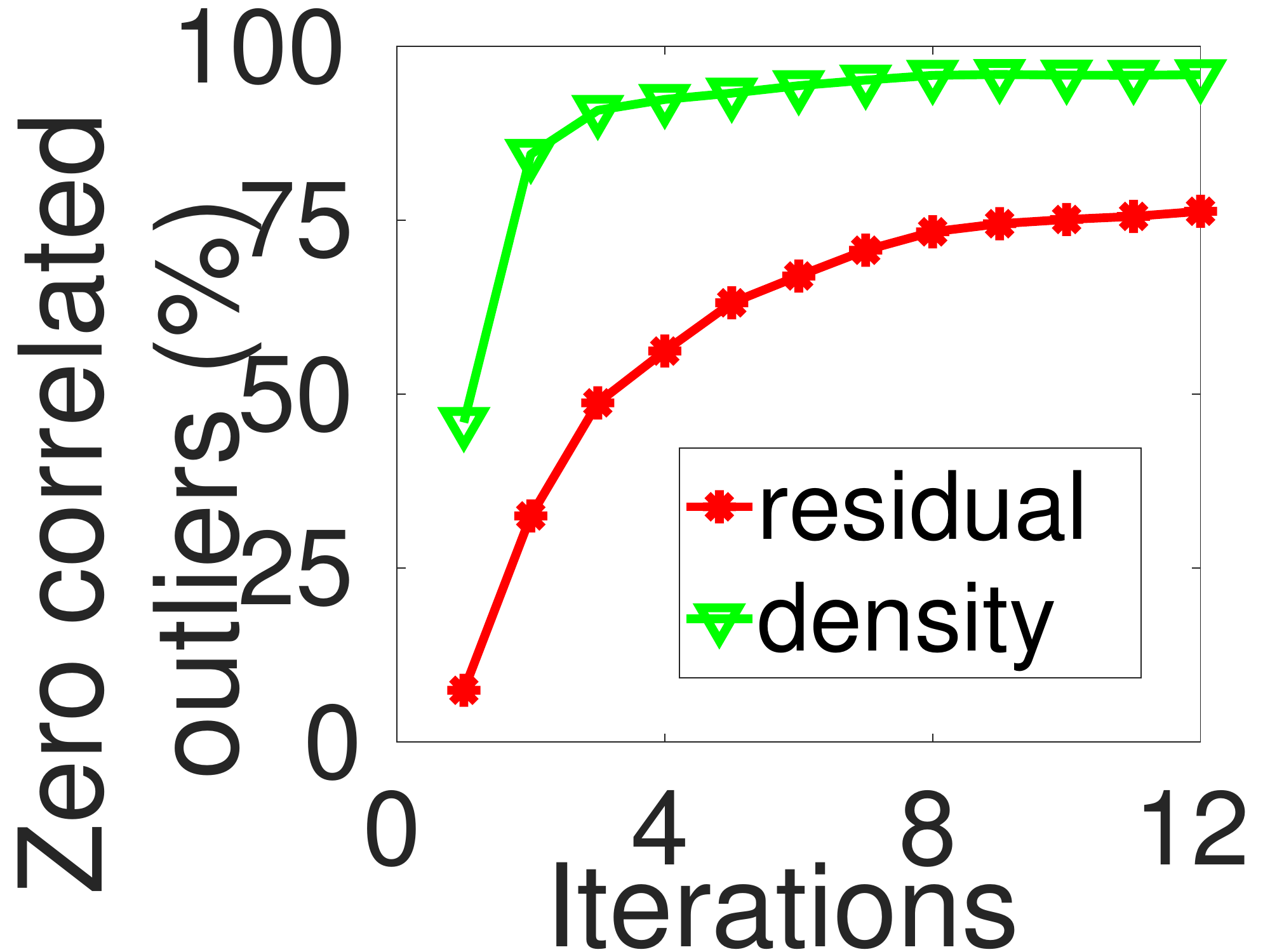}                \\
  \hspace{-0.5cm}  (c) bread      &  \hspace{-4.0cm} (d) cube        &  \hspace{-3cm}  (e)  chips            
\end{tabular}%
\caption{ \textbf{Residual vs. KRD based point correlation}: $breadcubechips$ example of AdelaideRMF  \cite{wong2011dynamic} dataset (a) Residual based and (b) Density based point correlation (PC) matrix. For better visualization the rows and columns in (a) and (b) are ordered by structure membership. For each of the three structures $bread$, $cube$ and $chips$, plots in (c), (d) and (e) show percentage of \emph{uncorrelated} outliers varying with iterations of (Algo. \ref{kdgs}) respectively.}
\label{fig:res_vs_den_PCM}
\end{figure}

%% file: KDGS_sec.tex
\section{KRD Guided Hypotheses Generation}
\label{sec:kdgs}

Hypothesis generation is the critical component of a multi-model fitting pipeline that operates in a \textit{hypothesize-and-test} framework. The quality of generated hypotheses significantly influences the final result of the multi-model fitting pipeline, i.e., classification of data point into their respective genuine structures. The purpose of a guided hypothesis generation process is to generate good hypotheses for all possible underlying genuine structures. Generating a hypothesis is equivalent to sampling its corresponding minimal sample set (MSS) followed by a deterministic step of the fitting model equation to the MSS elements. The goal of a guided sampling algorithm is to sample many \textit{pure} MSS (containing elements that are inliers to the same structure). 

Most of the multi-structure guided sampling algorithms take the following approach: sample the first element from a uniform distribution and sample the other elements from a conditional probability mass function (CPMF), which is usually derived from the PCM \cite{chin2012accelerated,wong2013simultaneous}. In the proposed Kernel Density Guided Sampling (KDGS) algorithm, we adopt a similar strategy except that we sample \textit{first element} of MSS \textit{deterministically} and the remaining from the CPMF derived from point correlation matrix (PCM) and potential good hypotheses of the first sampled element. 
The complete KDGS algorithm is illustrated in (Algo. \ref{kdgs}), which we describe below in detail.

\input{KDGS}

\subsection{Complete flow of KDGS}
\label{sec:kdgs_flow}
We first outline the complete flow of KDGS and then describe each component in detail. KDGS maintain an index set $\nu$ that contains indices of data point for which hypotheses is yet to be generated. KDGS starts with initializing the set $\nu=[1,...,n]$ (Algo. \ref{kdgs},~line \ref{kdgs:l3}), the PCM \textbf{C} and sampling weight matrix from potential good hypotheses \textbf{S} (described later) of size $n\times n$ with all elements equals to one (Algo. \ref{kdgs},~line \ref{kdgs:l2}). The initial \textbf{C} matrix represent that every point pair is equally correlated. We use routine \texttt{generateHyps} (Algo. \ref{kdgssub}), with inputs $\textbf{C},\textbf{S},\nu$ and $\eta$ to generate hypotheses. For each data point $j \in \nu$, the routine \texttt{generateHyps} generate a hypothesis by sampling $\eta$ points using conditional PMF derived from $\textbf{c}^j$ and $\textbf{s}^j$. The important steps in Algo. \ref{kdgssub} are in lines \ref{kdgssub:l4},\ref{kdgssub:l6}-\ref{kdgssub:l8}, where for each \jth~ data point ($j \in \nu$), we deterministically select the data point itself to be the first element of the MSS ($\mcalM$) (Algo. \ref{kdgssub},~line \ref{kdgssub:l4}) and the remaining $\eta-1$ elements of $\mcalM$ \textit{without replacement} (enforced by Algo. \ref{kdgssub},~line \ref{kdgssub:l7}) using the conditional PMF derived from $\textbf{c}^{j}$ and $\textbf{s}^j$ (Algo. \ref{kdgssub},~line \ref{kdgssub:l6}) obtained from KRD point correlation and potential good hypotheses (Algo. \ref{kdgs},~line \ref{kdgs:l18}). Using the data points with indices in the set $\mcalM$, we generate a model hypothesis $ \textbf{h} $ using the function {\tt fitModel} (Algo. \ref{kdgssub},~line \ref{kdgssub:l11}) , which is deterministic function and is known a priori for different model fitting tasks (plane, homography, vanishing point, etc.). We collect the $ |\nu| $ hypotheses in $ \textbf{H} $ (Algo.\ref{kdgssub},~line \ref{kdgssub:l12}) and compute the corresponding residual matrix $ \textbf{R} $ (Algo.\ref{kdgssub},~line \ref{kdgssub:l14})  and residual density matrix $ \textbf{D} $ (Algo.\ref{kdgssub},~line \ref{kdgssub:l15}).

We combine the outputs of the routine \texttt{generateHyps}, the generated hypotheses \textbf{H} and the corresponding residual (\textbf{R}) and kernel density matrix \textbf{D} in the current iteration with our previous set (Algo. \ref{kdgs}, line \ref{kdgs:l11}). With a slight abuse of notation we use set union even with the matrices, only to emphasize uniqueness of rows after the update. The density profiles of newly generated hypotheses are then normalized and scaled (Algo. \ref{kdgs},~line \ref{kdgs:l5}-\ref{kdgs:l7}). The updated density matrix is used to recompute the rows of point preference matrix $ \textbf{V} $ (Algo. \ref{kdgs}, line \ref{kdgs:l12}) as described in (Sec. \ref{sec:krd_based_pt_pref}). For each \jth~data point KDGS maintain an \textit{explanation score} $\tau^j$, which is the mean of its kernel densities \wrt~ all the potential good hypotheses in  $\theta^j$ (Algo. \ref{kdgs},~line \ref{kdgs:l14}). The \textit{explanation score} of all data points in the current iteration $\tau_{curr}$ is compared with the previous iteration $\tau_{prev}$, if it stop changing significantly we exclude those data points from the set $\nu$. For the remaining points in the set $\nu$ we update the corresponding pairwise correlation values in the point correlation matrix \textbf{C}. The updated correlation matrix and sampling weight $\textbf{s}^j,~\forall j \in \{1..n\}$ computed using density and residual based best potential good hypotheses (Algo. \ref{kdgs},~line \ref{kdgs:l14}-\ref{kdgs:l18}), is further passed to \texttt{generateHyps} routine. The whole process repeat until the set $\nu$ is empty. Finally, we retain the unique set of hypotheses which are in the top-$ 1 $ preference of all points and the corresponding residual and density matrices (Algo. \ref{kdgs},~lines \ref{kdgs:l25}-\ref{kdgs:l26}).  \\

\input{KRDGS_sub}

\noindent\textbf{Scale density profiles.} To make density profiles of all hypotheses in \textbf{H} comparable we normalize them sum to 1 (Algo. \ref{kdgs},~line \ref{kdgs:l5}. We further scale them with the disparity in the mean kernel density of top-$\beta$ and bottom-$\beta$ KRD based hypothesis preferences (Algo. \ref{kdgs},~line \ref{kdgs:l6}-\ref{kdgs:l7}). Similar to KRD based point preferences in (Sec. \ref{sec:krd_based_pt_pref}), we can find KRD based preferences of \ith~ hypothesis in vector $\textbf{w}_i = [w_{i}^{1},w_{i}^{2},...,w_{i}^{n}]$ such that $d_i^{w_{i}^{1}} \geq d_i^{w_{i}^{2}}\geq ,..., \geq d_i^{w_{i}^{n}}$. 

\noindent\textbf{Potential good hypotheses of a data point.} We define a hypothesis is potentially good for a data point (equivalently, it can describe the underlying structure to which the data point potentially be an inlier) if the point lies within $\beta$ neighborhood in its residual space. Let the set of all potential good hypotheses of $j^{th}$ data point is denoted by $\theta^j$ (Algo. \ref{kdgs},~line \ref{kdgs:l14}). We fix $\beta=2\eta$, it can be thought of the size of smallest \textit{genuine} structure. For fitting tasks like line, plane, circle, and vanishing point, where $\eta$ is too small ($<5$), we set $\beta$ to be at-least $15$. 

\input{KDGS_analysis}

\noindent\textbf{Conditional PMF.} For each \jth~ data point we first compute its potential good hypotheses $\theta^j$ (Algo. \ref{kdgs},~line \ref{kdgs:l14}). Let $i_{den} \in \theta^j$ be the index of the hypothesis to which that \jth~ data point has the maximum kernel density (Algo. \ref{kdgs},~line \ref{kdgs:l14}). Similarly, let $i_{res} \in \theta^j$ be the index of hypothesis which has minimum mean residual of top-$\beta$ data points in the residual space (Algo. \ref{kdgs},~line \ref{kdgs:l15}). With $i_{den}, i_{res}$, we get the best potential good hypotheses that can best describe the \jth data point at that instant of the hypothesis generation process. Note: It may be possible that $i_{den}=i_{res}$, but it does not affect the further process. Since, $i_{den}$ is obtained using density and $i_{res}$ using residual information, we derive conditional sampling weight using density profile and residual profile of $i_{den}$ and $i_{res}$ respectively and then combine them (Algo. \ref{kdgs},~line \ref{kdgs:l17}-\ref{kdgs:l18}). The conditional sampling weight using $i_{den}$ is directly proportional to the density values (higher is the density value more likely it is an inlier). This choice is made based on the fact: if \jth data point is the potential inlier of $i_{den}$ hypothesis, then the data points which has higher density values in the density profile of $i_{den}$ hypothesis are most likely also the inliers to the same structure to which \jth~ data point belongs. Hence, if we sample the \jth~data point in an MSS, then the remaining elements of MSS should be sampled using conditional sampling weights derived from a density profile of $i_{den}$. Similarly, if \jth~ point is the potential inlier of $i_{res}$ hypothesis, then the data points with smaller residuals are also most likely inliers of the same structure to which \jth~ data point belongs. 
The conditional sampling weights using $i_{den}$ and $i_{res}$ hypotheses are computed as  $\frac{d_{i_{den}}^{j}}{\sum_j d_{i_{den}}^{j}}$ and $\frac{\overline{r}_{i_{res}}^{j}}{\sum_j \overline{r}_{i_{res}}^{j}}, \forall j \in [1,...,n]$ respectively. Which is then combined in Algo. \ref{kdgs},~line \ref{kdgs:l18}) as below:

\begin{equation}
\textbf{s}^{j} = \frac{d_{i_{den}}^{j}}{\sum_j d_{i_{den}}^{j}} \times \frac{\overline{r}_{i_{res}}^{j}}{\sum_j \overline{r}_{i_{res}}^{j}}, \forall j \in [1,...,n] \end{equation}

Recall, KDGS aims to generate hypotheses for each data point. Therefore, for \jth~data point the conditional PMF for sampling points in MSS ($\mcalM$) is computed using its pairwise point correlation with all other data points in $\textbf{c}^j$ and conditional sampling weight $\textbf{s}^j$ as shown in Algo. \ref{kdgssub}, line \ref{kdgssub:l6}. The $\textbf{c}^j$ and the $\textbf{s}^j$ are the corresponding columns of \jth~ data point in the matrix \textbf{C} and \textbf{S} respectively.

\noindent\textbf{When to stop generating hypotheses?.} 
KDGS maintains an index set $\nu$ that contains indices of data point for which hypotheses are yet to be generated. KDGS starts with initializing the set $\nu=[1,...,n]$ (Algo. \ref{kdgs},~line \ref{kdgs:l3}) and stops when $\nu$ is empty. Next, we describe the process of updating the set $\nu$. 
In order to identify the points in $ \nu $ that are well explained by the hypotheses in $ \textbf{H} $, we introduce an \textit{explanation score} $ \tau $. For each \jth~data point, the $ \tau_j $ is computed as the mean of its kernel densities \wrt~all the hypotheses in its corresponding $\theta^{j}$ (Algo. \ref{kdgs},~line \ref{kdgs:l19}). If the explanation score does not change significantly (Algo. \ref{kdgs},~line \ref{kdgs:l21}, we fixed $\alpha=0.1$ for all our experiments), we assume the point is well explained and remove it from the set $ \nu $. Ideally, we expect the explanation score of each data point should be high and saturating when KDGSA terminates. We analyze this behavior in the next section. 

\subsubsection{Termination Analysis of KDGS}
\label{sec:term_kdgs}
We analyze the behaviour of \textit{explanation score} $\tau_{curr}$ and $\nu$ with every iteration of while loop of Algo. \ref{kdgs}, using two sequences each of motion segmentation (top 2 rows) and planar segmentation (bottom 2 rows) of AdelaideRMF dataset \cite{wong2011dynamic} in Fig. \ref{fig:kdgs_ana}. 
 
We show cardinality of set $\nu$  \wrt~iterations of the while loop of Algo. \ref{kdgs} in Fig. \ref{fig:kdgs_ana} (column 2). It can be seen that the cardinality of $\nu$ set decreases with every iteration of the while loop of Algo. \ref{kdgs} and reaches to zero. Simultaneously, we can see, $\tau_{corr}$ of data points of each structure (\eg~st-1,st-2,...), also saturates at the same iteration when KDGS terminates. For better illustration, we plotted the mean of the explanation scores $t_{curr}$ of the data points belonging to the same structure \wrt~ the iterations of KDGS. Another important observation is, along with explanation scores ($\tau_{curr}$), the number of good hypotheses generated for each structure also tends to saturate when the algorithm approaches the termination.

While it is difficult to \emph{guarantee} that termination of Algo. \ref{kdgs} with an \textit{upper bound on the number of generated hypotheses}, the strong empirical evidence shown in Fig. \ref{fig:kdgs_ana} indicates the KDGS terminates few iterations and generates hypotheses for all structures. In Sec. \ref{sec:exps}, we compare KDGS with other competitive guided sampling algorithms.

%% file: KDGS.tex
\begin{algorithm}[h!]
	\DontPrintSemicolon
	\SetAlgoLined
	\SetKwInOut{Input}{Input}\SetKwInOut{Output}{Output}
	\textbf{Input:} $ \mathbf{X},\eta$,~~ \textbf{Output:} \textbf{H}, \textbf{R}, \textbf{D}\\
    \textbf{Initialization:}~$\textbf{C} \leftarrow \mathbbm{1}_{n\times n}, \textbf{S} \leftarrow \mathbbm{1}_{n\times n}, \textbf{H} = \textbf{R} = \textbf{D} = \varnothing $ \label{kdgs:l2}\\
	$\nu \leftarrow \{1..n\}, \tau_{prev}^j=\tau_{curr}^j=0$  \label{kdgs:l3}\\
	\While{$\nu \neq \varnothing$}{    
		$[\widehat{\textbf{{H}}},\widehat{\textbf{{R}}},\widehat{\textbf{{D}}}] \leftarrow \texttt{generateHyps}(\textbf{C},\nu,\eta, \bS )$  \label{kdgs:l10}\\
		${\textbf{{H}}} \leftarrow \{{\textbf{{H}}} \cup \widehat{\textbf{{H}}}\}, ~ {\textbf{{R}}} \leftarrow \{{\textbf{{R}}} \cup \widehat{\textbf{{R}}}\}, ~ {\textbf{{D}}} \leftarrow \{{\textbf{{D}}} \cup \widehat{\textbf{{D}}}\}$  \label{kdgs:l11}\\
			\For{$i \leftarrow 1~ to ~m~(\text{no.of hypotheses in \textbf{H})}$}{
	$d_i^{j} = d_i^{j}/ \sum_{j=1}^{n}{d_i^{j}},~~\forall j \in [1,...,n]$ \label{kdgs:l5}\\
$\pi_{i} = mean(d_i^{\tw_i^{1}:\tw_i^{\beta}}) - mean(d_i^{\tw_i^{n-\beta}:\tw_i^{\beta}})$ \label{kdgs:l6} \\
    $d_i^{j} = d_i^{j} \times \pi_{i},~~ \forall j \in [1,...,n]$ \label{kdgs:l7} \\
    }
		$\textbf{V} \leftarrow \texttt{KRDPointPreferences}(\textbf{D})$ \label{kdgs:l12}\\ 
		\For{$j \leftarrow 1~ to ~n$}{
    $\theta^{j} = \{i,~~ s.t ~~r_i^{j} \leq \rho_i^{\beta} \},~~ \forall i \in [1,...,m]$ \label{kdgs:l14}\\
    $i_{den} = \arg max_{i~\in~\theta^j}~~d_i^{j}$ \label{kdgs:l15} \\
   $i_{res} = \arg min_{i~\in~ \theta^j}~ \big (\sum_{j=1}^{\beta} \rho_i^{j}/\beta \big)$ \label{kdgs:l16} \\
   $ \overline{r}_{i_{den}} = \frac{\max(r_{i_{res}})}{r_{i_{res}}^{j}}, \forall j \in [1,...,n] $ \label{kdgs:l17}\\
   $ s^{j} = \frac{d_{i_{den}}^{j}}{\sum_j d_{i_{den}}^{j}} \times \frac{\overline{r}_{i_{res}}^{j}}{\sum_j \overline{r}_{i_{res}}^{j}}, \forall j \in [1,...,n]$ \label{kdgs:l18}  \\
    $\tau^{j}_{curr} = \sum_{i\in \theta^{j}} d_i^{j}/|\theta^{j}|$ \label{kdgs:l19}\\
}

$\nu \leftarrow \{ j ~|~(\tau_{curr}^j-\tau_{prev}^j) \geq \alpha\tau_{curr}^j \} ~~\forall j \in \{1..n\}$ \label{kdgs:l21}\\    
		$\textbf{C} \leftarrow \texttt{updatePointCorrelation}(\textbf{C},\nu)$ \label{kdgs:l22} \\
		$\tau^{j}_{prev} \leftarrow \tau^{j}_{curr}$ \label{kdgs:l23}\\
	}  
			$  \Theta \leftarrow \bigcup_{k=1}^{n} v^k_1 $ \label{kdgs:l25}\\
	$\textbf{H} \leftarrow \textbf{H}_{\{\Theta\}}, ~ \textbf{R} \leftarrow \textbf{R}_{\{\Theta\}}, ~ \textbf{D} \leftarrow \textbf{D}_{\{\Theta\}}$  \label{kdgs:l26}\\     
	\caption{KRD Guided Sampling (KDGS)}
	\label{kdgs}
\end{algorithm} 

%% file: KRDGS_sub.tex
\begin{algorithm}[ht!]
	\DontPrintSemicolon
	\SetAlgoLined
	\SetKwInOut{Input}{Input}\SetKwInOut{Output}{Output}
	\textbf{Input:} $\textbf{C}, \nu, \eta, \bS$,~~\textbf{Output:}~~\textbf{H}, \textbf{R}, \textbf{D} \\
	\ForEach{$j \in \nu$}{    
		$\mcalM \leftarrow \varnothing$\\
		$\mcalM  \leftarrow \{\mcalM  \cup j \}$ \label{kdgssub:l4}\\
		\For{$k \leftarrow 2 ~ to ~ \eta$}{
			$\overline{\textbf{c}} \leftarrow \textbf{c}^j \times \textbf{s}^{j}$ \label{kdgssub:l6}\\
			$\overline{\textbf{c}}_{\{\mcalM \}} \leftarrow 0$ \label{kdgssub:l7}\\
			$\overline{\textbf{c}} \leftarrow  \frac{\overline{\textbf{c}}}{\Sigma \overline{\textbf{c}}}$ \label{kdgssub:l8}\\
			$\mcalM  \leftarrow \{ \mcalM  \cup \texttt{getSample}(\overline{\textbf{c}})\} $ \label{kdgssub:l9}
		}
		$\mathbf{h} \leftarrow \texttt{fitModel}(\mcalM) $ \label{kdgssub:l11} \\              
		$\textbf{H} =\{ \textbf{H} \cup ~\mathbf{h}\}$ \label{kdgssub:l12} \\
	}    
	$\textbf{R} \leftarrow \texttt{computeResiduals}(\textbf{H})$ \label{kdgssub:l14} \\
	$\textbf{D} \leftarrow \texttt{computeKernelResidualDensity}(\textbf{R})$ \label{kdgssub:l15} \\
	\caption{\texttt{generateHyps}}
	\label{kdgssub}
\end{algorithm}

%% file: KDGS_analysis.tex
\begin{figure*}[h!]
\centering
\resizebox{\textwidth}{!}{%
\begin{tabular}{cccc}
Data (left view only) &  \hspace{-0.8cm} $\nu$ vs Iterations  & \hspace{-1.0cm} $\tau_{curr}$ vs iteration & \hspace{-0.7cm}\# good hyps per structure \\
\includegraphics[width=5.2cm,height=3.8cm]{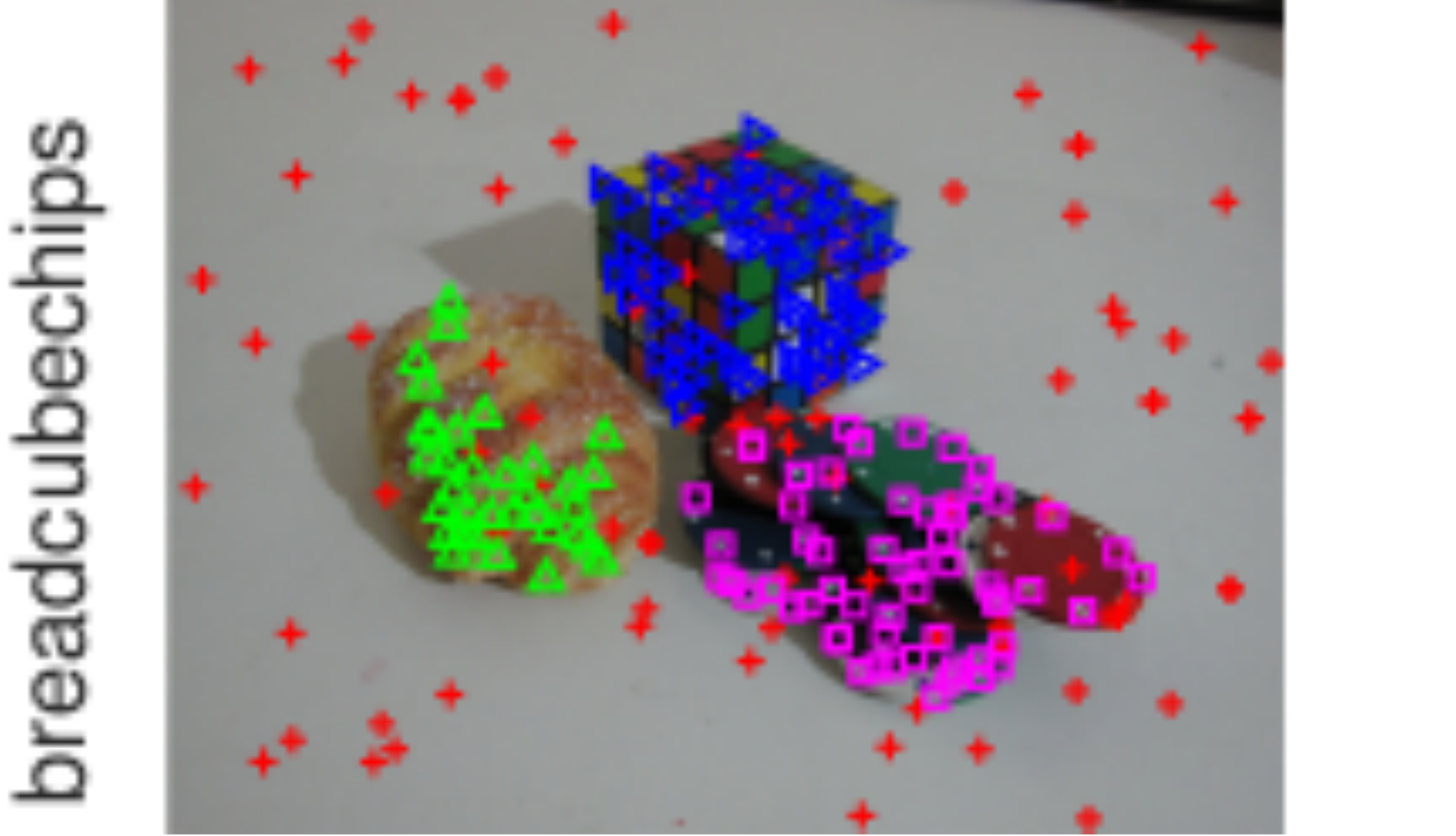} &\hspace{-0.8cm} 
\includegraphics[width=4.7cm]{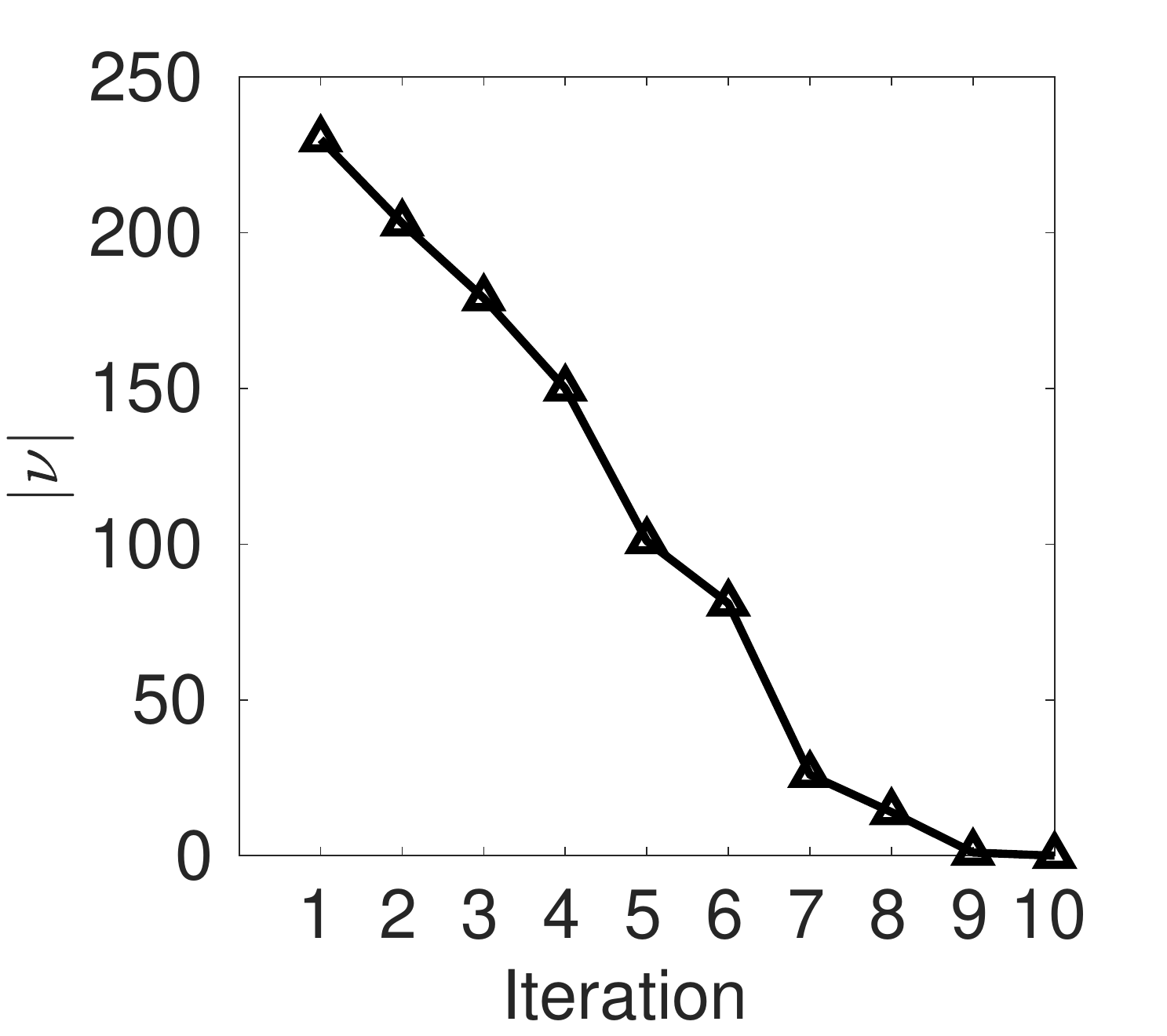}& \hspace{-0.6cm}\includegraphics[width=4.7cm]{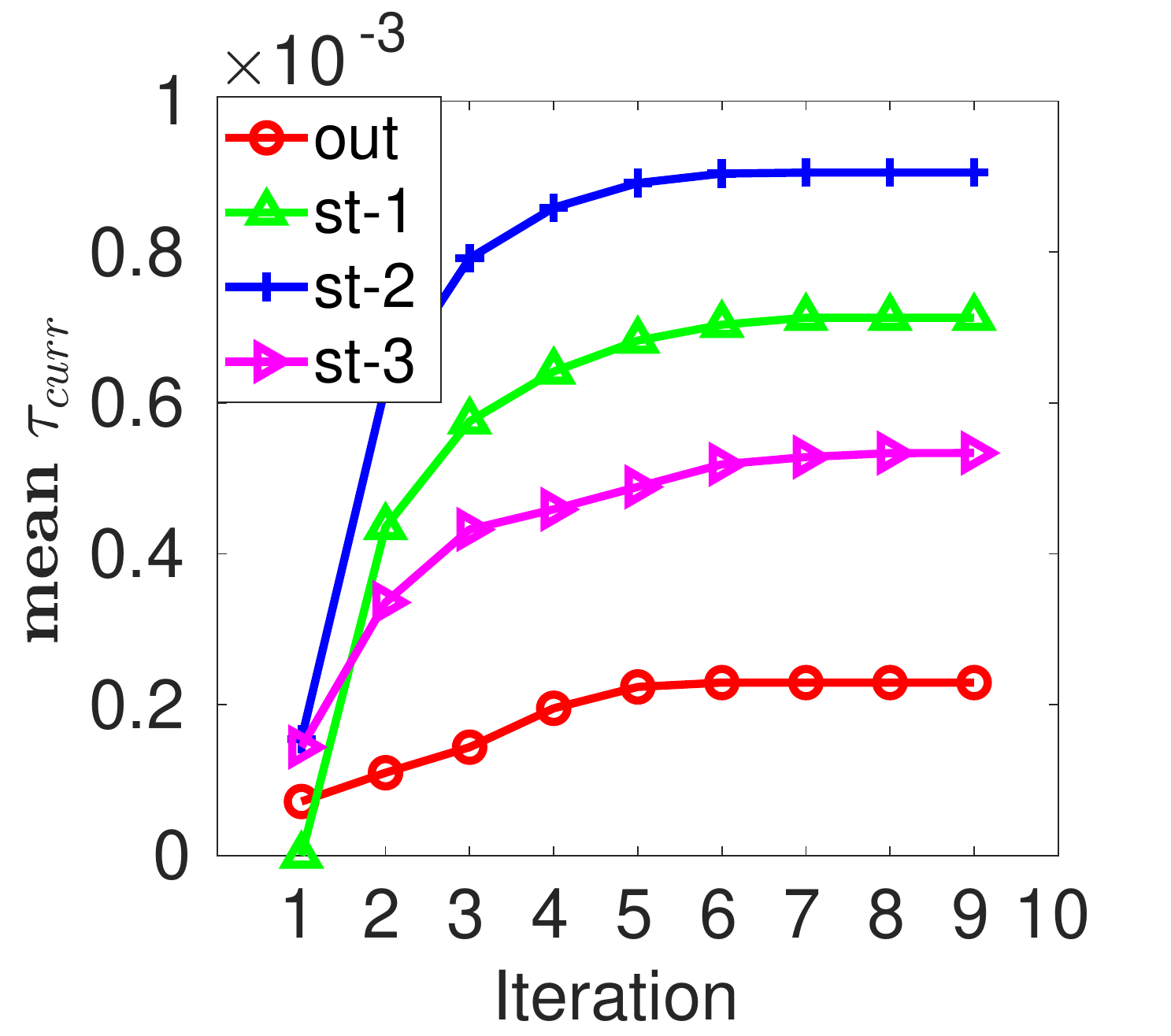}& \hspace{-0.7cm}\includegraphics[width=4.7cm]{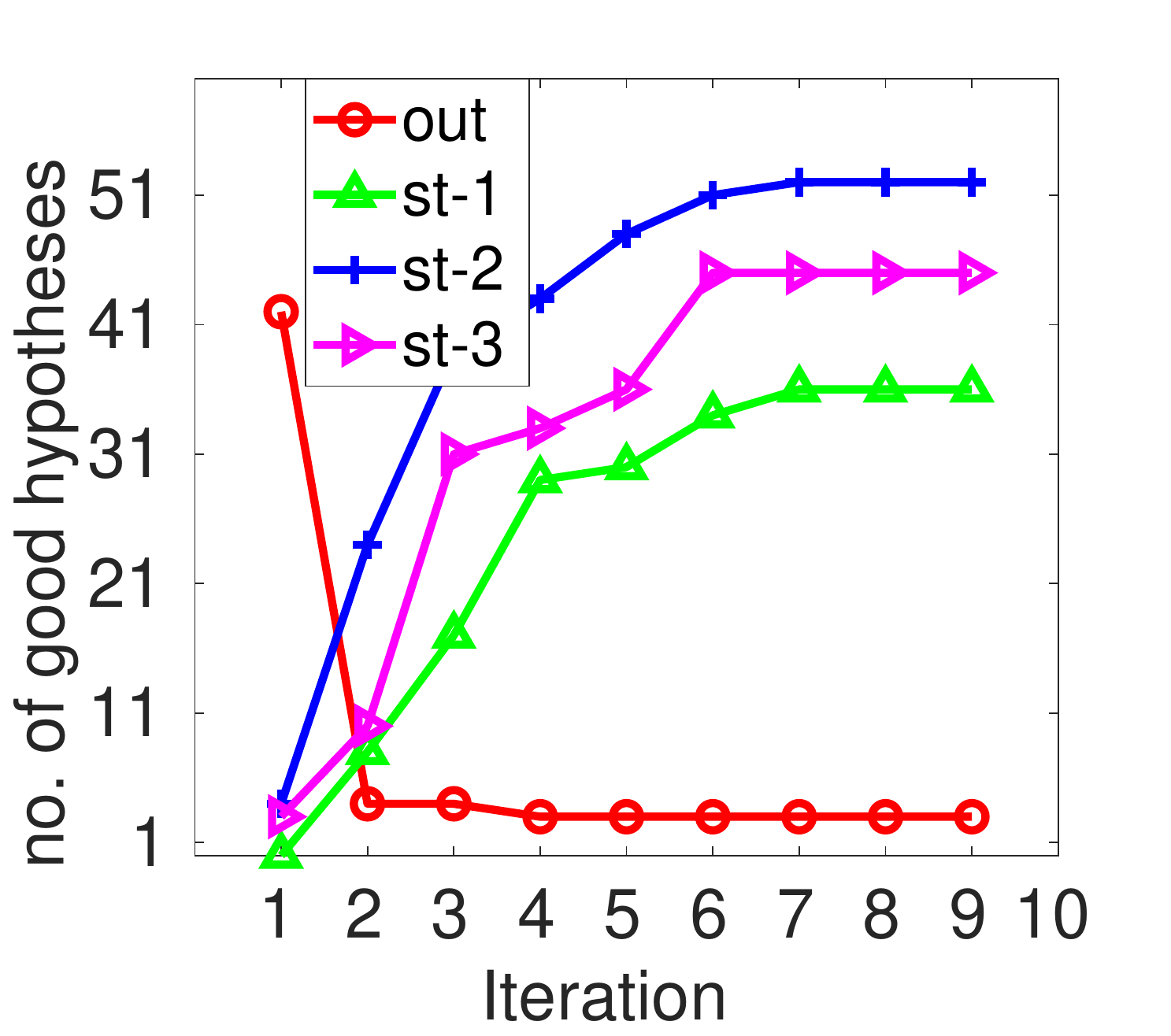} \\
\includegraphics[width=5.2cm,height=3.8cm]{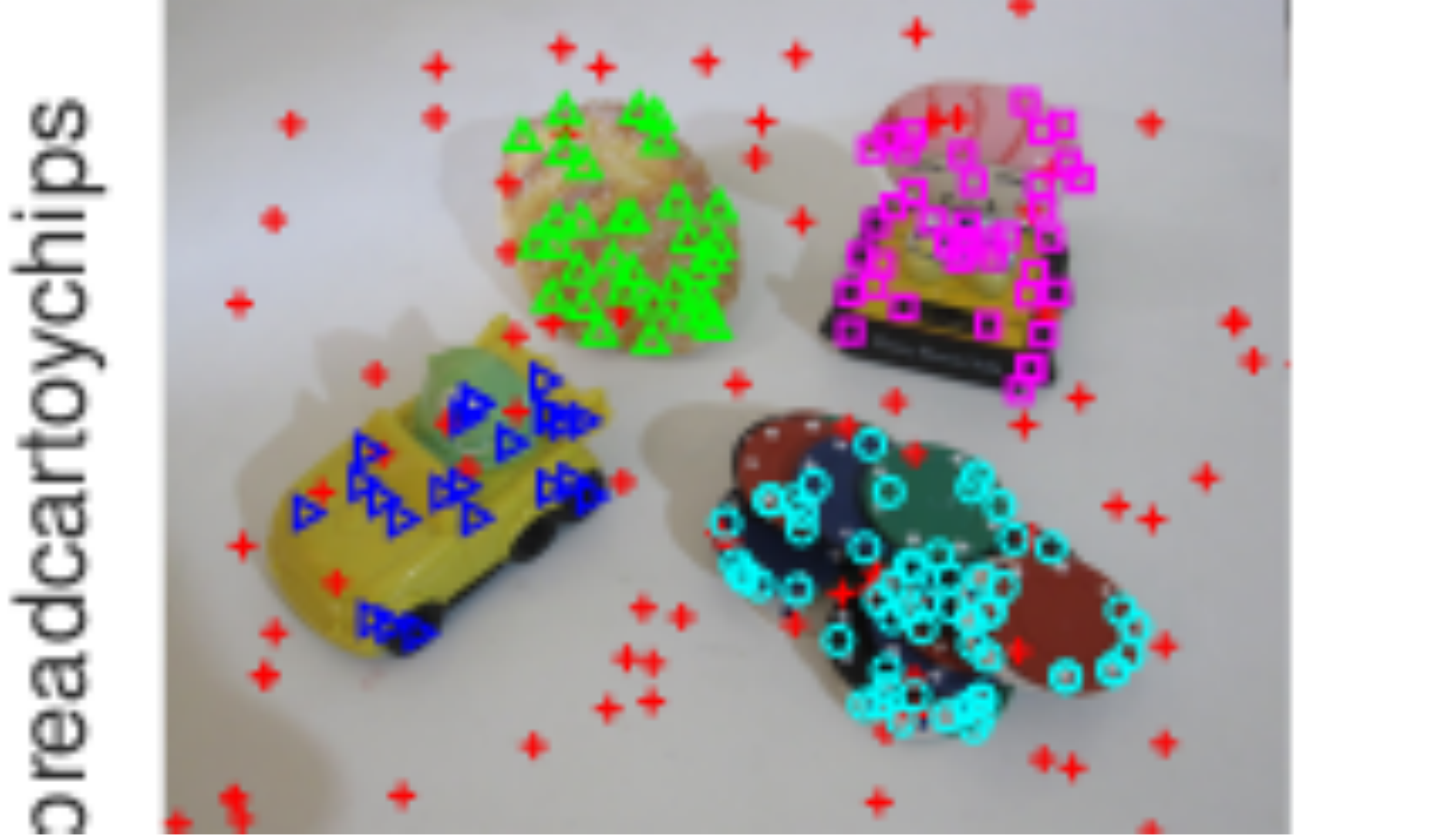} & \hspace{-0.8cm} 
\includegraphics[width=4.7cm]{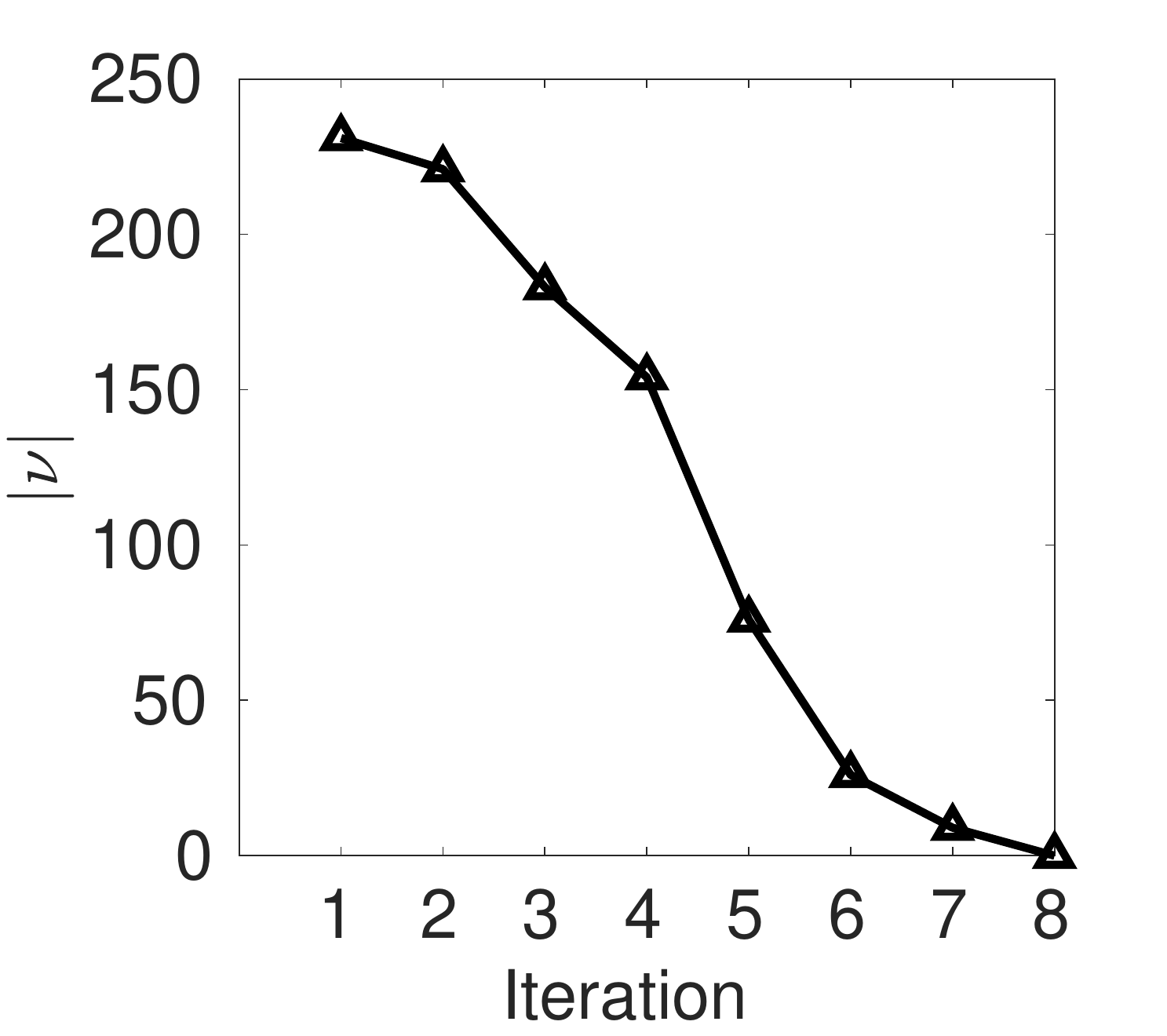}& \hspace{-0.8cm}\includegraphics[width=4.7cm]{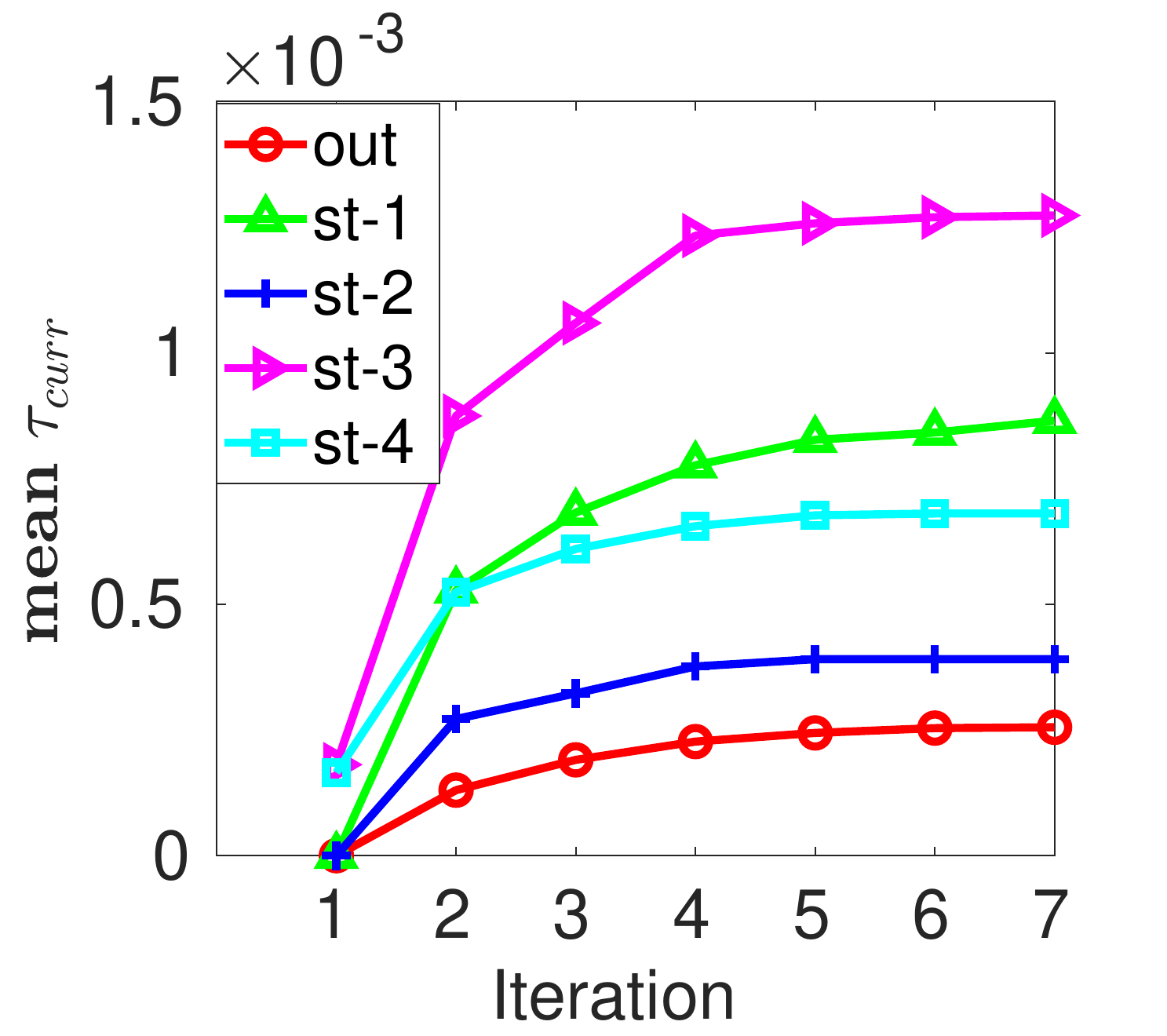}& \hspace{-0.7cm}\includegraphics[width=4.7cm]{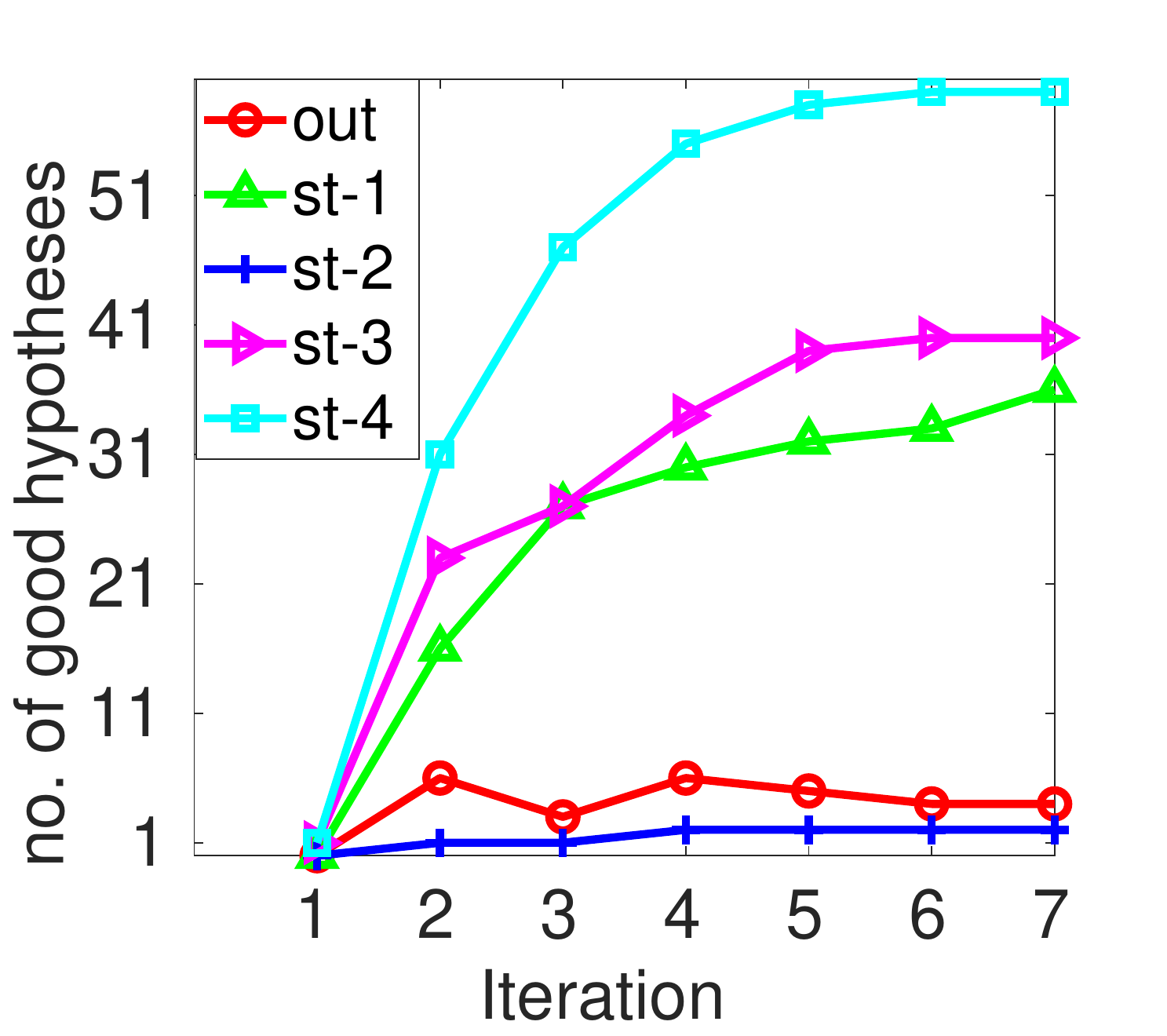} \\
\includegraphics[width=5.2cm,height=3.8cm]{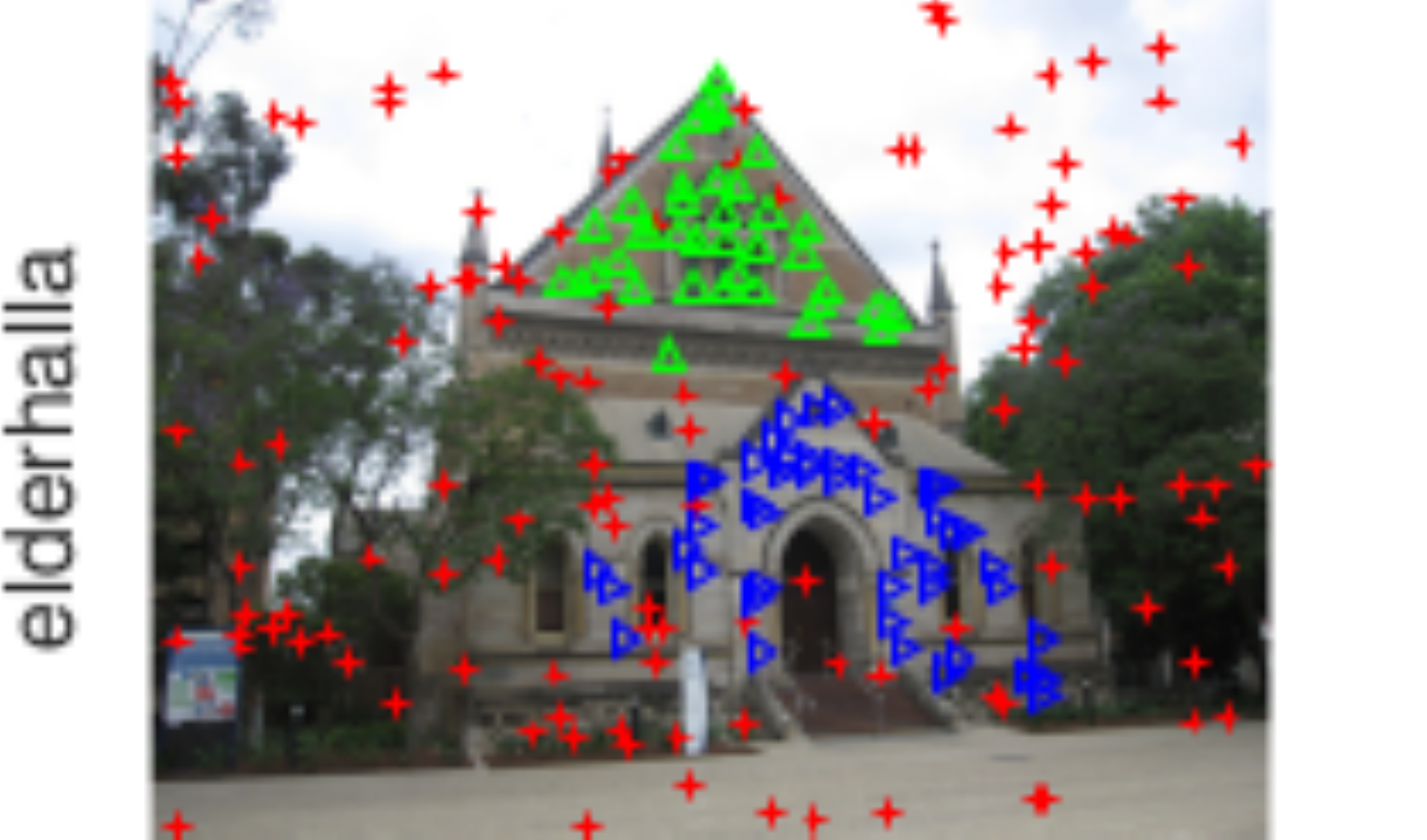} & \hspace{-0.8cm}
\includegraphics[width=4.7cm]{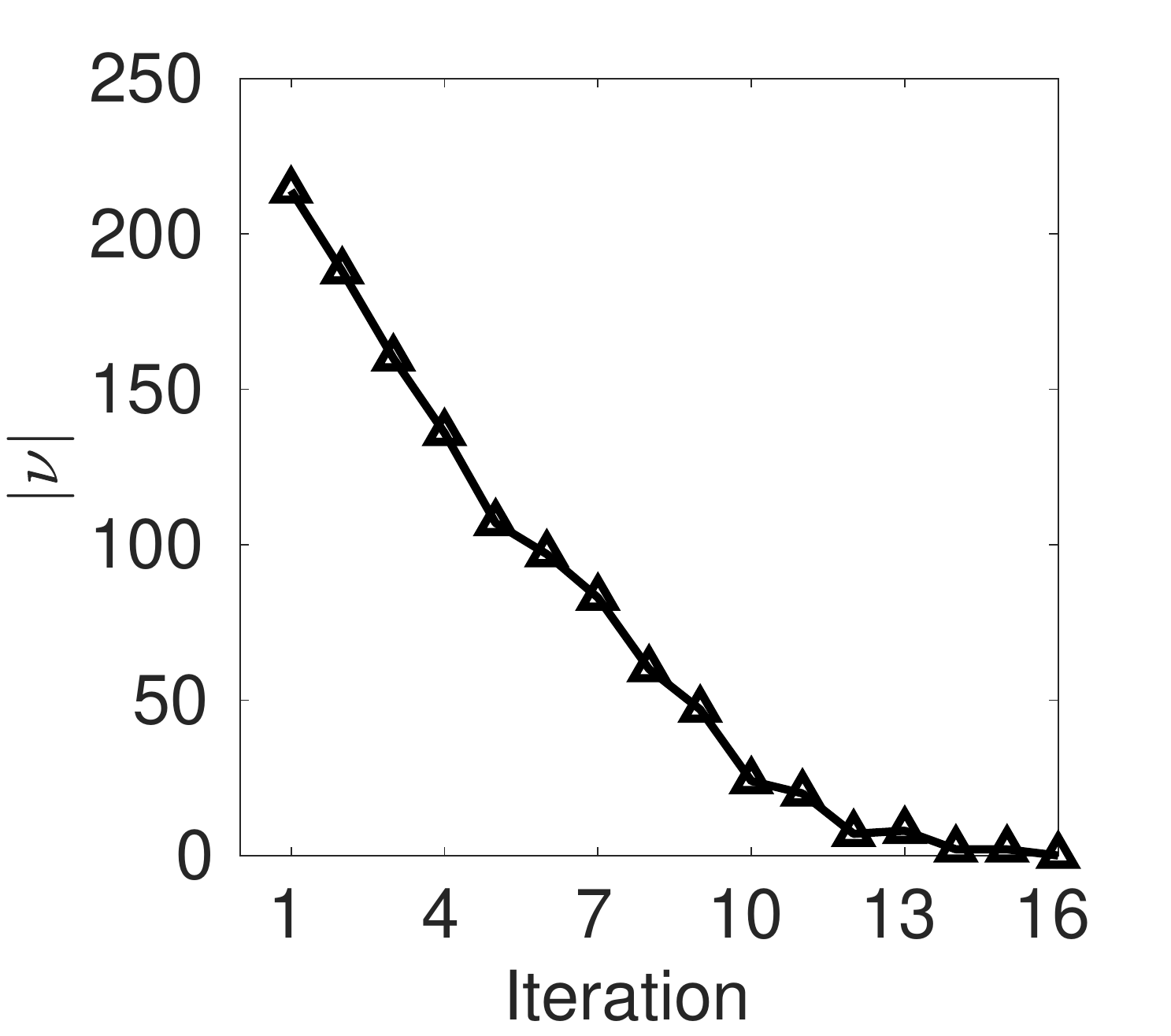}& \hspace{-0.7cm}\includegraphics[width=4.7cm]{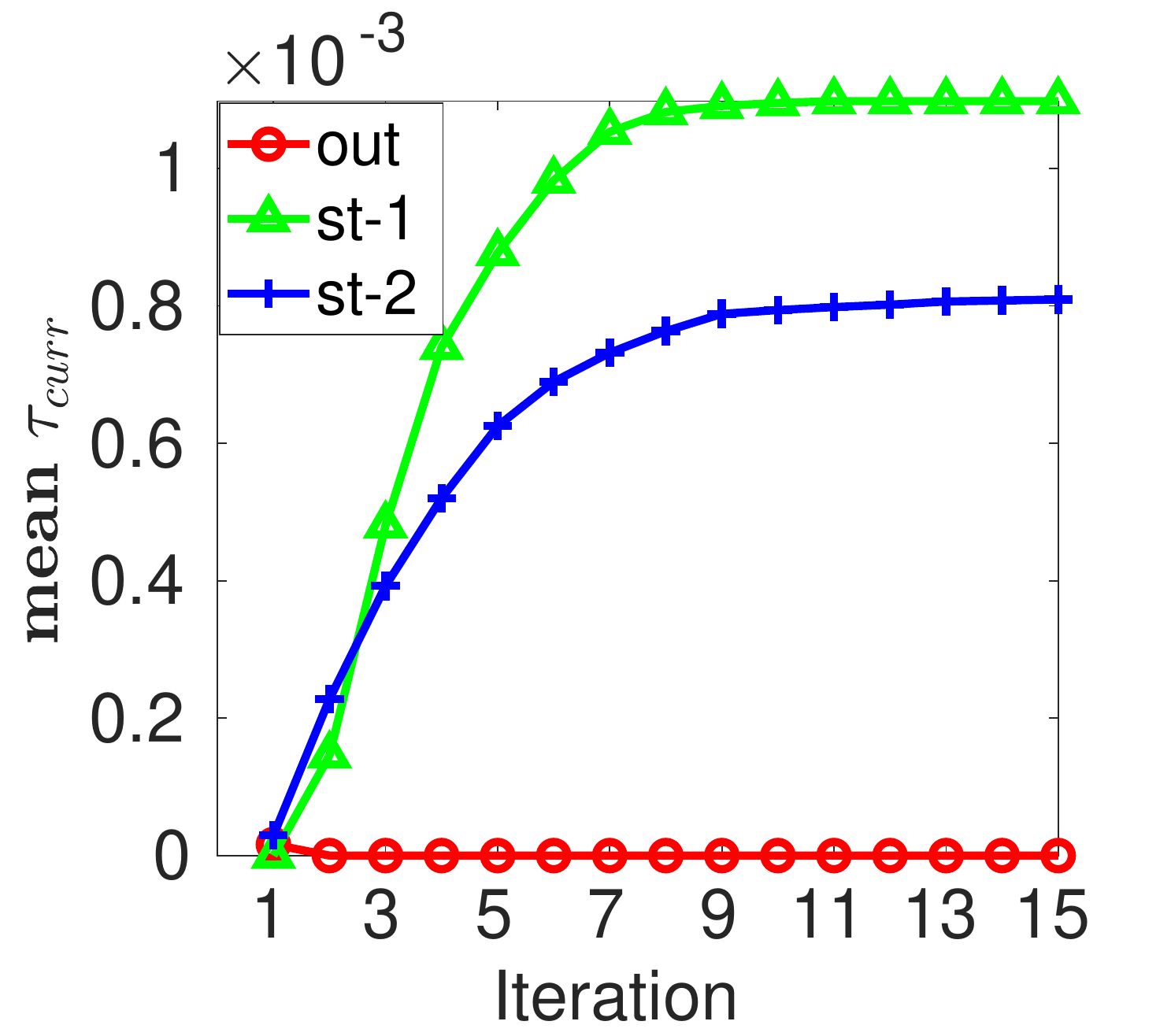}& \hspace{-0.7cm}\includegraphics[width=4.7cm]{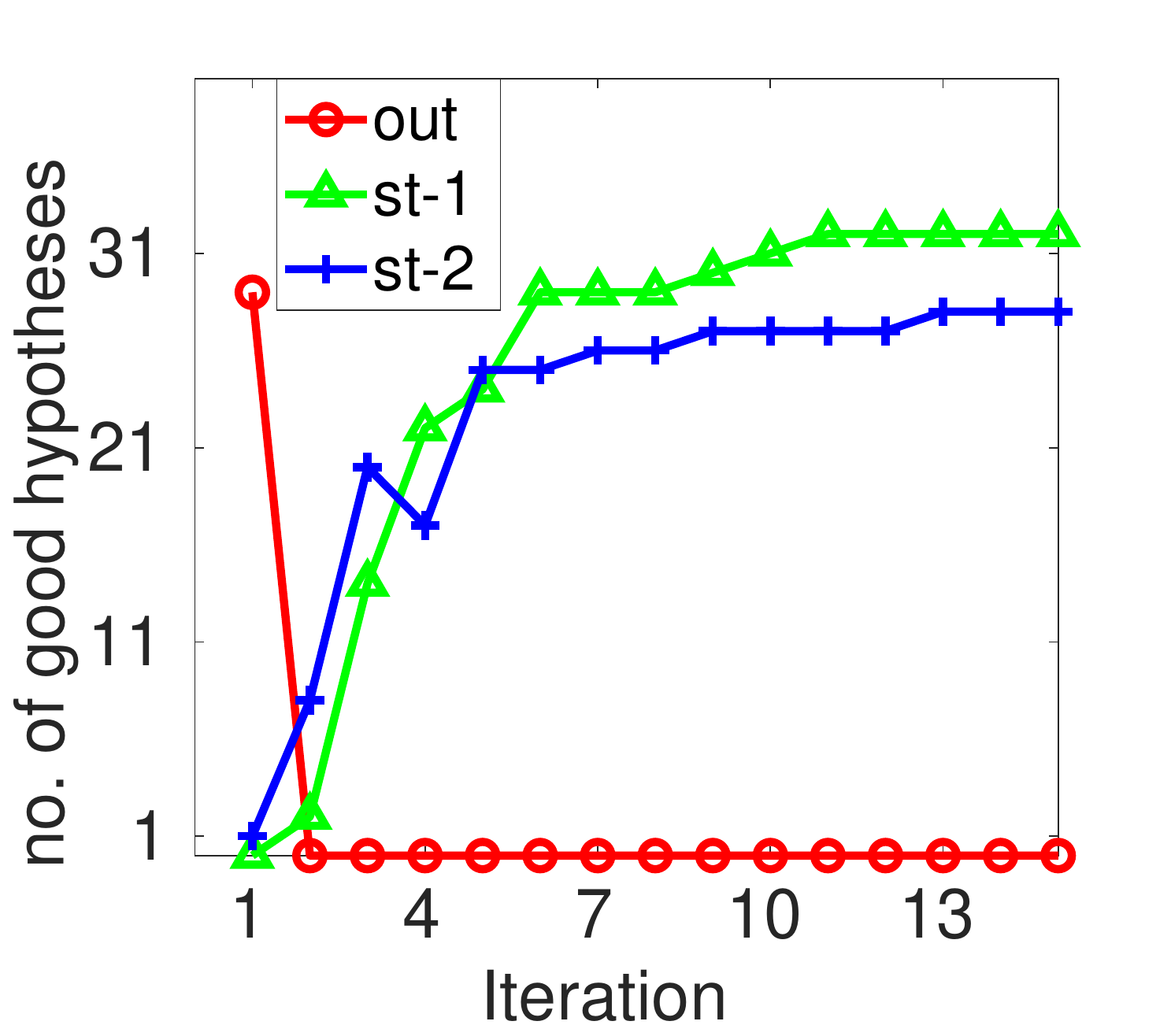} \\
\includegraphics[width=5.2cm,height=3.8cm]{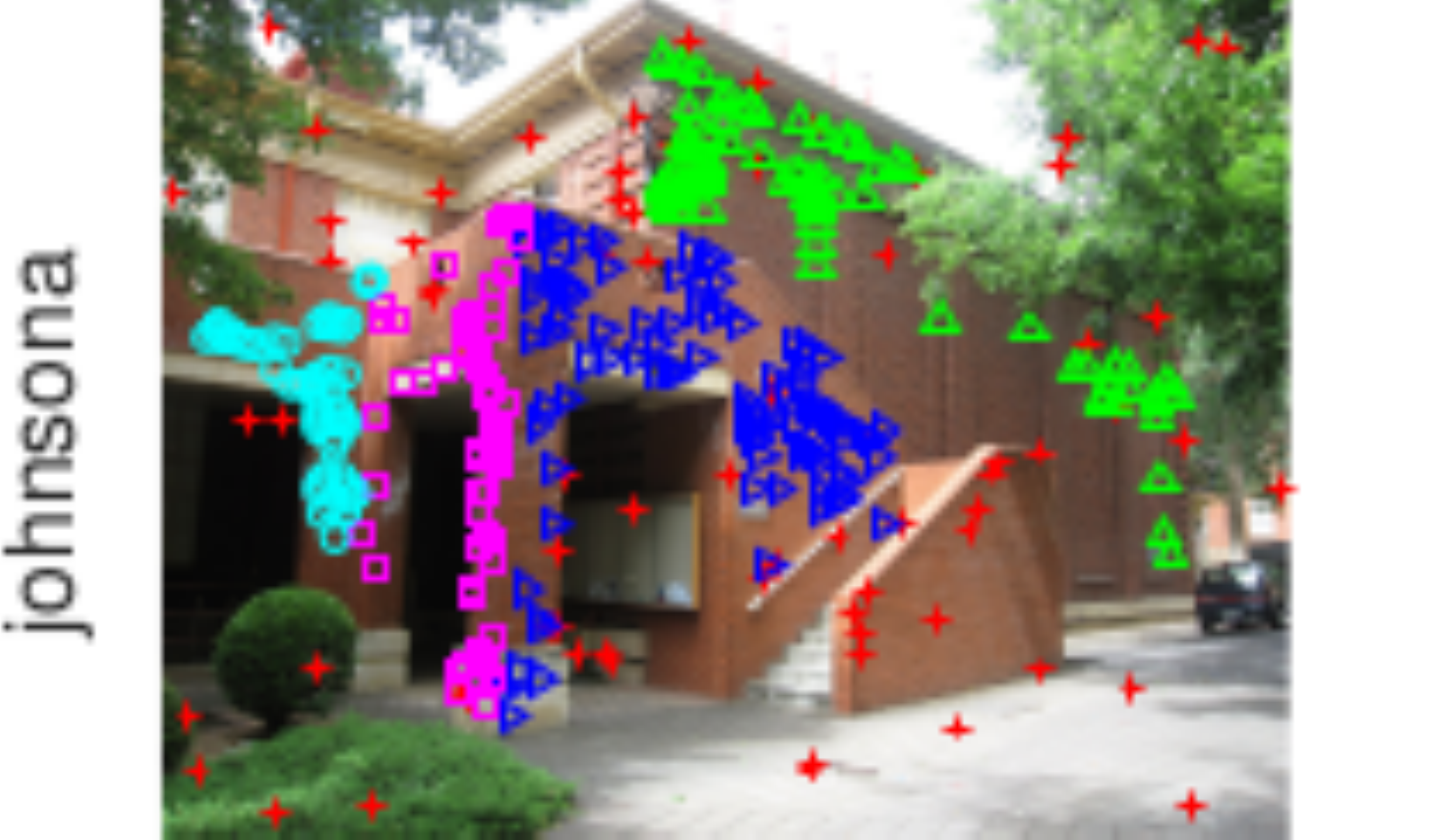} &  \hspace{-0.8cm}
\includegraphics[width=4.7cm]{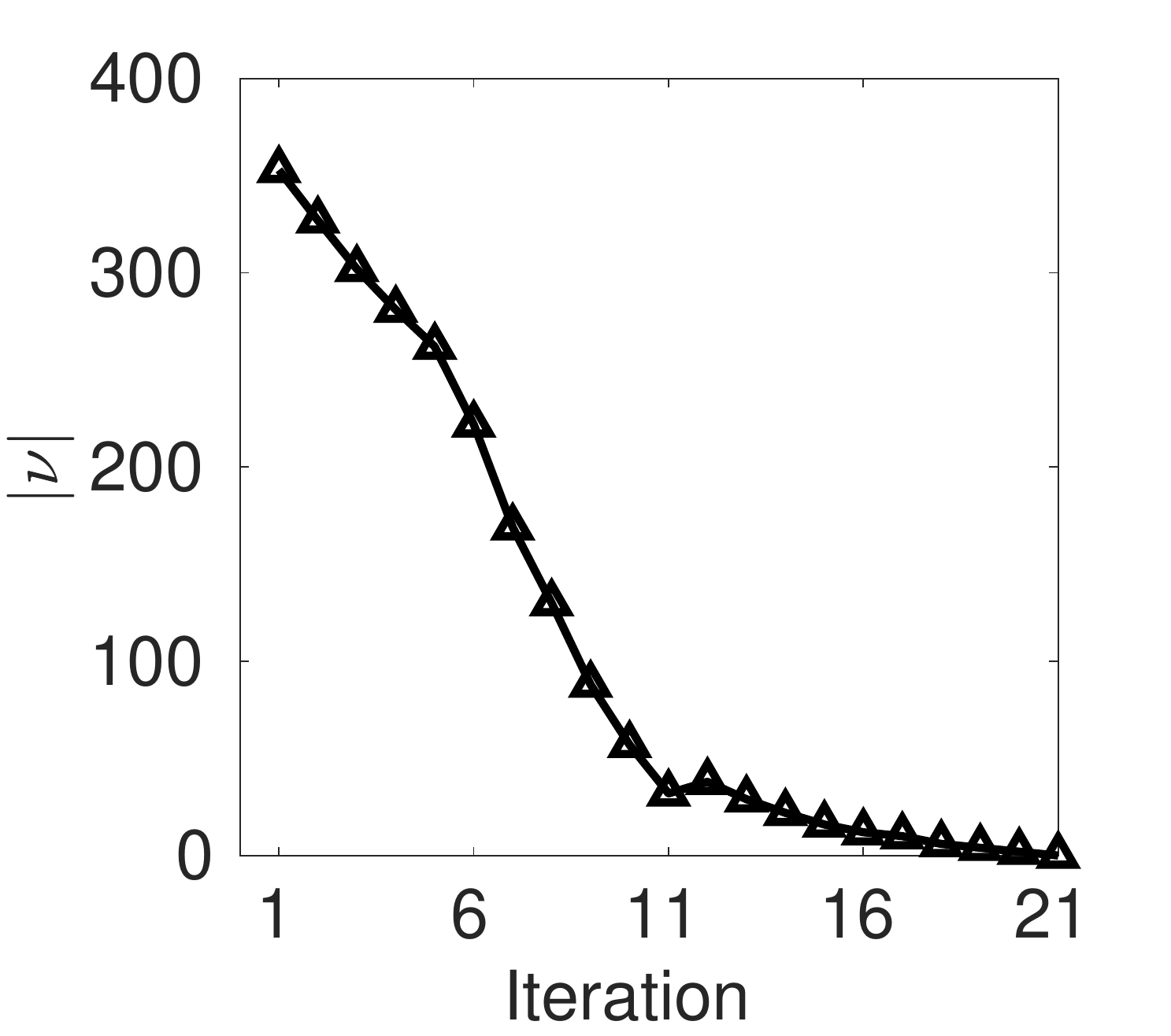}& \hspace{-0.8cm}\includegraphics[width=4.7cm]{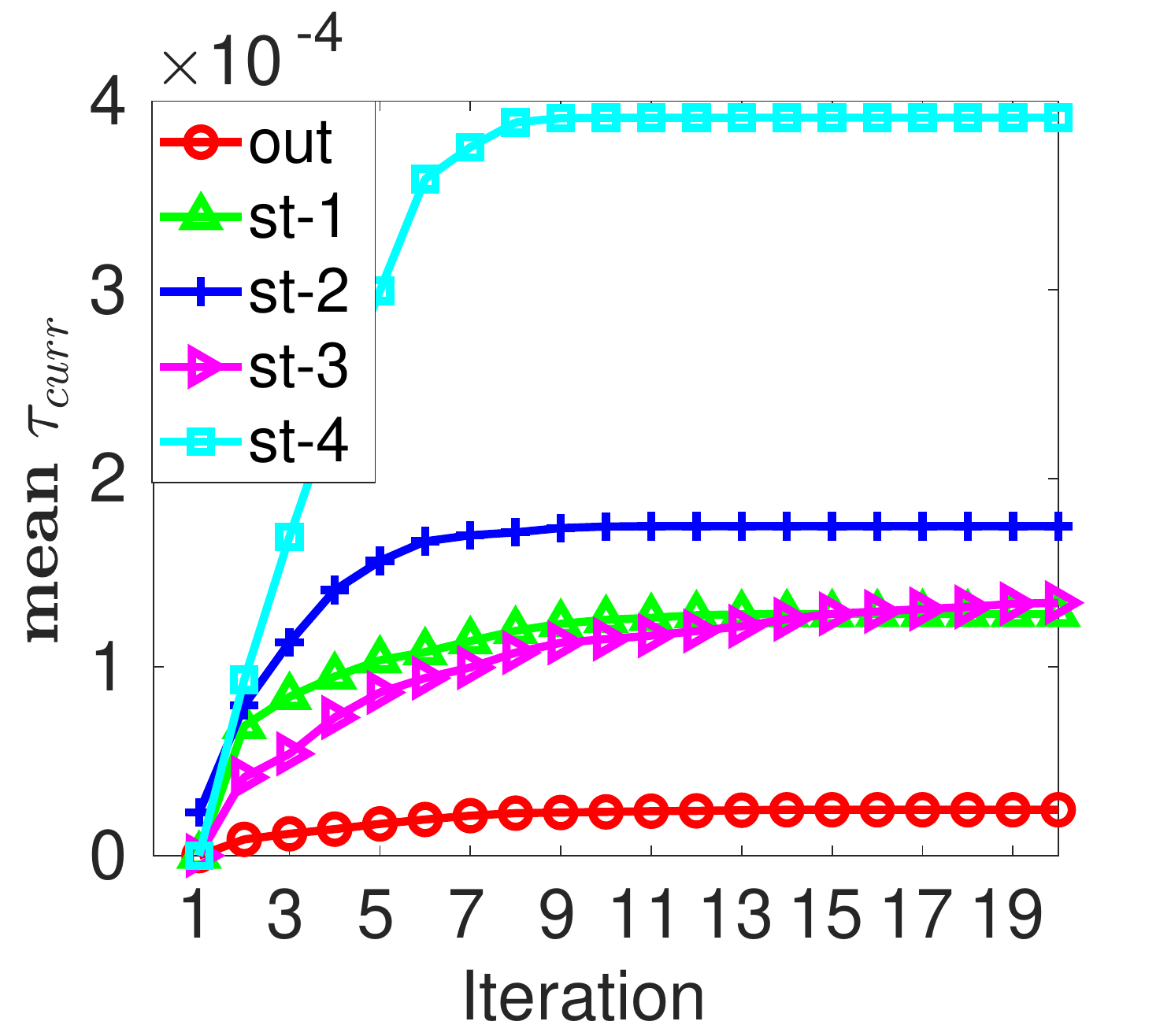}& \hspace{-0.7cm}\includegraphics[width=4.7cm]{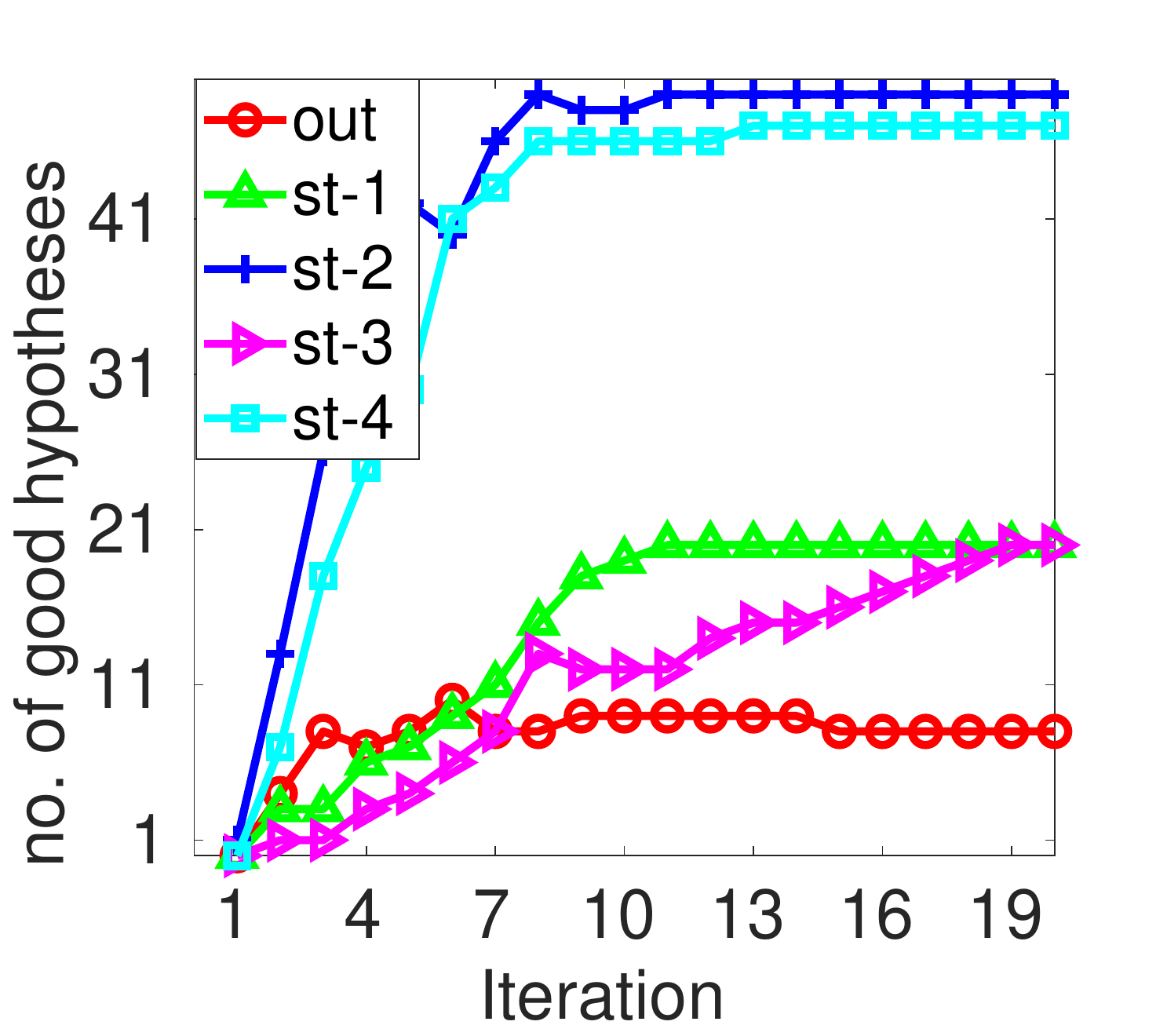} \\
\end{tabular}
}
\caption{Termination analysis of Kernel Density Guided Sampling (KDGS). Refer text in Sec. \ref{sec:term_kdgs} for detail.}
\label{fig:kdgs_ana}
\end{figure*}

%% file: KRD_Frac.tex
\section{KRD Based Inlier Noise Scale Estimation}
\label{sec:KRD_frac_est}
In \cite{tiwari2018dgsac}, an inlier fraction estimation algorithm \textit{estimateFraction}  is proposed. It takes residual density \cite{tiwari2018dgsac}, a sorted residual vector, the size of the MSS, and output the fraction of inliers $\fhat$. For each \ith~ hypothesis, we input its \textit{kernel Residual Density} (from Sec. \ref{sec:krd}) $\textbf{d}_i$, sorted residual vector $\boldsymbol{\rho}_i$ (from Sec. \ref{sec:hyp_pt_pref}) and size of MSS $\eta$ to the fraction estimation algorithm \textit{estimateFraction} and record the estimated fraction in $\fhat_i$. The \textit{estimateFraction} algorithm estimate the fraction by finding an inlier/outliers boundary using density disparity, and dispersion of residuals. For details we refer reader to \cite{tiwari2018dgsac}. From estimated fraction we can compute the number of estimated inliers ($t_i$) as $t_i = \floor{\fhat_i~n}$. The inlier noise scale  $\sigma_i$ is estimated using sorted residual vector $\boldsymbol{\rho}_i$ and the estimated number of inliers $t_i$ as below in Eq. \ref{eq:scale}.   The estimated fraction $\fhat_i$ to estimate noise scale $\sighat_i$. 
\begin{equation}
    \sighat_i = \sqrt{\frac{\sum_{j=1}^{t_i}(~\rho_{i}^{j}-mean(\rho_{i}^{1:t_i})~)^{2}}{t_{i}-1}}
    \label{eq:scale}
\end{equation}

\noindent\textit{Note:} we estimate noise scale for each hypothesis independently, hence cover those cases where different structures may have different noise scale.

%% file: GO_model.tex
\section{Greedy/Optimal Model Selection}
\label{sec:GO_model}
In this section, we propose two variants of model selection: \textit{greedy} and a \textit{quadratic program based optimal model selection}. Before describing model selection algorithms, we first detail a few preliminaries.

\noindent\textbf{Preliminaries.} Using number of estimated inliers $t_i$ from Sec. \ref{sec:KRD_frac_est}, we can obtain \emph{estimated} inlier set of each hypothesis $\textbf{h}_i$ as $ \mathcal{I}_i = \{q_i^{1},q_i^{2},...,q_i^{f_i} \} $ from their respective hypothesis preference set $\textbf{q}_i$ (Sec. \ref{sec:hyp_pt_pref}).
Due to the nature of the hypothesis generation process, there may be multiple model hypotheses that explain the same inlier structure. The goal of model selection is to retain the most representative model hypothesis and discard the redundant ones. To identify the best model, we need a measure to quantify the goodness of a model. Since the number of structures is \emph{not known} a priori, therefore, we need to measure the redundancy between model hypotheses before discarding them. For the latter, we estimate the pairwise correlation between hypotheses by computing the Spearman-Footrule ($\mathcal{SF}$) distance \cite{wong2013simultaneous,topk} between their corresponding inlier only preference lists. 
\subsection{Hypothesis Correlation using Spearman-Footrule}
\label{sec:hyp_corr_sf}
For each hypothesis pair $\textbf{h}_i$ and $\textbf{h}_k$, let their respective top-t inlier only preference list are denoted by $\bar{\textbf{q}}_i=[q_i^{1},q_i^{2},...,q_i^{t}]$ and $\bar{\textbf{q}}_k=[q_k^{1},q_k^{2},...,q_k^{t}]$, where, $t = min(t_i,t_k)$ (\ie~ minimum of the number of estimated inliers). The Spearman-Footrule distance is computed using (\ref{eqn:sf}), where $Y(\bar{\textbf{q}}_i)$  denotes the data points with indexes in $\bar{\textbf{q}}_i$ and $j^{\bar{\textbf{q}}_i}$ denotes the position of the data point ($j$) in the preference list $\bar{\textbf{q}}_i$. We use $j+1$ for $j^{\bar{\textbf{q}}_i}$ if $j \notin Y(\bar{\textbf{q}}_i)$. Variables for $\bar{\textbf{q}}_k$ are similarly defined.
\begin{eqnarray}\label{eqn:sf}
\mathcal{SF}(\bar{\textbf{q}}_i,\bar{\textbf{q}}_k)= \Sigma_{j\in Y(\bar{\textbf{q}}_i)\cup Y(\bar{\textbf{q}}_k)} ~~ |j^{\bar{\textbf{q}}_i} - j^{\bar{\textbf{q}}_k} | \\
z_i^k =1 - \frac{1}{t \times (t+1)}\mathcal{SF}(\bar{\textbf{q}}_i,\bar{\textbf{q}}_k) \label{eqn:hc}
\end{eqnarray}

\noindent We compute pairwise hypothesis correlation between $\textbf{h}_i$ and $\textbf{h}_k$ is computed using (\ref{eqn:hc}) as $z^k_i\in [0,1]$. A perfect correlation of $z_k^i =1$ indicates that both $\textbf{h}_i$ and $\textbf{h}_k$ have identical inlier only preference lists, while  $z_k^i =0$ indicates completely dissimilar. We construct a binary similarity matrix \textbf{B} by thresholding $z_k^i\geq\delta$. We say a pair of hypotheses are similar, \ie~ $b_k^i =1$, if $z_k^i \geq\delta$, else dissimilar \ie~$b_k^i =0$. We construct a binary similarity matrix $\textbf{B}$ by thresholding $z_j^i\geq\delta$.
\subsection{Model Hypothesis Goodness Measure}
For each model hypothesis $\textbf{h}_i$ we measure it goodness $g_i$, it is defined as the ratio of median density of estimated inliers and top-$\beta$ \textit{closest outliers} weighted by inverse estimated inlier noise scale $\sighat_i$. The goodness score $g_i$ is computed as shown below in Eq. \ref{eq:good}, where $\gamma_i=[{t_{i}+1},...,{t_i+\beta}]$ are the indices of top-$\beta$ \textit{closest outliers} in the residual space.
\begin{equation}
    g_i = \frac{\text{median}(d_i^{1:t_i})}{\text{median}(d_i^{\gamma_i})} \times \frac{1}{\sighat_i}
    \label{eq:good}
\end{equation}
\noindent We use goodness scores $\textbf{g}=[g_1,...,g_m]$ of all the model hypothesis in our model selection algorithms, which we describe in the next sections.

\subsection{Greedy Model Selection (GMS)}
\label{sec:gms}
The complete greedy model selection (GMS) algorithm is explained in Algo. \ref{alg:gms}. It starts with initializing the index set $\ell=\{1,..,m\}$, which, contains the indices of all generated hypotheses in \textbf{H}. The Algo. \ref{alg:gms} takes model goodness scores $\textbf{g}=[g_1,g_2,...,g_m]$ and binary similarity matrix \textbf{B} (Sec. \ref{sec:hyp_corr_sf}) and output the indices of final selected models in $\vartheta$. The algorithm begins with selecting the hypothesis with maximum goodness score (Algo. \ref{alg:gms},~line \ref{alg:gms:l4}) (say the hypothesis with index $k$ (\ie~$\textbf{h}_k$) has the maximum goodness score). Next, it identifies the hypotheses similar to $k^{th}$ hypothesis in the set $\varrho$ and remove them from the set $\ell$. The removed hypotheses are high likely the representative hypotheses the structure already explained by \kth~hypothesis. The process is repeated until the set $\ell$ is empty. The final set of fitted models are in $ \vartheta  $ (Algo. \ref{alg:gms}, line \ref{alg:gms:l6}).

\input{GMS}

\subsection{Global Optimal Model Selection}
\label{sec:oms}
In this section, we formulate the model selection as a standard quadratic program for global optimal model selection. The quadratic program formulation is given in Eq. \ref{eq:qpms}, where,  $\textbf{g}=[g_1,g_2,...,g_m]$ contains the model goodness score of all the hypotheses in \textbf{H}, $\lambda$ is the regularization constant and \textbf{Q} is the symmetric matrix derived from symmetric hypothesis correlation matrix \textbf{Z} (Sec. \ref{sec:hyp_corr_sf}), diagonal penaly matrix \textbf{P} (Sec. \ref{sec:dpm}) and \textbf{g}. The solution of the quadratic program is a $m$ dimensional vector $\textbf{y}\in [0,1]^{m\times 1}$, where $m$ is the number of hypotheses in \textbf{H}. 
We use \textit{trust region reflective} algorithm \cite{conn2000trust,coleman1996interior} to solve the quadratic program in Eq. \ref{eq:qpms}. The trust region is defined by the linear bounds $0 \leq y_i \leq 1,~ \forall i \in \{1,...,m\}$. We use \textit{qudprog} solver of Matlab\footnote{https://in.mathworks.com/help/optim/ug/quadprog.html} to apply \textit{trust region reflective} algorithm. 
The initial point $\textbf{y}_{0}$ for the optimization is set to $\textbf{y}_{0} = \textbf{0.5}^{m \times 1}$, which is a $m$ length vector with all entries equal to 0.5. The final set of fitted models $ \vartheta $ are obtained as $ \vartheta =\{i~|~y_i \geq \pi\}$, where, $\pi$ is the threshold we use to hard select the final set of hypotheses. We set $\pi=1e\text{-}3$ for all our experiments.

\begin{equation}
    \begin{aligned}
       \max_{\textbf{y}} \quad & \textbf{g}^{T}\textbf{y} - \lambda~\textbf{y}^{T}\textbf{Q}\textbf{y}\\
       \text{s.t.} \quad & \textbf{y} \in [0,1]^{m\times 1}\\
    \end{aligned}
    \label{eq:qpms}
\end{equation}
We next describe the process of constructing the diagonal penalty matrix \textbf{P}.

\subsubsection{Diagonal Penalty Matrix}
\label{sec:dpm}
\begin{figure}
    \centering
    \includegraphics[width=7cm]{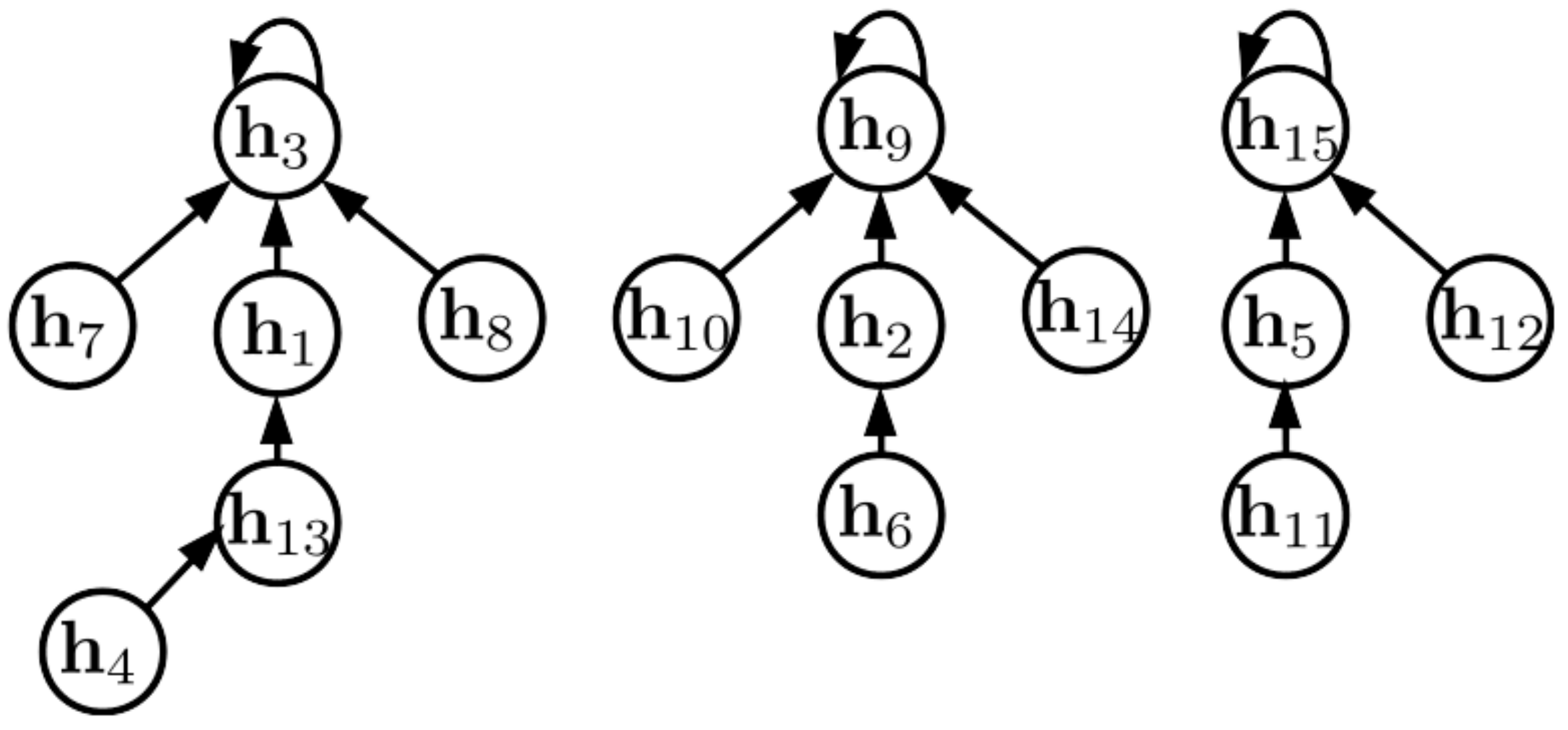}
    \caption{If there are a total of 15 hypotheses present in \textbf{H}, a possible output of Algo. \ref{alg:dpm}, lines \ref{alg:dpm:l4}-\ref{alg:dpm:l7} looks like this. \textit{Note:}  This is just for the illustration purpose, the number of trees and hypotheses may vary. }
    \label{fig:DPMsnap}
\end{figure}

The steps for constructing a diagonal penalty matrix is explained in Algo. \ref{alg:dpm}. We adopt a tree traversal based construction of diagonal penalty matrix similar to proposed \cite{yu2011global}.  We first iteratively construct trees by connecting edges between nodes. The node corresponds to the hypothesis and the directed edge between hypotheses $h_i$ to $h_k$ indicates $g_k$ geq $g_i$ and $z^i_k > 0.5$. The complete process as follows: for each \ith~ hypothesis we select the maximum correlated hypothesis (other than itself) (Algo. \ref{alg:dpm},~line \ref{alg:dpm:l4}). If the maximum correlated hypothesis (say \kth~ hypothesis) has high goodness score and they are 50\% correlated (Algo. \ref{alg:dpm},~line\ref{alg:dpm:l5}), we add an edge from $h_i$ to $h_k$ (Algo. \ref{alg:dpm},~line \ref{alg:dpm:l6}), otherwise, we add an edge to itself (Algo. \ref{alg:dpm},~line\ref{alg:dpm:l7}). We repeat this process for all the hypothesis in \textbf{H}. A snapshot of the output of Algo. \ref{alg:dpm},~lines \ref{alg:dpm:l4}-\ref{alg:dpm:l7} is shown in Fig. \ref{fig:DPMsnap}. For each node (equivalently hypothesis $\textbf{h}_i$), we find its root node (Algo. \ref{alg:dpm},~line \ref{alg:dpm:l9}). For example, refer Fig. \ref{fig:DPMsnap}, the root node of $\textbf{h}_4$, $\textbf{h}_8$ $\textbf{h}_6$ and $\textbf{h}_{15}$ are $\textbf{h}_3$, $\textbf{h}_3$, $\textbf{h}_9$ and $\textbf{h}_{15}$ respectively. Let the root node of $h_i$ is indexed by $i_r$ (Algo. \ref{alg:dpm},~line \ref{alg:dpm:l10}). For each \ith~ hypothesis, if $i\neq i_r$, we compute the respective diagonal entry in the diagonal penalty matrix \textbf{P}, \ie~$p^i_i = \max(~\textbf{g}~) \times (~z^{i}_{i_r} g_i)$ (Algo. \ref{alg:dpm},~line \ref{alg:dpm:l13}). The diagonal matrix is then added to the scaled hypothesis correlation matrix \textbf{Z} as shown below in Eq. \ref{eq:pm} to enforce model diversity.
\begin{eqnarray}
   \textbf{Q} =  \max(\textbf{g})\times \textbf{Z} + \textbf{P}  \label{eq:pm}
\end{eqnarray}

\input{Penalty_matrix}

%% file: GMS.tex
\begin{algorithm}[h!]
\DontPrintSemicolon
\SetAlgoLined
\textbf{Input:} $\textbf{g}, \textbf{B}$,~~\textbf{Output:}$~~\vartheta$ \\
\textbf{Initialization:}~$ \vartheta \leftarrow \varnothing , \ell \leftarrow \{1,..,m\}$ \label{alg:gms:init}\\   
\While{$\ell \neq \varnothing$ }{
$k  \leftarrow \text{argmax}_i~ g_i, ~~~~ \forall i \in \ell$ \label{alg:gms:l4}\\
$ \varrho \leftarrow \{ i ~ |~ b_{k}^{i} = 1 \}, ~~~~ \forall i \in \ell$ \label{alg:gms:l5}\\
$\vartheta \leftarrow \{\vartheta \cup \kappa\}$ \label{alg:gms:l6}\\

$\ell \leftarrow  \{ \ell \setminus \varrho \}$ \label{alg:gms:l7}\\
}
\caption{Greedy Model Selection (GMS)}\label{alg:gms}
\end{algorithm}

%% file: Penalty_matrix.tex
\begin{algorithm}[h!]
	\DontPrintSemicolon
	\SetAlgoLined
	\SetKwInOut{Input}{Input}\SetKwInOut{Output}{Output}
	\textbf{Input:} $ \textbf{H}, \textbf{Z}, \textbf{g}$,~~\textbf{Output:} \textbf{P} \\
	\textbf{Initialization:}~$\textbf{P} \leftarrow \textbf{0}_{n\times n}$\\
		\For{$i \leftarrow 1 ~ to ~ m$}{
		 $k \leftarrow \text{argmax}_j~z_j^i$ \label{alg:dpm:l4} \\
		 \textbf{if} $k \neq i,~~~ \text{and}~~ g_k \geq g_i~~ \text{with}~~ z_k^i \geq 0.5$ \label{alg:dpm:l5}\\
		 \textbf{then} \text{add an edge from} $\textbf{h}_i~~to~~ \textbf{h}_k$~~~~$(\textbf{h}_i\rightarrow \textbf{h}_k)$\label{alg:dpm:l6}\\
		 \textbf{else} \text{add an edge to itself} from $\textbf{h}_i~~\text{to} ~~\textbf{h}_i$~~~~$(\textbf{h}_i\rightarrow \textbf{h}_i)$\label{alg:dpm:l7}\\
		} \label{alg:dpm:l8}
\text{for each node ($\textbf{h}_i$), perform tree traversal and find its}  \label{alg:dpm:l9}\\
\text{root node $\textbf{h}_{i_{r}}$ indexed by $i_{r}$}. \label{alg:dpm:l10}\\
	\For{$i \leftarrow 1 ~ to ~ m$}{
	\If{$i \neq i_r$}{
	$p_{i}^{i} = \max(\textbf{g})\times(z_{i_r}^{i}  g_{i_r}) $ \label{alg:dpm:l13}
	}
	}
	\caption{\texttt{Diagonal Penalty Matrix}}
	\label{alg:dpm}
\end{algorithm}

%% file: point-to-model_assignment.tex
\section{Point-to-Model Assignment}
\label{sec:p2m}
After model selection either  \textit{greedy} (GMS) or \textit{optimal} (QMS), we get indices of our final set of selected model hypotheses in $\vartheta$ and their associated inlier sets $\mathcal{I}_i, i \in \vartheta$. At this stage, some of the data points may be members of multiple sets $ \mathcal{I}_i$ and $ \mathcal{I}_j$ for $i,j\in\vartheta $. This is acceptable for soft partitioning, however, we reassign the points based on the kernel density to achieve hard partitioning of data points. That is, a point (say $x^j$) can be associated to \textit{only one} inlier set (say $\mathcal{I}_k$), provided $d_k^{j} \geq d_i^{j}~~ \forall i \in \{\vartheta \setminus k\}$. We refine the inlier sets following above kernel density based point-to-model assignment strategy.

%% file: experiments.tex
\section{Experimental Analysis}
\label{sec:exps}
In this section, we evaluate our proposed DGSAC pipeline. We first present an experimental evaluation of our KRD Guided Sampling (KDGS). Next, we evaluate the full DGSAC pipeline. We present the evaluation of two variants of our full pipeline. One with guided sampling (KDGS) with greedy model selection (dubbed as DGSAC-G), other is a guided sampling (KDGS) with optimal (QMS) model selection (dubbed as DGSAC-O). We evaluate full DGSAC pipeline on wide variety of applications: \emph{planar segmentation}, \emph{motion segmentation}, \emph{vanishing point estimation/lines classification}, \emph{plane fitting to 3D point cloud}, \emph{line} and \emph{circle fitting}. \\

\noindent\textbf{Datasets:} The datasets we use in this paper for the respective applications are as follows: \\

\noindent\underline{Planar and Motion Segmentation}: We use standard AdelaideRMF  \cite{wong2011dynamic} dataset. It consists of 19 each sequence for planar and motion segmentation. The dataset provides labeled SIFT point correspondences in two-view and the ground-truth labeling for each point correspondence. We use ground-truth only for the evaluation purpose. \\
\noindent\underline{Vanishing Point Estimation}: We use the York urban line segment  \cite{york} and the Toulouse Vanishing Points  \cite{tvp} data sets. The York Urban and Toulouse Vanishing Point data sets comprise 102 and 110 images of indoor and outdoor urban scenes. \\
\noindent\underline{Plane Fitting to 3D Point Cloud}: We use two real examples \textit{CastelVechio} and \textit{PozzoVeggiani} of SAMANTHA  \cite{SAMANTHA} data set. Since the ground truth labeling of data points is not available. We show qualitative results for this application. \\
\noindent\underline{Line and Circle Fitting}: We use \textit{Star5} and \textit{Circle5} dataset from  \cite{jlink}. These examples are synthetically generated with Gaussian noise $\sigma=0.0075$ and 50\% outliers.

\input{hyp_gen_combined}

\subsection{Experimental Analysis of KDGS}
We compare our guided sampling algorithm with other state-of-the-art methods like DHF  \cite{wong2013simultaneous}, Multi-GS \cite{chin2012accelerated}, ITKSF  \cite{wong2013simultaneous} for which the authors released the implementations \footnote{We thank Hoi Sim Wong for providing the source code of DHF and ITKSF.}. The competing methods DHF, ITKSF, and Multi-GS, works in a time budget framework and require a user-specified inlier threshold. In contrast, our method (KDGS) is a non-time budget, self-terminating, and does not require an inlier-outlier threshold. Since KDGS is an automated guided sampling method, for a fair comparison, we first run KDGS and record the time taken for each data sequence. We run all three competing methods for the same time budget and evaluate KDGS using AdelaideRMF  \cite{wong2011dynamic} dataset for planar and motion segmentation. 
\input{hyp_gen_evol}
\noindent\textbf{Metrics.} The competing methods have defined a good model hypothesis as a hypothesis fitted on an all inlier MSS. However, it may be possible that a hypothesis fit on an all inlier MSS results in a bad hypothesis due to the inherent noise scale \cite{tennakoon2015robust}. Therefore, in addition to the all inlier MSS criteria, we define another strong criterion for defining a good model hypothesis: a hypothesis is a called a good hypothesis if its estimated inliers have at-least 80\% overlap with the true-inliers. \\

\noindent\textbf{Results.} We report quantitative results in Tab. \ref{tab:quant_kdgs}. We tabulate the total number of hypotheses generated by the respective methods (\#H), the percentage of good hypotheses based on all inliers MSS (\#HM(\%)), and the percentage of hypothesis satisfying the 80\% overlapping criteria (\#HI(\%)). The total time taken $tim$ (in seconds) by the KDGS algorithm is reported.  While we have also reported the running time, it may not be a strictly fair comparison as some parts of the competing methods are implemented in C programming language. DGSAC is fully implemented in Matlab; therefore, further improvement in running time is possible with an optimized implementation.\\

\noindent \textbf{Analysis.} On the one hand, most of the guided sampling algorithms use $10$ seconds as a reasonable time budget for running the guided sampling algorithm. On the other hand, the proposed KDGS self-terminates after generating hypotheses explaining all the data points. From the results in Tab. \ref{tab:quant_kdgs} it can be seen that compared to other competing methods, KDGS can generate a high percentage of good hypothesis and self-terminate within a time which is an order of magnitude smaller (in most of the cases) than the usual time budget (\ie~10s). In some cases, \textit{johnsona}, and \textit{oldclassicswing}, the time taken by KDGS is slightly more than 10s, while generating a high percentage of good quality hypotheses. Any time budget used by the sampling methods is merely a guess, and there exist no time budget constraints that ensure at least one good hypothesis generation for all genuine structures. While KDGS uses an explanation score to stop the sampling process, which is a data-driven approach than any time budget guess. \\
Consider \textit{breadtoy} example, where, \#HM is smaller than \#HI, which indicates that not all hypothesis fitted on all inlier MSS are good. It happens due to the inherent large noise scale within the true structure, which leads to a bad hypothesis capable of describing less than 80\% of true inliers. \\
\noindent\textbf{DHF vs KDGS.} The closest method to our KDGS is the DHF. DHF also follows a per point iterative guided sampling approach. We show using the multiple-circle fitting example on \textit{Circle5} dataset to show the quality of hypotheses generated by both KDGS and DHF \wrt~ the sampling iterations. The qualitative analysis is shown in Tab. \ref{tab:dhf-vs-kdgs}. It is evident, KDGS quickly starts sampling within true structures, hence, generating meaningful good hypotheses. \\
\begin{figure}[h!]
    \centering
    \includegraphics[width=8.5cm]{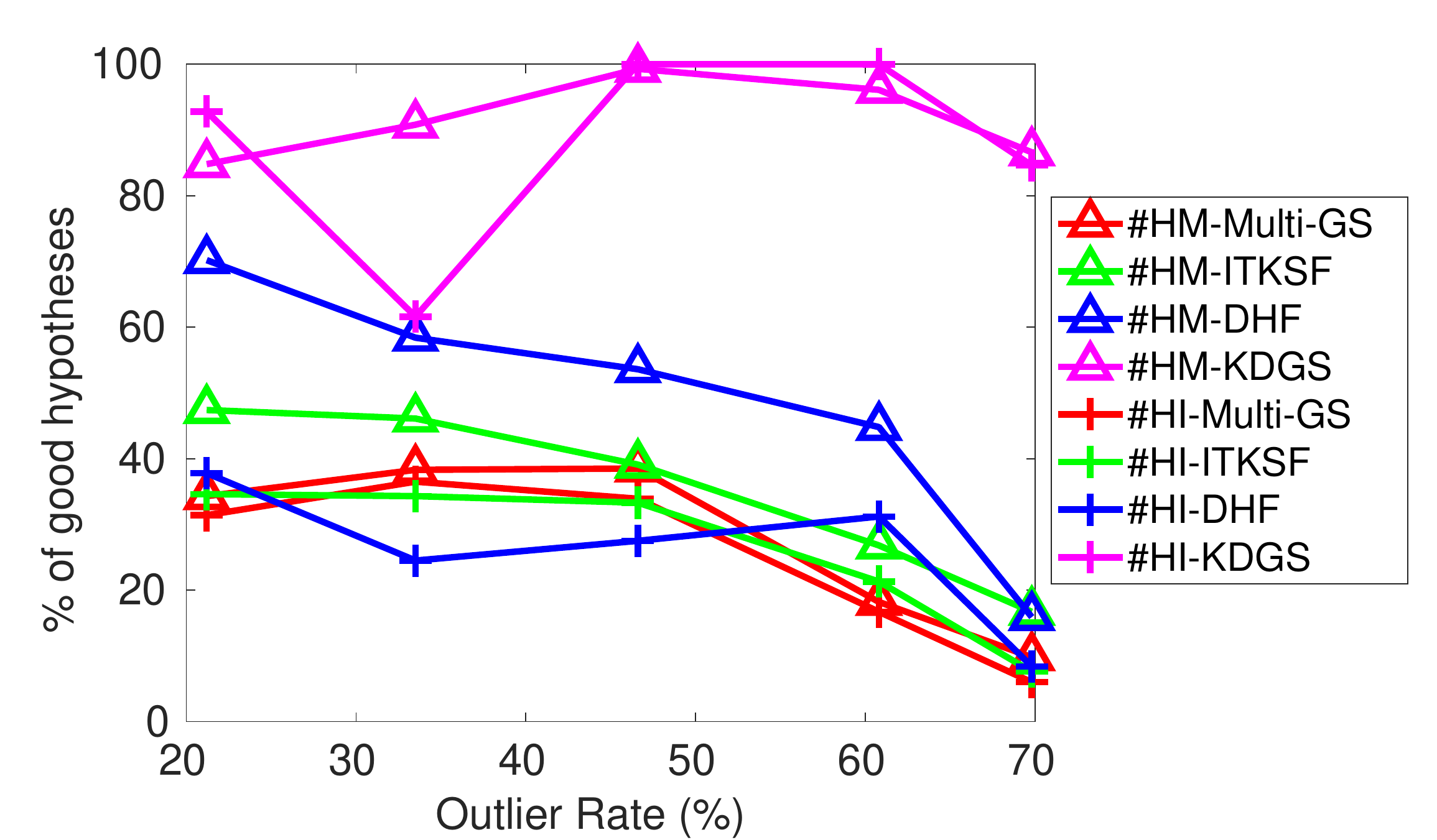}
    \caption{\% of good hypotheses \wrt~ the outlier rate (\%).}
    \label{fig:out_vs_good}
\end{figure}
\noindent\textbf{Outlier rate(\%) vs. \% of good hypotheses.} We analyze the effect of the outlier rate (\%) on the performance of DHF, ITKSF, Multi-GS, and KDGS. We have plotted the \#HM and \#HI metrics against the outlier rate(\%) in Fig. \ref{fig:out_vs_good}. We can observe a drop in the performance \ie~ percentage of good hypotheses generated by DHF, ITKSF, and Multi-GS with the increase in the outlier rate(\%). In contrast, KDGS performance does not affect by the outlier rate. A similar decrease in the number of good hypotheses generated by DHF, ITKSF, and Multi-GS is observed in  \cite{wong2013simultaneous}. 
\input{depndency_tble}
\subsection{Evaluation of full DGSAC Pipeline}
In this section, we evaluate two variants of our DGSAC pipeline: \textit{DGSAC-G} (KDGS with Greedy Model Selection) \textit{DGSAC-O} (KDGS with Optimal Quadratic Program based Model Selection).\\
\noindent \textbf{Competing Methods.} We compare our full DGSAC pipeline on variety of tasks with state-of-the-art multi-model fitting methods for which the source code is publicly released by the respective authors, like J-Linkage (Jlink) \cite{jlink}, T-Linkage (Tlink) \cite{magri14tl}, RPA  \cite{magri15rpa}, DPA \cite{tiwari2016robust}, RCM \cite{pham2014random}, RansaCov (Cov) \cite{magri16set}, NMU  \cite{tepper2016nonnegative}, QP-MF \cite{yu2011global}, 
Prog-X \cite{barath2019progressive}, L1-NMF \cite{tepper2016fast}. We follow the guidelines mentioned by the authors in their papers and provide the necessary parameters like \textit{user specified inlier threshold}, \textit{number of models}, or \textit{both}. Our DGSAC-G/O is the only method that does not require an inlier threshold or the number of models. A comparison of competing methods based on the dependency on user inputs is shown in Tab. \ref{tbl:userip}, these user inputs are mostly computed from the ground truth. \\
\noindent\textbf{Metrics.} We use Classification Accuracy (CA) as an evaluation metric, i.e., the percentage of data points correctly assigned to their respective true structures or gross outliers category. All results are averaged over 10 runs. While we have also
reported the running time, it is not a strictly fair comparison
as the programming language varies across the competing
approaches. DGSAC is implemented in Matlab therefore
further improvement in running time is possible with an optimized implementation.
\input{motion_seg_table}
\input{planar_seg_table}
\subsubsection{ Motion and Planar Segmentation}
We use AdelaideRMF \cite{wong2011dynamic} to evaluate DGSAC-G/O on motion and planar segmentation tasks. The quantitative results are reported in Tab. \ref{tbl:fun}. NMU achieves the highest accuracy for both motion and planar segmentation task, but it takes \textit{highest running time} and \textit{requires user-specified inlier threshold}. Our DGSAC-G achieves the next best accuracy, lagging by a margin of $< 1\%$, and DGSAC-O gives competitive results without any user input (inlier threshold and the number of structures (refer Tab. \ref{tbl:userip}) at all. Moreover, in terms of average time to run, DGSAC-G and DGSAC-O  are respectively nearly $68\times$ and $75\times$ faster than the NMU in motion segmentation, and nearly $21\times$ faster in planar segmentation, with the median run times being even better. We also compare with recently proposed optimization-based method Prog-X and another global optimal quadratic program based method QP-MF \cite{yu2011global}. The Prog-X require inlier-threshold along with other user inputs (refer \cite{barath2019progressive}), while QP-MF require both \textit{user specified inlier threshold} and \textit{number of models} as an inputs. Our optimal DGSAC-O outperforms both Prog-X\footnote{https://github.com/danini/progressive-x. Only planar segmentation implementation is available.} and QP-MF with a significant margin, without any dependency on the user-specified inputs. We report some sample qualitative results of both motion and planar segmentation tasks in Fig. \ref{fig:quan_fun_hom} where point membership is color-coded.
\input{qual_fun}
\input{vp_results}
\input{qual_vp_tvp.tex}
\input{qual_plane}
\input{qual_line}
\input{qual_circle}
\subsubsection{Vanishing Point Estimation and Line Classification} We use the York urban line segment \cite{york} and the Toulouse Vanishing Points  \cite{tvp} dataset to evaluate DGSAC-G/O. We compare with the RANSAC like state-of-the-art vanishing point estimation methods. The quantitative results are reported in Tab. \ref{tab:quan_vp}. DGSAC-G/O outperforms in both the data sets. Sample qualitative results are reported in Fig. \ref{fig:qual_vp_tvp}, where point membership is color-coded, \ie~ lines with the same color belong to the same vanishing point direction.
\subsubsection{Plane Fitting to 3D Point Cloud} We use two real examples \textit{CastelVechio} and \textit{PozzoVeggiani} from SAMANTHA \cite{SAMANTHA} dataset to evaluate DGSAC-G/O. The ground-truth labeling is not provided with the data set. Therefore, we report qualitative results in Fig. \ref{fig:qual_plane}, where point membership is color-coded. Only, J-Linkage and DGSAC-G/O can recover planes correctly.
\subsubsection{Line Fitting}
We use \textit{Star5} dataset from \cite{jlink}. The qualitative and quantitative results are reported in Fig. \ref{fig:qual_line}. While all the competing methods can recover all the five-line structures. Our DGSAC-G and DGSAC-O can classify $96\%$ of the total data points to their respective classes \ie~their true structures or gross outliers. Both DGSAC-G and DGSAC-O output the same set of final hypotheses using greedy model selection and optimal model selection algorithm, hence the same CA(\%).
\subsubsection{Circle Fitting}
We use \textit{Circle5} dataset \cite{jlink}. The qualitative and quantitative results are reported in Fig. \ref{fig:qual_circle}. It can be seen, only DGSAC-G and DGSAC-O are able to recover all the five circles. The next best performing method is RansaCov, which is able to recover 4 out of 5 structures. J-Linkage leads to over-segmentation of structures, while T-Linkage is able to recover only 2 structures.

%% file: hyp_gen_combined.tex
\begin{table*}[h!]

\centering
\scriptsize
\caption{\textbf{Qualitative Evaluation of KDGS.} \#H is the total number of hypotheses generated by each method, \#HM(\%) is the percentage of good hypotheses satisfying the all inlier MSS criteria, \#HI(\%) percentage of good hypotheses having at-least 80\% overlap of their estimated inliers with the true inliers, $n$ is the number of point correspondences, O(\%) is the outlier percentage. The $\mathcal{T}=[\mathcal{T}_1,\mathcal{T}_2,...,\mathcal{T}_{\kappa}]$ vector shows the number of true inliers of all $\kappa$ genuine structures. \textit{tim}(s) shows the total time taken in seconds. Image shows two-view visual ground-truth inliers with color coded structural membership. Outliers are in red. Best result is in \textbf{bold} and second best is \underline{underlined}.}
\setlength{\tabcolsep}{2pt}
\resizebox{\textwidth}{!}{%
\begin{tabular}{|c|ccccc?c|ccccc|}
\hline
\multicolumn{6}{|c?}{\textbf{Planar Segmentation}}                                                                                                                                                                                                          & \multicolumn{6}{c|}{\textbf{Motion Segmentation}}                                                                                                                                                                                                       \\ \hline
                                                          Data      &     &  MGS                                    &       ITKSF                            &           DHF                          &   DGS                                      &                                  Data                               &     &                           MGS     &                           ITKSF     &          DHF                               & DGS                                        \\ \hline
                                                                & \#H & { 620}          & { 218}        & { 55}           & { 34}               &                                                                 & \#H & { 380}     & { 213}     & { 68}               & { 154}              \\
                                                                & \#HM & { 9.8}          & { {\ul 16.7}} & { 15.9}         & { \textbf{86.6}}    &  & \#HM & { 32.15} & { 57.98} & { {\ul 85.23}}    & {\textbf{86.39}} \\   & \#HI & { 6.0}          & { 7.6}        & { {\ul 8.4}}    & { \textbf{84.6}}    &                 & \#HI & { 55.69} & { 59.41} & { {\ul 73.79}}    & { \textbf{97.30}} \\ 
\multirow{-4}{*}{\begin{tabular}[c]{@{}c@{}}
     \includegraphics[width=3.8cm,height=1.2cm]{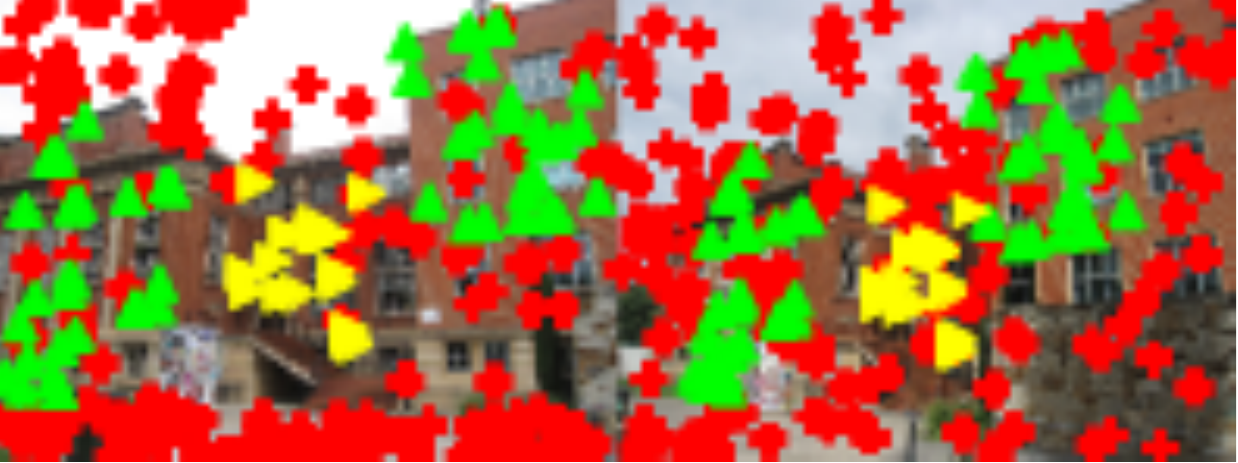}  \\
     barrsmith, $n$~=~235, O~=~69.8\%
    \\
     $\mathcal{T}$~=~[~50,21~]\\
\end{tabular}} & tim & { 3.27}            & { 3.27}          & { 3.27}            & { 3.27}                & \multirow{-4}{*}{\begin{tabular}[c]{@{}c@{}}
     \includegraphics[width=3.8cm,height=1.2cm]{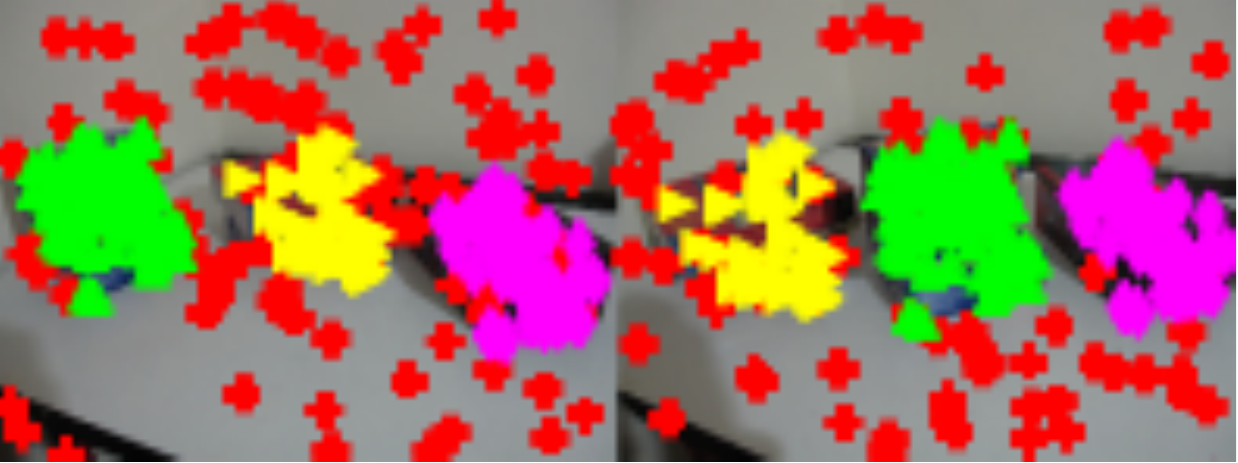} \\
     biscuitbookbox, $n$~=~258, O~=~37.2\%
     \\
     $\mathcal{T}$~=~[~67,41,54~]\\
\end{tabular}} & tim & { 4.38}       & { 4.38}       & { 4.38}                & { 4.38}                \\
      & &          &         &           &              &                                                                 & &      &      &               &               \\
            & &          &         &           &              &                                                                 & &      &      &               &               \\
                  & &          &         &           &              &                                                                 & &      &      &               &               \\
                                    
       \hline

                                              & \#H & { 463}          & { 201}        & { 61}           & { 57}               &                                                                 & \#H & { 328}     & { 169}     & { 71}               & { 157}              \\ 
                                                                & \#HM & { 18.3}       & { 26.8}     & { {\ul 44.8}} & { \textbf{96.1}}  &                                                                 & \#HM & { 21.67} & { 32.26} & { {\ul 68.34}}    & { \textbf{69.41}} \\ 
                                                                & \#HI & { 16.7}       & { 21.3}     & { {\ul 31.2}} & { \textbf{100.0}} &                                                                 & \#HI & { 44.92} & { 41.50} & { {\ul 75.53}}    & { \textbf{95.83}} \\ 
\multirow{-4}{*}{\begin{tabular}[c]{@{}c@{}}
     \includegraphics[width=3.8cm,height=1.2cm]{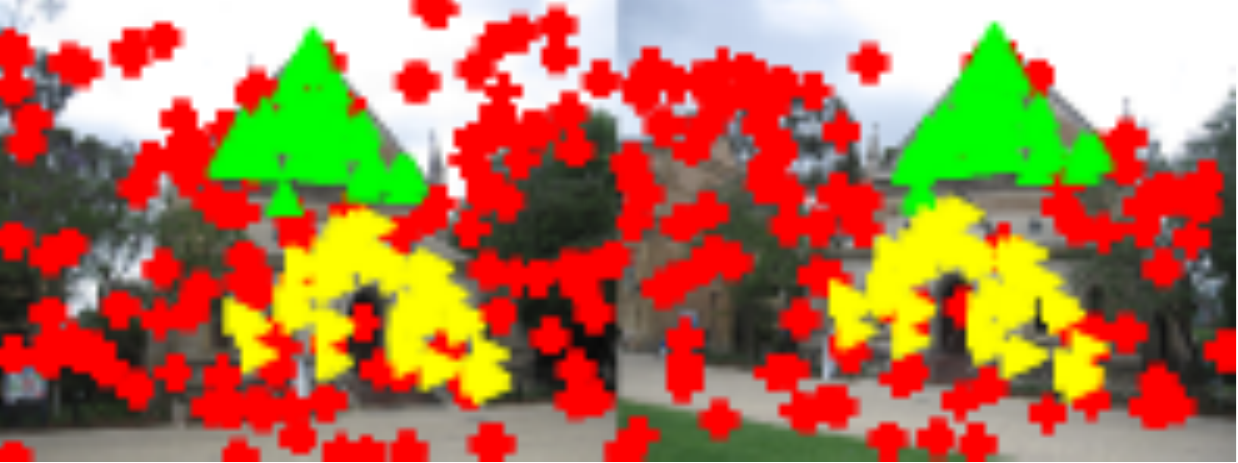}  \\
     elderhalla,~~~$n$~=~214,~O~=~60.8\%
    \\
     $\mathcal{T}$~=~[~38,~46~]\\
\end{tabular}}                                              & tim & { 1.90}            & { 1.90}          & { 1.90}            & { 1.90}                & \multirow{-4}{*}{\begin{tabular}[c]{@{}c@{}}
     \includegraphics[width=3.8cm,height=1.2cm]{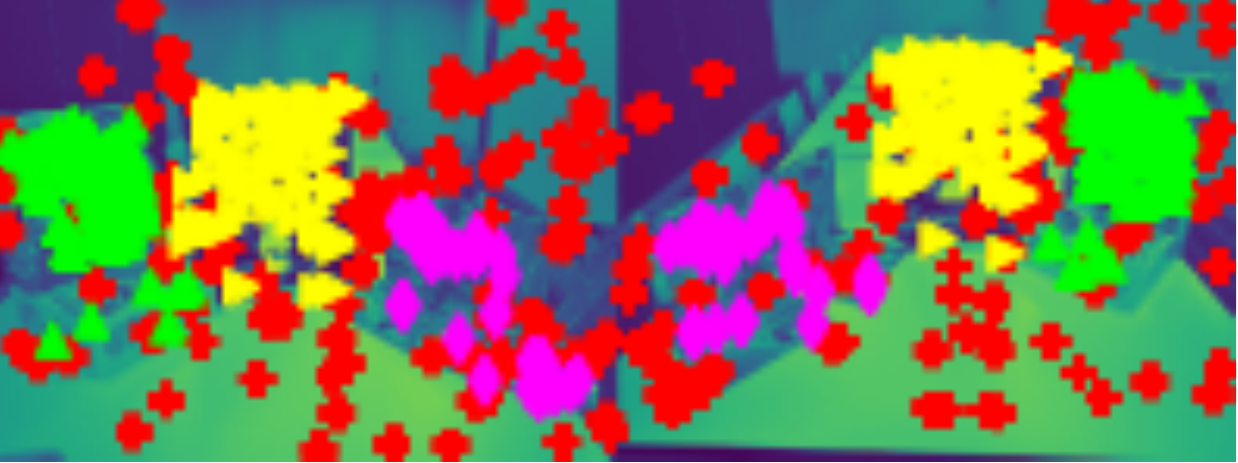}  \\
     boardgame, $n$~=~266,~O~=~42.5\%
     \\
     $\mathcal{T}$~=~[~63,~63,~27~]\\
\end{tabular}}                                              & tim & { 3.33}       & { 3.33}       & { 3.33}                & { 3.33}                \\       & &          &         &           &              &                                                                 & &      &      &               &               \\
            & &          &         &           &              &                                                                 & &      &      &               &               \\
                  & &          &         &           &              &                                                                 & &      &      &               &               \\

       \hline
                                                                & \#H & { 611}          & { 231}        & { 56}           & { 64}               &                                                                 & \#H & { 264}     & { 182}     & { 68}               & { 130}              \\ 
                                                                & \#HM & { 16.0}       & { 23.9}     & { {\ul 27.5}} & { \textbf{81.8}}  &                                                                 & \#HM & { 21.79} & { 50.14} & { \textbf{72.31}} & { {\ul 67.53}}    \\ 
                                                                & \#HI & { 11.7}       & { 12.5}     & { {\ul 15.0}} & { \textbf{82.2}}  &            & \#HI & { 46.00} & { 73.66} & { {\ul 94.45}}    & { \textbf{96.87}} \\ 
\multirow{-4}{*}{\begin{tabular}[c]{@{}c@{}}
     \includegraphics[width=3.8cm,height=1.2cm]{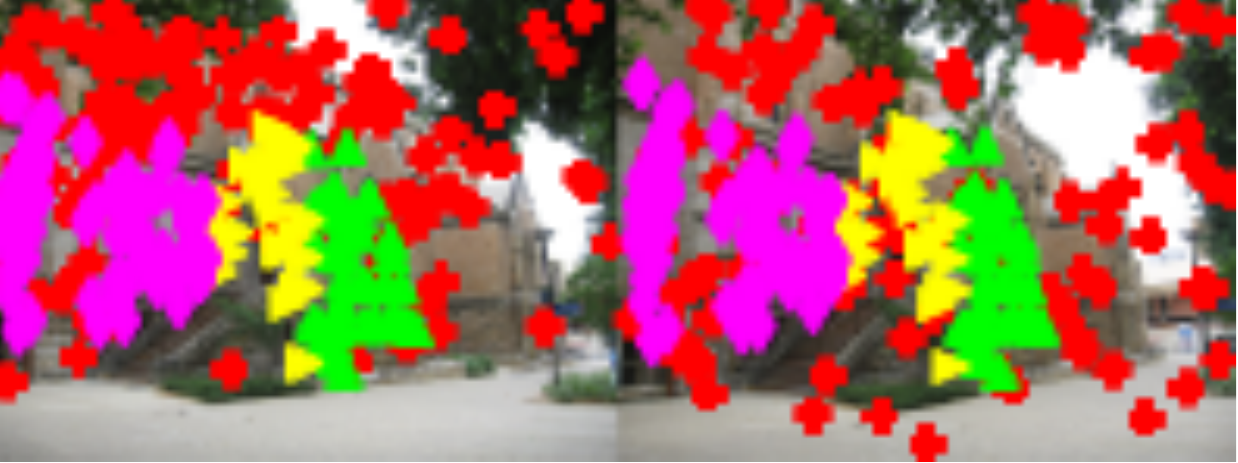} \\
     elderhallb, $n$~=~245,~O~=~49.8\%
    \\
     $\mathcal{T}$~=~[~	38,~25,~60~]\\
\end{tabular}}                                              & tim & { 3.43}            & { 3.43}          & { 3.43}            & { 3.43}                & \multirow{-4}{*}{\begin{tabular}[c]{@{}c@{}}
     \includegraphics[width=3.8cm,height=1.2cm]{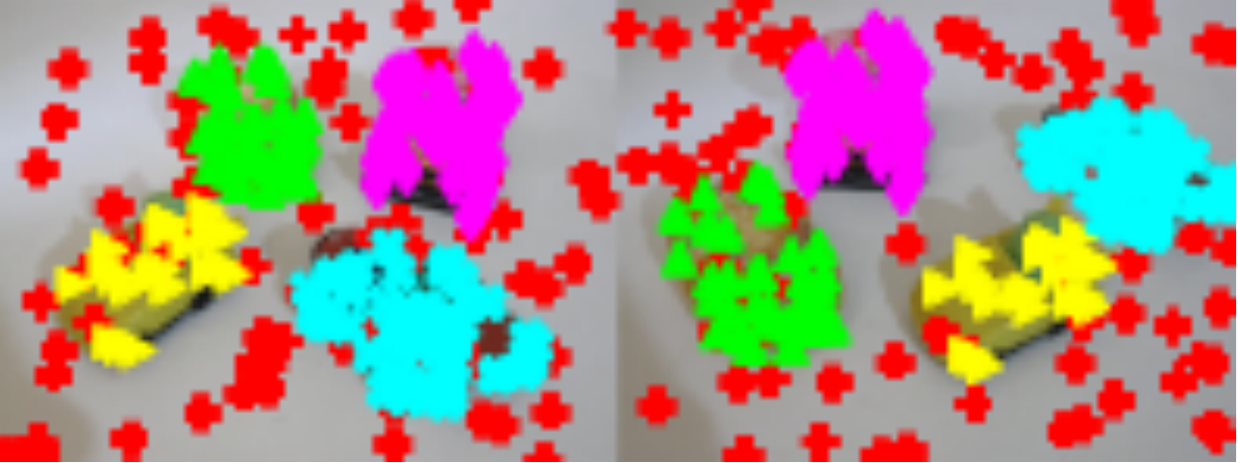} \\
     breadcartoychips, $n$~=~231, O~=~35.5\%
    \\
     $\mathcal{T}$~=~[~33,~23,~39,~54~]\\
\end{tabular}}                                              & tim & { 2.37}       & { 2.37}       & { 2.37}                & { 2.37}                \\       & &          &         &           &              &                                                                 & &      &      &               &               \\
            & &          &         &           &              &                                                                 & &      &      &               &               \\
                  & &          &         &           &              &                                                                 & &      &      &               &               \\

       \hline
                                                                & \#H & { 956}          & { 331}        & { 68}           & { 140}              &                                                                 & \#H & { 258}     & { 181}     & { 70}               & { 137}              \\  
                                                                & \#HM & { 34.3}       & { 47.4}     & { {\ul 70.2}} & { \textbf{84.8}}  &                                                                 & \#HM & { 27.26} & { 55.98} & { \textbf{86.98}} & { {\ul 76.68}}    \\ 
                                                                & \#HI & { 31.4}       & { 34.6}     & { {\ul 37.8}} & { \textbf{92.8}}  &                                                                 & \#HI & { 52.40} & { 73.70} & { {\ul 89.16}}    & { \textbf{96.24}} \\ 
\multirow{-4}{*}{\begin{tabular}[c]{@{}c@{}}
     \includegraphics[width=3.8cm,height=1.2cm]{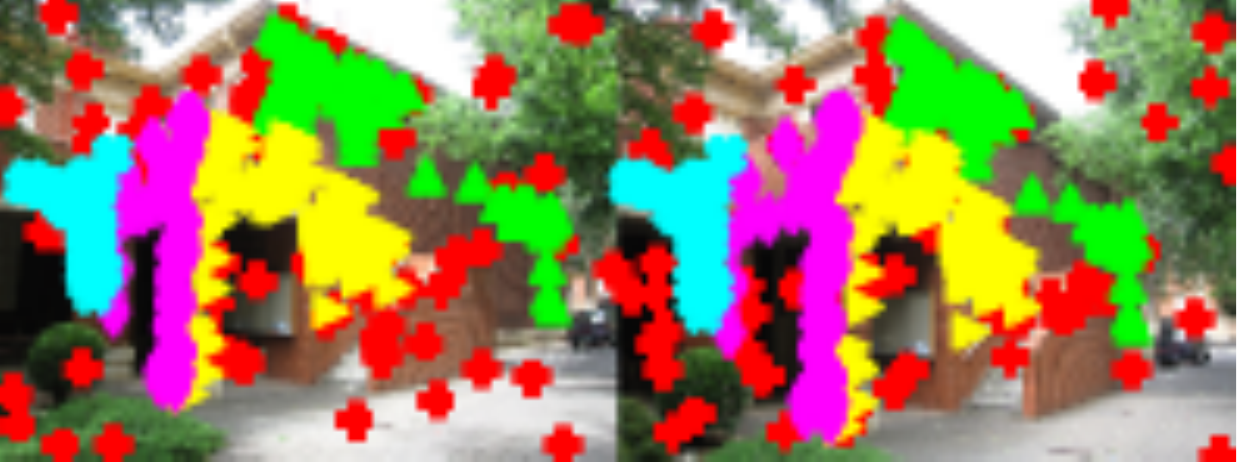} \\
     johnsona, $n$~=~353,~O~=~21.2\%
     \\
     $\mathcal{T}$~=~[~76,~91,~61,~50~]\\
\end{tabular}}                                              & tim & { 11.60}            & { 11.60}          & { 11.60}            & { 11.60}                & \multirow{-4}{*}{\begin{tabular}[c]{@{}c@{}}
     \includegraphics[width=3.8cm,height=1.2cm]{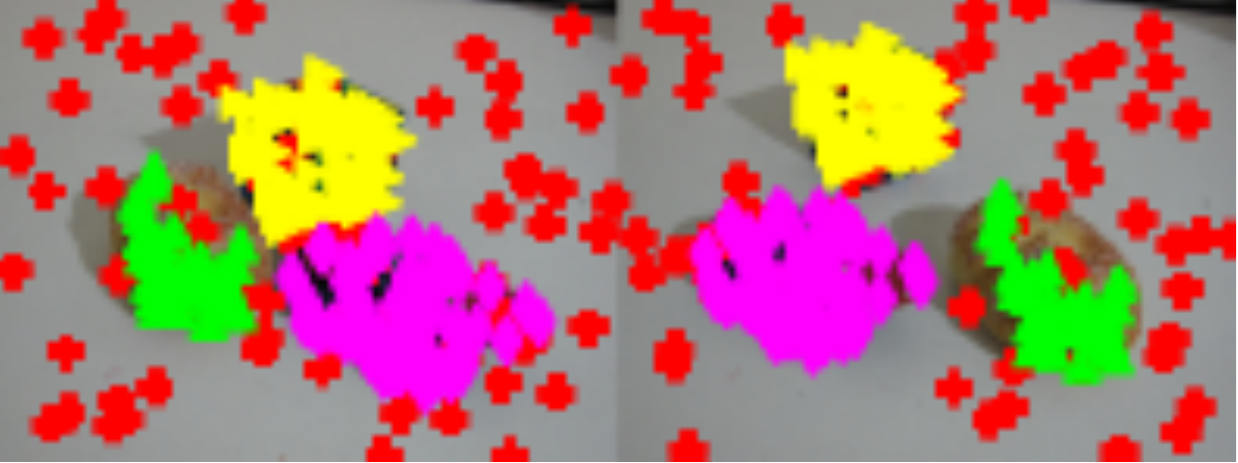}  \\
     breadcubechips, $n$~=~230, O~=~35.2\%
     \\
     $\mathcal{T}$~=~[~34,~57,~58~]\\
\end{tabular}}                                              & tim & { 2.15}       & { 2.15}       & { 2.15}                & { 2.15}                \\       & &          &         &           &              &                                                                 & &      &      &               &               \\
            & &          &         &           &              &                                                                 & &      &      &               &               \\
                  & &          &         &           &              &                                                                 & &      &      &               &               \\

       \hline
                                                                & \#H & { 579}          & { 215}        & { 59}           & { 37}               &                                                                 & \#H & { 415}     & { 237}     & { 76}               & { 81}               \\ 
                                                                & \#HM & { 38.3}       & { 46.1}     & { {\ul 58.4}} & { \textbf{90.8}}  &                                                                 & \#HM & { 28.30} & { 51.56} & { \textbf{85.66}} & { {\ul 80.29}}    \\                    & \#HI & { {\ul 36.5}} & { 34.3}     & { 24.5}       & { \textbf{61.6}}  &                                                                 & \#HI & { 58.80} & { 67.77} & { {\ul 90.56}}    & { \textbf{97.29}} \\ 
\multirow{-4}{*}{\begin{tabular}[c]{@{}c@{}}
     \includegraphics[width=3.8cm,height=1.2cm]{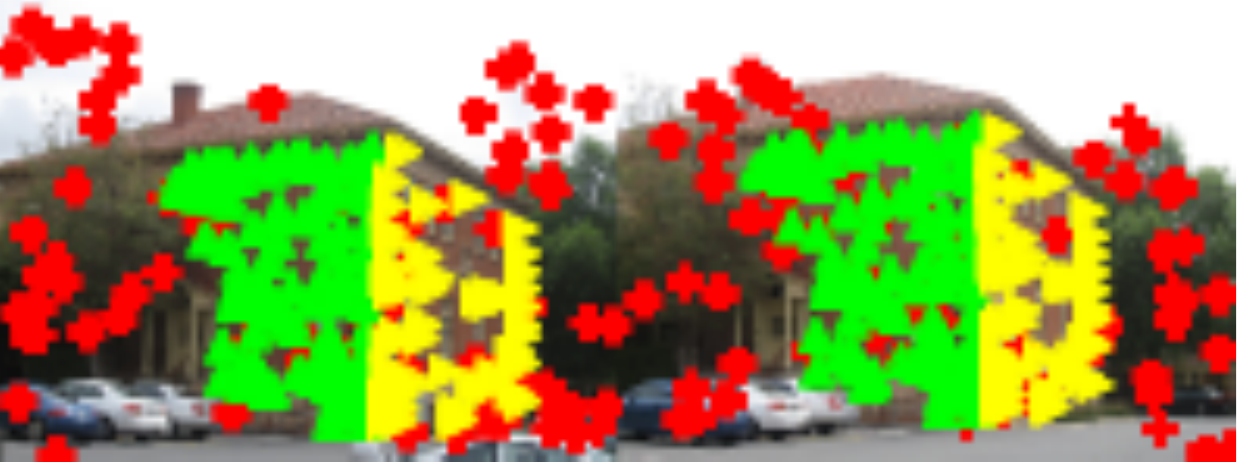}  \\
     ladysymon, $n$~=~227, O~=~33.5\%
     \\
     $\mathcal{T}$~=~[~102,~49~]\\
\end{tabular}}                                              & tim & { 2.65}            & { 2.65}          & { 2.65}            & { 2.65}                & \multirow{-4}{*}{\begin{tabular}[c]{@{}c@{}}
     \includegraphics[width=3.8cm,height=1.2cm]{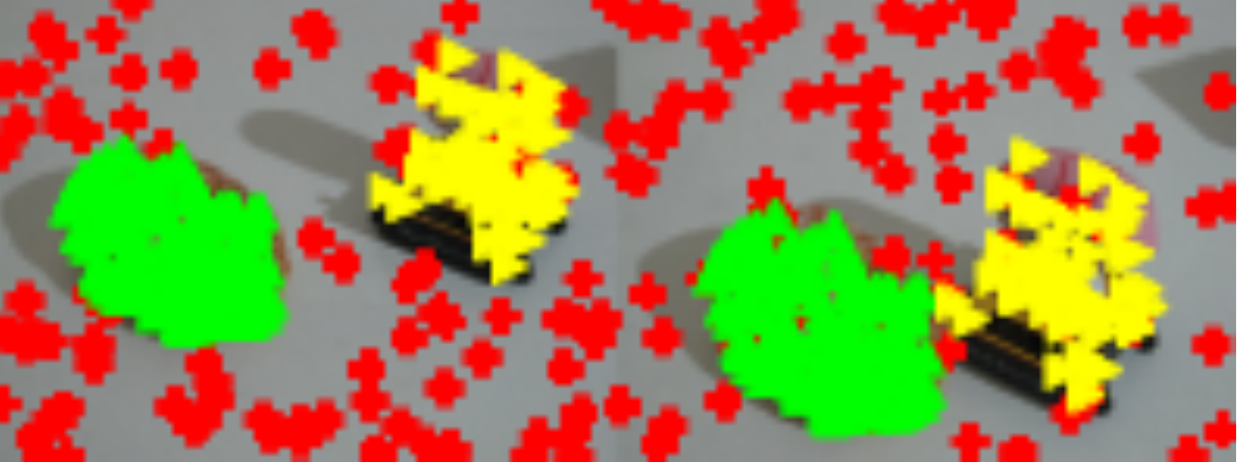} \\
     breadtoy, $n$~=~278, O~=~37.4\%
    \\
     $\mathcal{T}$~=~[~119,~55~]\\
\end{tabular}}                                              & tim & { 4.57}       & { 4.57}       & { 4.57}                & { 4.57}                \\       & &          &         &           &              &                                                                 & &      &      &               &               \\             & &          &         &           &              &                                                                 & &      &      &               &               \\
                  & &          &         &           &              &                                                                 & &      &      &               &               \\
       \hline
                                                                & \#H & { 546}          & { 220}        & { 56}           & { 66}               &                                                                 & \#H & { 197}     & { 135}     & { 57}               & { 88}               \\ 
                                                                & \#HM & { 23.6}       & { 30.3}     & { {\ul 42.7}} & { \textbf{56.3}}  &                                                                 & \#HM & { 21.55} & { 50.67} & { {\ul 71.31}}    & { \textbf{75.12}} \\ 
                                                                & \#HI & { 24.6}       & { 25.0}     & { {\ul 28.3}} & { \textbf{83.1}}  &                                                                 & \#HI & { 40.44} & { 66.58} & { {\ul 82.22}}    & { \textbf{96.82}} \\ 
\multirow{-4}{*}{\begin{tabular}[c]{@{}c@{}}
     \includegraphics[width=3.8cm,height=1.2cm]{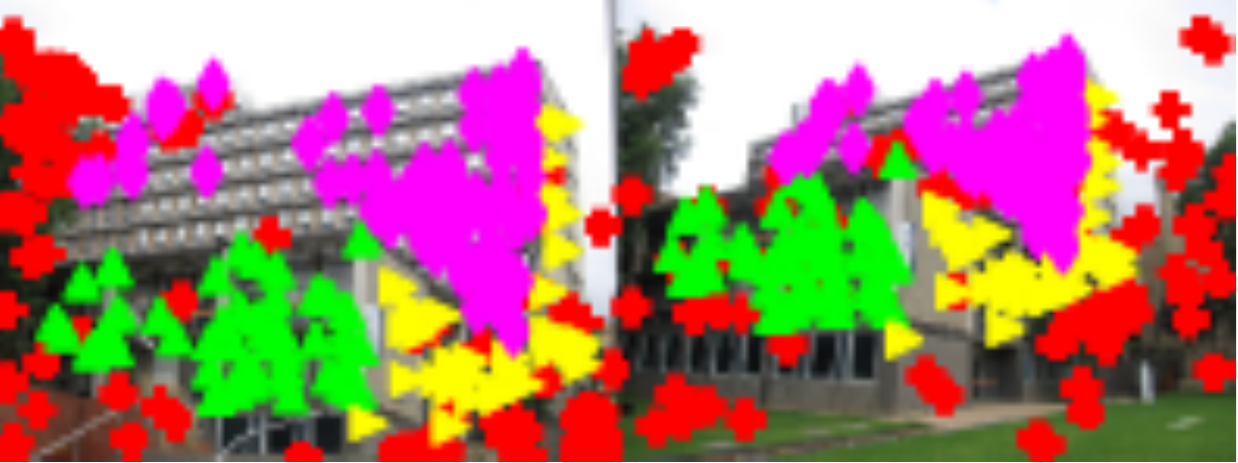} \\
     napierb, $n$~=~237, O~=~37.1\%
\\
     $\mathcal{T}$~=~[~46,~33,~70~]\\
\end{tabular}}                                              & tim & { 2.43}            & { 2.43}          & { 2.43}            & { 2.43}                & \multirow{-4}{*}{\begin{tabular}[c]{@{}c@{}}
     \includegraphics[width=3.8cm,height=1.2cm]{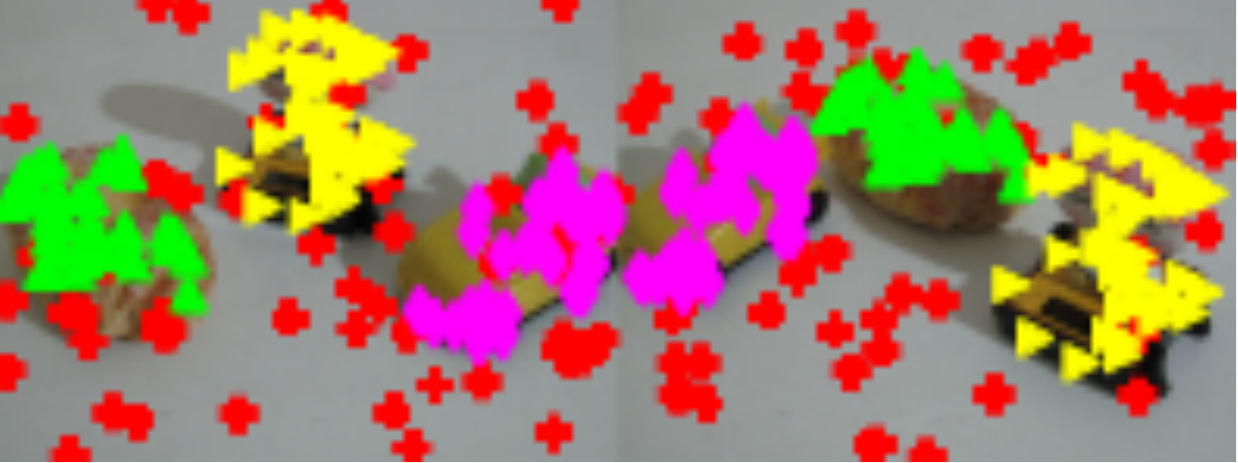}  \\
     breadtoycar, $n$~=~164, O~=~34.1\%
   \\
     $\mathcal{T}$~=~[~37,~39,~32~]\\
\end{tabular}}                                              & tim & {1.24}       & { 1.24}       & { 1.24}                & { 1.24}                \\       & &          &         &           &              &                                                                 & &      &      &               &               \\
                                    & &          &         &           &              &                                                                 & &      &      &               &               \\
                                                                        & &          &         &           &              &                                                                 & &      &      &               &               \\
     
       \hline
                                                                & \#H & { 939}          & { 343}        & { 66}           & { 32}               &                                                                 & \#H & { 176}     & { 138}     & { 57}               & { 84}               \\ 
                                                                & \#HM & { 48.8}       & { 55.2}     & { {\ul 74.8}} & { \textbf{95.7}}  &                                                                 & \#HM & { 23.64} & { 50.37} & { {\ul 69.42}}    & { \textbf{70.21}} \\ 
                                                                & \#HI & { {\ul 51.9}} & { 49.6}     & { 50.0}       & { \textbf{95.2}}  &                                                                 & \#HI & { 50.37} & { 69.12} & { {\ul 84.95}}    & { \textbf{92.86}} \\ 
\multirow{-4}{*}{\begin{tabular}[c]{@{}c@{}}
     \includegraphics[width=3.8cm,height=1.2cm]{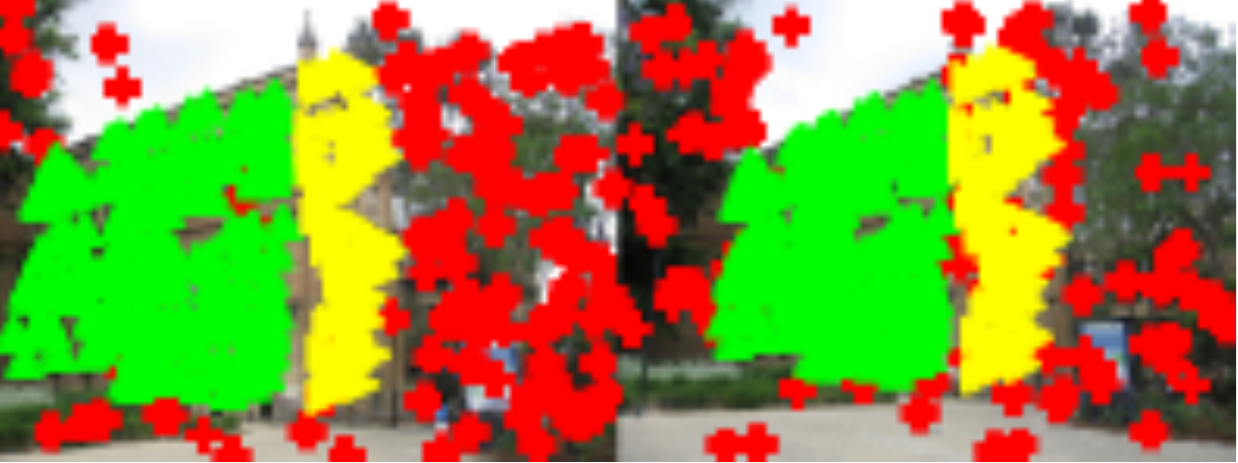}  \\
     oldclassicswing, $n$~=~363, O~=~32.2\%
  \\
     $\mathcal{T}$~=~[~181,~65~]\\
\end{tabular}}                                              & tim & { 11.30}            & { 11.30}          & { 11.30}            & { 11.30}                & \multirow{-4}{*}{\begin{tabular}[c]{@{}c@{}}
     \includegraphics[width=3.8cm,height=1.2cm]{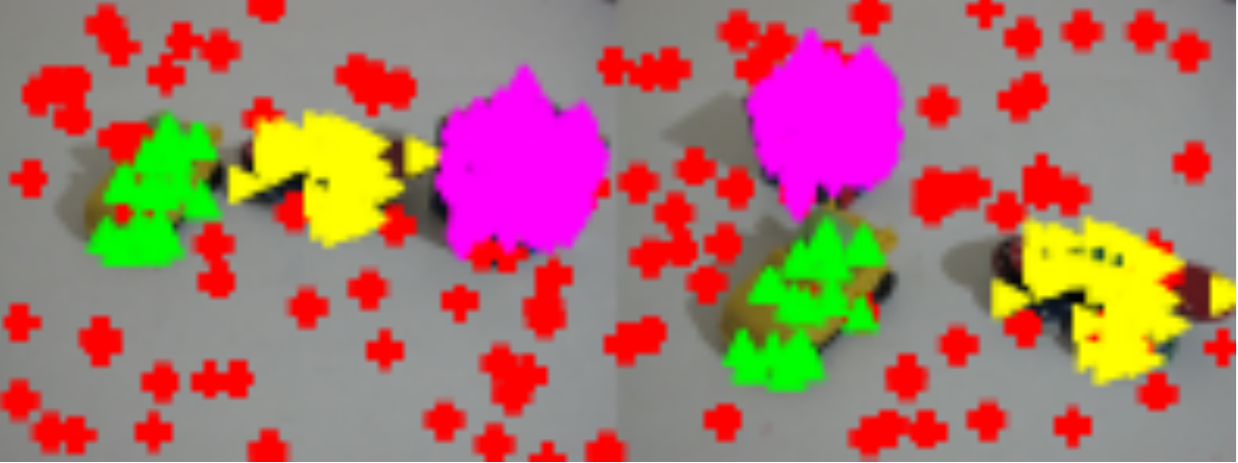}  \\
     carchipscube, $n$~=~164, O~=~36.6\%
    \\
     $\mathcal{T}$~=~[~18,~33,~53~]\\
\end{tabular}}                                              & tim & { 1.23}       & { 1.23}       & { 1.23}                & { 1.23}                \\ & &          &         &           &              &                                                                 & &      &      &               &               \\
            & &          &         &           &              &                                                                 & &      &      &               &               \\
                  & &          &         &           &              &                                                                 & &      &      &               &               \\
       \hline
                                                                & \#H & { 141}          & { 95}         & { 47}           & { 10}               &                                                                 & \#H & { 371}     & { 191}     & { 73}               & { 170}              \\  
                                                                & \#HM & { 38.5}       & { 39.1}     & { {\ul 53.6}} & { \textbf{99.3}}  &                                                                 & \#HM & { 21.53} & { 36.11} & { \textbf{86.73}} & { {\ul 74.94}}    \\ 
                                                                & \#HI & { {\ul 33.9}} & { 33.3}     & { 27.5}       & { \textbf{100.0}} &                                                                 & \#HI & { 42.67} & { 48.17} & { {\ul 81.37}}    & { \textbf{98.10}} \\ 
\multirow{-4}{*}{\begin{tabular}[c]{@{}c@{}}
     \includegraphics[width=3.8cm,height=1.2cm]{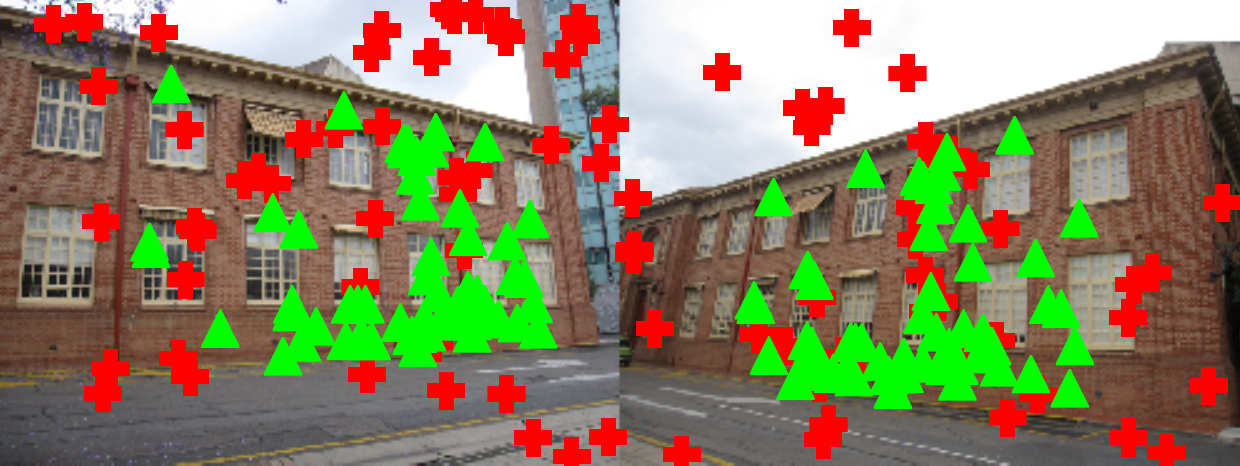}  \\
     physics, $n$~=~103, O~=~46.6\%
    \\
     $\mathcal{T}$~=~[~55~]\\
\end{tabular}}                                              & tim & { 1.01}            & { 1.01}          & { 1.01}            & { 1.01}                & \multirow{-4}{*}{\begin{tabular}[c]{@{}c@{}}
     \includegraphics[width=3.8cm,height=1.2cm]{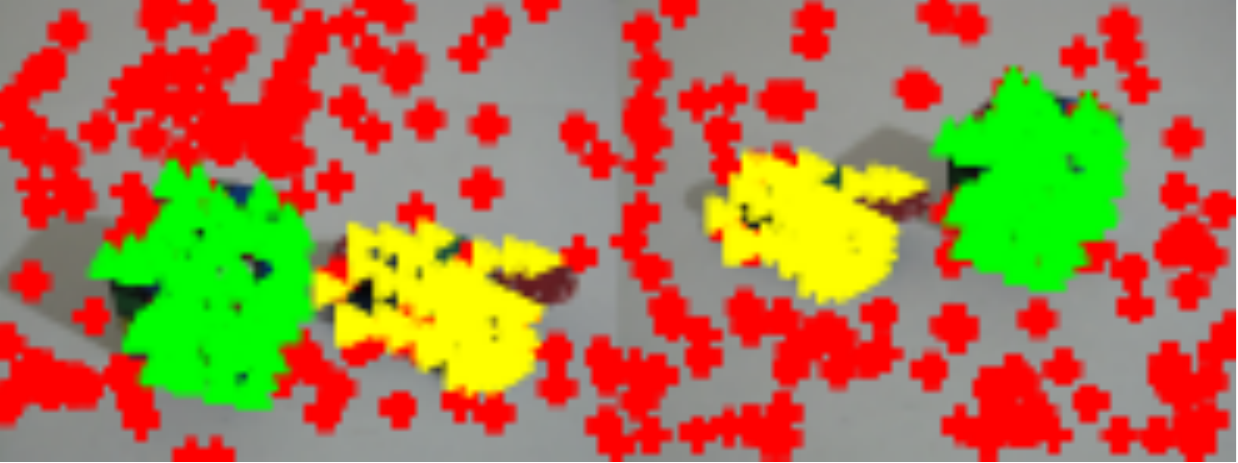}  \\
     cubechips, $n$~=~277, O~=~51.6\%
     \\
     $\mathcal{T}$~=~[~81,~53~]\\
\end{tabular}}                                              & tim & { 3.73}       & { 3.73}       & {3.73}                & {3.73}                \\       & &          &         &           &              &                                                                 & &      &      &               &               \\
            & &          &         &           &              &                                                                 & &      &      &               &               \\
                  & &          &         &           &              &                                                                 & &      &      &               &               \\

       \hline
\end{tabular}
}
\label{tab:quant_kdgs}
\end{table*}

%% file: hyp_gen_evol.tex
\begin{table*}[h!]
\centering
\setlength{\tabcolsep}{1.5pt}

\caption{\textbf{Comparison of KDGS with DHF.} Both DHF and KDGS focus on generating hypothesis for each data point, the plotted hypotheses (in red) are of the inliers of five genuine structures (s1,...,s5) shown in column 1. $n$ is the total number of data points, O\% is the ground-truth (GT) percentage of gross outliers. Ground-truth structure is shown in column 1 (in magenta) and outliers are in green. For better illustration we have plotted hypotheses of each of the five structures separately in the five different rows. }
\begin{tabular}{|c|cccccc?cccccc|}
\hline
\multicolumn{1}{|c|}{{GT, n=500}}   & \multicolumn{6}{|c?}{{DHF}}                                                                                                                                                          & \multicolumn{6}{c}{{KDGS}}                                                                                                                                    \\ 
\multicolumn{1}{|c|}{{O\%=43.2}} & \multicolumn{1}{c}{{iter=1}} & \multicolumn{1}{c}{{iter=25}} & \multicolumn{1}{c}{{iter=50}} & \multicolumn{1}{c}{{iter=75}} & \multicolumn{1}{c}{{iter=100}} & \multicolumn{1}{c?}{{iter=120}} & \multicolumn{1}{c}{{iter=1}} & \multicolumn{1}{c}{{iter=2}} & \multicolumn{1}{c}{{iter=3}} & \multicolumn{1}{c}{{iter=4}} & \multicolumn{1}{c}{{iter=5}} & \multicolumn{1}{c}{{iter=6}} \\ 
\rotatebox{90}{\hspace{.2cm}  {st1, n1=61}}  \includegraphics[width=1.2cm, height=1.7cm, angle=0]{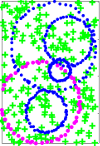} &
\includegraphics[width=1.2cm, height=1.7cm, angle=0]{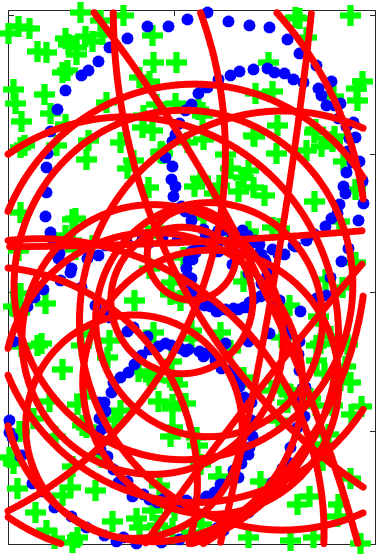}  &
\includegraphics[width=1.2cm, height=1.7cm, angle=0]{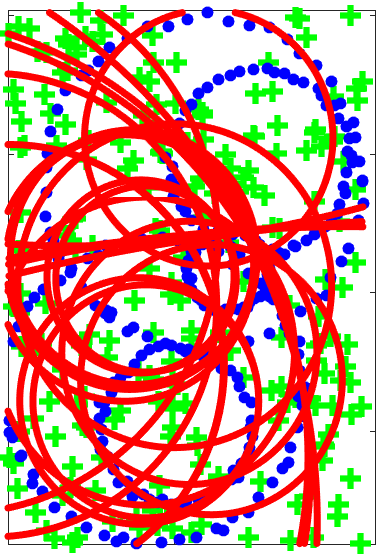}  &
\includegraphics[width=1.2cm, height=1.7cm, angle=0]{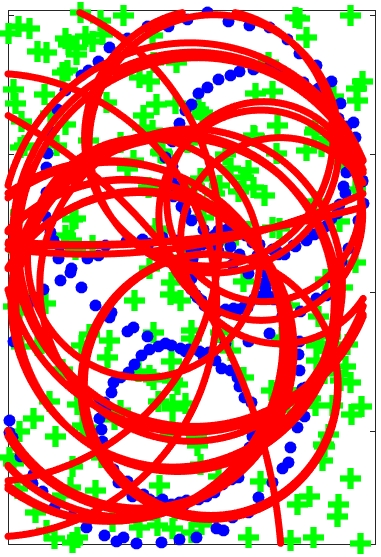} &
\includegraphics[width=1.2cm, height=1.7cm, angle=0]{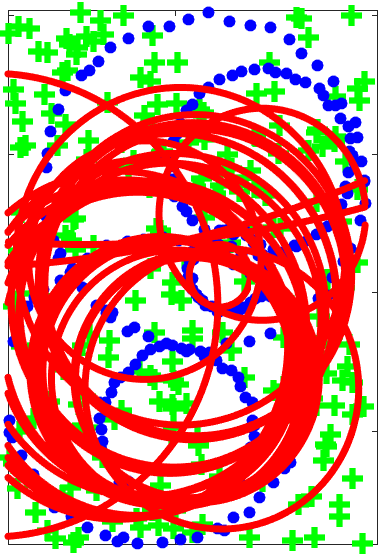}  &
\includegraphics[width=1.2cm, height=1.7cm, angle=0]{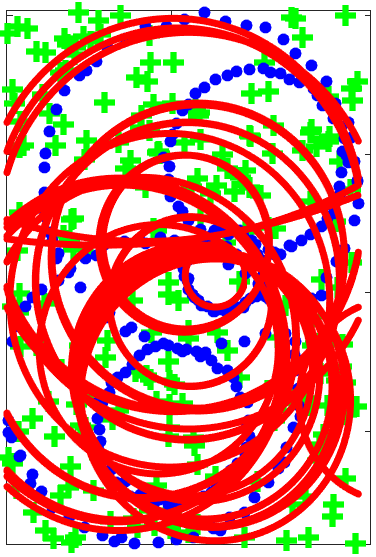}  &
\includegraphics[width=1.2cm, height=1.7cm, angle=0]{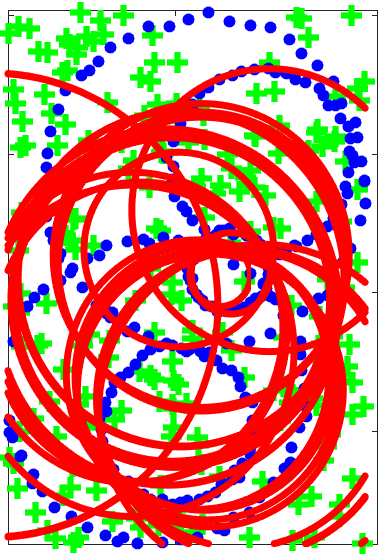}  &
\includegraphics[width=1.2cm, height=1.7cm, angle=0]{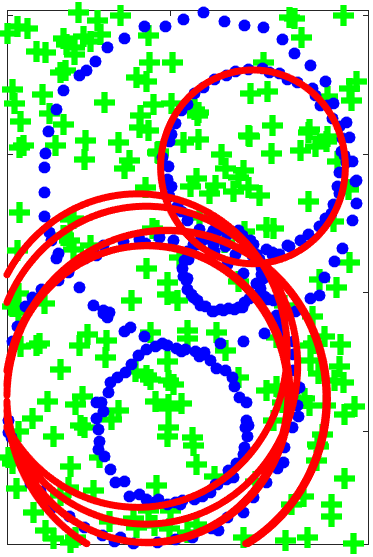}  &
\includegraphics[width=1.2cm, height=1.7cm, angle=0]{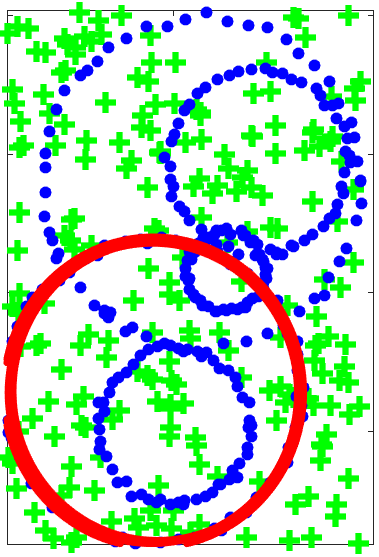}  &
\includegraphics[width=1.2cm, height=1.7cm, angle=0]{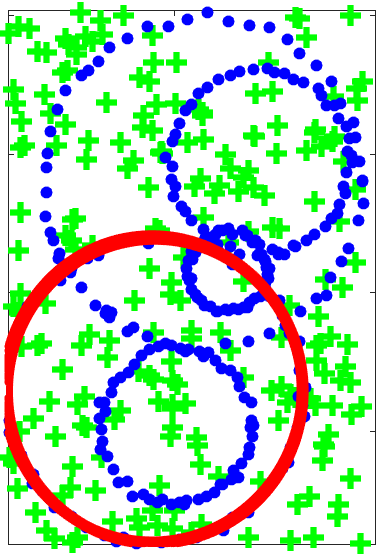} &
\includegraphics[width=1.2cm, height=1.7cm, angle=0]{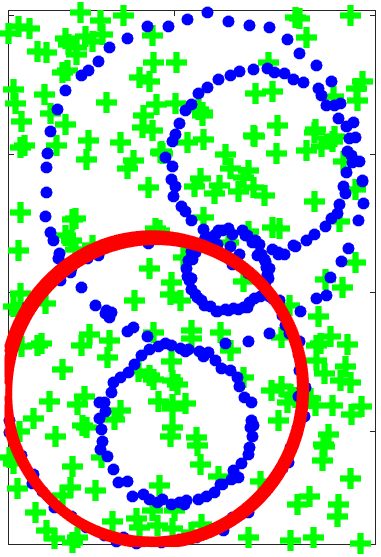}  &
\includegraphics[width=1.2cm, height=1.7cm, angle=0]{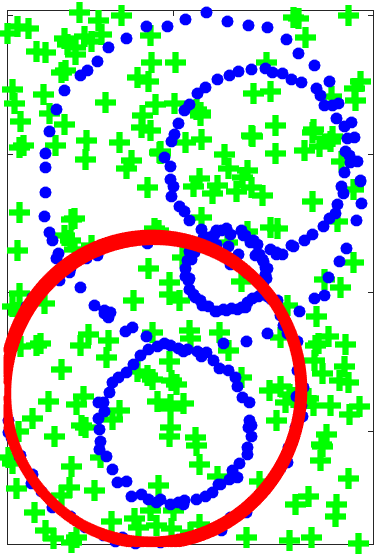}  &
\includegraphics[width=1.2cm, height=1.7cm, angle=0]{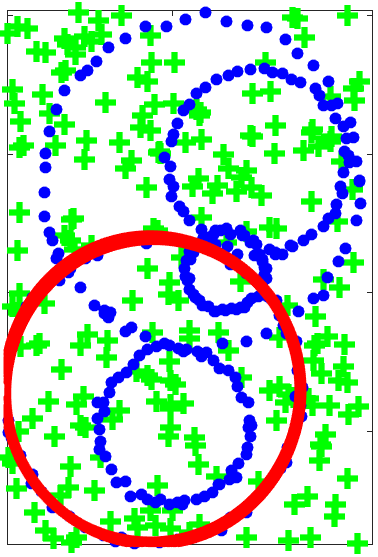} \\ 
\rotatebox{90}{\hspace{.2cm}  {st2, n2=55}}  \includegraphics[width=1.2cm, height=1.7cm, angle=0]{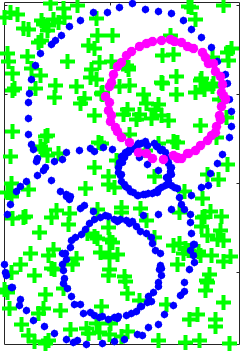} &
\includegraphics[width=1.2cm, height=1.7cm, angle=0]{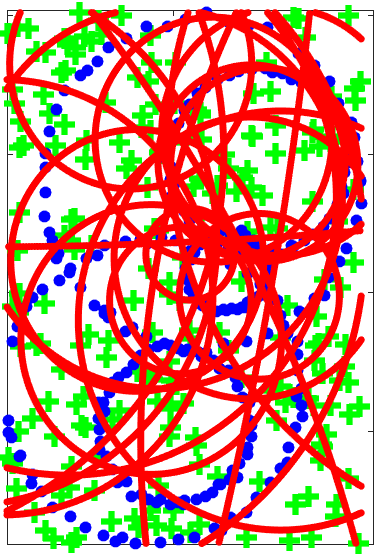}  &
\includegraphics[width=1.2cm, height=1.7cm, angle=0]{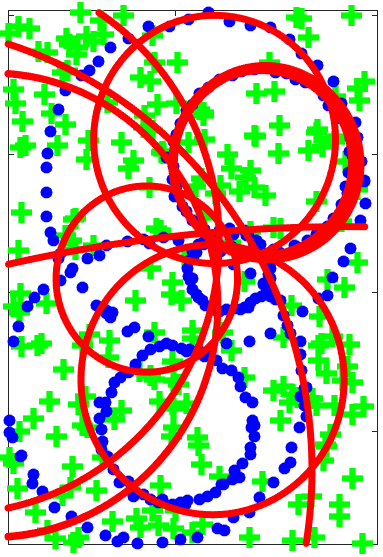}  &
\includegraphics[width=1.2cm, height=1.7cm, angle=0]{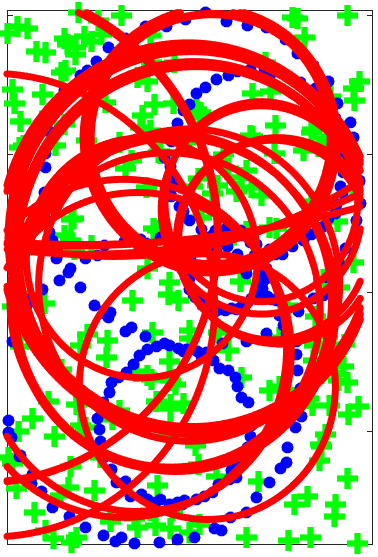} &
\includegraphics[width=1.2cm, height=1.7cm, angle=0]{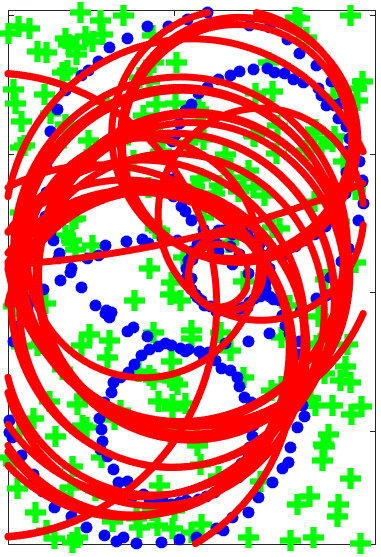}  &
\includegraphics[width=1.2cm, height=1.7cm, angle=0]{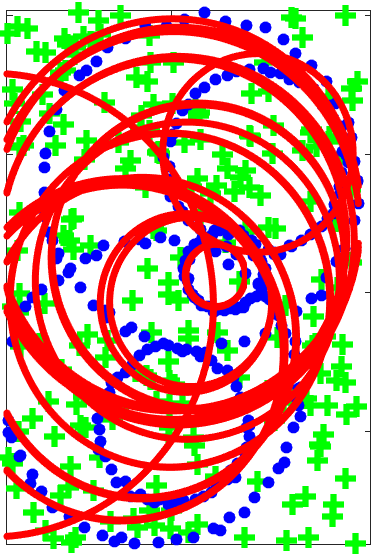}  &
\includegraphics[width=1.2cm, height=1.7cm, angle=0]{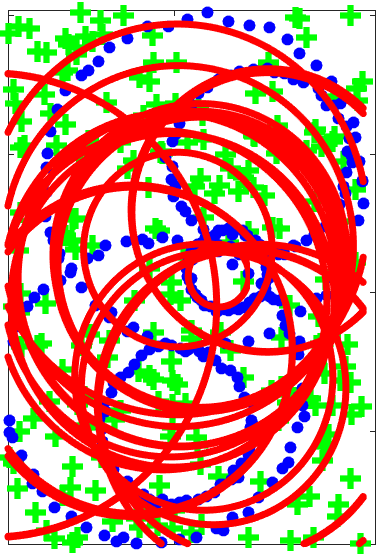}  &
\includegraphics[width=1.2cm, height=1.7cm, angle=0]{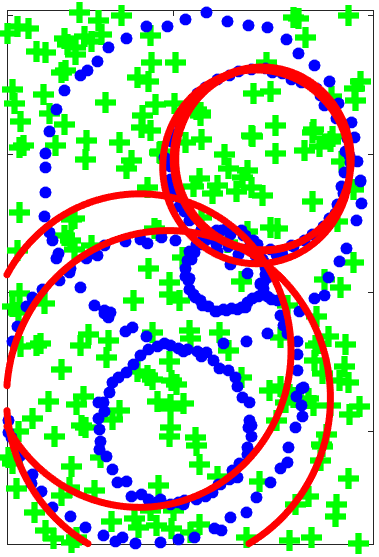}  &
\includegraphics[width=1.2cm, height=1.7cm, angle=0]{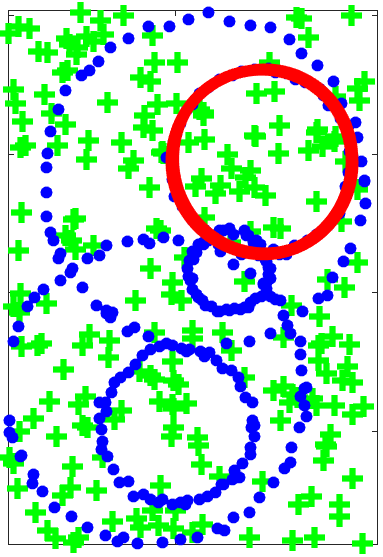}  &
\includegraphics[width=1.2cm, height=1.7cm, angle=0]{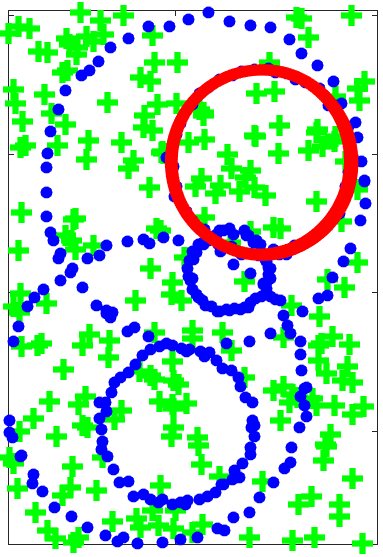} &
\includegraphics[width=1.2cm, height=1.7cm, angle=0]{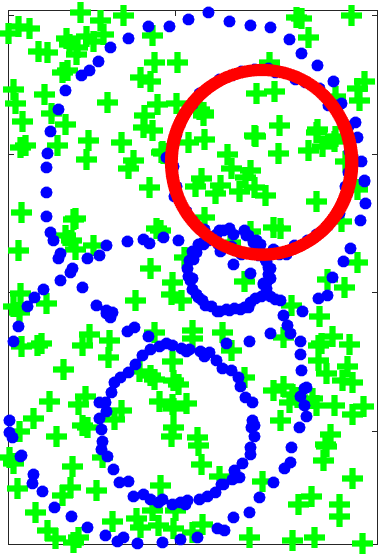}  &
\includegraphics[width=1.2cm, height=1.7cm, angle=0]{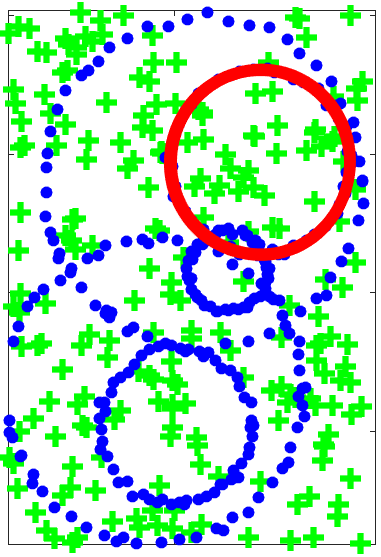}  &
\includegraphics[width=1.2cm, height=1.7cm, angle=0]{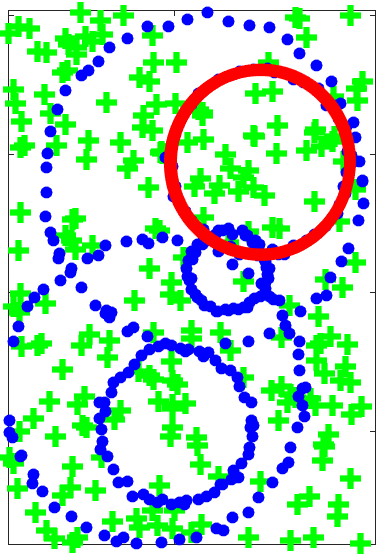} 
 \\ 
\rotatebox{90}{\hspace{.2cm}  {st3, n3=54}}  \includegraphics[width=1.2cm, height=1.7cm, angle=0]{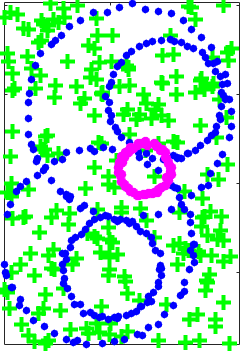} &
\includegraphics[width=1.2cm, height=1.7cm, angle=0]{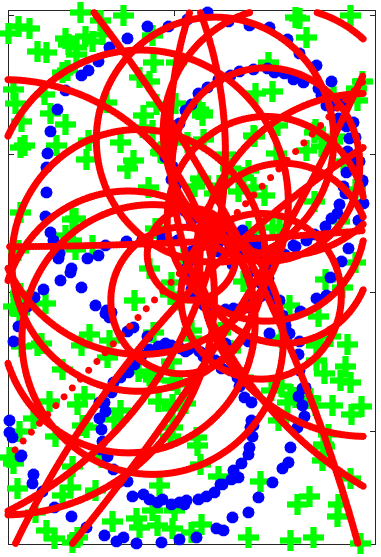}  &
\includegraphics[width=1.2cm, height=1.7cm, angle=0]{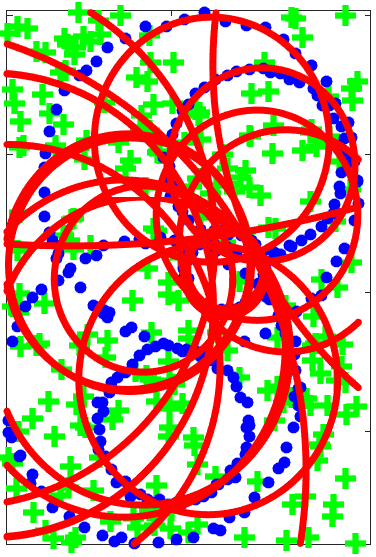}  &
\includegraphics[width=1.2cm, height=1.7cm, angle=0]{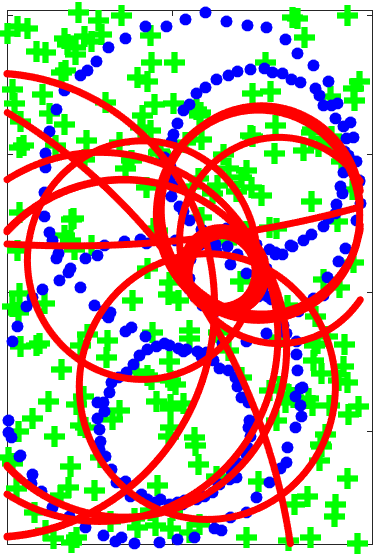} &
\includegraphics[width=1.2cm, height=1.7cm, angle=0]{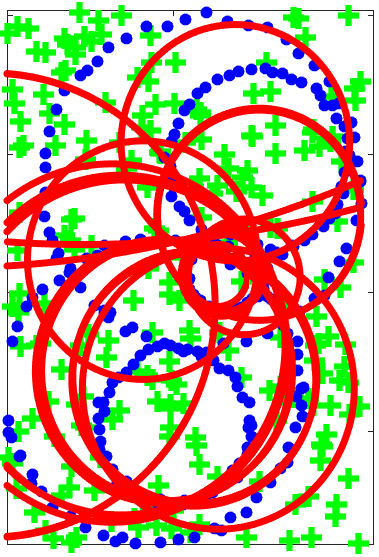}  &
\includegraphics[width=1.2cm, height=1.7cm, angle=0]{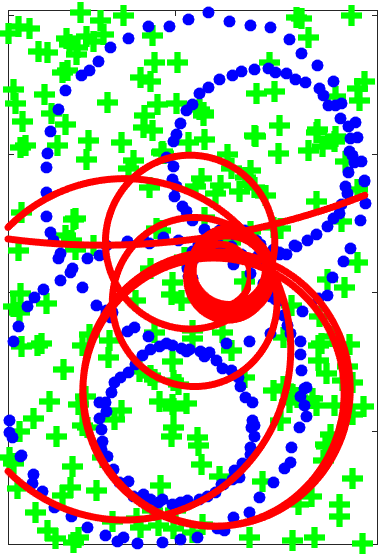}  &
\includegraphics[width=1.2cm, height=1.7cm, angle=0]{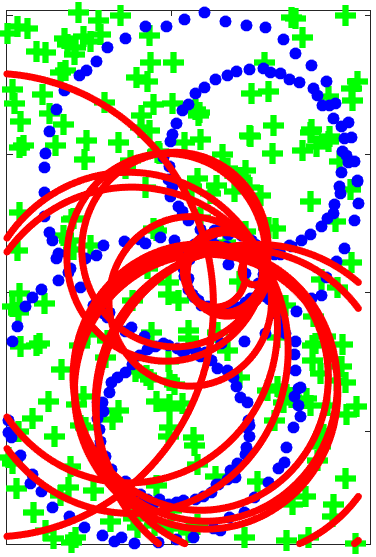}  &
\includegraphics[width=1.2cm, height=1.7cm, angle=0]{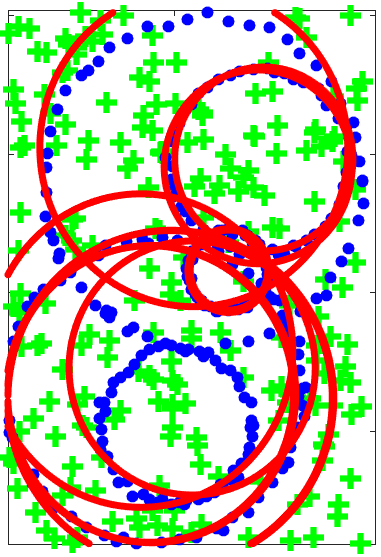}  &
\includegraphics[width=1.2cm, height=1.7cm, angle=0]{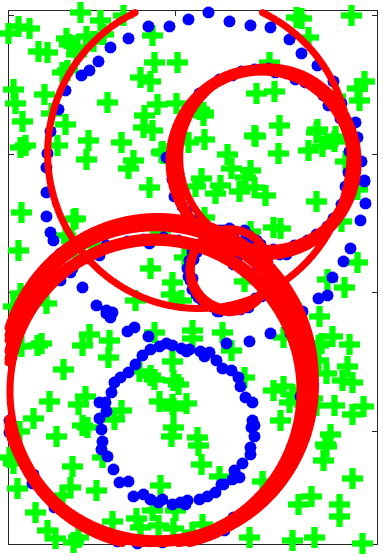}  &
\includegraphics[width=1.2cm, height=1.7cm, angle=0]{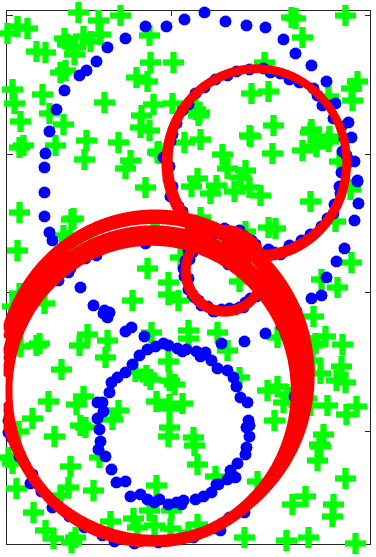} &
\includegraphics[width=1.2cm, height=1.7cm, angle=0]{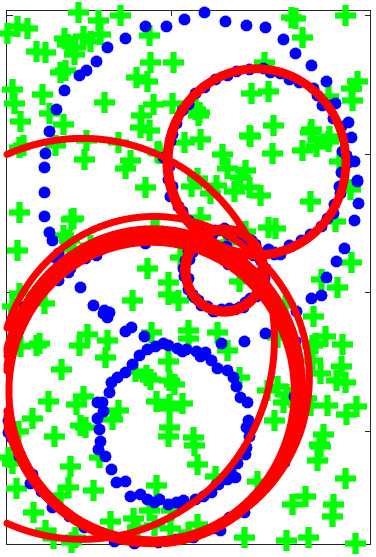}  &
\includegraphics[width=1.2cm, height=1.7cm, angle=0]{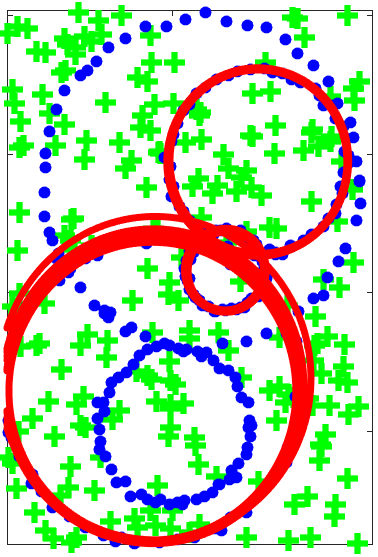}  &
\includegraphics[width=1.2cm, height=1.7cm, angle=0]{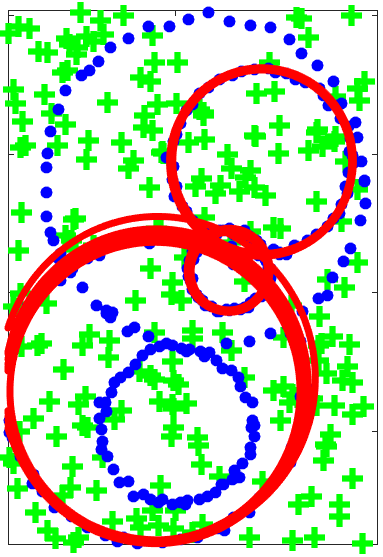}\\
\rotatebox{90}{\hspace{.2cm}  {st4, n4=57}}  \includegraphics[width=1.2cm, height=1.7cm, angle=0]{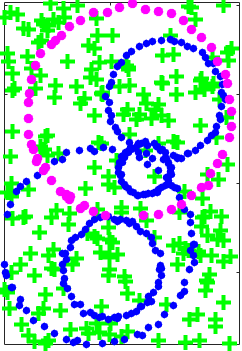} &
\includegraphics[width=1.2cm, height=1.7cm, angle=0]{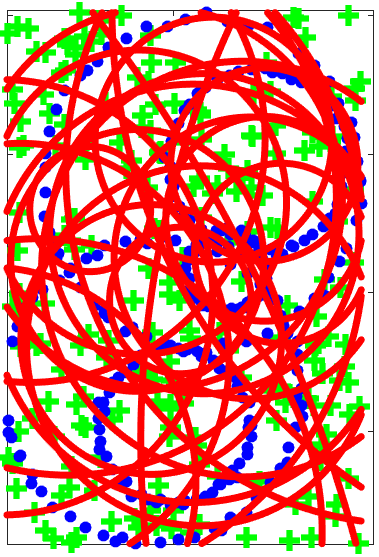}  &
\includegraphics[width=1.2cm, height=1.7cm, angle=0]{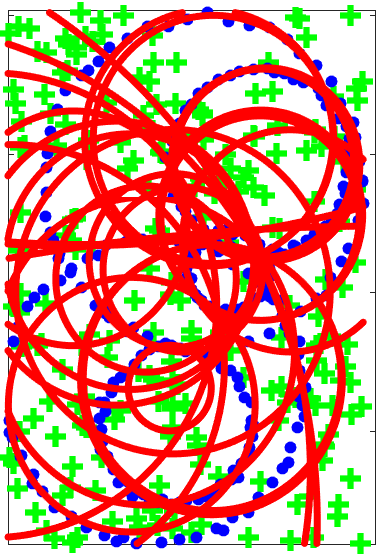}  &
\includegraphics[width=1.2cm, height=1.7cm, angle=0]{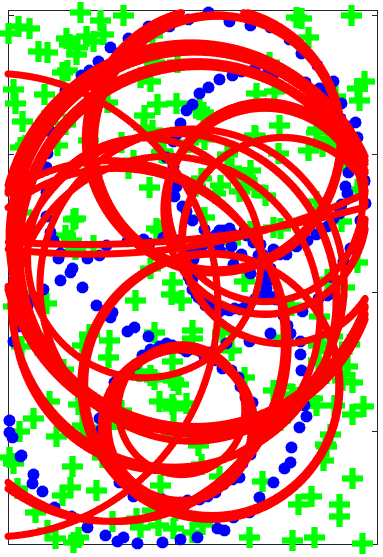} &
\includegraphics[width=1.2cm, height=1.7cm, angle=0]{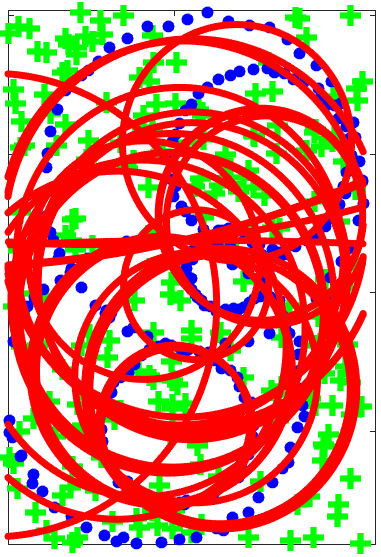}  &
\includegraphics[width=1.2cm, height=1.7cm, angle=0]{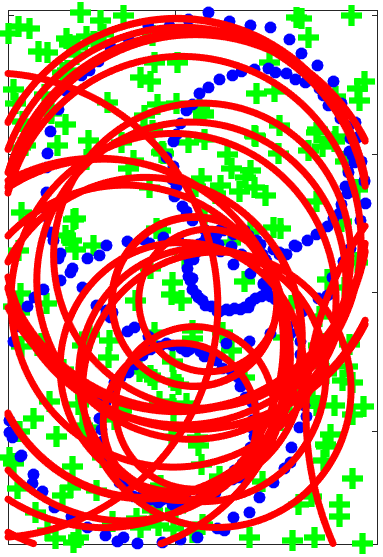}  &
\includegraphics[width=1.2cm, height=1.7cm, angle=0]{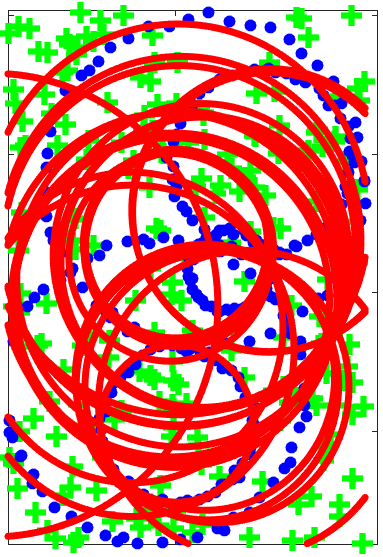}  &
\includegraphics[width=1.2cm, height=1.7cm, angle=0]{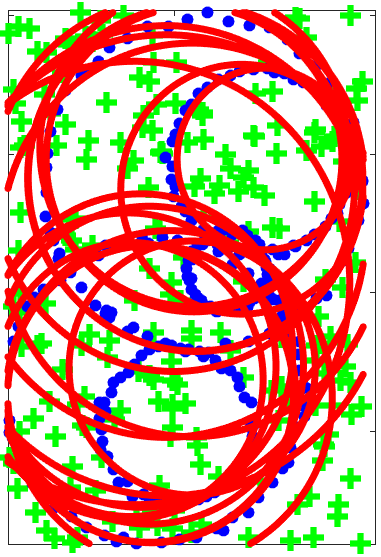}  &
\includegraphics[width=1.2cm, height=1.7cm, angle=0]{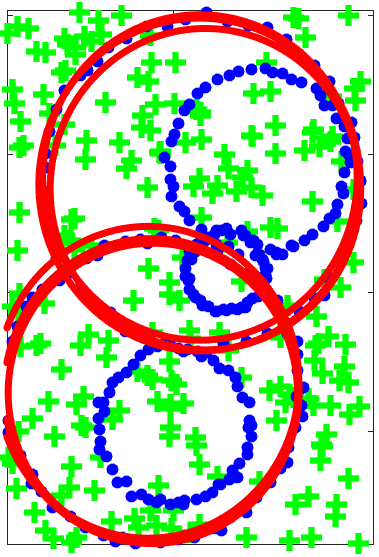}  &
\includegraphics[width=1.2cm, height=1.7cm, angle=0]{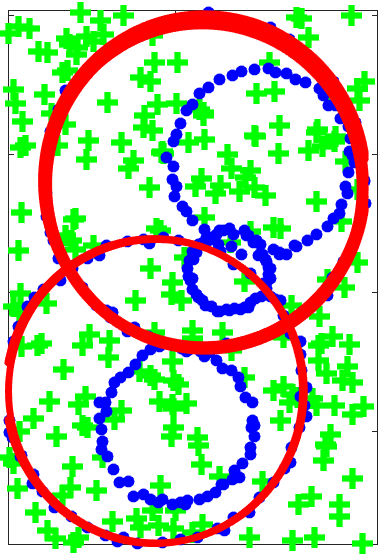} &
\includegraphics[width=1.2cm, height=1.7cm, angle=0]{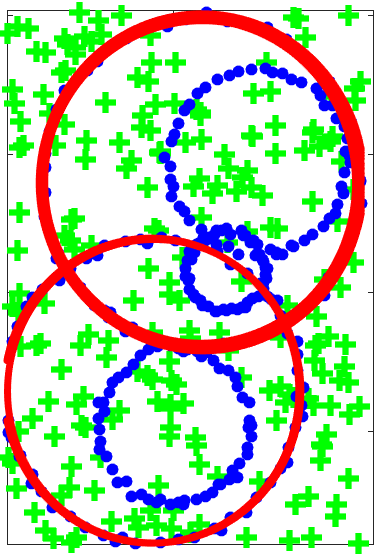}  &
\includegraphics[width=1.2cm, height=1.7cm, angle=0]{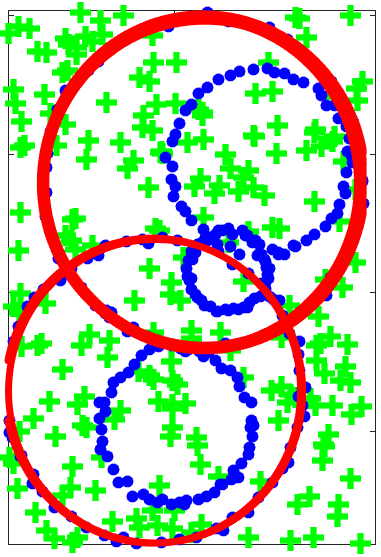}  &
\includegraphics[width=1.2cm, height=1.7cm, angle=0]{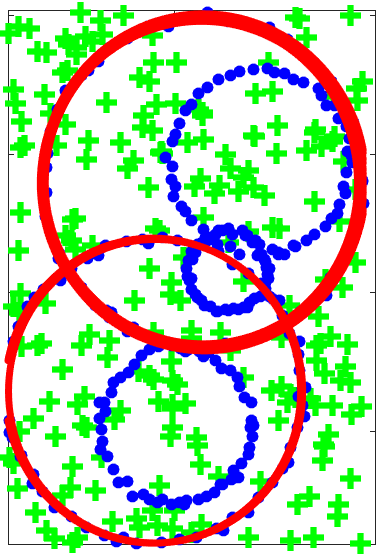}\\
\rotatebox{90}{\hspace{.2cm}  {st5, n5=57}}  \includegraphics[width=1.2cm, height=1.7cm, angle=0]{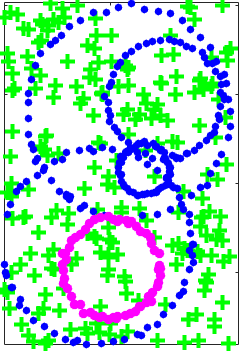} &
\includegraphics[width=1.2cm, height=1.7cm, angle=0]{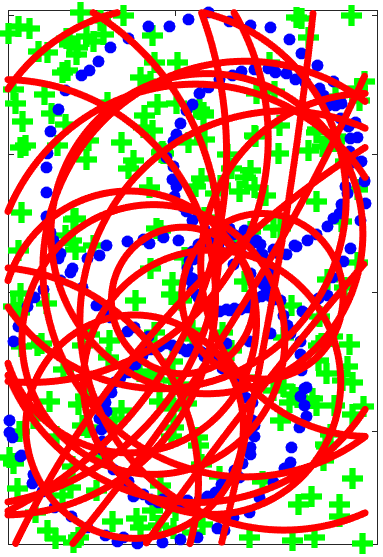}  &
\includegraphics[width=1.2cm, height=1.7cm, angle=0]{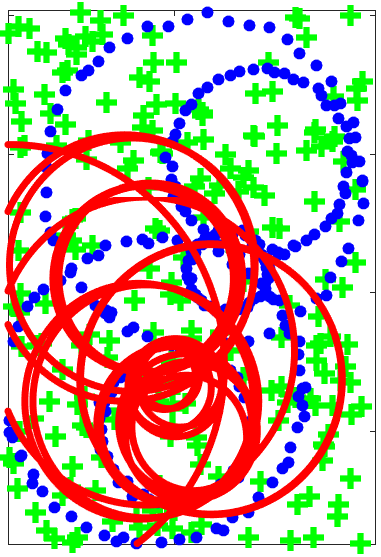}  &
\includegraphics[width=1.2cm, height=1.7cm, angle=0]{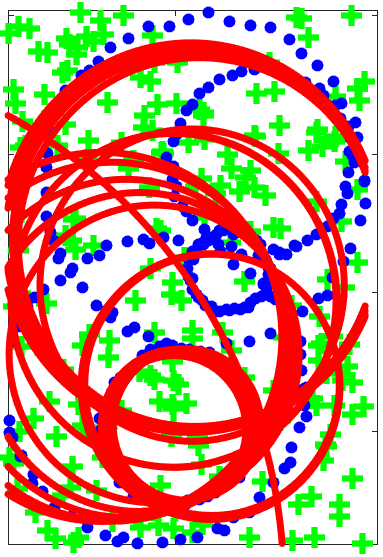} &
\includegraphics[width=1.2cm, height=1.7cm, angle=0]{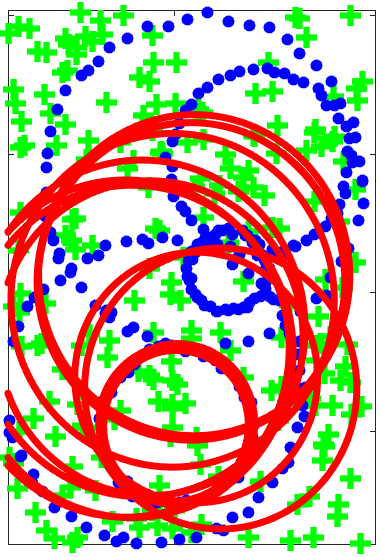}  &
\includegraphics[width=1.2cm, height=1.7cm, angle=0]{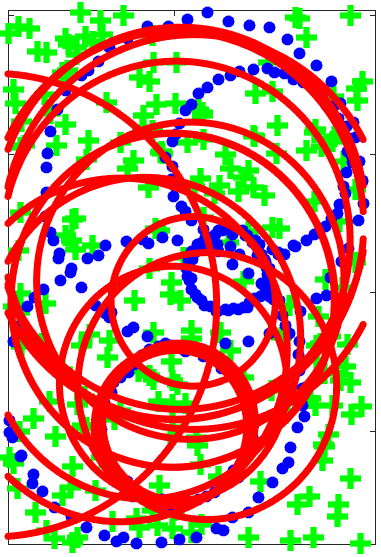}  &
\includegraphics[width=1.2cm, height=1.7cm, angle=0]{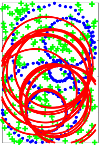}  &
\includegraphics[width=1.2cm, height=1.7cm, angle=0]{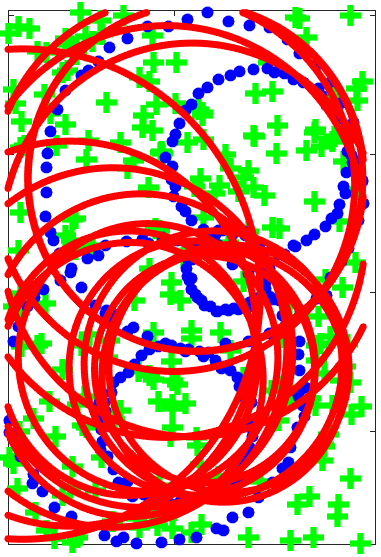}  &
\includegraphics[width=1.2cm, height=1.7cm, angle=0]{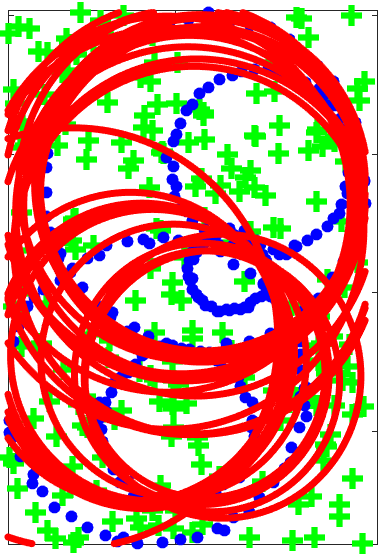}  &
\includegraphics[width=1.2cm, height=1.7cm, angle=0]{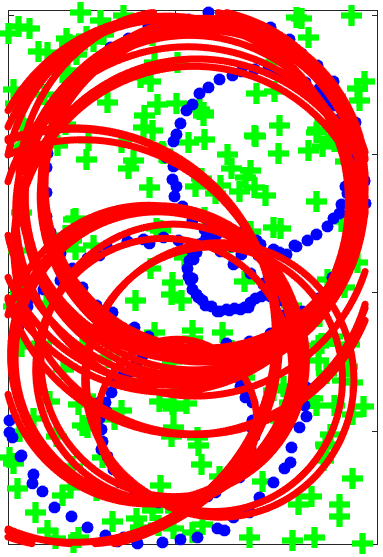} &
\includegraphics[width=1.2cm, height=1.7cm, angle=0]{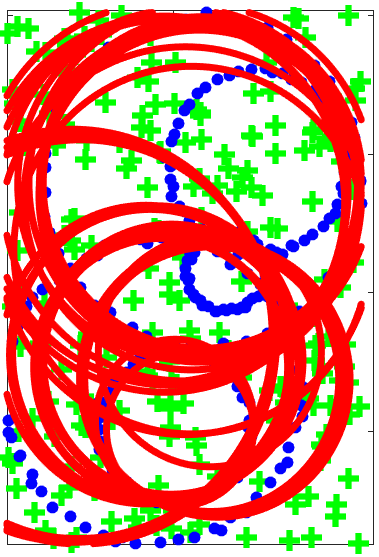}  &
\includegraphics[width=1.2cm, height=1.7cm, angle=0]{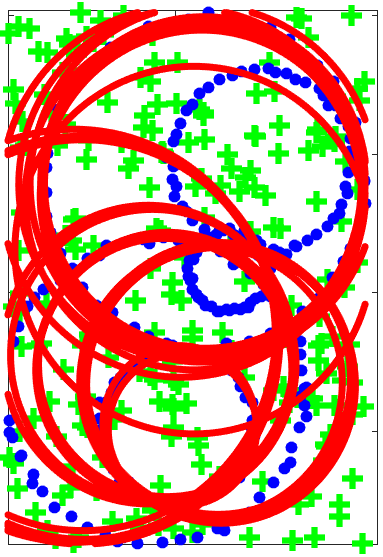}  &
\includegraphics[width=1.2cm, height=1.7cm, angle=0]{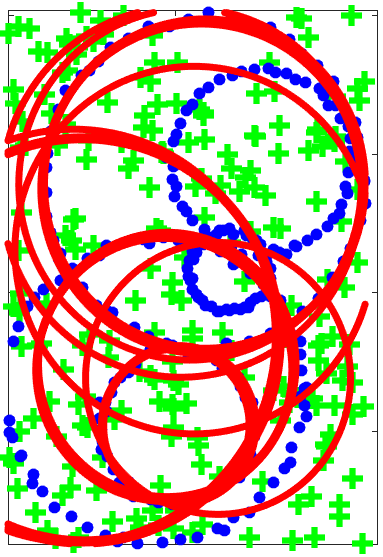}\\ \hline  
\end{tabular}
\label{tab:dhf-vs-kdgs}
\end{table*}

%% file: depndency_tble.tex
\begin{table}[h!]
\setlength{\tabcolsep}{2pt}
\fontsize{8}{9}\selectfont
\centering
\caption{{Required user inputs ($\epsilon$= inlier/outlier threshold,$\kappa$ = no. of structures): (\checkmark= Required, $\times$= Not Required)} . In addition to $\epsilon$, Prog-X require more user-defined thresholds, details are in \cite{barath2019progressive}.}
\begin{tabular}{cccccccccc}
&\textbf{RPA} &\textbf{Tlink}&\textbf{RCM}&\textbf{DPA}&\textbf{Cov}&\textbf{QP-MF}&\textbf{NMU}&\textbf{Prog-X}&\textbf{DGSAC-G/O} \\
                                                       \hline
$\epsilon$     & \checkmark   & \checkmark      & \checkmark     & \checkmark   & \checkmark & \checkmark         & \checkmark & \checkmark   &  $\times$     \\
$\kappa$ & \checkmark   & \checkmark      & $\times$      & $\times$    & \checkmark       &\checkmark       & $\times$ & $\times$    & $\times$     \\
\hline
\label{tbl:userip}
\end{tabular}
\end{table} 

%% file: motion_seg_table.tex
\begin{table*}[h!]
\centering
\setlength{\tabcolsep}{1.4pt}
\fontsize{7.5}{8}\selectfont
{
\caption{\label{tbl:fun}\textbf{Quantitative Analysis on Motion Segmentation.} Classification Accuracy(CA) in (\%), Total time taken including both KDGS and model selection in seconds, \textbf{O}(\%)= Outliers Percentage, $\kappa$ = number of true structures, \textbf{$\mu$}=mean, \textit{med}=median. * is the DGSAC version proposed in \cite{tiwari2018dgsac}.}}
\begin{tabular}{c|c|c|c|c|c|c|c|c|c|c|c|c|c|c|c|c|c|c|c||c|c|c|c|}
\cline{2-24}
& \rotatebox{69}{{biscuit}}
& \rotatebox{69}{{biscuitbook}}
& \rotatebox{69}{{bisbookbox}} 
& \rotatebox{69}{{boardgame}}
& \rotatebox{62}{{book}}
& \rotatebox{75}{{brdcartoychips}}
& \rotatebox{69}{{breadcube}}
& \rotatebox{69}{{brdcubechips}}
& \rotatebox{69}{{breadtoy}}
& \rotatebox{69}{{breadtoycar}}
& \rotatebox{69}{{carchipscube}}
& \rotatebox{62}{{cube}}
& \rotatebox{75}{{cubebrdtoychips}}
& \rotatebox{69}{{cubechips}}
& \rotatebox{69}{{cubetoy}} 
& \rotatebox{69}{{dinobooks}} 
& \rotatebox{62}{{game}}
& \rotatebox{69}{{gamebiscuit}}
& \rotatebox{69}{{toycubecar}} 
& \multicolumn{2}{c|}{\textbf{CA(\%)}}                 
& \multicolumn{2}{c|}{\textbf{Time(s)}}                 \\ \hline
\multicolumn{1}{|c|}{}   &   &
&
&
&
&
&
&
&
&
&
&
&
&
&
&
&
&
&
&                   
&                  
&                   
& \vspace{-0.25cm}\\ 

\multicolumn{1}{|c|}{$n$} & 
319&	341&	258&	266&	185&	231&	233&	230&	278&	164&	164&	295&	314&	277&	239&	339&	230&	324&	198&
\multirow{2}{*}{\textbf{$\mu$}} & 
\multirow{2}{*}{\textit{med}} & 
\multirow{2}{*}{\textbf{$\mu$}} & 
\multirow{2}{*}{\textit{med}}         \\ 
\multicolumn{1}{|c|}{\textbf{O}(\%)} & 57.2 &
47.5&
37.2&
42.5&
21.5&
35.2&
32.2&
35.2&
37.4&
34.2&
36.6&
69.5&
28.0&
51.6&
41.4&
44.5&
73.5&
51.5&
36.4&
&
&
&\\
\multicolumn{1}{|c|}{$\kappa$}   &  1 &
2&
3&
3&
1&
4&
2&
3&
2&
3&
3&
1&
4&
2&
2&
3&
1&
2&
3&                   
&                  
&                   
&                   \\ \hline 
\multicolumn{1}{|c|}{}   &   &
&
&
&
&
&
&
&
&
&
&
&
&
&
&
&
&
&
&                   
&                  
&                   
& \vspace{-0.18cm}\\ 
\multicolumn{1}{|l|}{\textbf{Tlink}}     &  83.1 &
97.8 &
88.8 &
83.7&
82.6&
80.5&
85.6&
82.0&
96.8&
84.7&
88.0&
46.3&
80.2&
95.1&
78.8&
78.6&
77.6&
70.6&
70.7&
81.7&
82.6&
12.8&
11.7  \\
\multicolumn{1}{|l|}{}   &   &
&
&
&
&
&
&
&
&
&
&
&
&
&
&
&
&
&
&                   
&                  
&                   
& \vspace{-0.18cm}\\   
\multicolumn{1}{|l|}{\textbf{RCM}}      & 95.2 &
92.5&
83.7&
78.5&
94.0&
78.8&
87.3&
83.2&
78.4&
83.1&
78.9&
87.9&
81.6&
90.3&
89.6&
72.3&
90.8&
85.4&
83.5& 
85.0&
83.7&
04.6&
03.8  \\ 
\multicolumn{1}{|l|}{}   &   &
&
&
&
&
&
&
&
&
&
&
&
&
&
&
&
&
&
&                   
&                  
&                   
& \vspace{-0.18cm}\\ 
\multicolumn{1}{|l|}{\textbf{RPA}}      & 
98.4&
96.4&
95.8&
87.5&
97.5&
91.7&
96.0&
95.6&
97.2&
92.2&
94.3&
97.2&
93.2&
96.5&
96.3&
84.8&
95.9&
96.9&
91.7&
94.5&
95.9&
39.3&
38.8  \\ 
\multicolumn{1}{|l|}{}   &   &
&
&
&
&
&
&
&
&
&
&
&
&
&
&
&
&
&
&                   
&                  
&                   
& \vspace{-0.18cm}\\ 
\multicolumn{1}{|l|}{\textbf{DPA}}      & 82.1 &
97.2&
95.1&
83.7&
90.2&
91.6&
94.1&
94.6&
90.6&
88.7&
86.3&
96.9&
87.3&
92.9&
93.6&
84.2&
97.5&
90.9&
85.6&
90.6&
90.9&
50.3&
46.8  \\ 
\multicolumn{1}{|l|}{}   &   &
&
&
&
&
&
&
&
&
&
&
&
&
&
&
&
&
&
&                   
&                  
&                   
& \vspace{-0.18cm}\\ 
\multicolumn{1}{|l|}{\textbf{Cov}}      &  
98.4&
97.6&
94.0&
77.8&
97.2&
87.3&
95.9&
88.6&
82.4&
89.2&
88.7&
97.1&
90.7&
93.6&
95.5&
68.7&
92.4&
95.5&
82.1&
90.1&
92.4&
54.7&
47.3  \\ 
\multicolumn{1}{|c|}{}   &   &
&
&
&
&
&
&
&
&
&
&
&
&
&
&
&
&
&
&                   
&                  
&                   
& \vspace{-0.18cm}\\ 
\multicolumn{1}{|l|}{\textbf{NMU}}      &  97.6 &
98.8&
98.1&
82.8&
100&
94.9&
97.1&
97.4&
97.9&
92.2&
97.6&
98.0&
87.2&
98.6&
98.0&
84.4&
98.7&
92.1&
91.5&
94.9&
97.6&
399&
399  \\ 
\multicolumn{1}{|l|}{}   &   &
&
&
&
&
&
&
&
&
&
&
&
&
&
&
&
&
&
&                   
&                  
&                   
& \vspace{-0.18cm}\\ 
\multicolumn{1}{|l|}{\textbf{QP-MF}}    &  
55.8&
52.5&
62.5&
64.4&
56.2&
65.4&
68.2&
64.8&
63.2&
66.3&
67.9&
67.9&
73.1&
60.9&
60.2&
56.9&
73.0&
60.7&
66.5&
63.5&
64.4&
20.3&
20.2
  \\ 
\multicolumn{1}{|l|}{}   &   &
&
&
&
&
&
&
&
&
&
&
&
&
&
&
&
&
&
&                   
&                  
&                   
& \vspace{-0.18cm}\\ 
\cdashline{1-24}
\multicolumn{1}{|l|}{}   &   &
&
&
&
&
&
&
&
&
&
&
&
&
&
&
&
&
&
&                   
&                  
&                   
& \vspace{-0.18cm}\\ 
\multicolumn{1}{|l|}{\textbf{DGSAC{*}}}    &  
98.2&
98.6&
97.6&
82.6&
99.0&
87.9&
97.7&
93.0&
90.7&
89.5&
85.5&
96.8&
88.6&
97.4&
97.3&
83.9&
95.0&
98.2&
90.4&
93.1&
95.0&
24.2&
20.2
  \\ 
\multicolumn{1}{|l|}{}   &   &
&
&
&
&
&
&
&
&
&
&
&
&
&
&
&
&
&
&                   
&                  
&                   
& \vspace{-0.18cm}\\ 
\multicolumn{1}{|c|}{\textbf{DGSAC-G}}    &  
98.1&	98.5&	97.3&	89.5&	99.3&	89.0&	96.7&	97.7&	96.1&	89.9&	88.9&	95.9&	91.9&	97.3&	98.03&	86.5&	97.5&	99.0&	91.9&	94.7& 96.7& 5.7&	5.4
  \\
\multicolumn{1}{|l|}{}   &   &
&
&
&
&
&
&
&
&
&
&
&
&
&
&
&
&
&
&                   
&                  
&                   
& \vspace{-0.18cm}\\ 
\multicolumn{1}{|c|}{\textbf{DGSAC-O}}    &  
88.3&	94.8&	98.1&	83.3&	89.3&	85.3&	89.4&	97.8&	97.7&	93.3&	86.6&	88.4&	94.0&	97.4&	96.9&	85.4&	98.7&	93.8&	89.9&	92.1&	93.3&5.3 &	4.9
  \\ \hline
\end{tabular}
\end{table*}

%% file: planar_seg_table.tex
\begin{table*}[h!]
\centering
\setlength{\tabcolsep}{1.4pt}
\fontsize{7.5}{8}\selectfont
{\small
\caption{ \label{tbl:hom} \textbf{Quantitative Analysis on Planar Segmentation.} Notations are same as of table \ref{tbl:fun}.  }}
\begin{tabular}{c|c|c|c|c|c|c|c|c|c|c|c|c|c|c|c|c|c|c|c||c|c|c|c|}
\cline{2-24}
& \rotatebox{63}{{barrsmith}}
& \rotatebox{63}{{bonhall}}
& \rotatebox{63}{{bonython}} 
& \rotatebox{63}{{elderhalla}}
& \rotatebox{63}{{elderhallb}}
& \rotatebox{63}{{hartley}}
& \rotatebox{63}{{johnsona}}
& \rotatebox{63}{{johnsonb}}
& \rotatebox{63}{{ladysymon}}
& \rotatebox{63}{{library}}
& \rotatebox{63}{{napiera}}
& \rotatebox{63}{{napierb}}
& \rotatebox{63}{{neem}}
& \rotatebox{63}{{nese}}
& \rotatebox{63}{{oldclassics}}
& \rotatebox{63}{{physics}}
& \rotatebox{63}{{sene}}
& \rotatebox{63}{{unihouse}}
& \rotatebox{63}{{unionhouse}} 
& \multicolumn{2}{c|}{\textbf{CA(\%)}}                 
& \multicolumn{2}{c|}{\textbf{Time(s)}}                 \\ \hline
\multicolumn{1}{|c|}{}   &   &
&
&
&
&
&
&
&
&
&
&
&
&
&
&
&
&
&
&                   
&                  
&                   
& \vspace{-0.18cm}\\ 
\multicolumn{1}{|c|}{$n$} & 235&	948&	193&	214&	245&	315&	353&	624&	227&	212&	292&	237&	230&	241&	363&	103&	236&	1784&	321& 
\multirow{2}{*}{\textbf{$\mu$}} & 
\multirow{2}{*}{\textit{med}} & 
\multirow{2}{*}{\textbf{$\mu$}} & 
\multirow{2}{*}{\textit{med}}         \\ 
\multicolumn{1}{|c|}{\textbf{O}(\%)}   & 68.9     & 06.2   & 73.7    & 60.7      & 47.8      & 61.6   & 20.9    & 12.0    & 32.5 & 55.3   & 62.9   & 39.5   & 36.5 & 33.5 & 32.5 & 45.3   & 47.2 & 16.6    & 76.5&                   
&                  
&                   
&                   \\ 
\multicolumn{1}{|c|}{$\kappa$}   & 2         & 6       & 1        & 2          & 3          & 2       & 4        & 7        & 2     & 2       & 2       & 3       & 3     & 2     & 2     & 1       & 2     & 5        & 1&                   
&                  
&                   
&                   \\
\hline 
\multicolumn{1}{|c|}{}   &   &
&
&
&
&
&
&
&
&
&
&
&
&
&
&
&
&
&
&                   
&                  
&                   
& \vspace{-0.18cm}\\ 
\multicolumn{1}{|l|}{\textbf{Tlink}}     & 57.9       & 60.4     & 64.3     &  69.5  &  57.8     &  71.6       &  57.8         & 70.7    &77.7      & 82.5      & 81.3        &  67.7       &  53.0       & 53.7      & 73.8      & 68.5            &  84.3     &  71.9        &  77.3 &
69.0&
70.7&
492&
81.3  \\
\multicolumn{1}{|l|}{}   &   &
&
&
&
&
&
&
&
&
&
&
&
&
&
&
&
&
&
&                   
&                  
&                   
& \vspace{-0.18cm}\\   

\multicolumn{1}{|l|}{\textbf{RCM}}  & 84.8&
81.7&  
87.3&  
75.2&
71.5&
77.4&
83.0&
79.4&
75.3&
77.0&
70.7&
74.3&
71.9&
77.6&
92.5&
54.5 &
71.7 &
97.0 &
90.1 &
78.6 &
77.4&
5.3&
3.4       \\ 
\multicolumn{1}{|l|}{}   &   &
&
&
&
&
&
&
&
&
&
&
&
&
&
&
&
&
&
&                   
&                  
&                   
& \vspace{-0.18cm}\\   

\multicolumn{1}{|l|}{\textbf{RPA}}      & 
62.9& 
52.9&
84.3&
99.1&
82.0&
81.4&
91.1&
66.8&
79.2&
63.5&
73.3&
75.1&
78.5&
99.2&
76.7&
100&
99.4&
88.0&
76.1& 
80.5&
79.2&
967&
247      \\ 
\multicolumn{1}{|l|}{}   &   &
&
&
&
&
&
&
&
&
&
&
&
&
&
&
&
&
&
&                   
&                  
&                   
& \vspace{-0.18cm}\\   

\multicolumn{1}{|l|}{\textbf{DPA}}      &  97.7 &
78.0& 
96.6&
96.2&
85.9&
96.9&
87.1&
74.4&
90.5&
95.2&
80.6&
83.6&
80.2&
97.4&
96.3&
98.4&
99.8&
93.2&
98.3&
90.9&
95.2&
37.7&
30.1       \\ 
\multicolumn{1}{|l|}{}   &   &
&
&
&
&
&
&
&
&
&
&
&
&
&
&
&
&
&
&                   
&                  
&                   
& \vspace{-0.18cm}\\   

\multicolumn{1}{|l|}{\textbf{Cov}}      &    70.7 &
68.6& 
99.7&
77.9&
82.8&
91.8&
86.1&
65.2&
93.8&
92.9&
86.6&
74.1&
72.6&
90.8&
79.2&
99.5&
80.4&
91.2&
99.5&
84.2&
82.8&
145&
53.2      \\ 
\multicolumn{1}{|l|}{}   &   &
&
&
&
&
&
&
&
&
&
&
&
&
&
&
&
&
&
&                   
&                  
&                   
& \vspace{-0.18cm}\\   

\multicolumn{1}{|l|}{\textbf{NMU}}      &   
89.6& 
84.3&
98.5&
98.1&
86.7&
98.4&
90.6&
75.0&
96.2&
98.1&
94.7&
78.4&
95.9&
97.6&
98.4&
79.3&
99.6&
94.7&
99.2 &
92.3&
96.0&
499&
298      \\ 
\multicolumn{1}{|l|}{}   &   &
&
&
&
&
&
&
&
&
&
&
&
&
&
&
&
&
&
&                   
&                  
&                   
& \vspace{-0.18cm}\\  
\multicolumn{1}{|l|}{\textbf{Prog-X}}    &  
88.4&
74.4&
98.3&
78.9&
80.9&
96.9&
91.5&
82.9&
96.2&
97.3&
87.1&
79.2&
74.0&
96.4&
97.7&
65.0&
97.7&
97.4&
98.7&
88.4&
88.7&
1.53&
1.48
  \\ 
  \multicolumn{1}{|l|}{}   &   &
&
&
&
&
&
&
&
&
&
&
&
&
&
&
&
&
&
&                   
&                  
&                   
& \vspace{-0.18cm}\\ 
\multicolumn{1}{|l|}{\textbf{QP-MF}}    &  
63.9&
71.6&
73.7&
72.9&
82.4&
55.9&
62.2&
61.0&
60.3&
55.4&
54.0&
73.8&
78.8&
66.5&
77.8&
54.7&
79.6&
80.4&
76.5&
68.5&
71.6&
25.2&
20.1
  \\ 
  \multicolumn{1}{|l|}{}   &   &
&
&
&
&
&
&
&
&
&
&
&
&
&
&
&
&
&
&                   
&                  
&                   
& \vspace{-0.18cm}\\ 
\cdashline{1-24}
  \multicolumn{1}{|l|}{}   &   &
&
&
&
&
&
&
&
&
&
&
&
&
&
&
&
&
&
&                   
&                  
&                   
& \vspace{-0.18cm}\\ 
\multicolumn{1}{|l|}{\textbf{DGSAC{*}}}    &   69.6 &
73.0&
98.2&
96.8&
88.3&
97.8&
94.9&
77.5&
91.9&
94.2&
92.8&
82.6&
90.6&
99.2&
94.2&
99.4&
98.6&
92.9&
97.1&
91.0&
94.2&
115&
23.2
  \\ 
  \multicolumn{1}{|c|}{}   &   &
&
&
&
&
&
&
&
&
&
&
&
&
&
&
&
&
&
&                   
&                  
&                   
& \vspace{-0.18cm}\\   
\multicolumn{1}{|c|}{\textbf{DGSAC-G}}    &  

89.4 &	73.4&	99.0&	99.1&	87.6&	97.2&	94.3&	77.6&	96.5&	97.7&	93.8&	83.5&	88.9&	99.6&	94.2&	100&	99.2&	92.7&	98.8&	92.8&	94.3& 23.8	&3.5
  \\ 
  \multicolumn{1}{|c|}{}   &   &
&
&
&
&
&
&
&
&
&
&
&
&
&
&
&
&
&
&                   
&                  
&                   
& \vspace{-0.18cm}\\   
 \multicolumn{1}{|c|}{\textbf{DGSAC-O}}    &   
 89.4&	72.8&	98.4&	99.1&	87.6&	97.5&	94.3&	77.9&	87.4&	98.6&	89.3&	83.5&	80.4&	99.6&	91.8&	100&	99.2&	92.1&	98.8&	91.5&	92.1& 24.9&	3.4
  \\ \hline
\end{tabular}
\end{table*}

%% file: qual_fun.tex
\begin{figure*}[h!]
\centering
\setlength{\tabcolsep}{2pt}
\begin{tabular}{cccccc}
 & biscuitbook & biscuitboobox & breadcubechips & breadtoy & gamebiscuit \\
 \rotatebox{90}{\hspace{.2cm}  \textbf{GroundTruth}} &    \includegraphics[width=2.6cm]{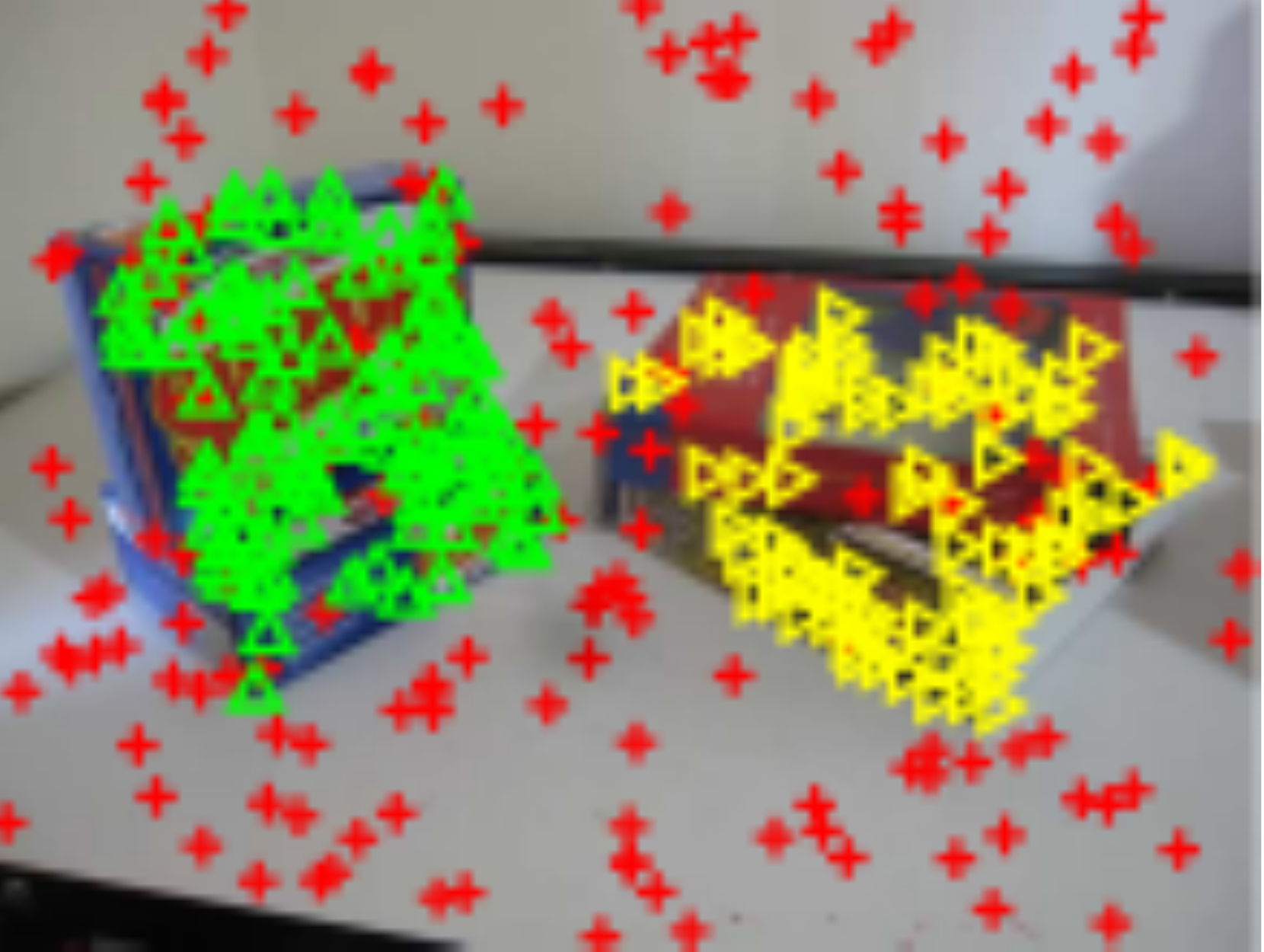}&
     \includegraphics[width=2.6cm]{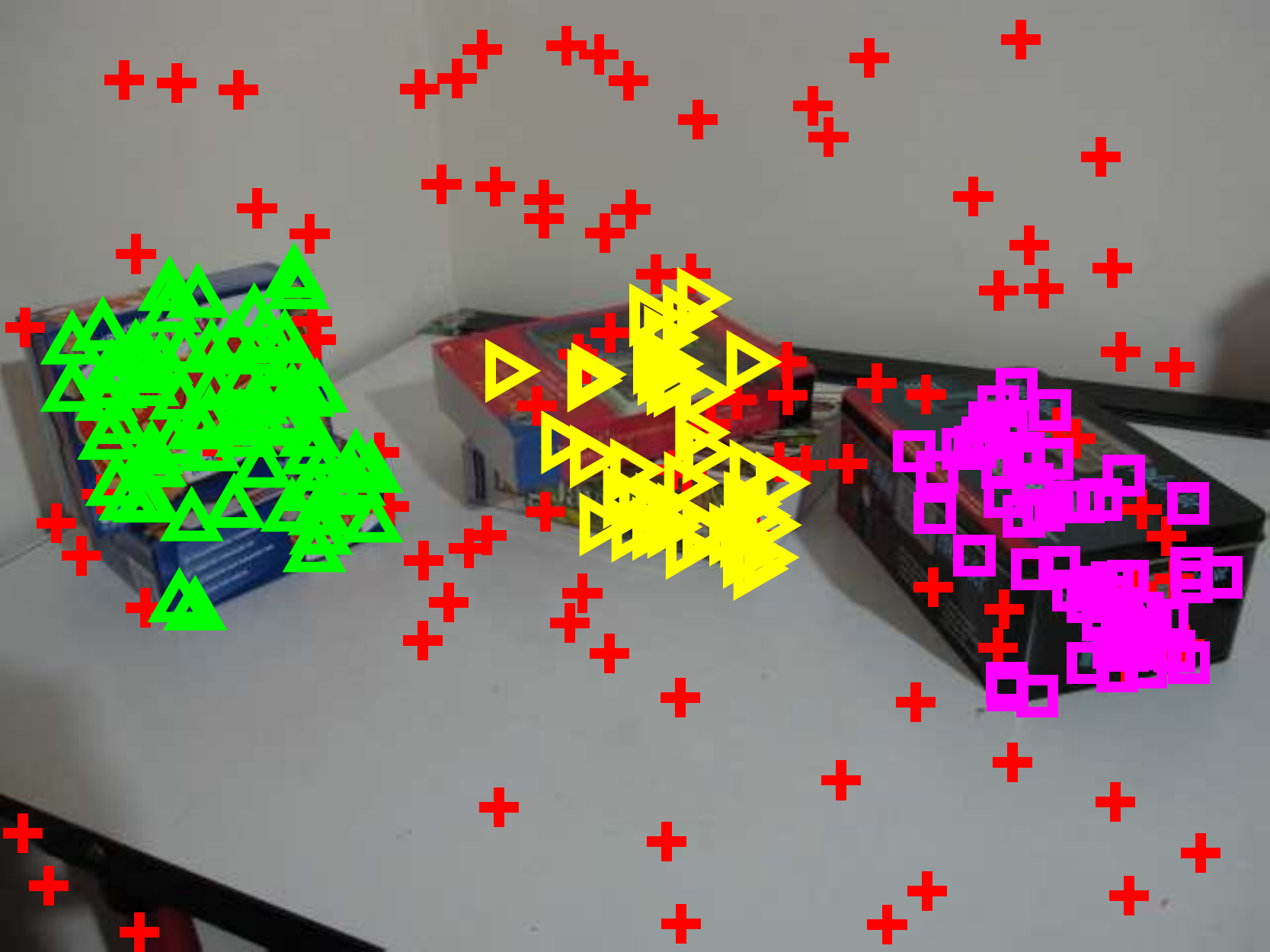}&
     \includegraphics[width=2.6cm]{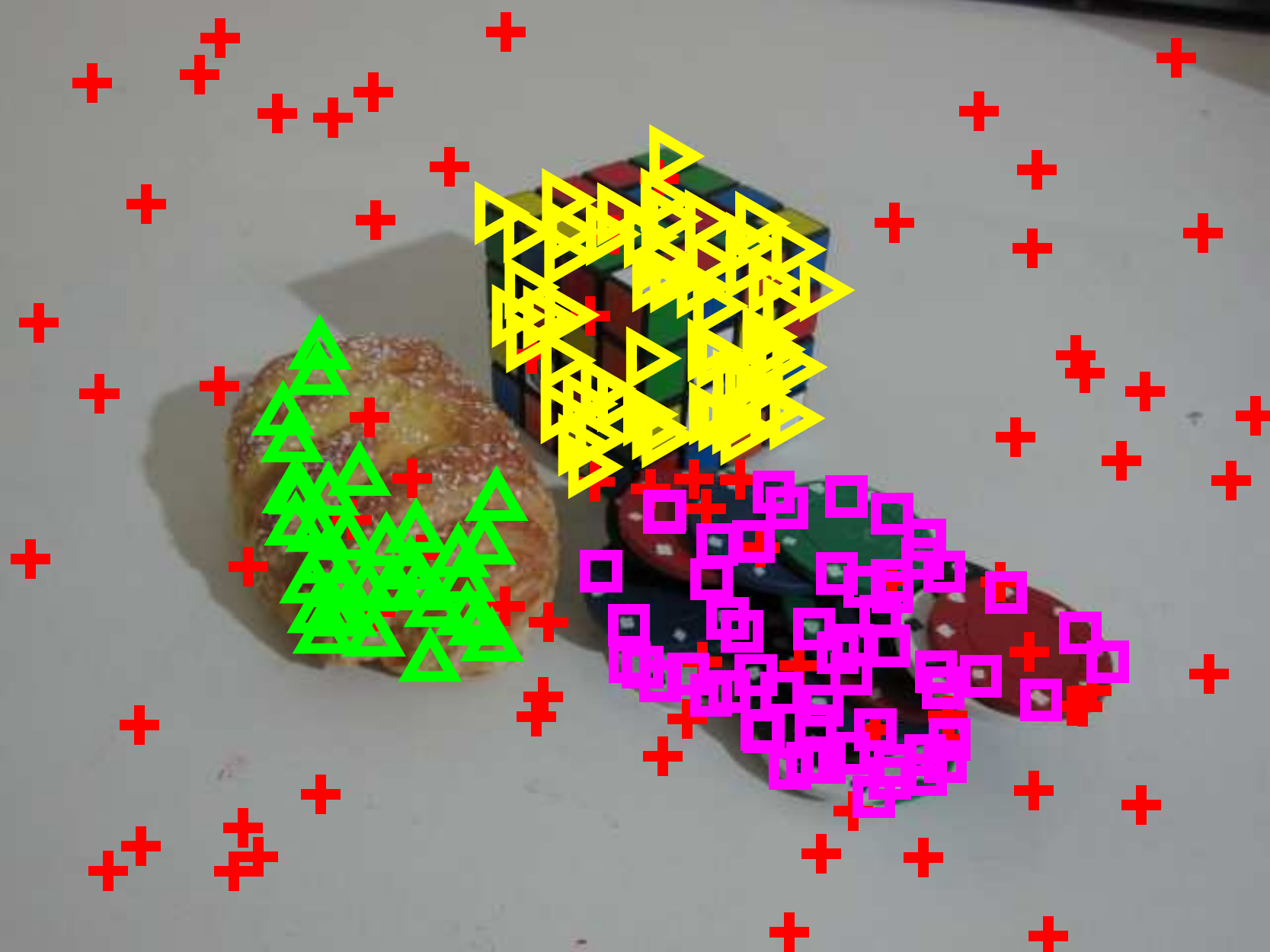}&
     \includegraphics[width=2.6cm]{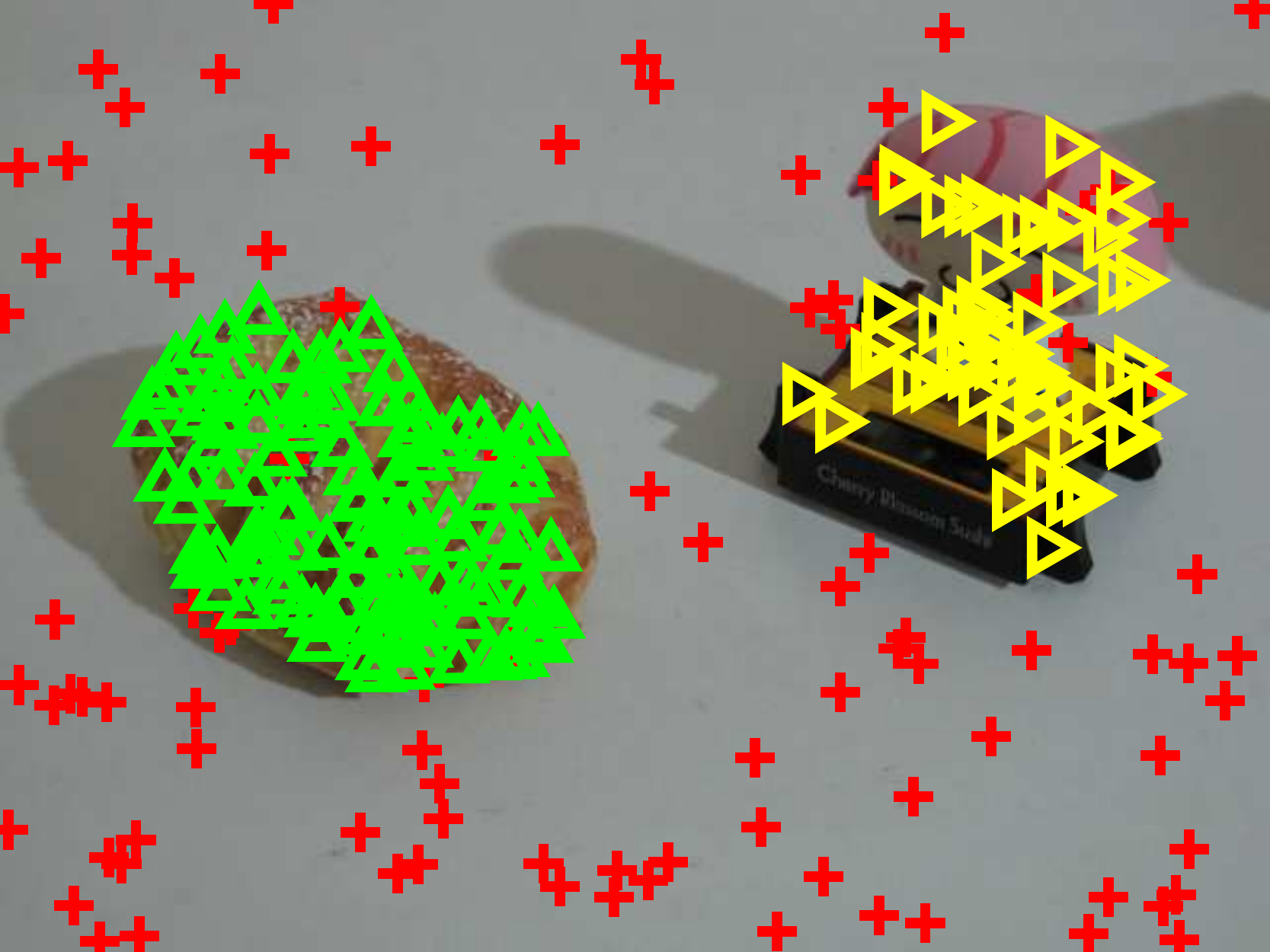}&
     \includegraphics[width=2.6cm]{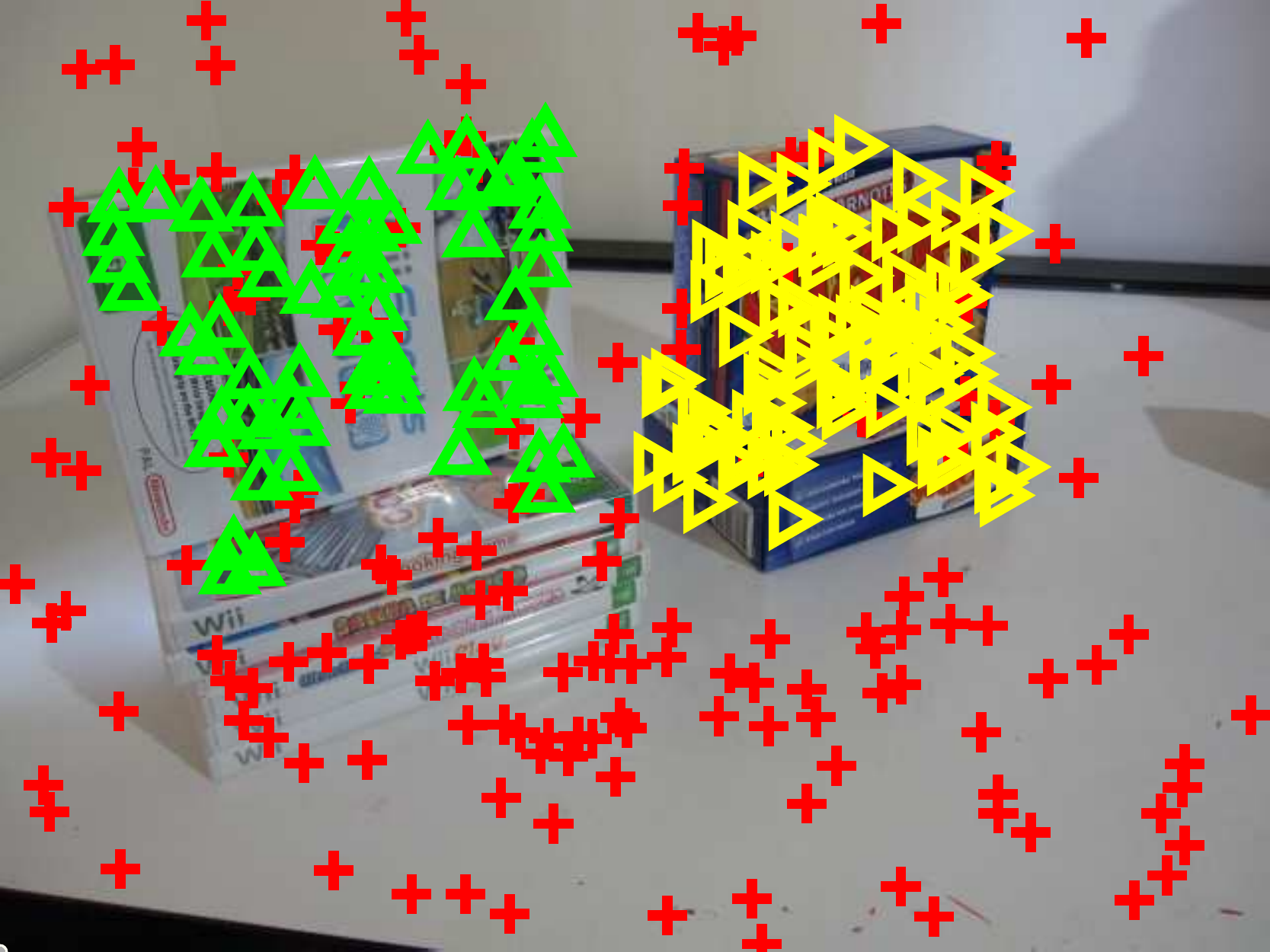} \\
  \rotatebox{90}{\hspace{.2cm}  \textbf{DGSAC-G}}   & \includegraphics[width=2.6cm]{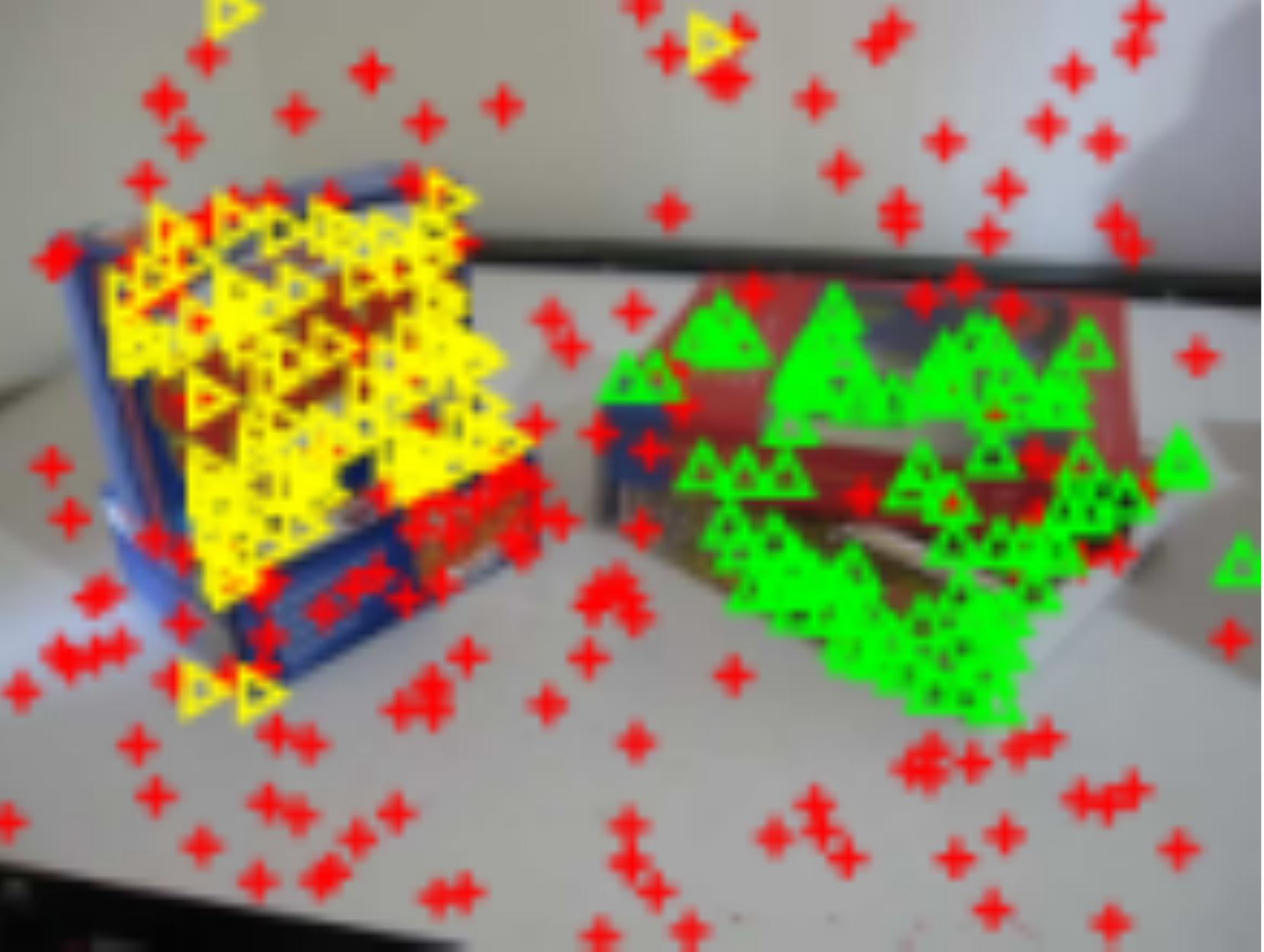}&
     \includegraphics[width=2.6cm]{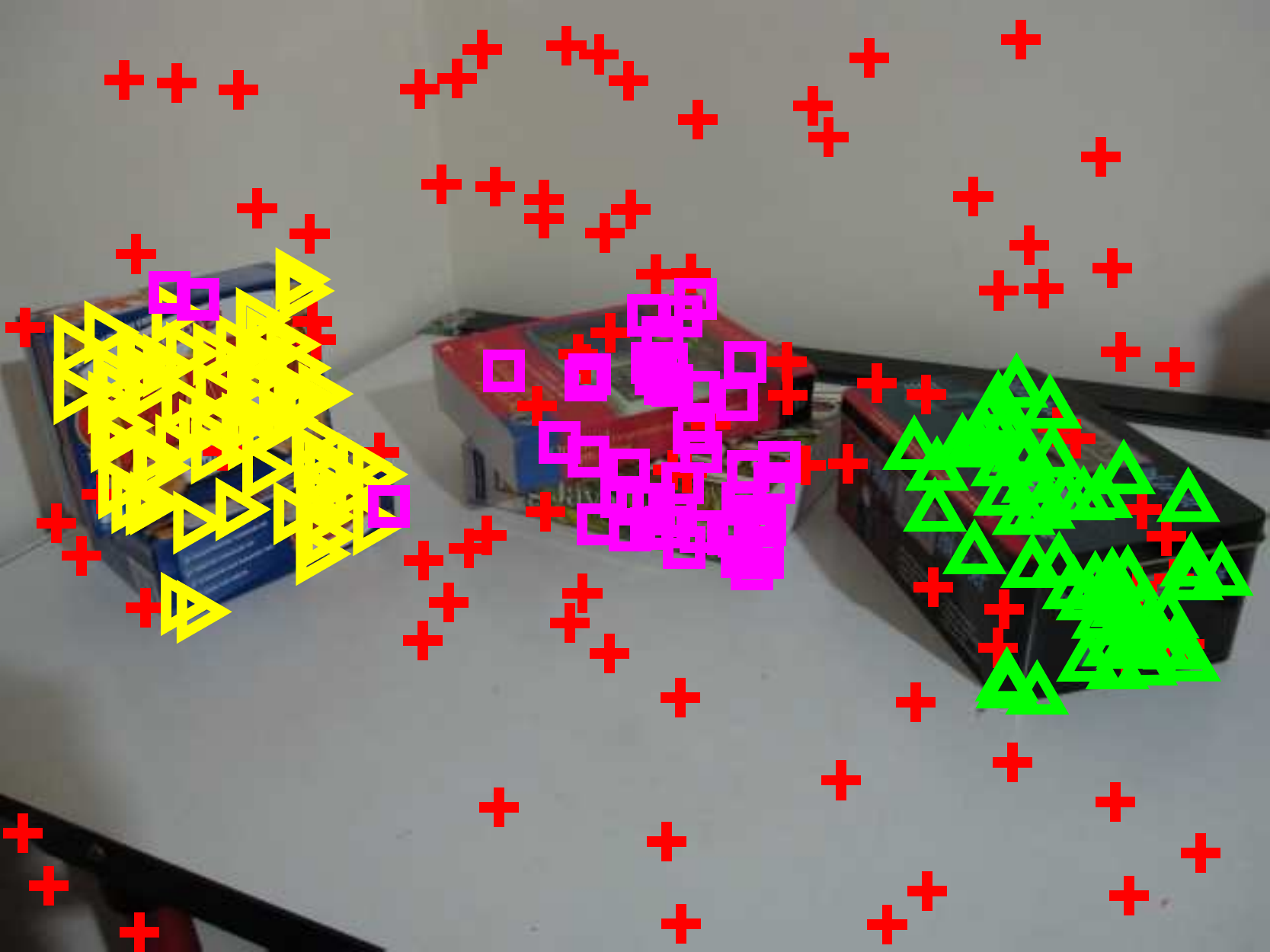}&
     \includegraphics[width=2.6cm]{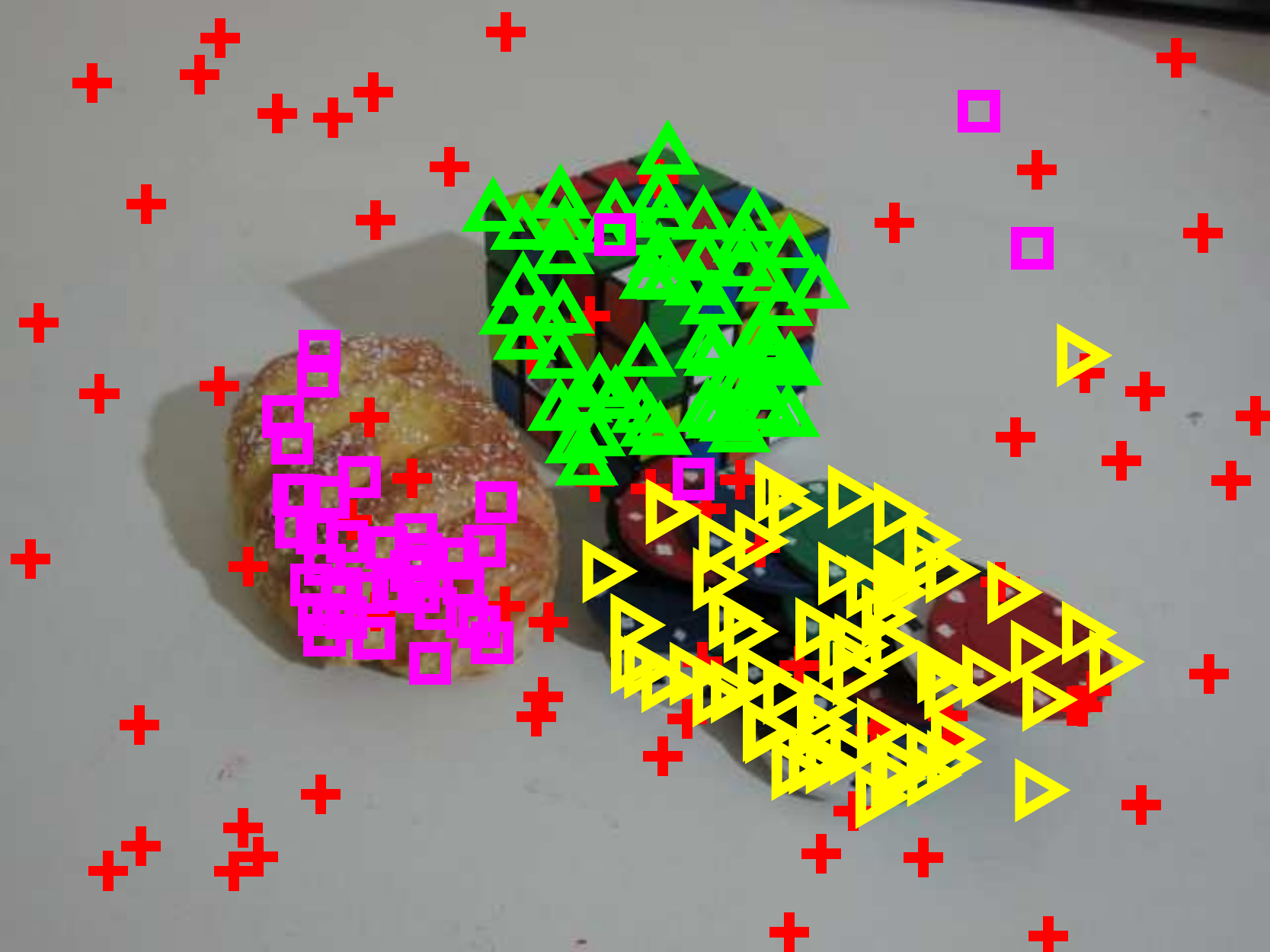}&
     \includegraphics[width=2.6cm]{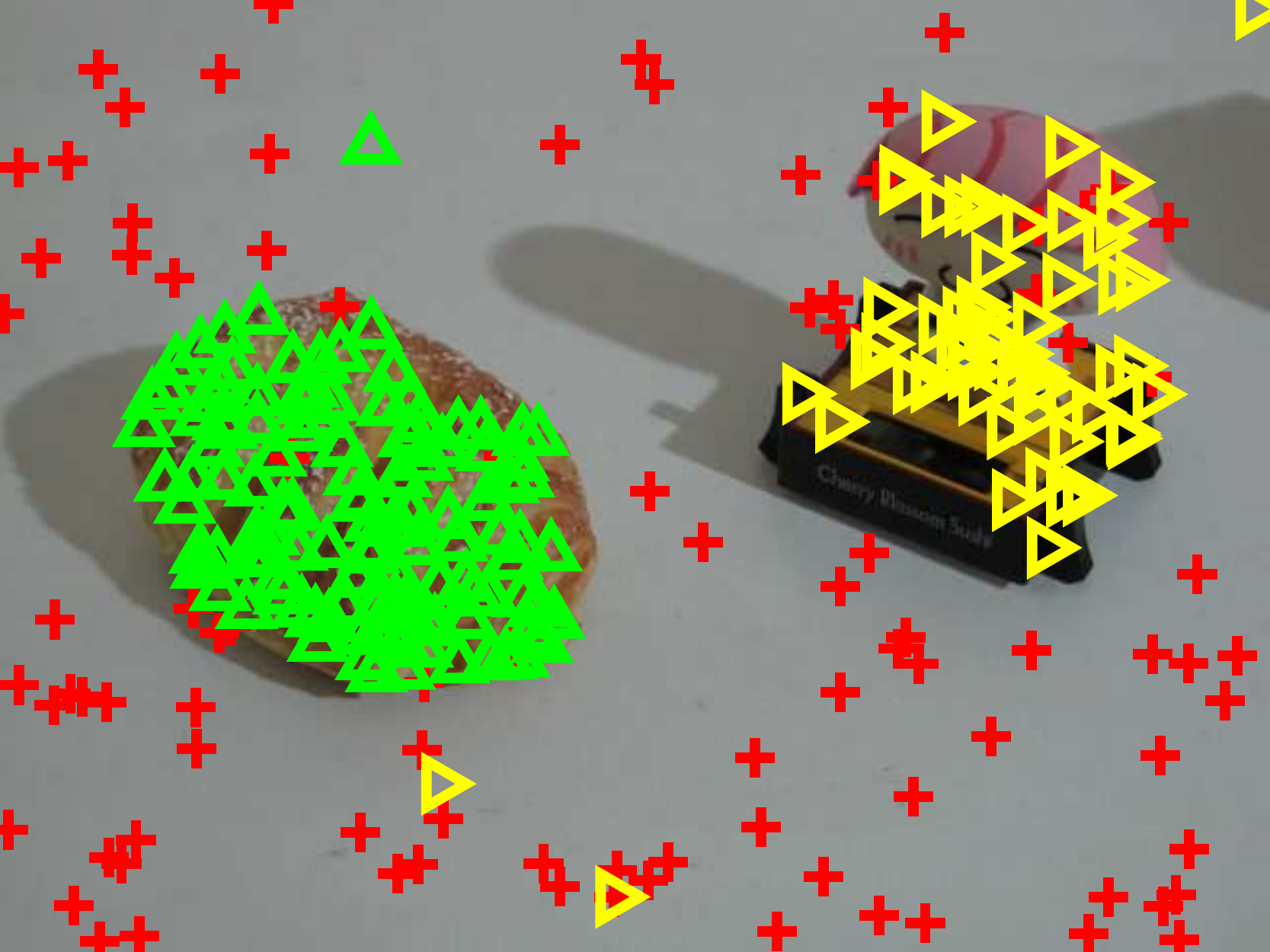}&
     \includegraphics[width=2.6cm]{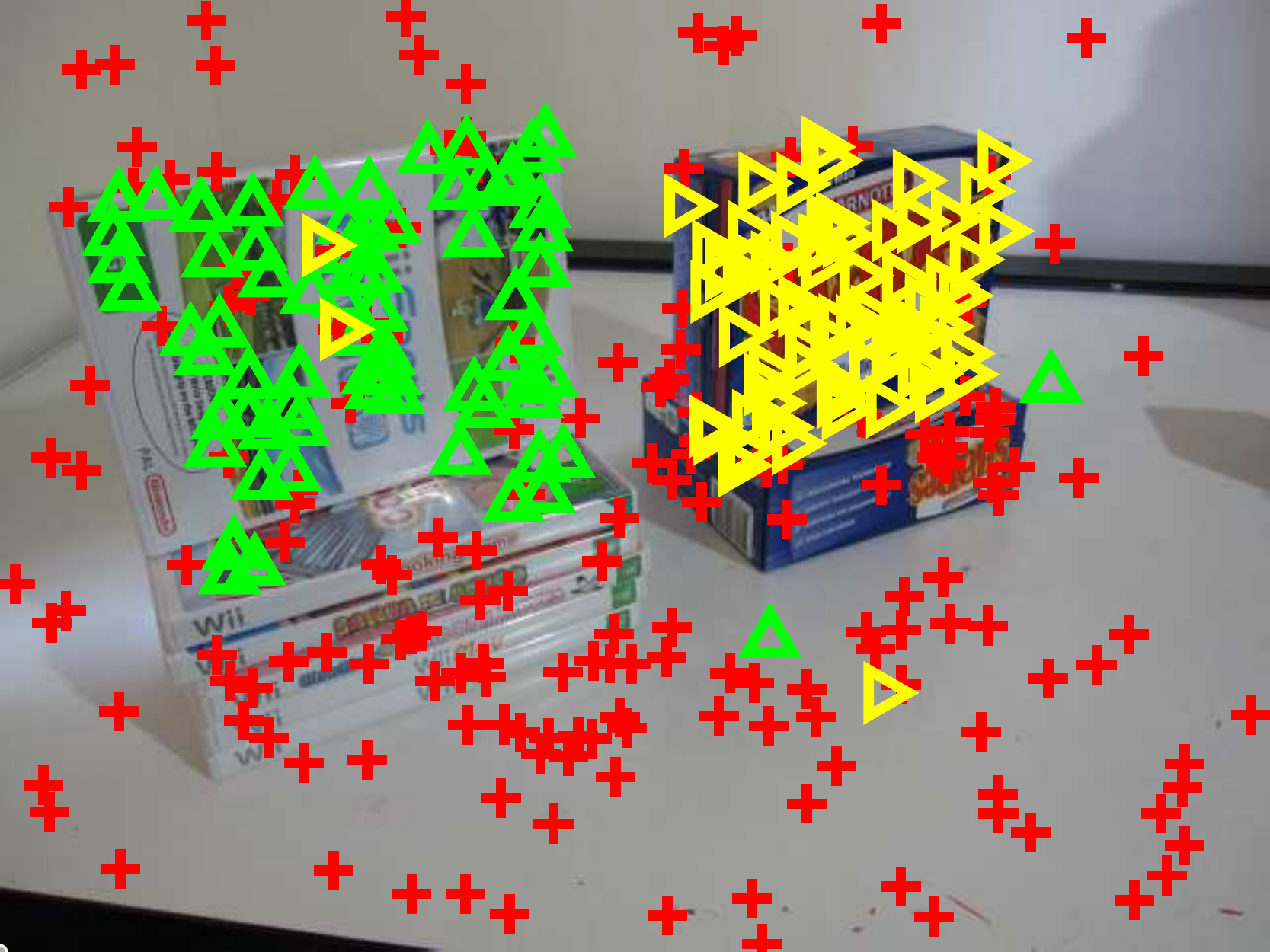} \\
   \rotatebox{90}{\hspace{.2cm}  \textbf{DGSAC-O}}      &\includegraphics[width=2.6cm]{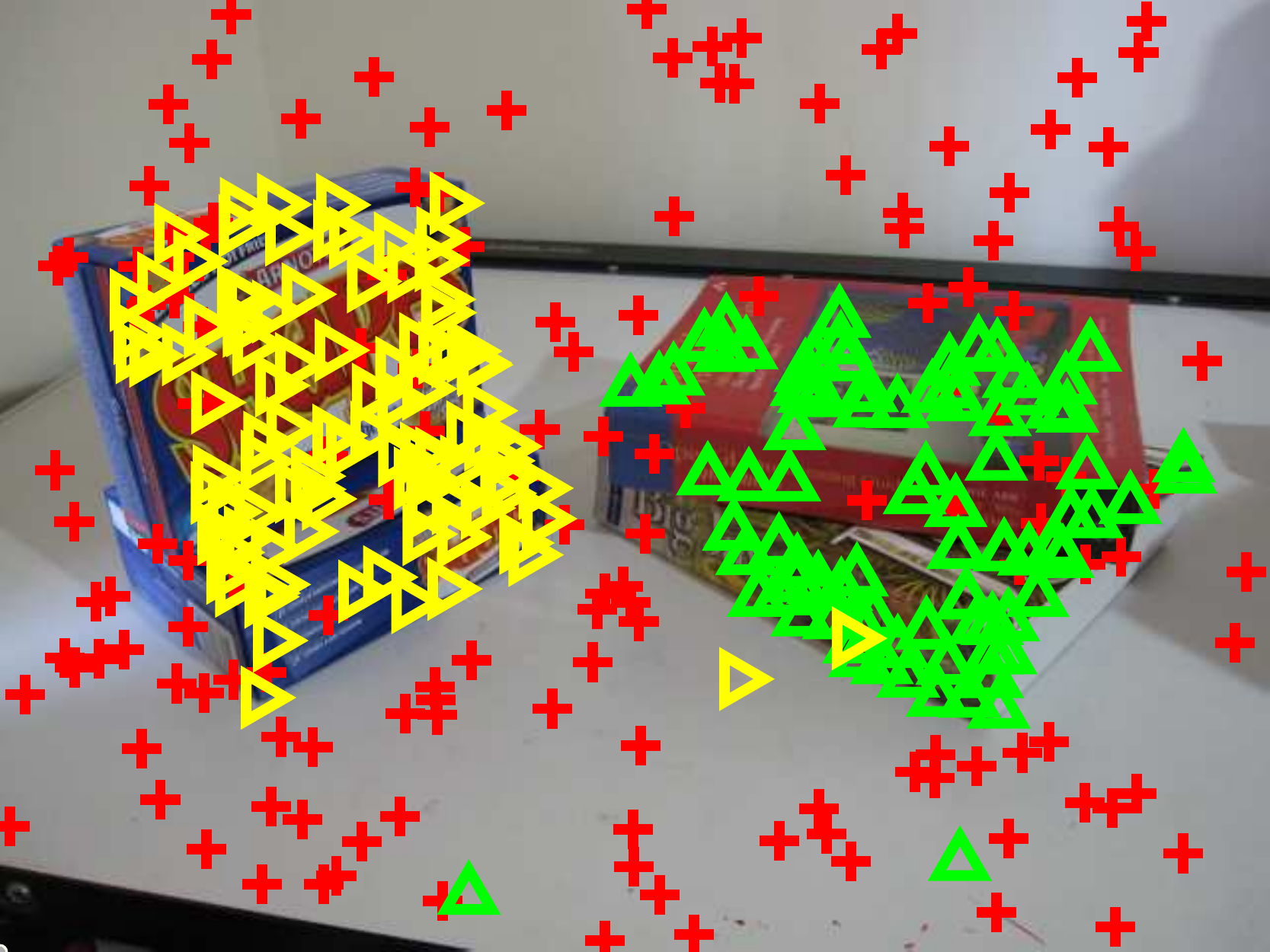}&
     \includegraphics[width=2.6cm]{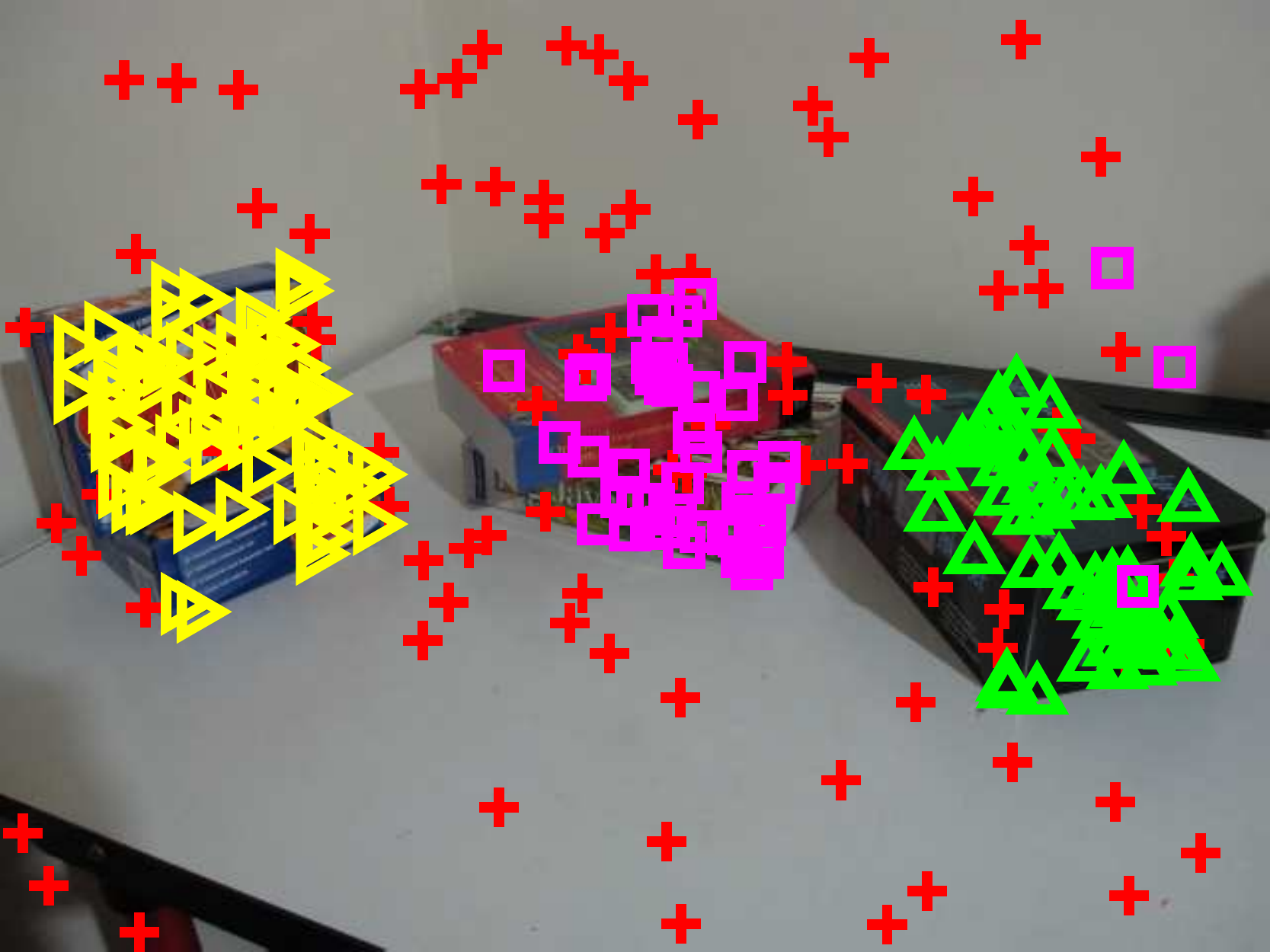}&
     \includegraphics[width=2.6cm]{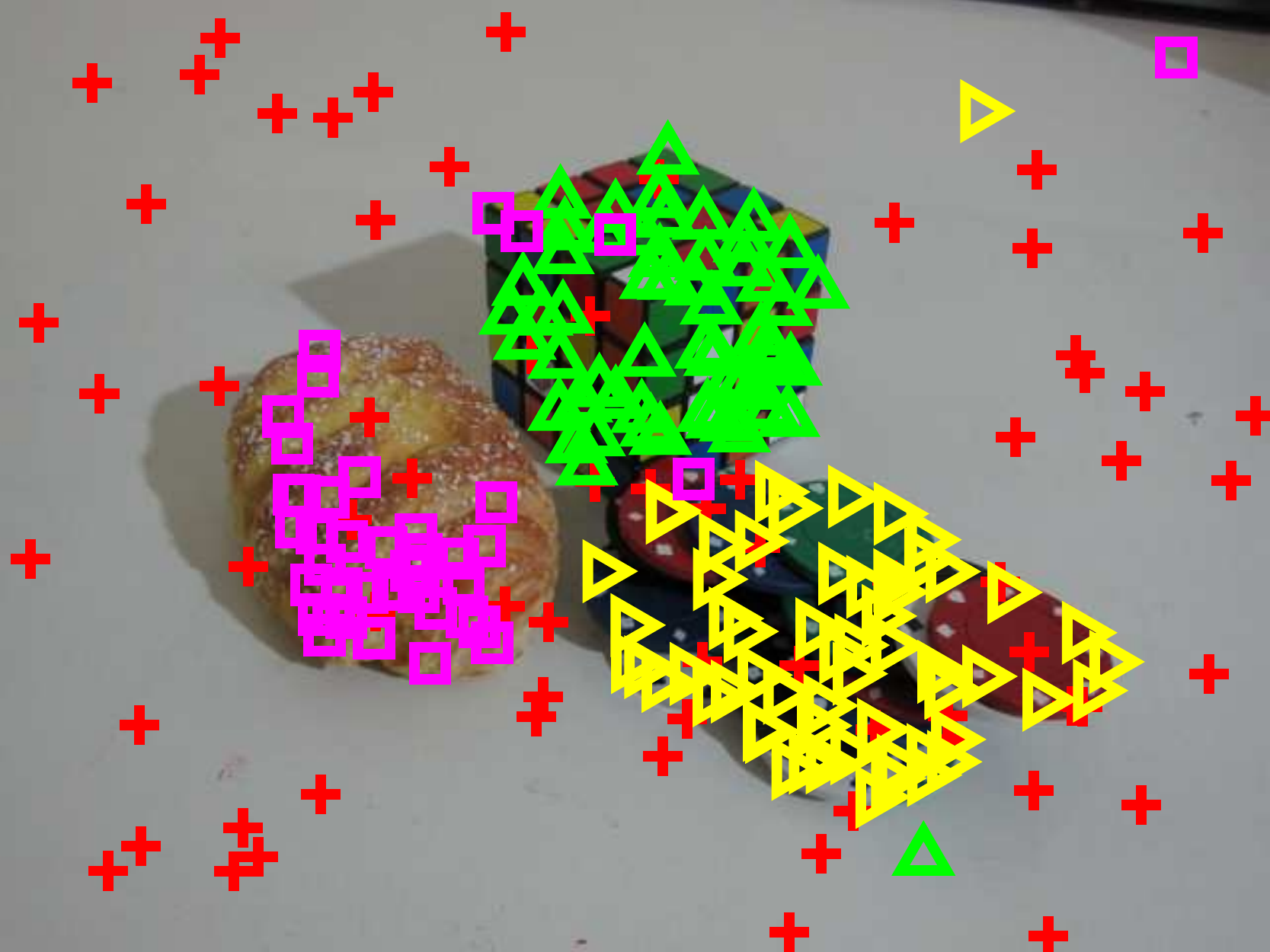}&
     \includegraphics[width=2.6cm]{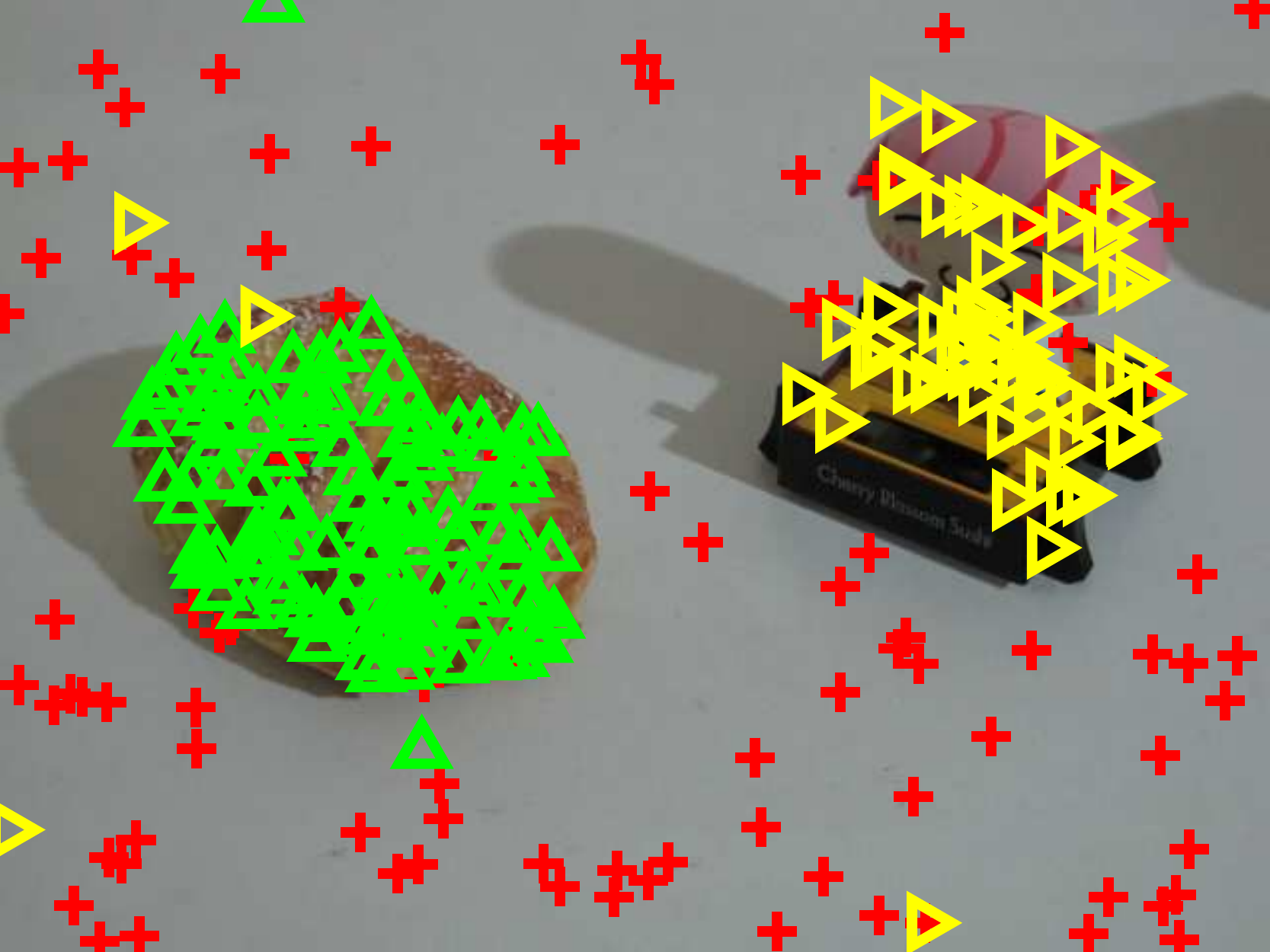}&
     \includegraphics[width=2.6cm]{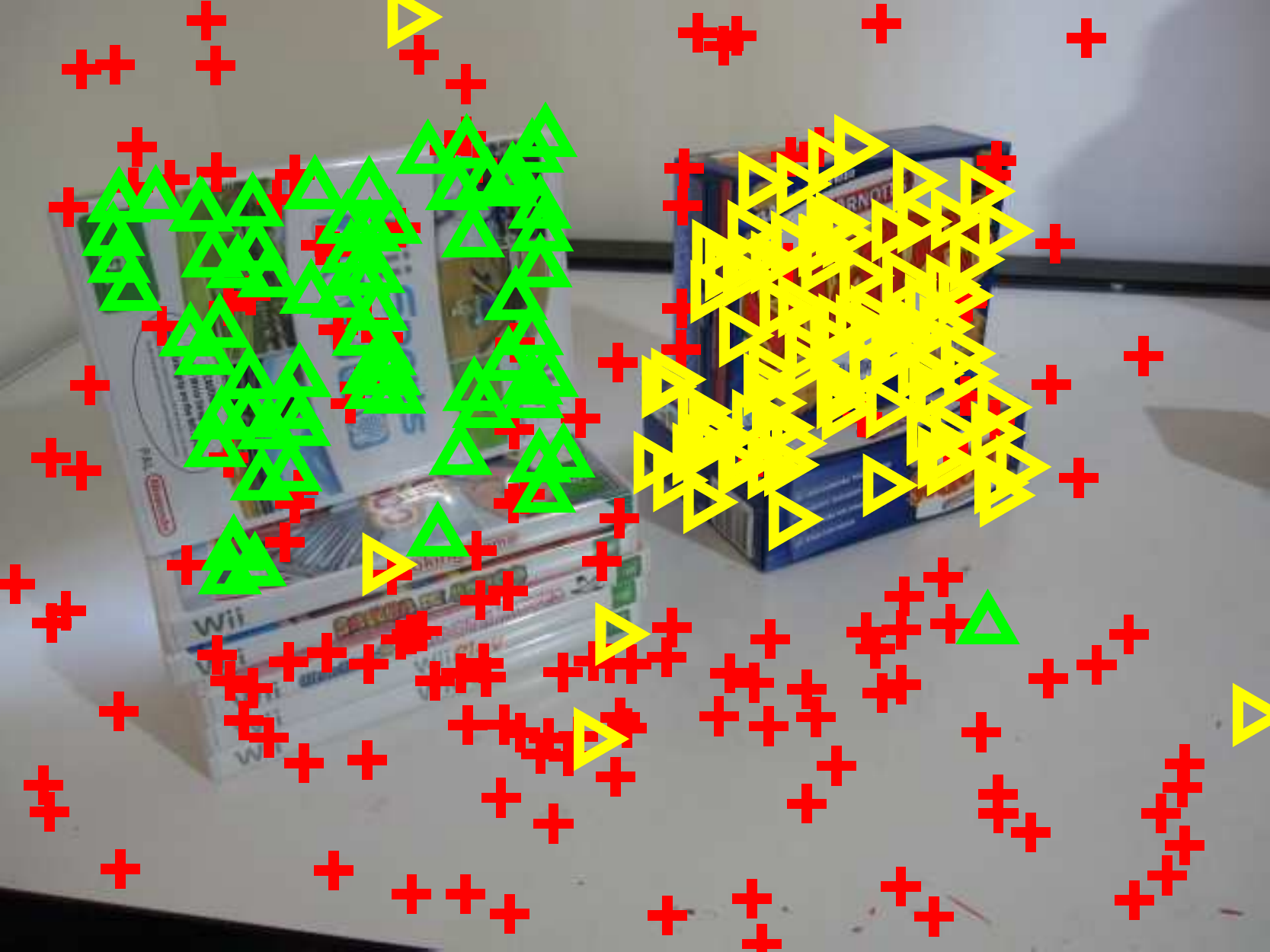} \\
     \hline
     & elderhalla & johnsona & library & ladysymon & sene \\
 \rotatebox{90}{\hspace{.2cm}  \textbf{GroundTruth}} &    \includegraphics[width=2.6cm]{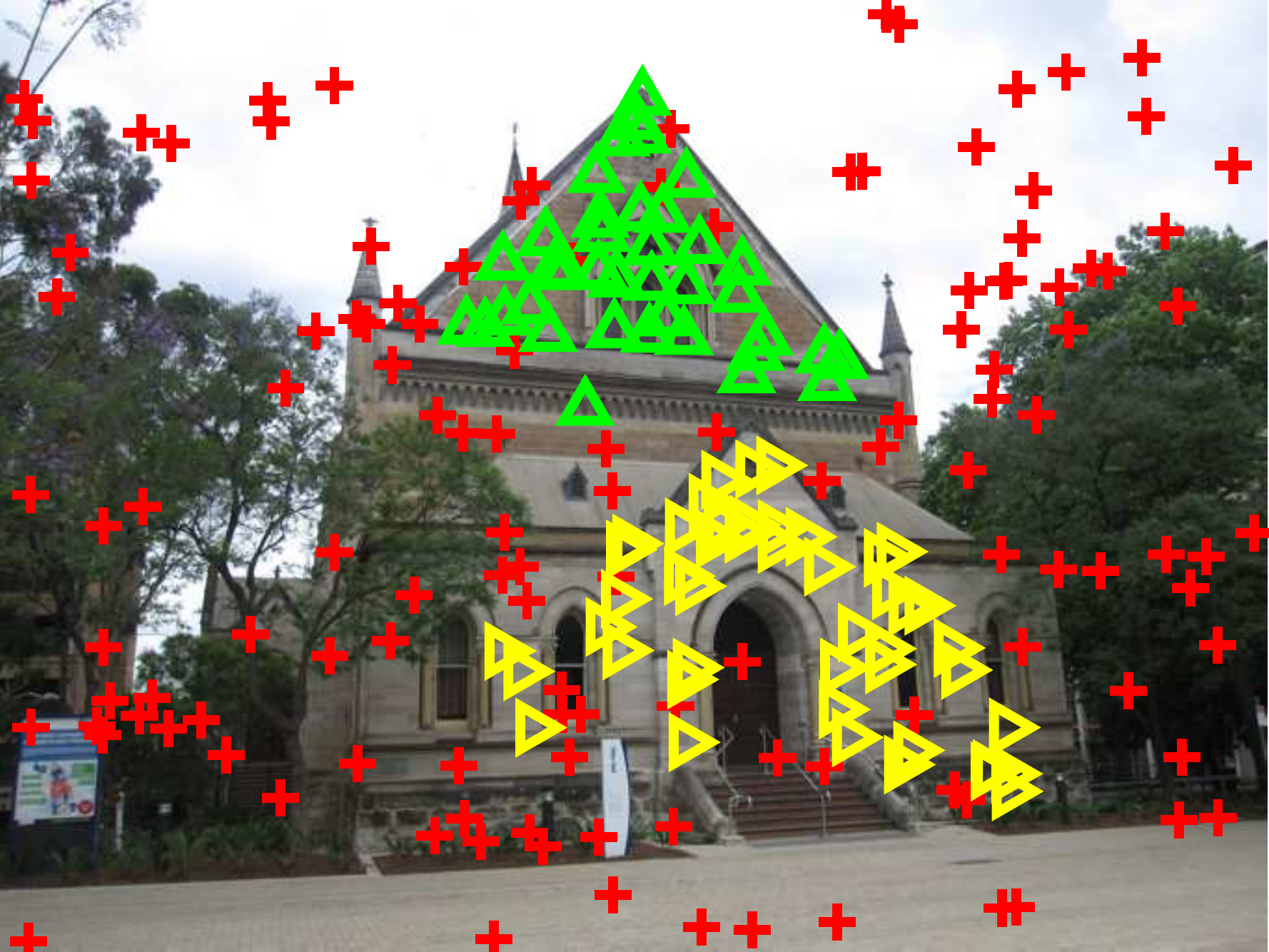}&
     \includegraphics[width=2.6cm]{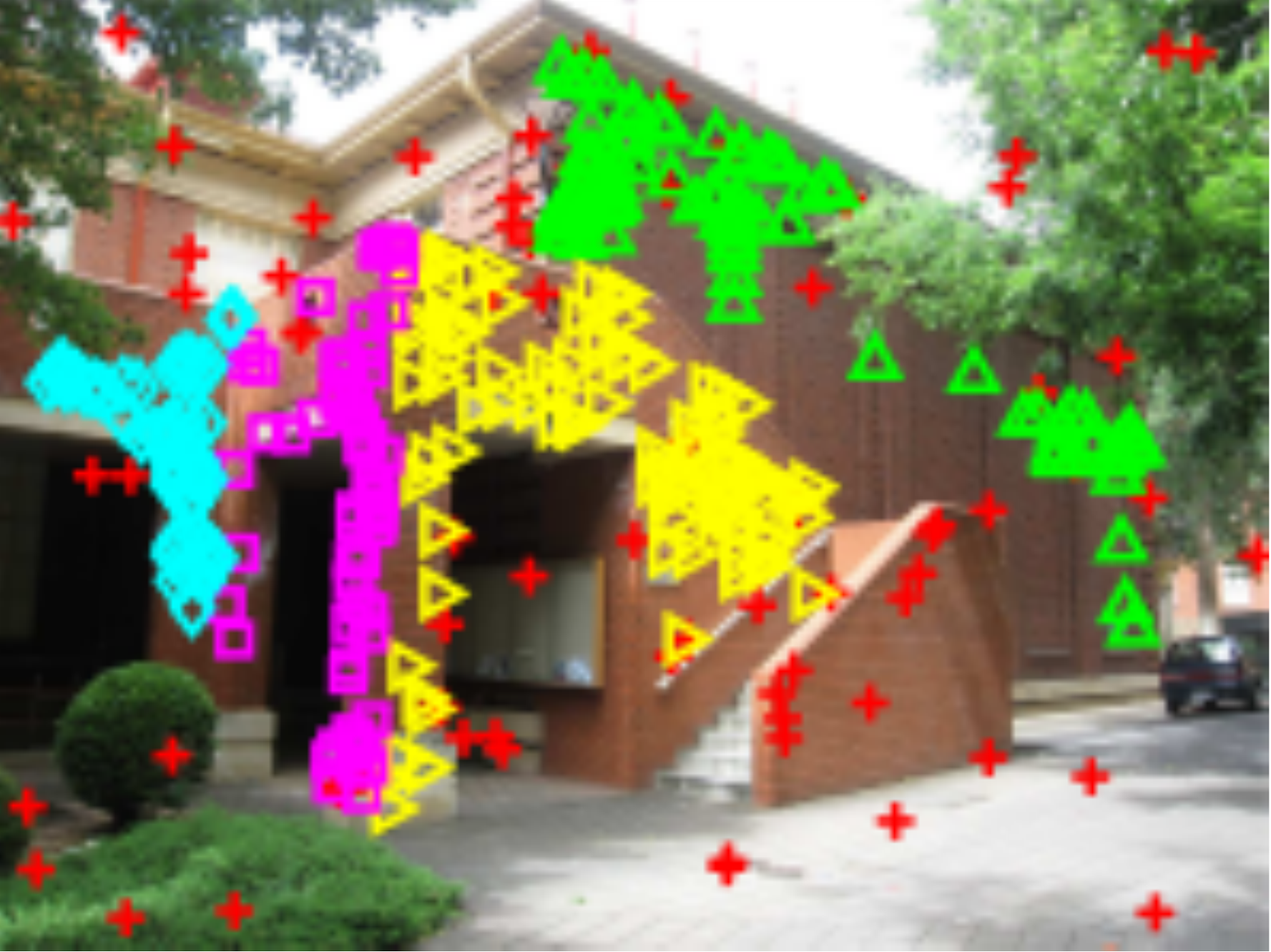}&
     \includegraphics[width=2.6cm]{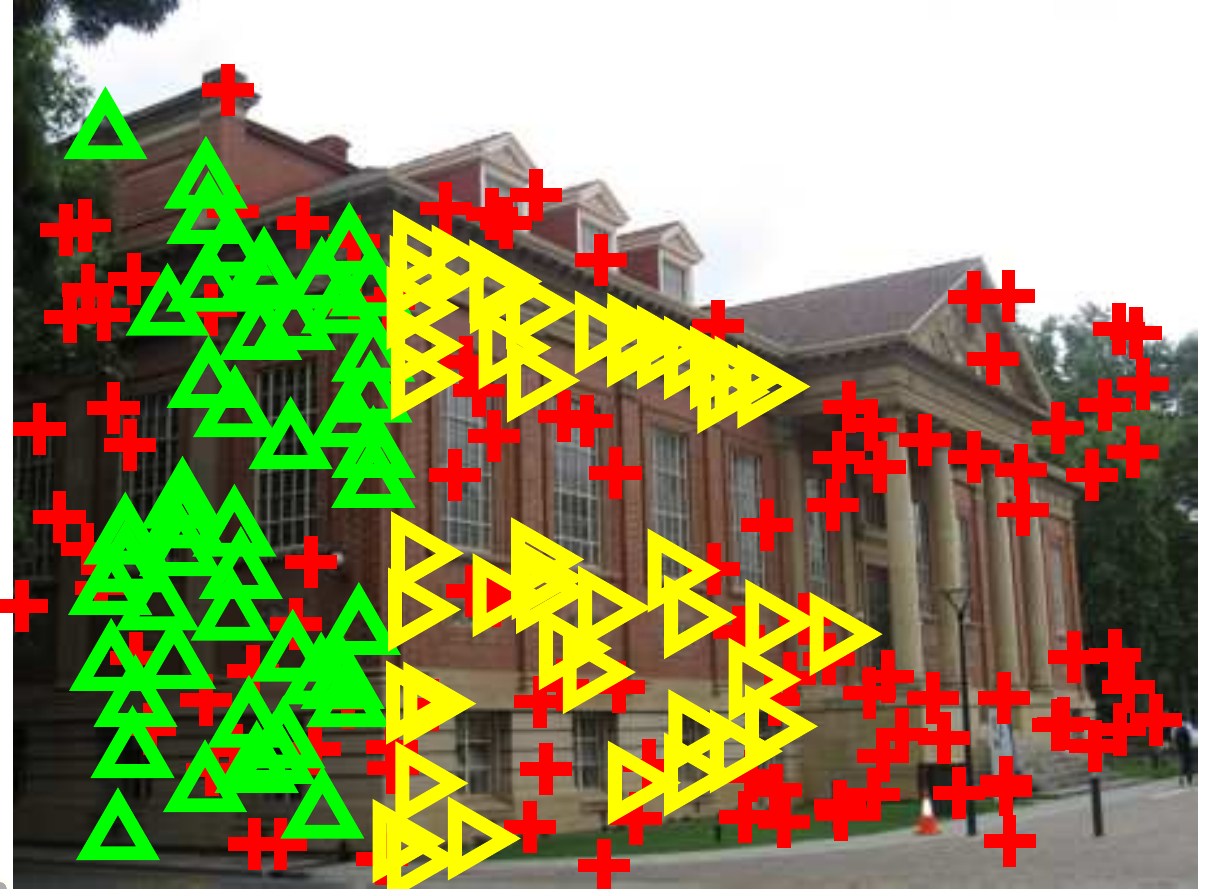}&
     \includegraphics[width=2.6cm]{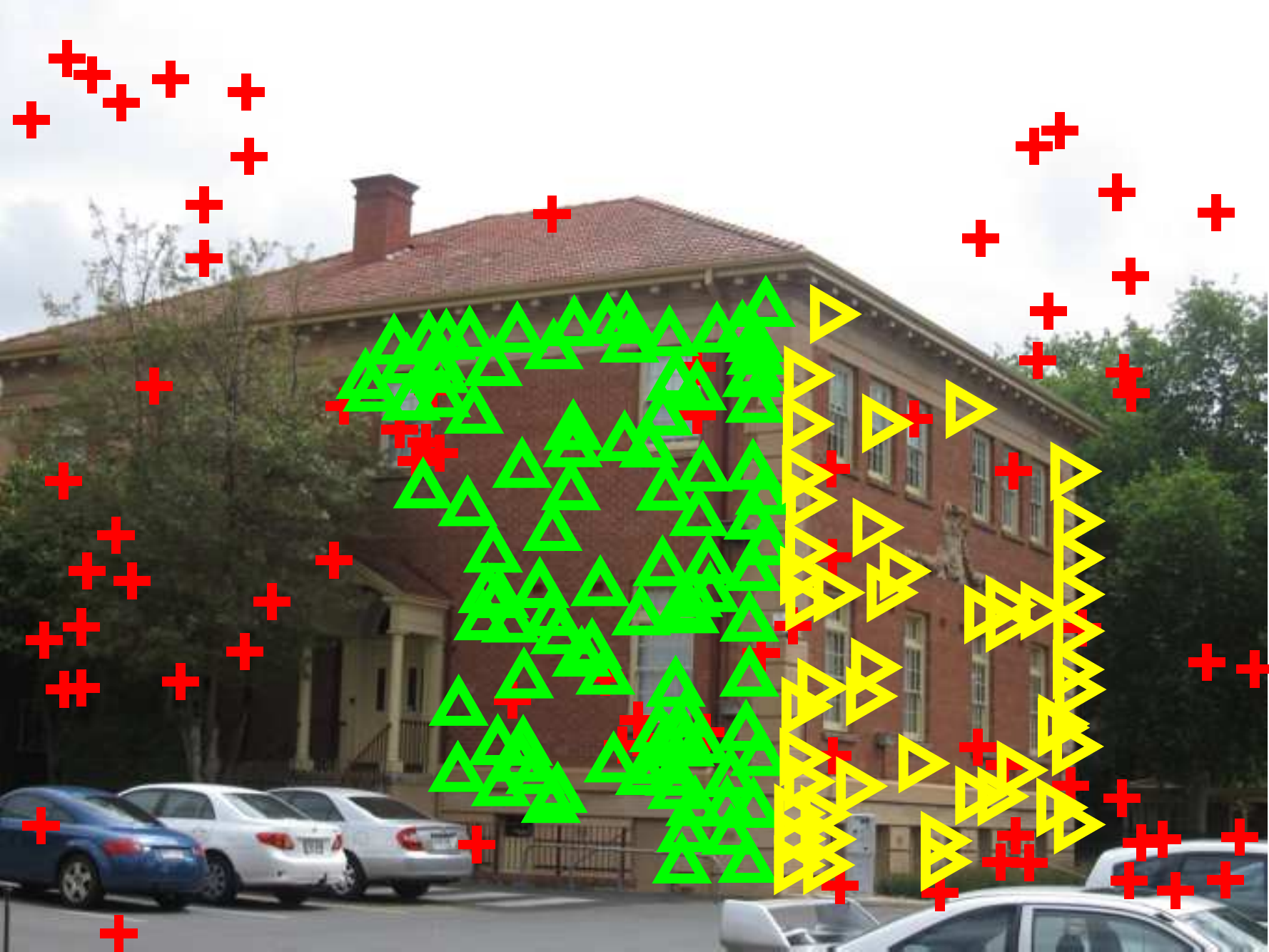}&
     \includegraphics[width=2.6cm]{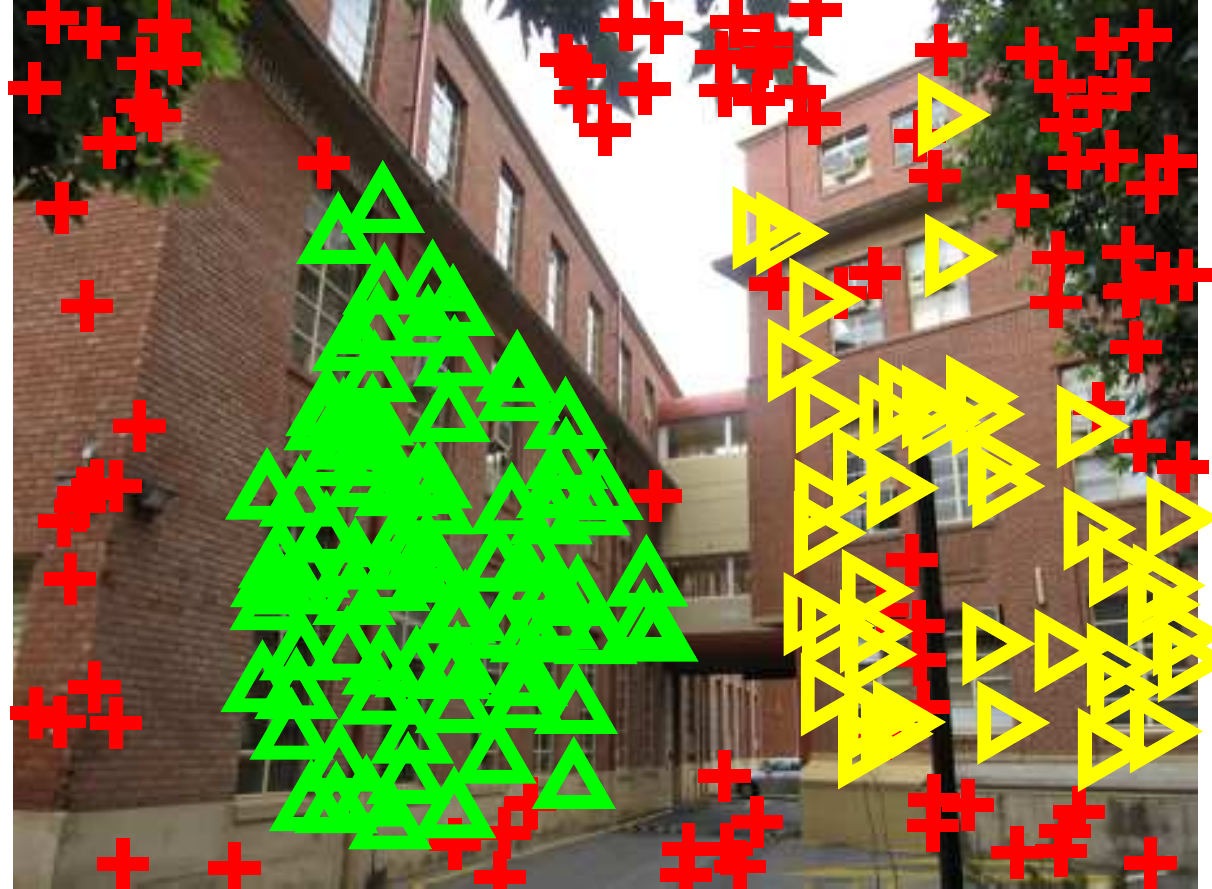} \\
  \rotatebox{90}{\hspace{.2cm}  \textbf{DGSAC-G}}   & \includegraphics[width=2.6cm]{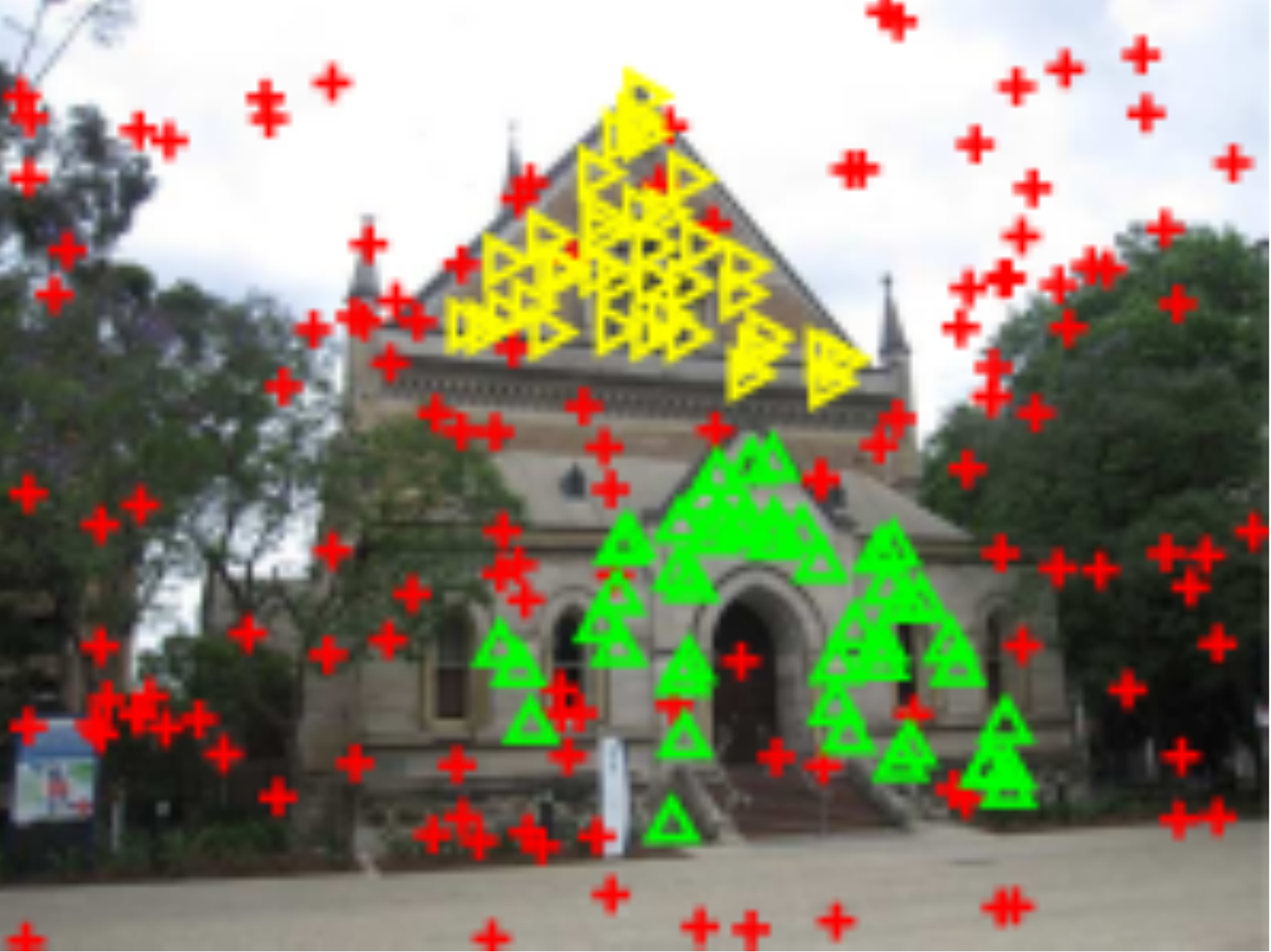}&
     \includegraphics[width=2.6cm]{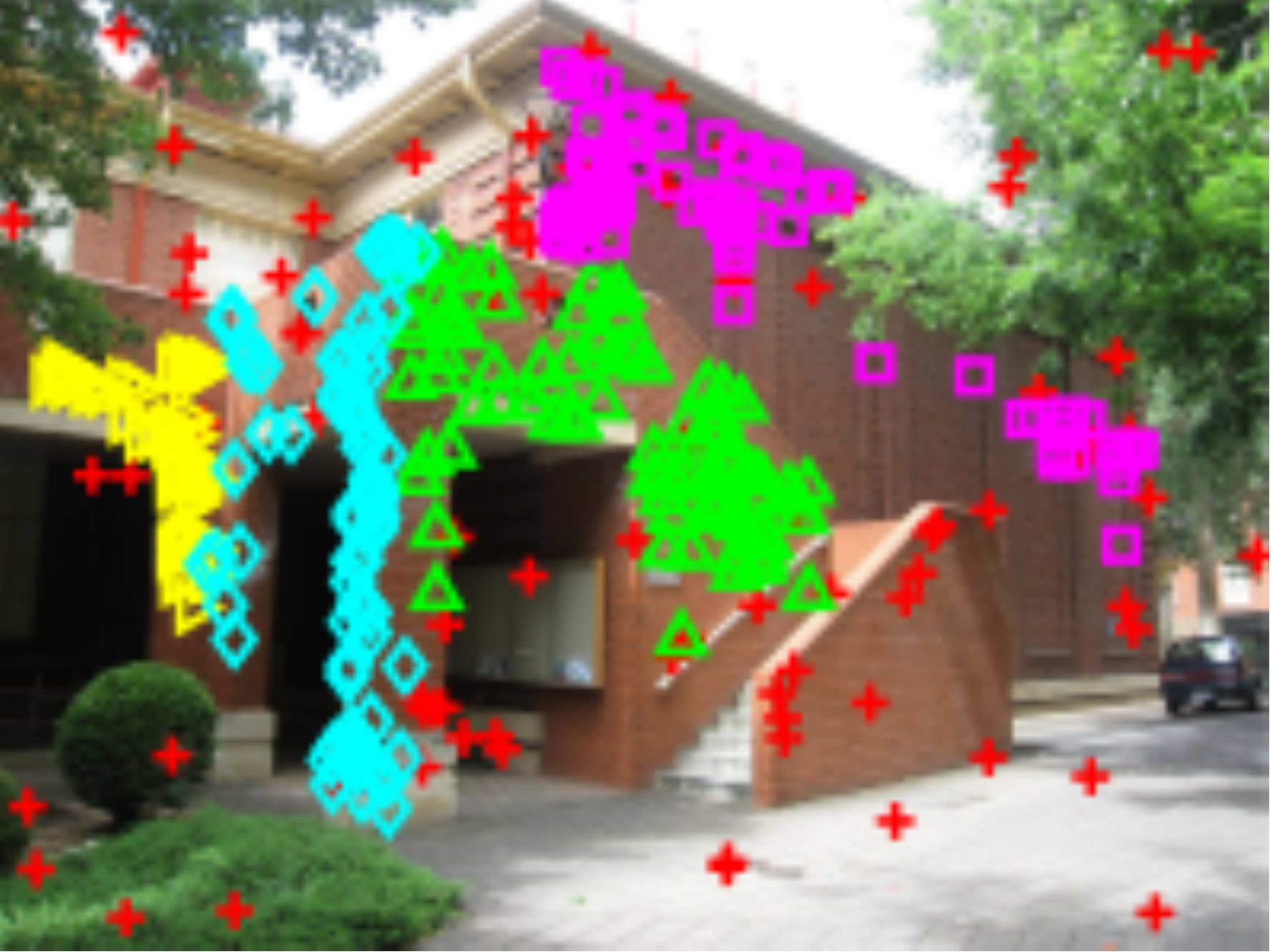}&
     \includegraphics[width=2.6cm]{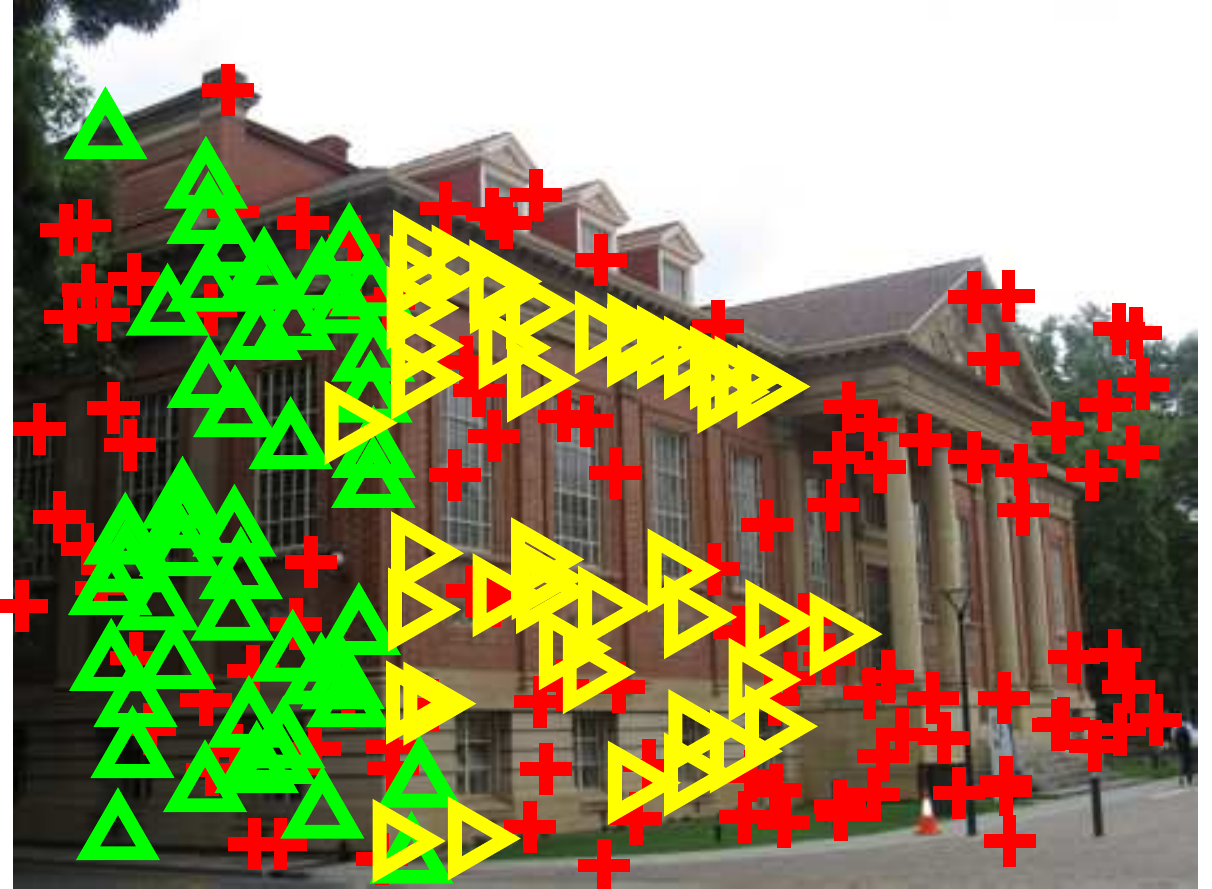}&
     \includegraphics[width=2.6cm]{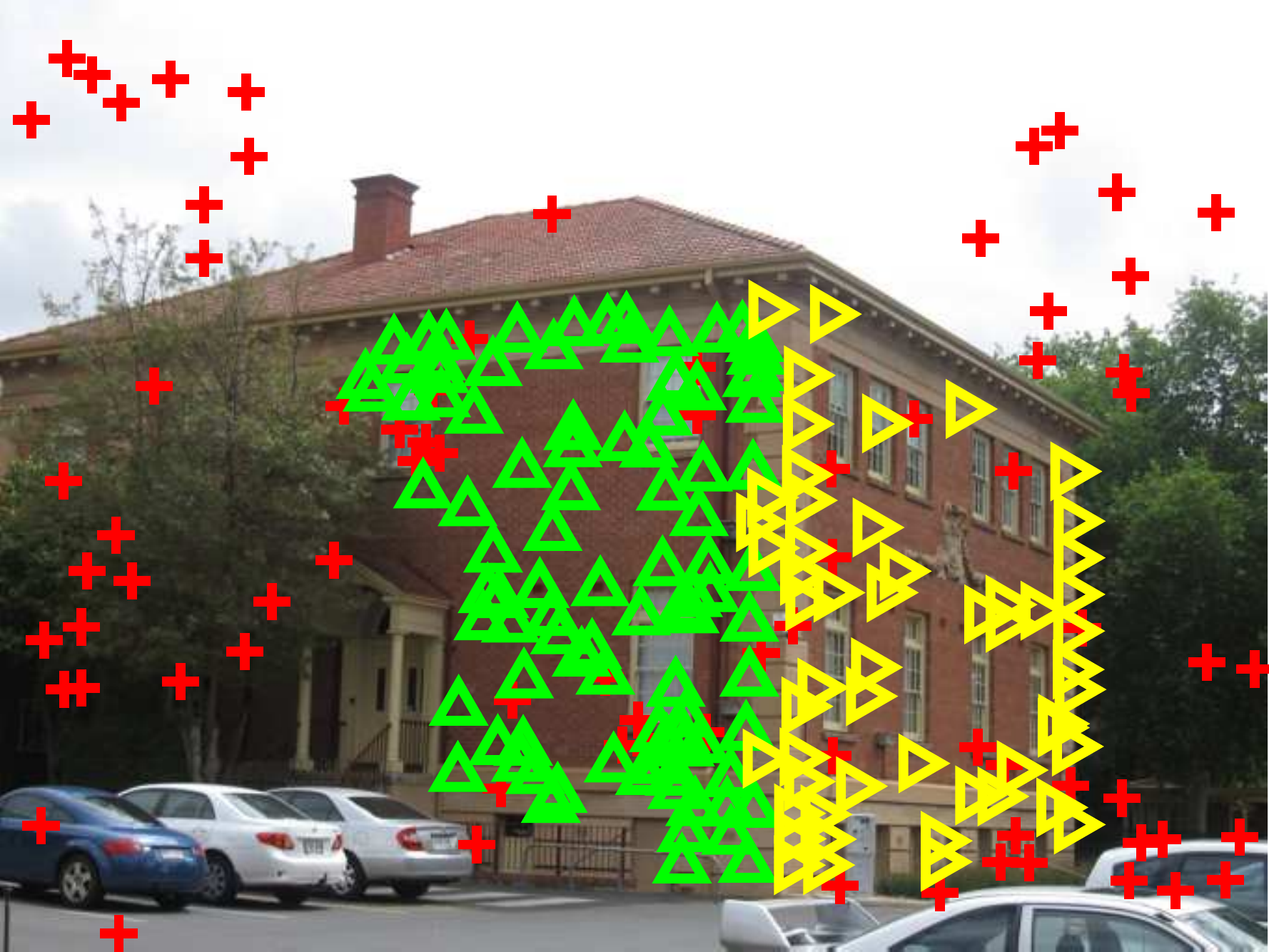}&
     \includegraphics[width=2.6cm]{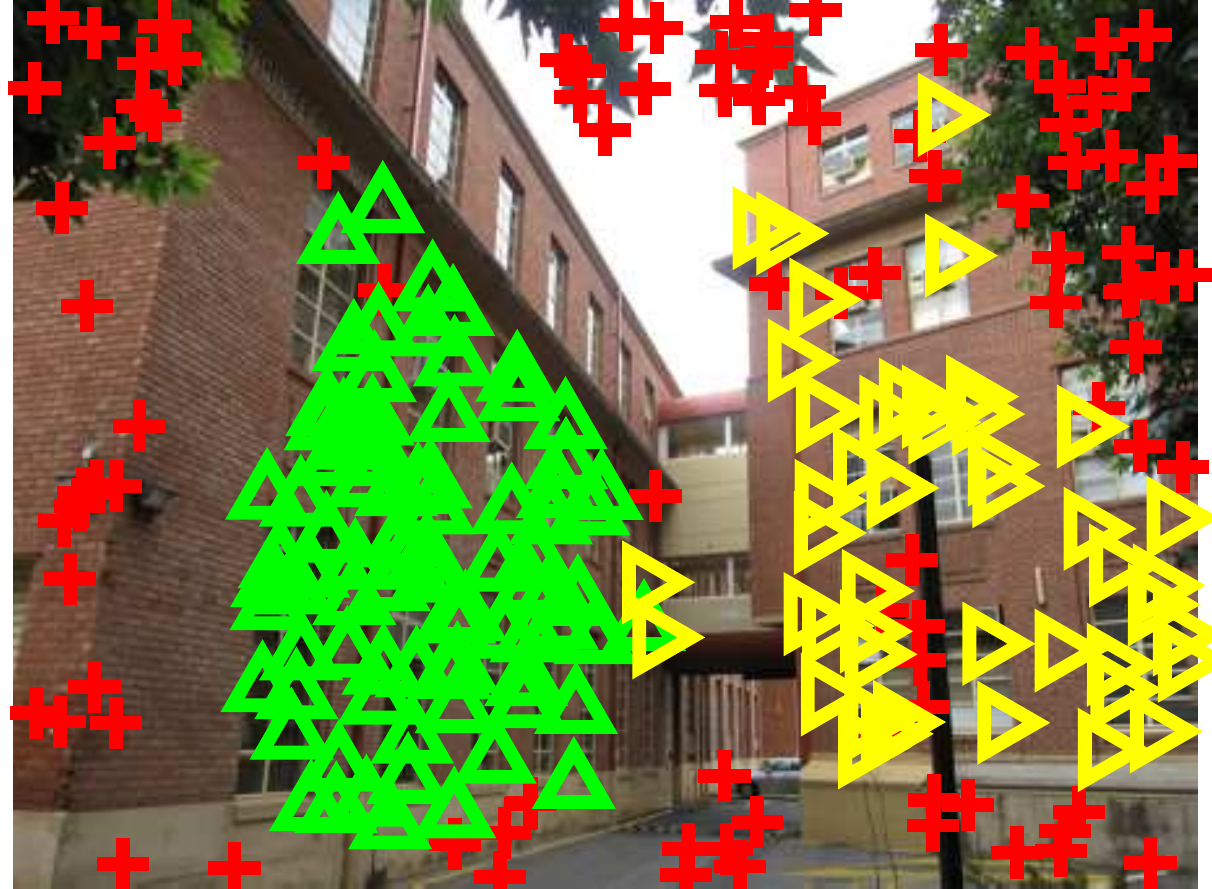} \\
   \rotatebox{90}{\hspace{.2cm}  \textbf{DGSAC-O}}      &\includegraphics[width=2.6cm]{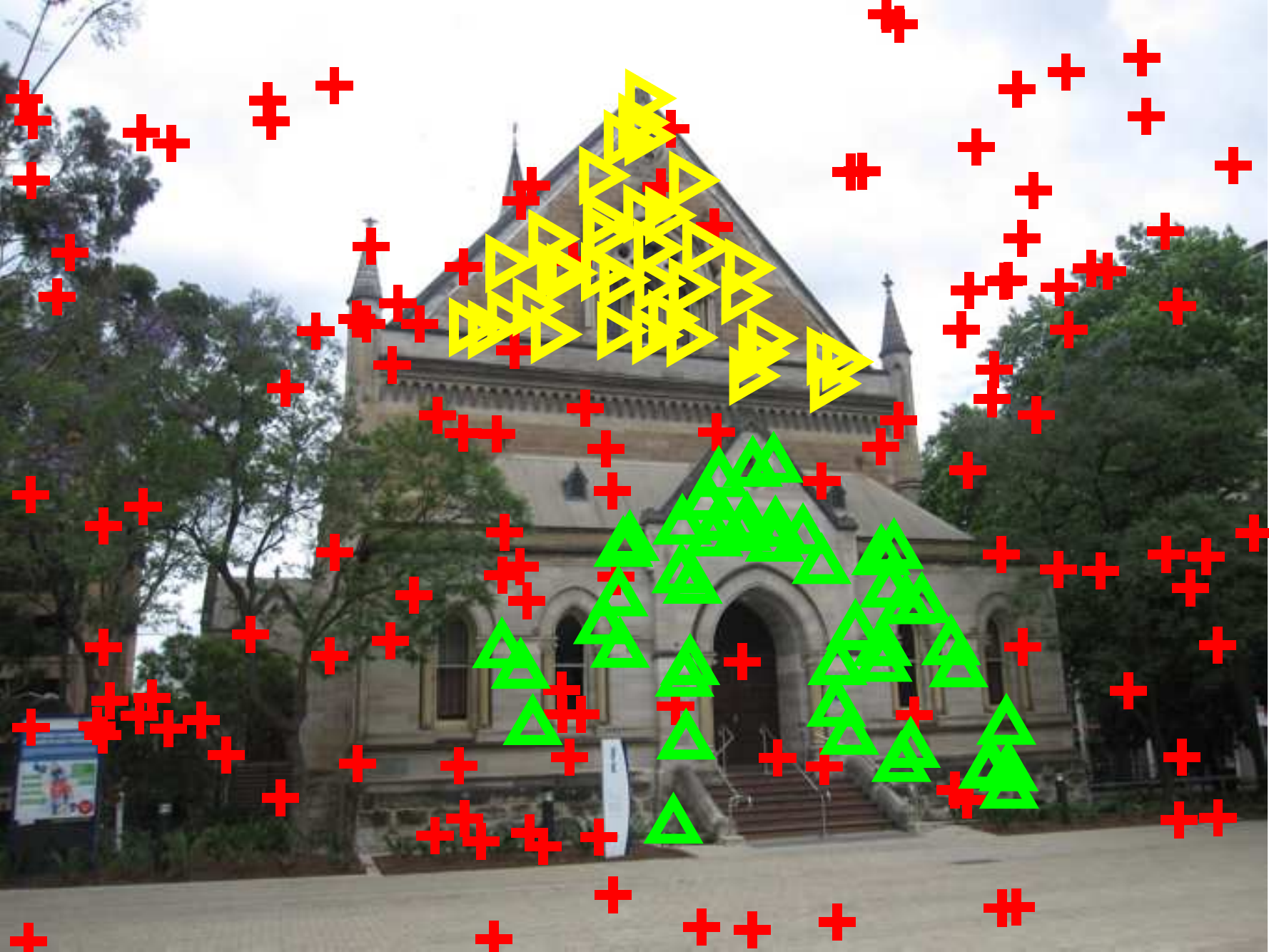}&
     \includegraphics[width=2.6cm]{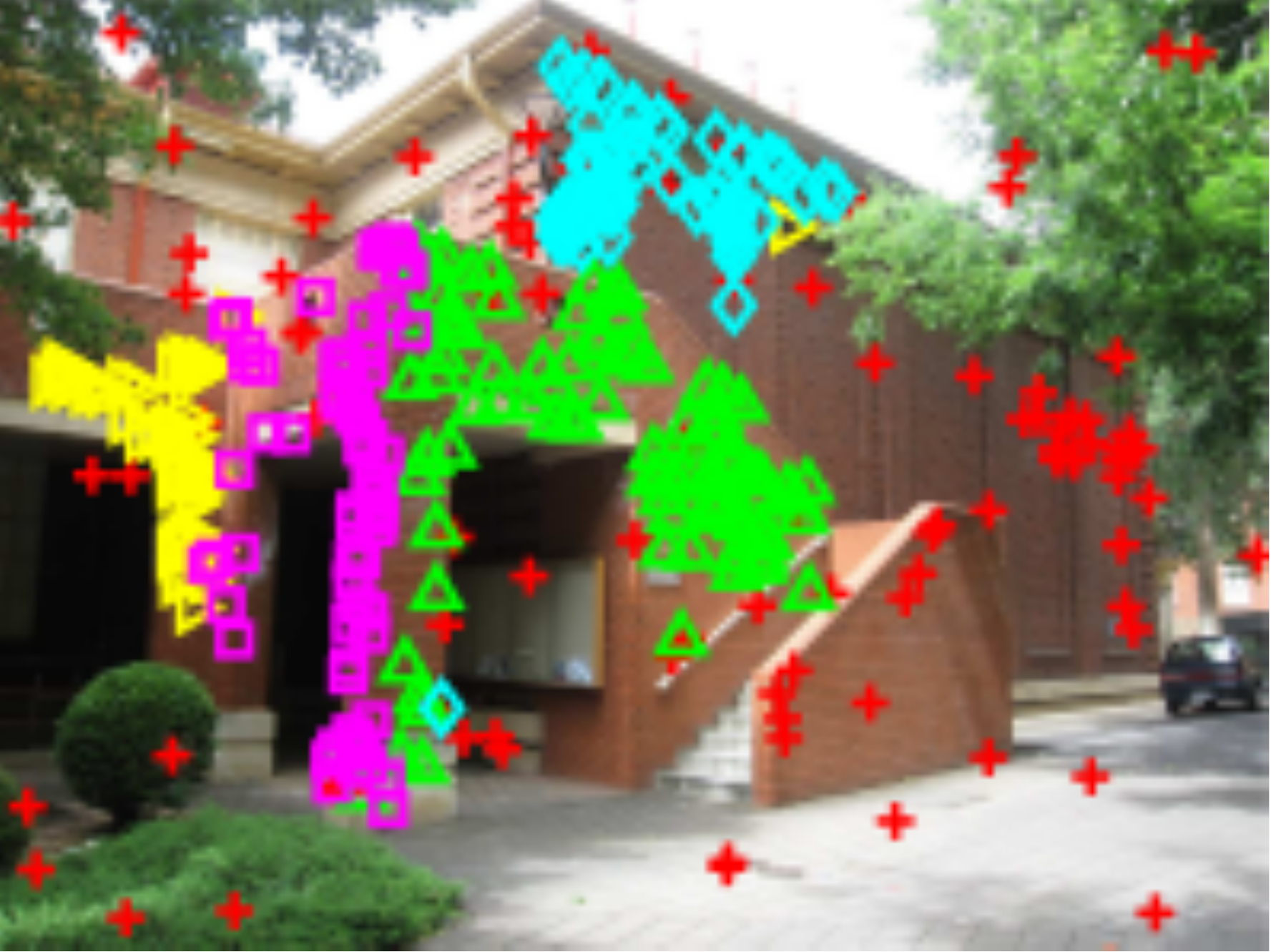}&
     \includegraphics[width=2.6cm]{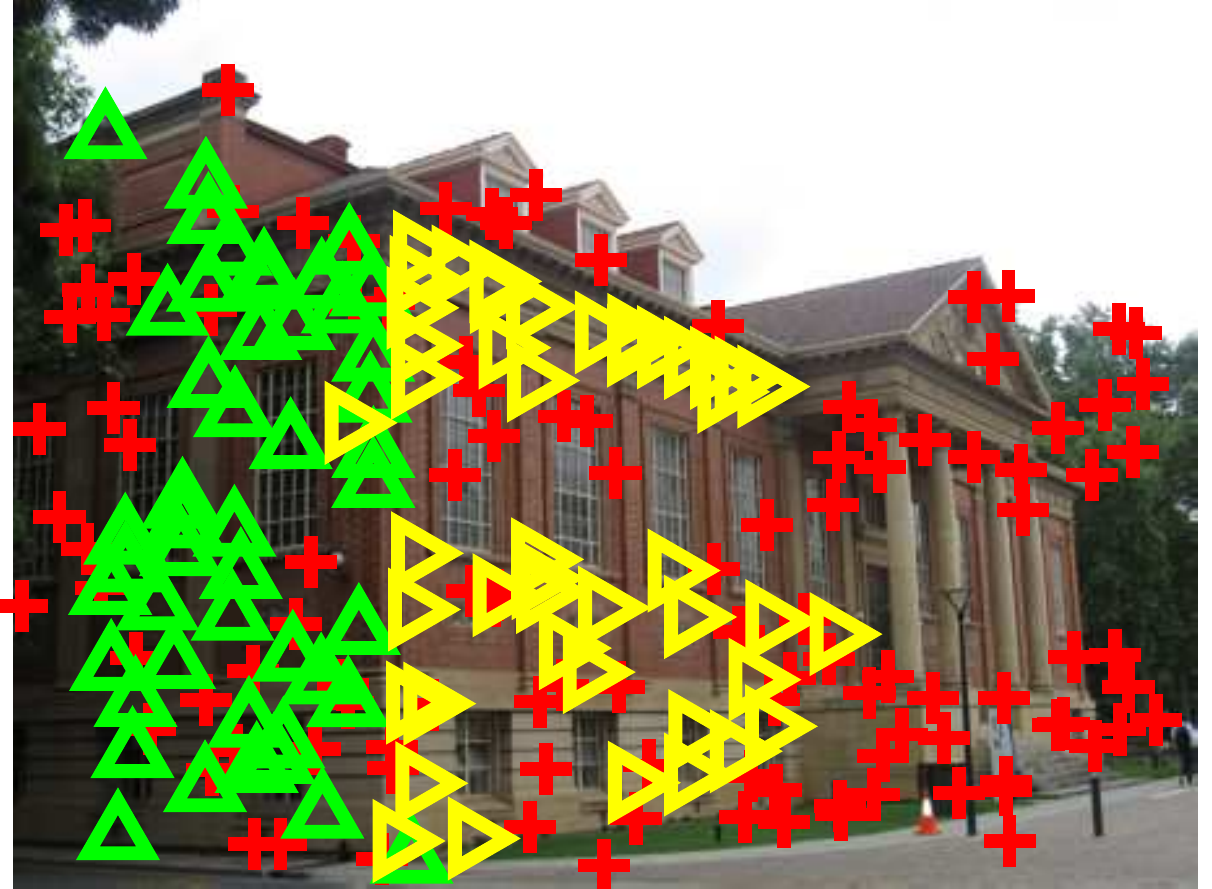}&
     \includegraphics[width=2.6cm]{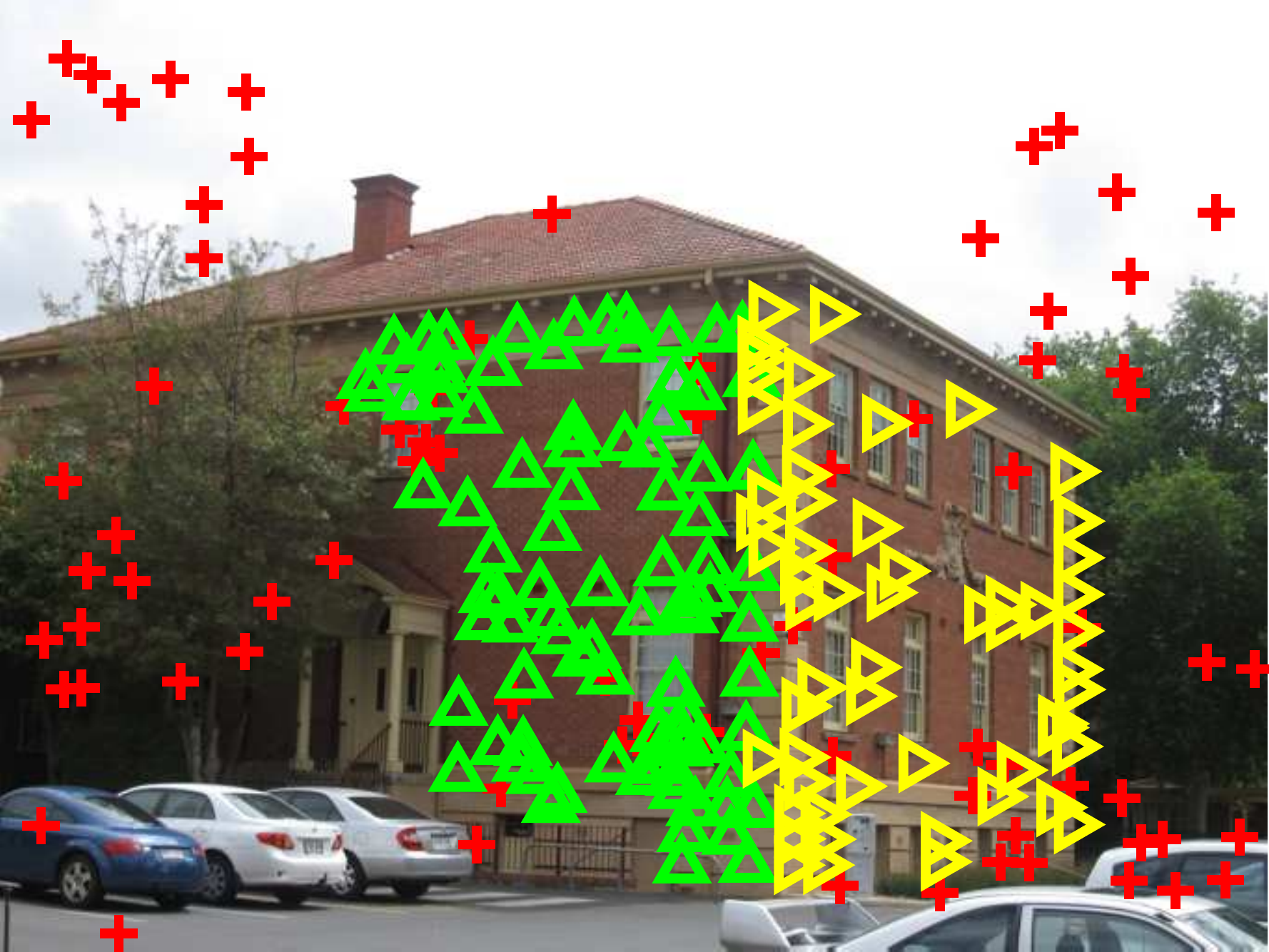}&
     \includegraphics[width=2.6cm]{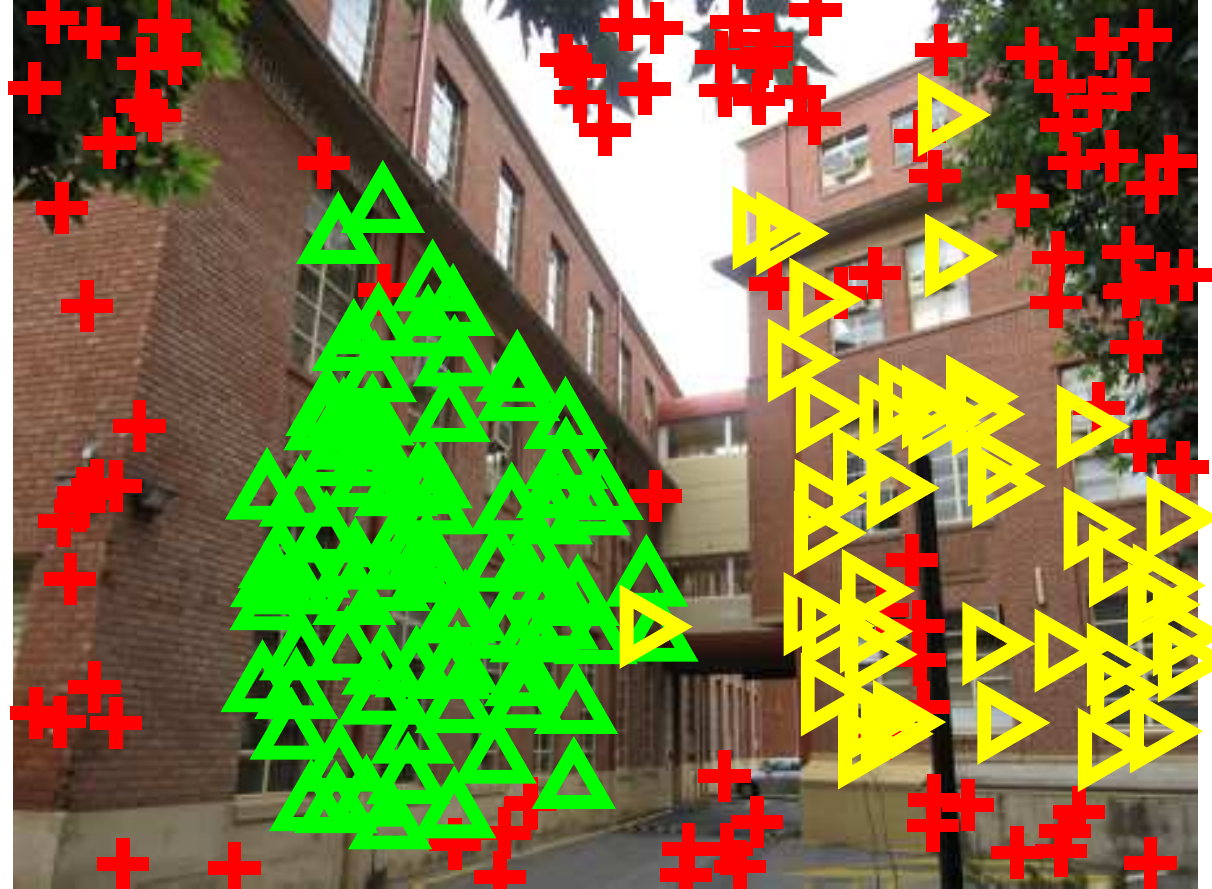} \\
\end{tabular}
\caption{\textbf{Some Qualitative Results of Motion and Planar Segmentation.} Qualitative results are shown for some examples from AdelaideRMF \cite{wong2011dynamic} dataset. Motion segmentation: Top-3 rows, Planar Segmentation: Bottom-3 rows. Point membership is color coded. Gross outliers are in red.}
\label{fig:quan_fun_hom}
\end{figure*}

%% file: vp_results.tex
\begin{table}[h]
\setlength{\tabcolsep}{2pt}
\centering
\caption{\textbf{Qualitative Evaluation on VP Estimation.} Notations are same as in table \ref{tbl:fun}.}
\label{tab:quan_vp}
\fontsize{8}{9}\selectfont
\begin{tabular}{l|c|c|c|c|c|c|c|c|}
\cline{2-9}
                            & \multicolumn{4}{c|}{York Dataset}                           & \multicolumn{4}{c|}{Toulouse Dataset}                            \\ \cline{2-9} 
                            & \multicolumn{2}{c|}{\textbf{CA(\%)}} & \multicolumn{2}{c|}{\textbf{Time(s)}} & \multicolumn{2}{c|}{\textbf{CA(\%)}} & \multicolumn{2}{c|}{\textbf{Time(s)}} \\ \cline{2-9} 
                            & \textbf{$\mu$}          & \textit{med}         & \textbf{$\mu$}           & \textit{med}          & \textbf{$\mu$ }         & \textit{med}         & \textbf{$\mu$}           & \textit{med}          \\ \hline
\multicolumn{1}{|l|}{\textbf{RPA}}   &95.4            &97.9            &04.4             &02.4             &55.3            &54.6            &00.8             &00.72             \\ 
\multicolumn{1}{|l|}{\textbf{Cov}}   &95.6            &97.4            &01.24             &00.26             &51.8            &50.0            &00.04             &00.07             \\ 
\multicolumn{1}{|l|}{\textbf{L1-NMF}}   &94.1            &96.7            &00.71             &00.31             &74.1            &75.0            &00.07             &00.54             \\ 
\multicolumn{1}{|l|}{\textbf{DGSAC-G}} &96.0            &98.0            &01.29             &01.25             &92.1            &95.9            &00.03             &00.03             \\
\multicolumn{1}{|l|}{\textbf{DGSAC-O}} &95.8            &97.9            &01.30             &01.26             &91.9            &95.6            &00.05             &00.05             \\ \hline
\end{tabular}
\end{table}

%% file: qual_vp_tvp.tex
\begin{figure*}
\centering
\setlength{\tabcolsep}{2pt}
\resizebox{\textwidth}{!}{%
\begin{tabular}{cccccc}
          \multicolumn{6}{c}{\textbf{York Urban Vanishing Point Dataset} \cite{york}}  \\
 \rotatebox{90}{\hspace{0.4cm}  \textbf{GT}} &    \includegraphics[width=2.1cm]{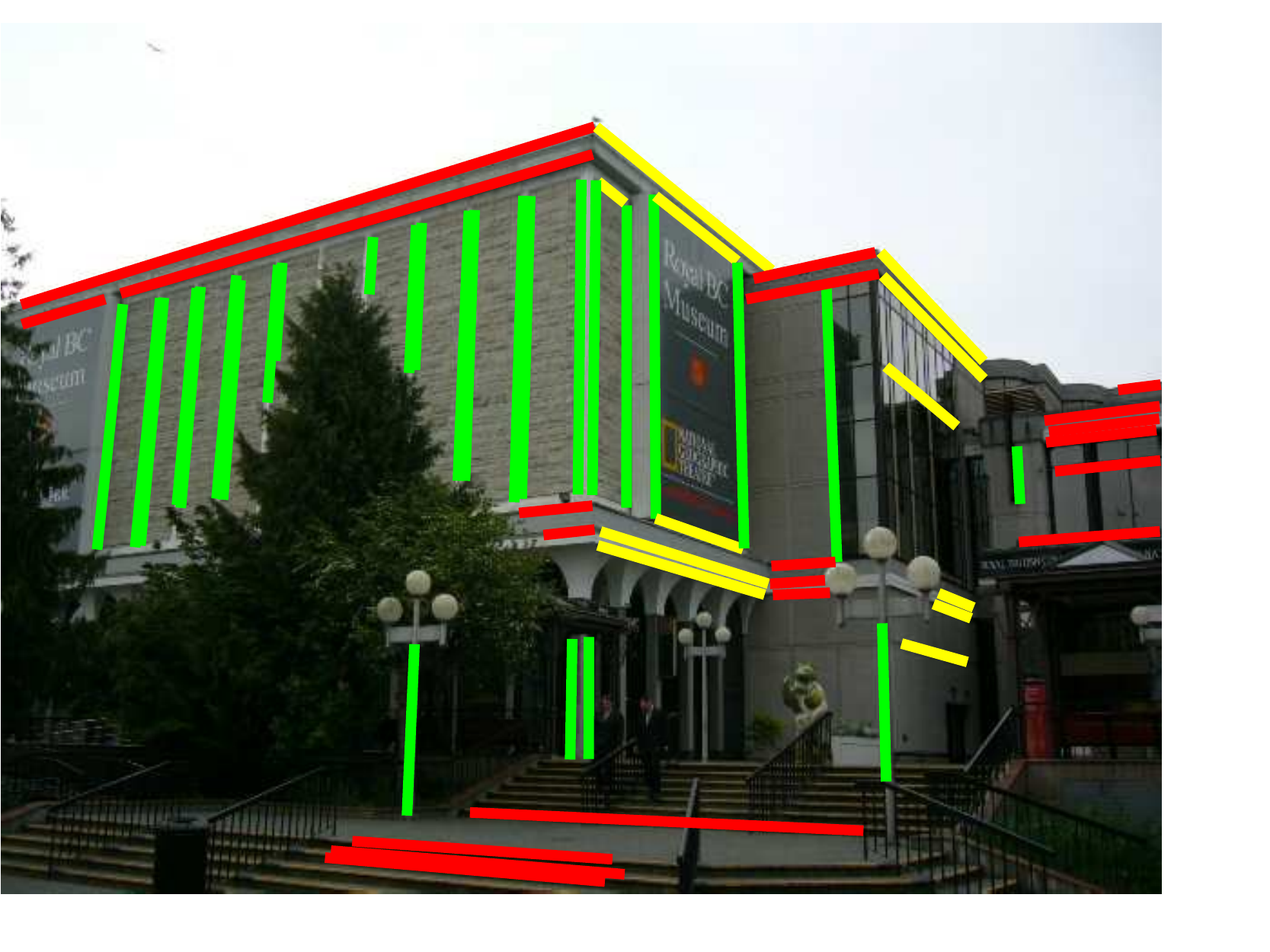}&
     \includegraphics[width=2.1cm]{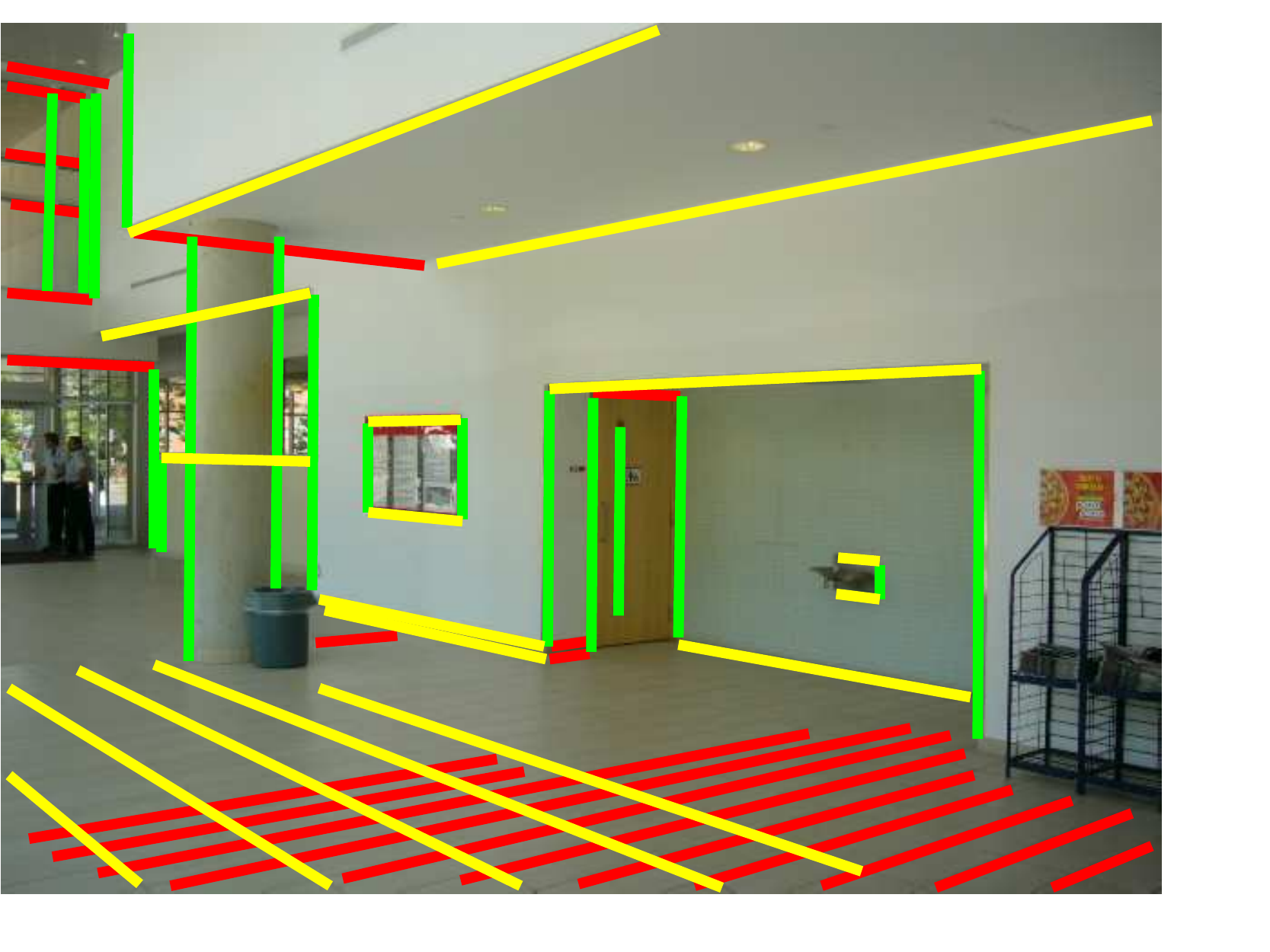}&
     \includegraphics[width=2.1cm]{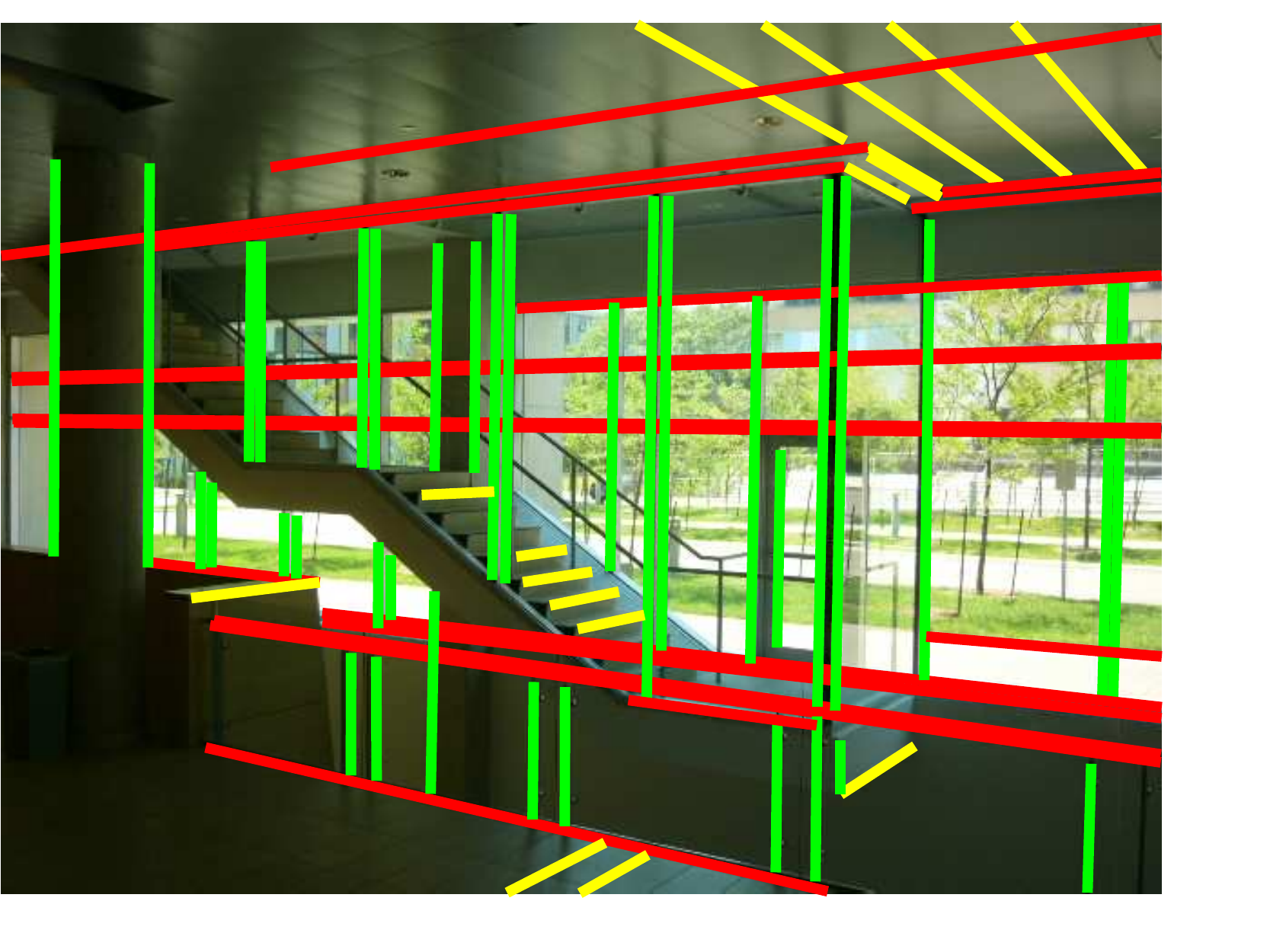}&
     \includegraphics[width=2.1cm]{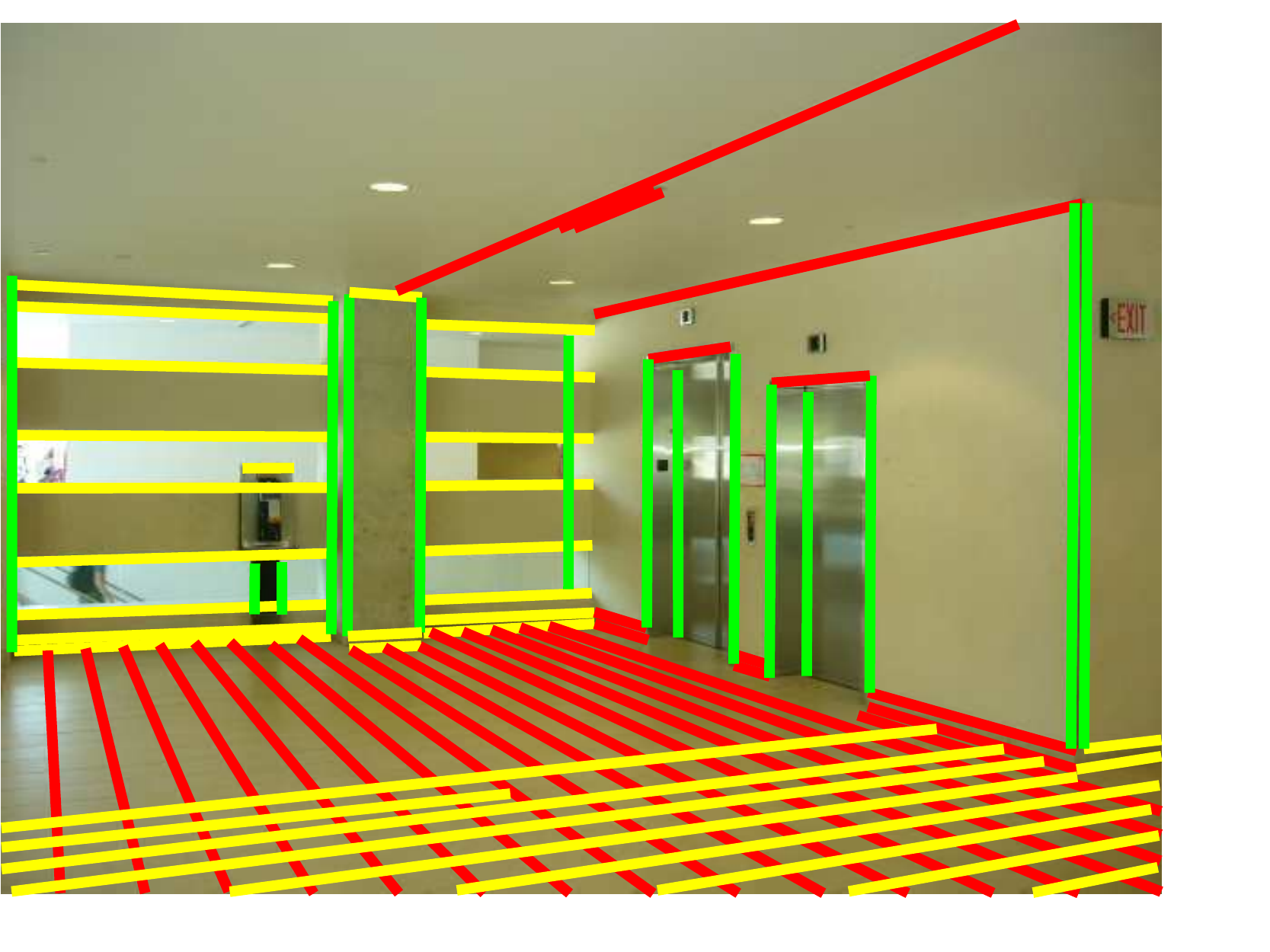}&
     \includegraphics[width=2.1cm]{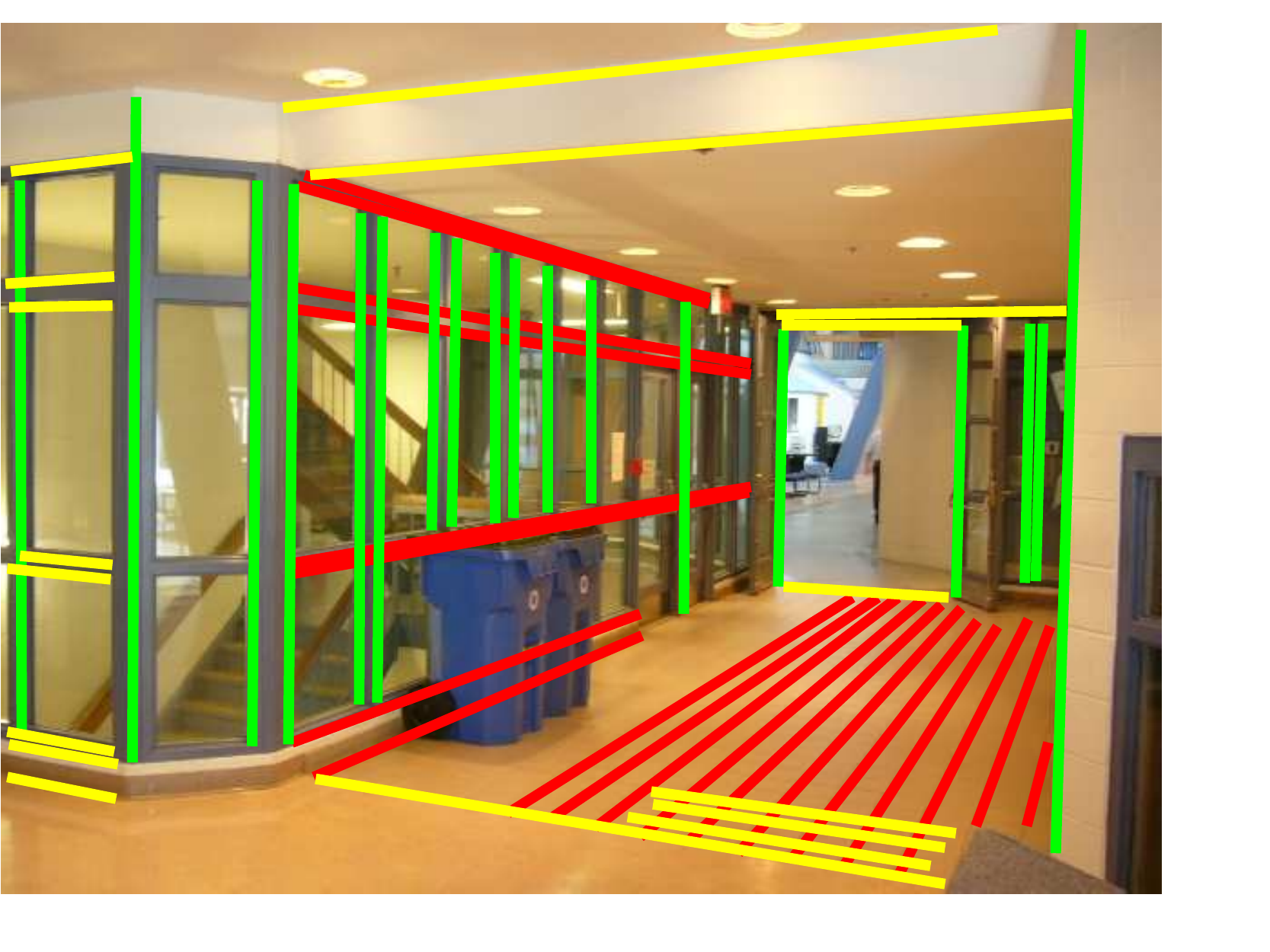} \\
  \rotatebox{90}{\hspace{.1cm}  \textbf{DGSAC-G}}   & \includegraphics[width=2.1cm]{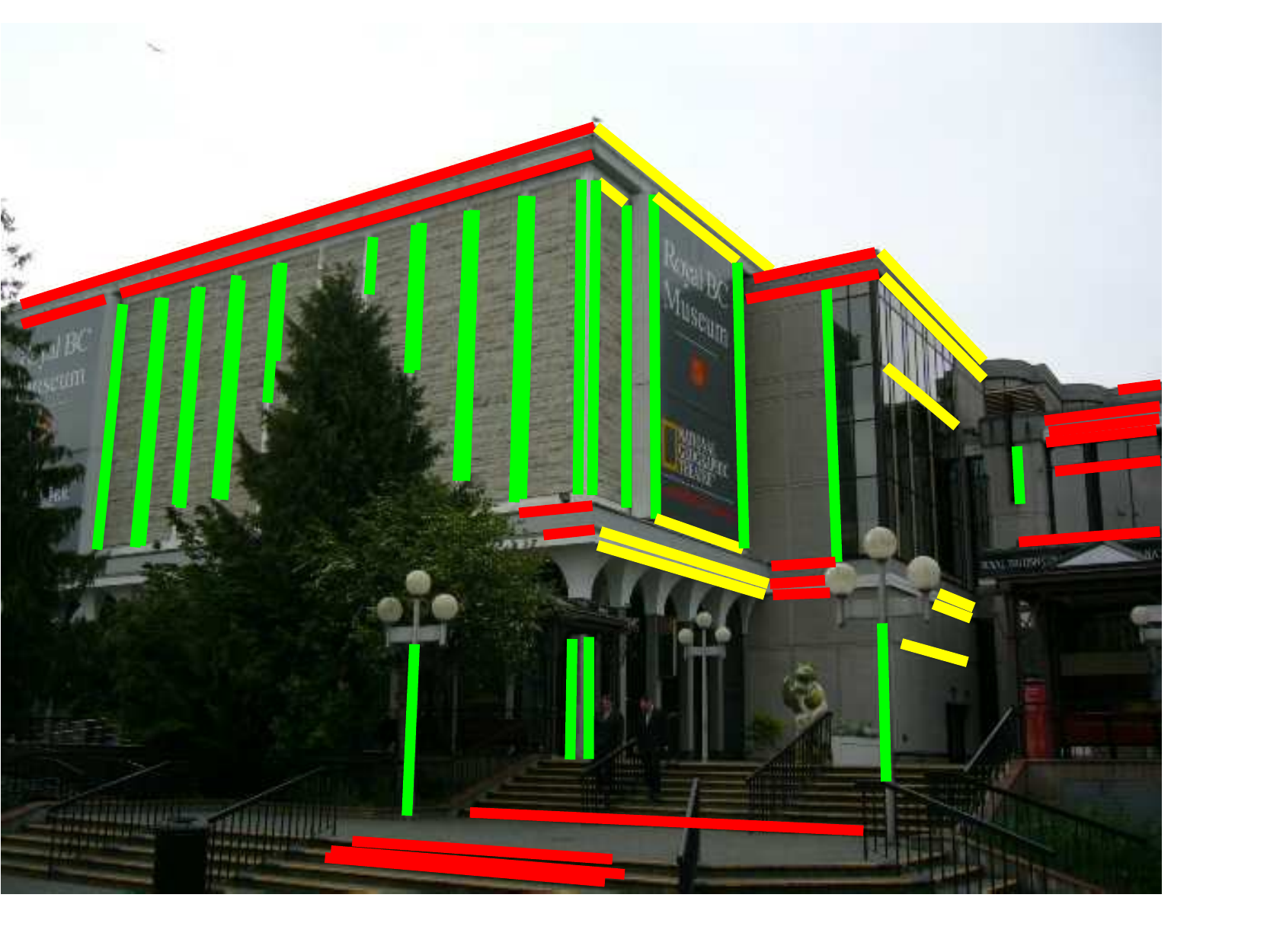}&
     \includegraphics[width=2.1cm]{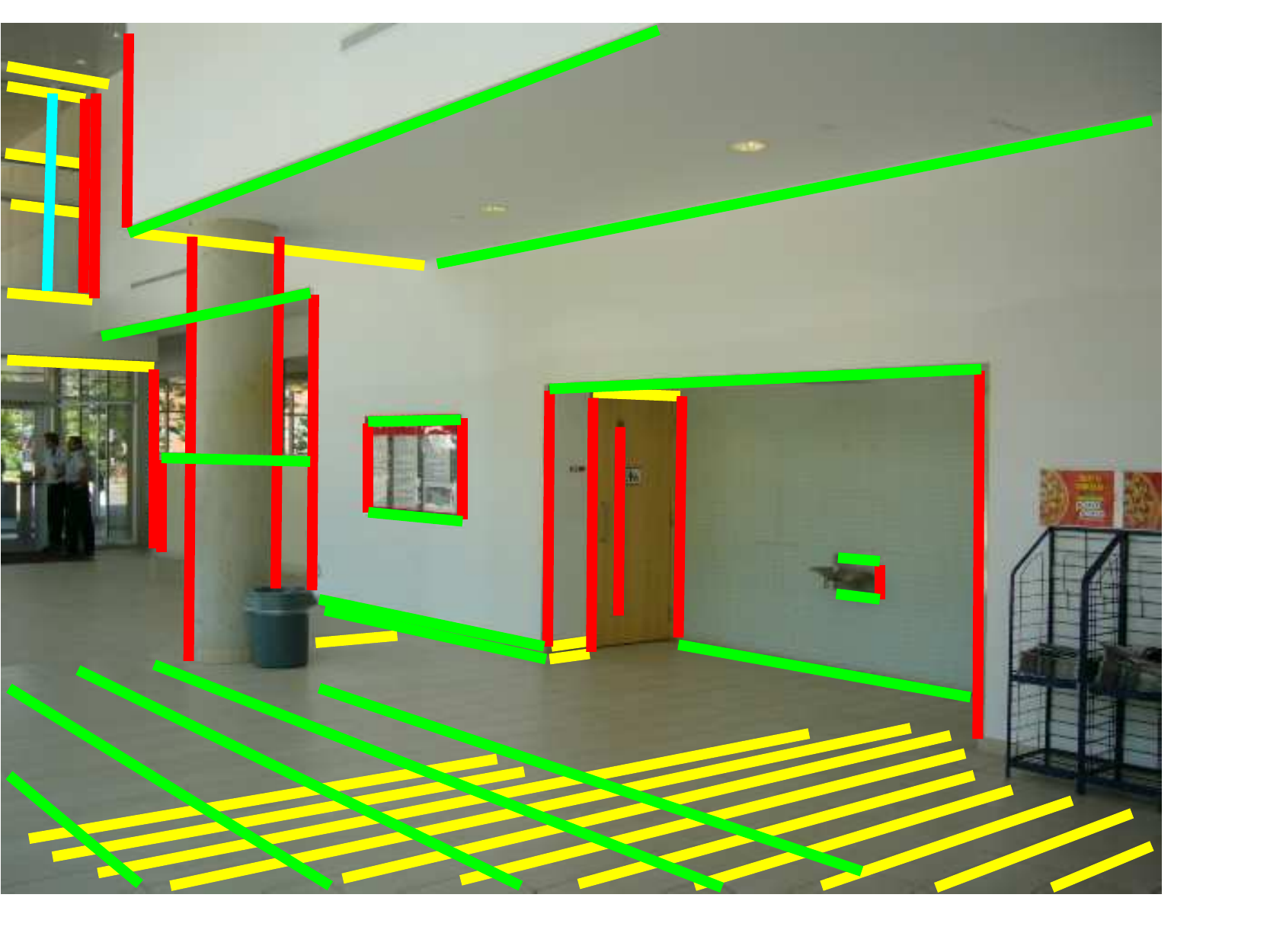}&
     \includegraphics[width=2.1cm]{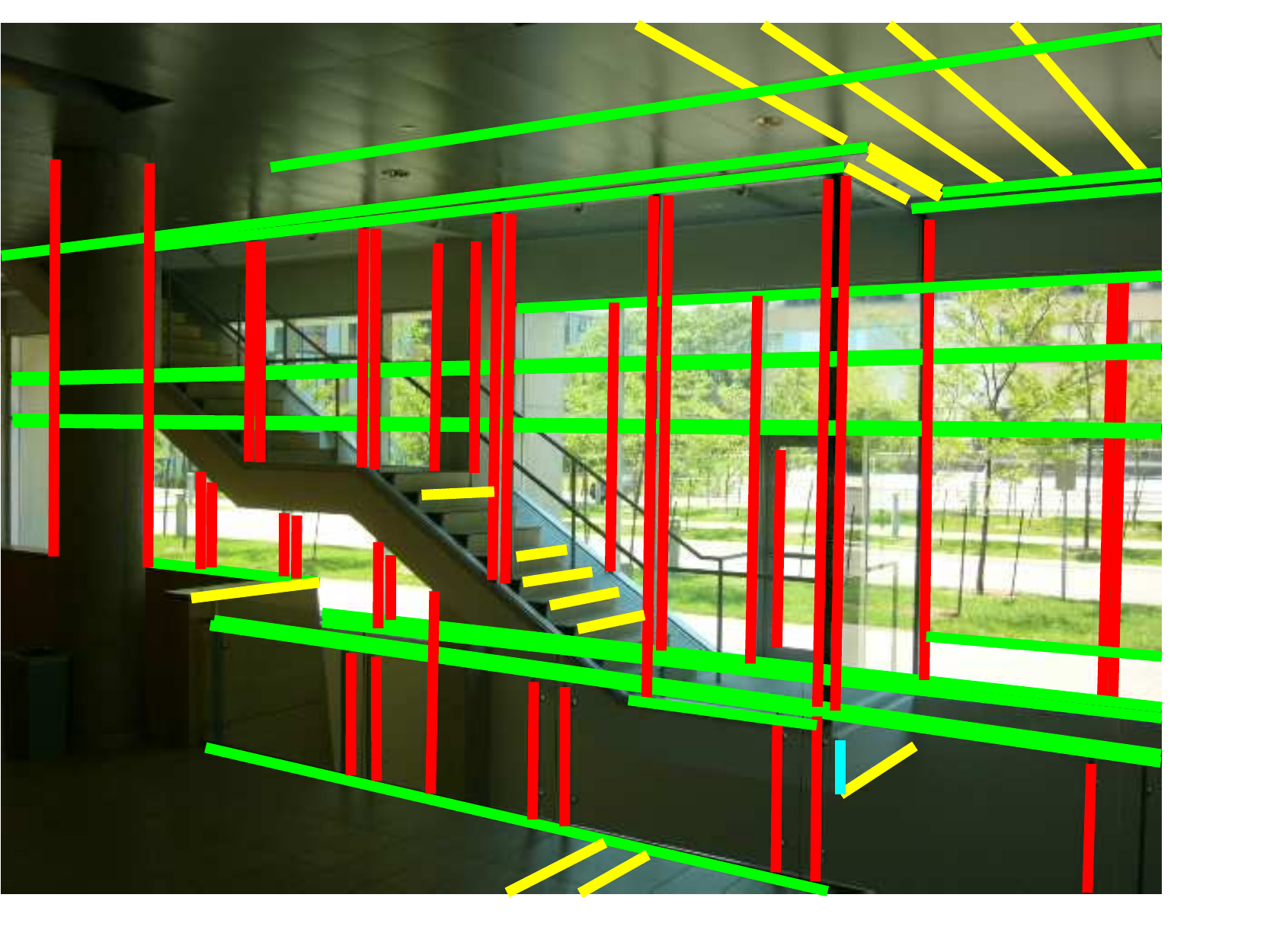}&
     \includegraphics[width=2.1cm]{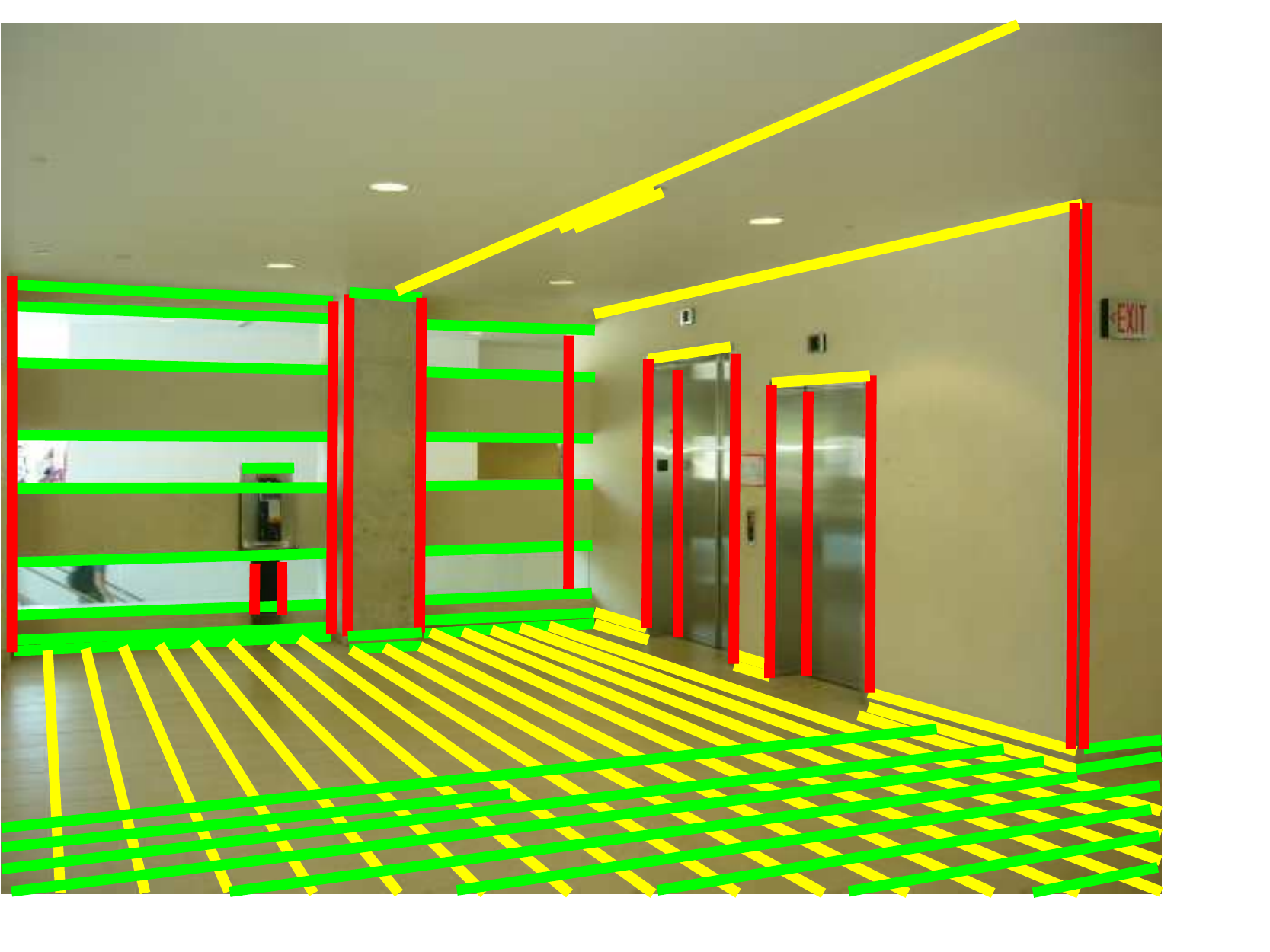}&
     \includegraphics[width=2.1cm]{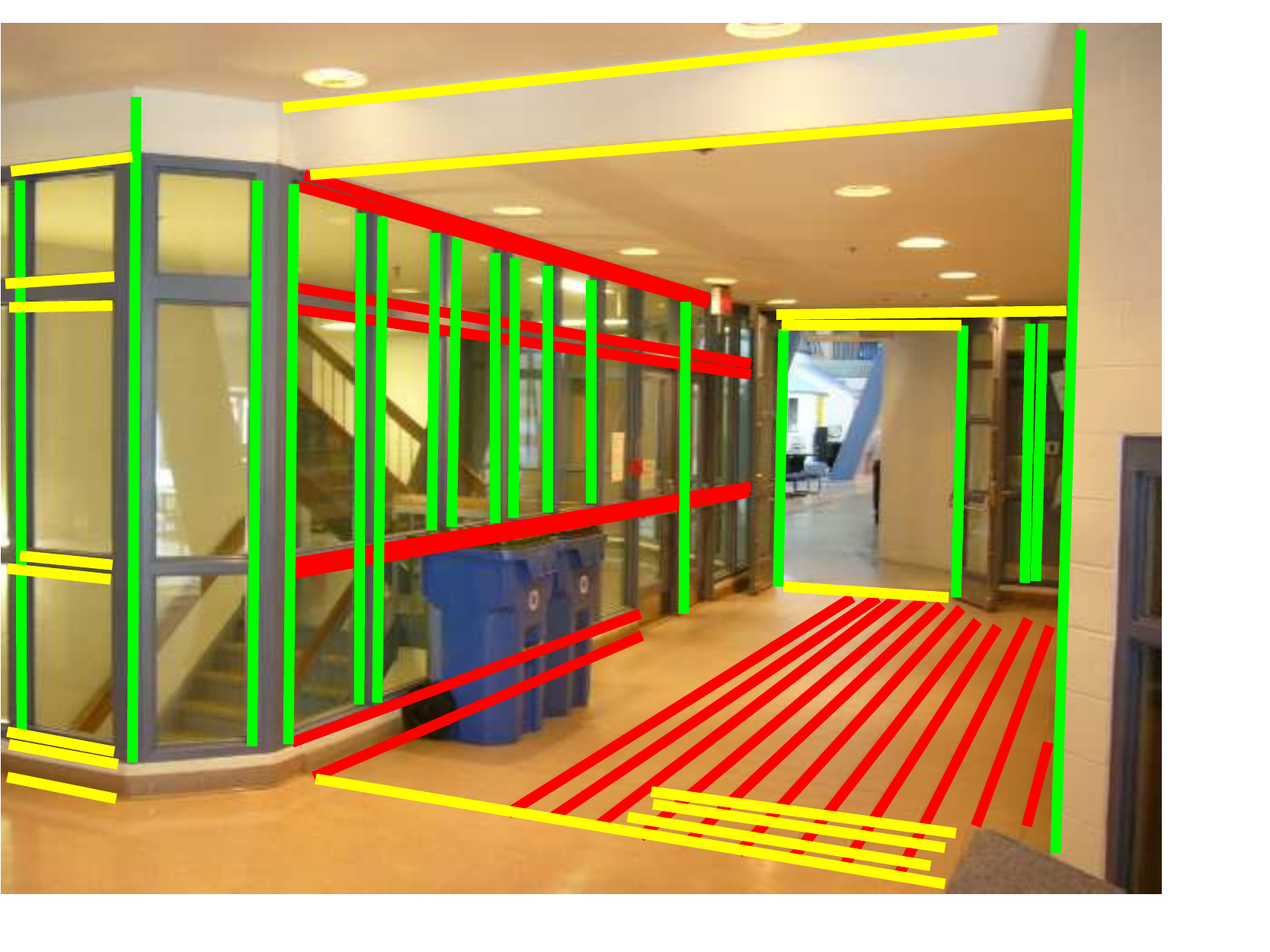} \\
   \rotatebox{90}{\hspace{.1cm}  \textbf{DGSAC-O}}      &\includegraphics[width=2.1cm]{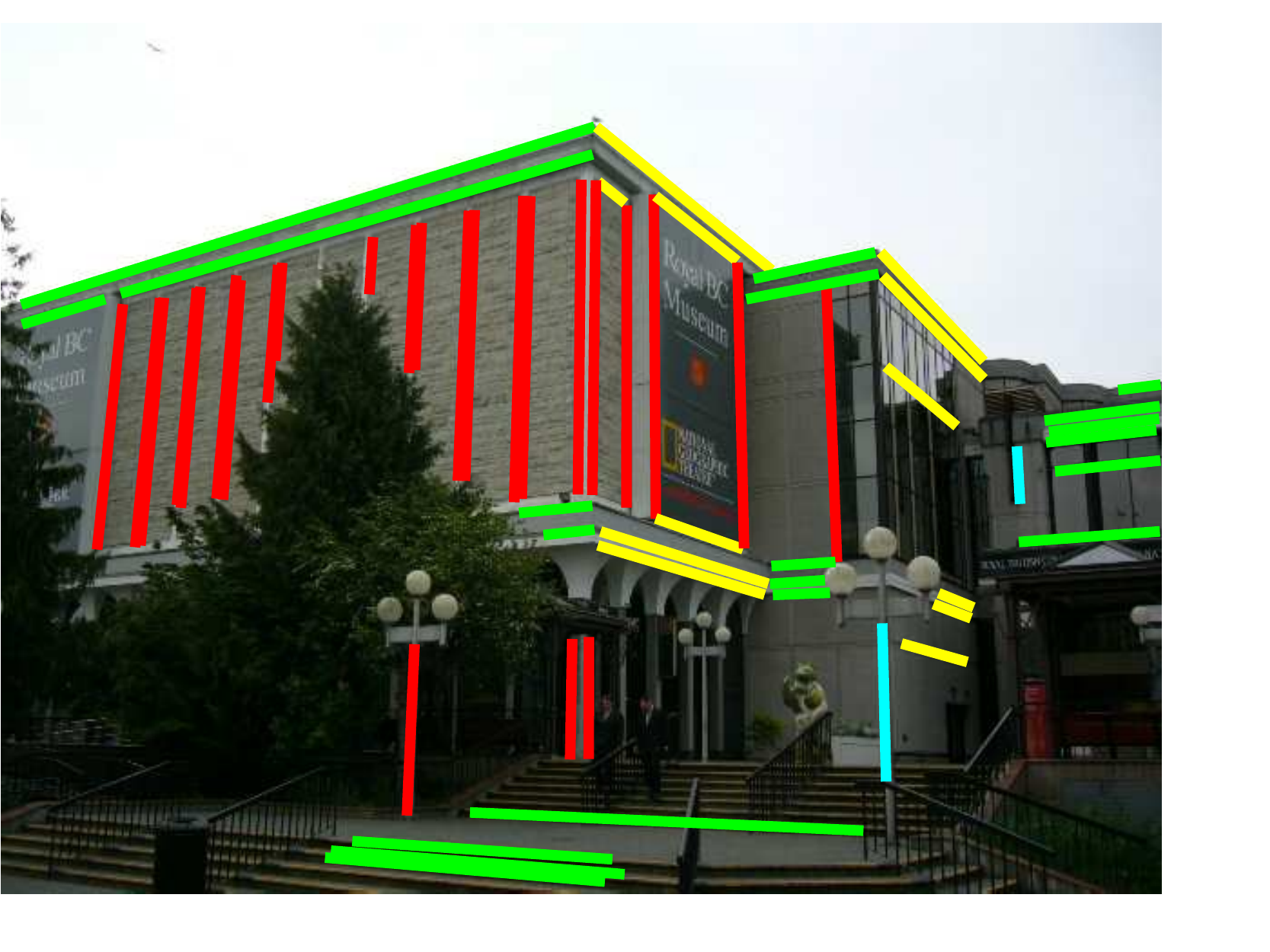}&
     \includegraphics[width=2.1cm]{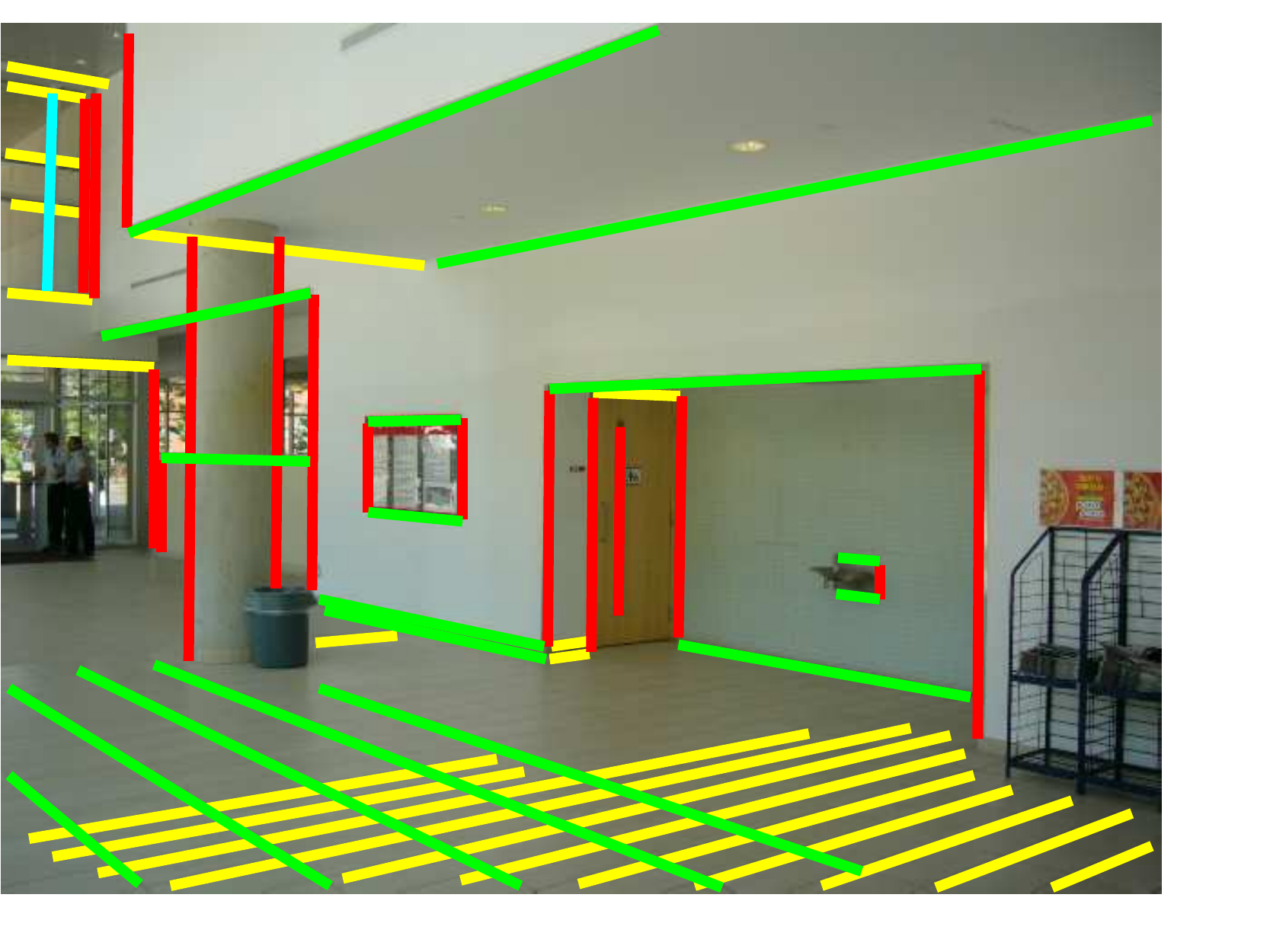}&
     \includegraphics[width=2.1cm]{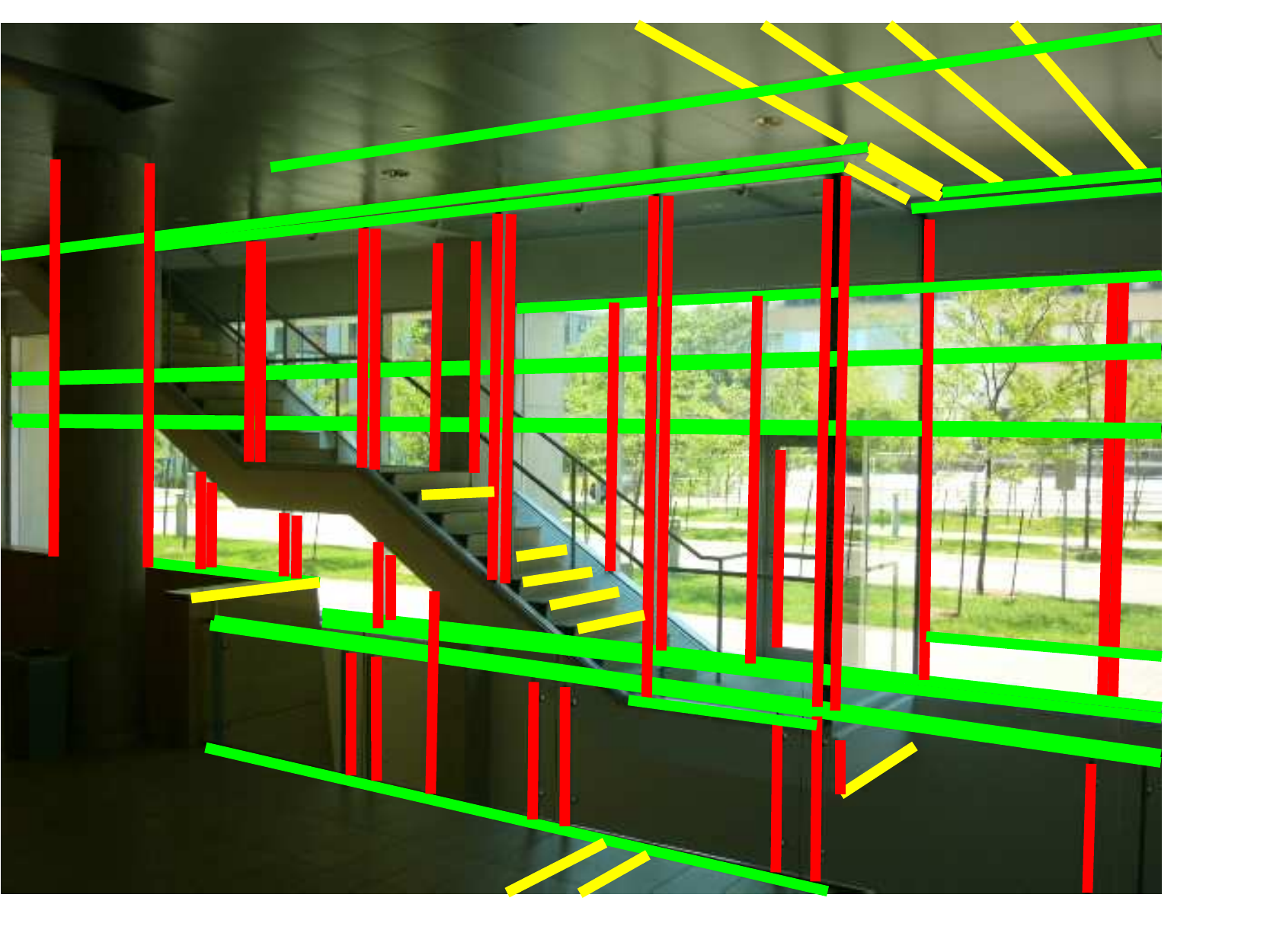}&
     \includegraphics[width=2.1cm]{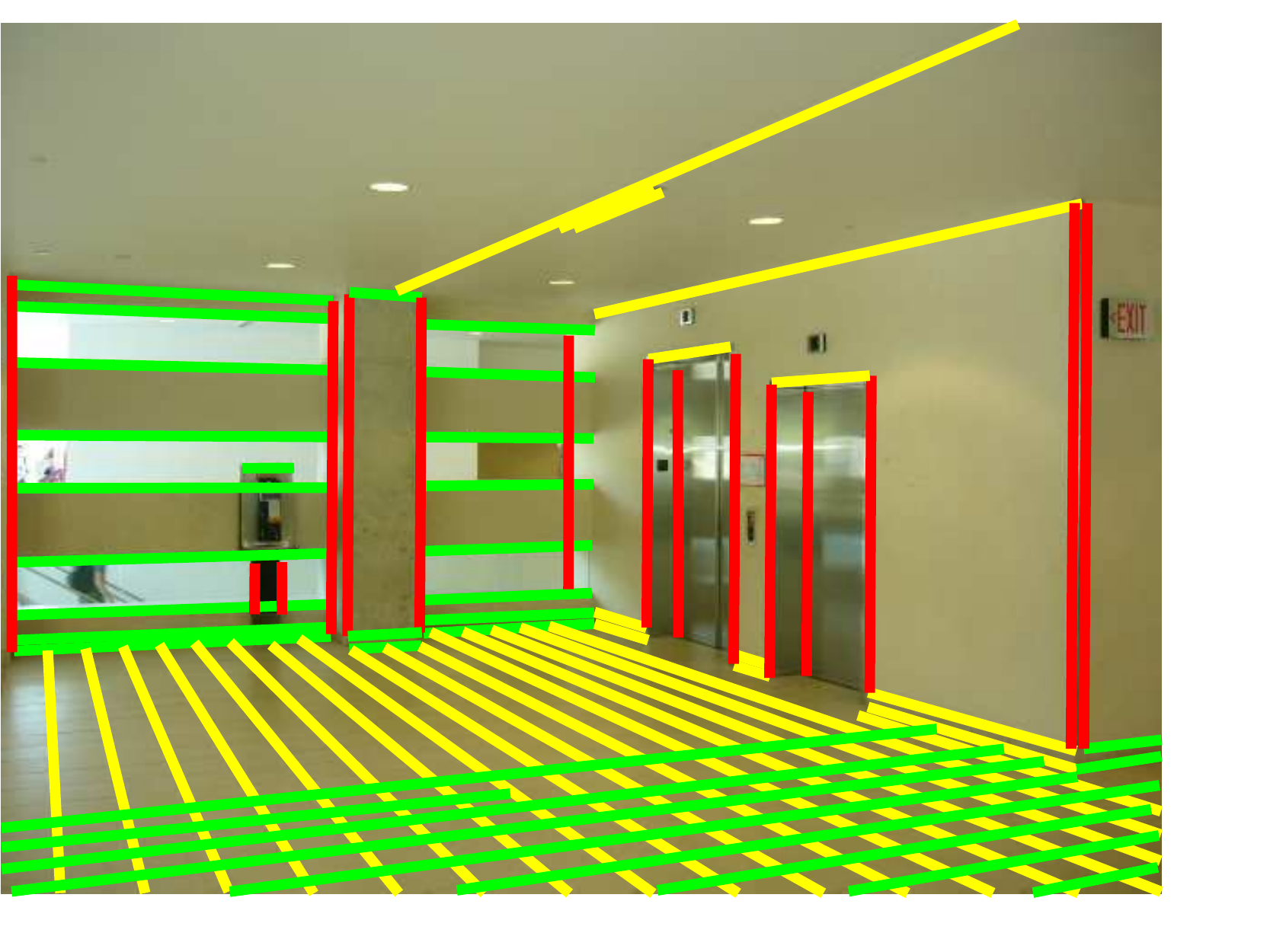}&
     \includegraphics[width=2.1cm]{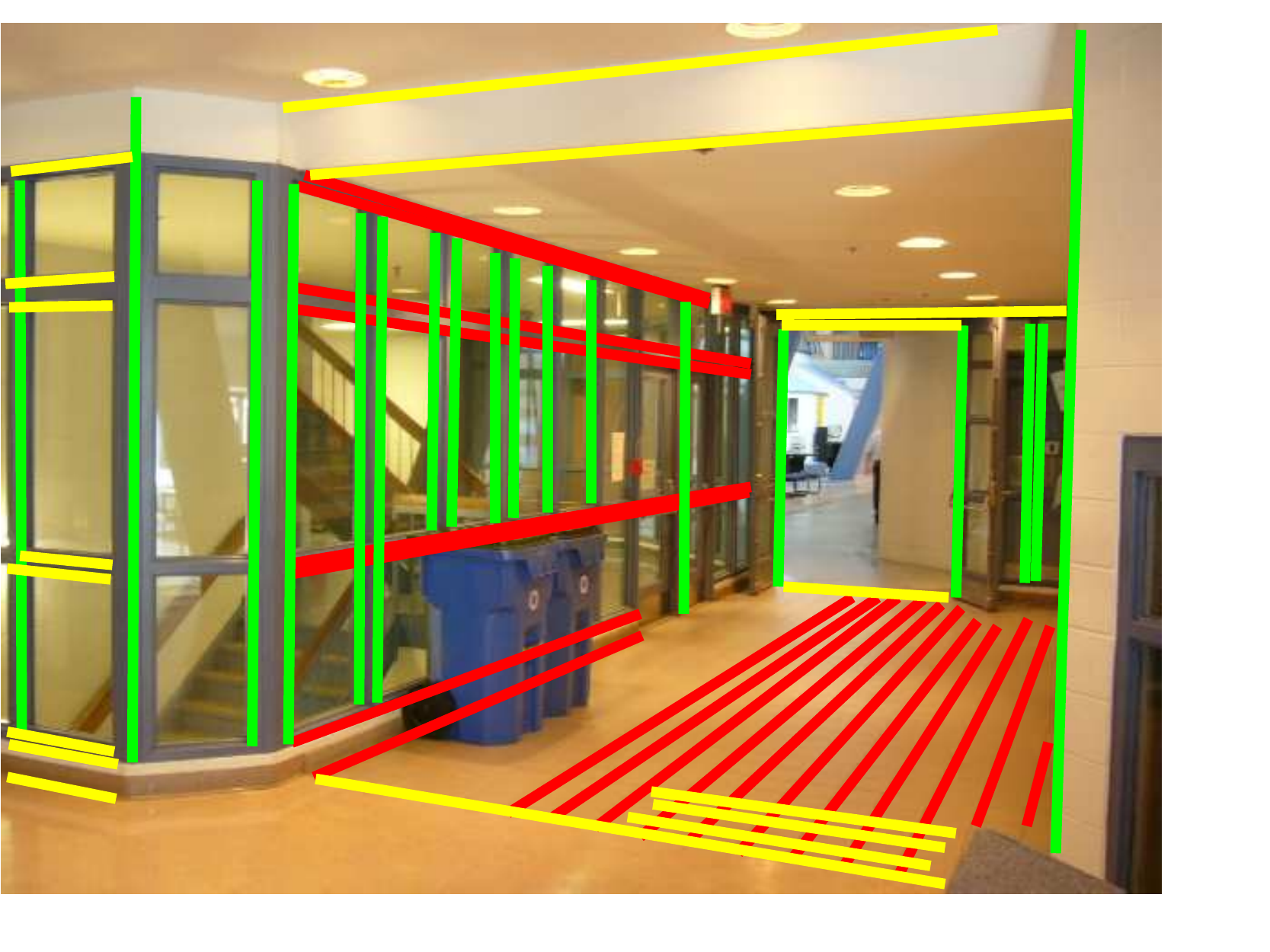} \\
     \hline
          \multicolumn{6}{c}{\textbf{Toulouse Vanishing Point Dataset} \cite{tvp}}  \\
     \rotatebox{90}{\hspace{0.4cm}  \textbf{GT}} &    \includegraphics[width=2.1cm]{qualitative_imgs/york_1_gt.pdf}&
     \includegraphics[width=2.1cm]{qualitative_imgs/york_2_gt.pdf}&
     \includegraphics[width=2.1cm]{qualitative_imgs/york_3_gt.pdf}&
     \includegraphics[width=2.1cm]{qualitative_imgs/york_4_gt.pdf}&
     \includegraphics[width=2.1cm]{qualitative_imgs/york_5_gt.pdf} \\
  \rotatebox{90}{\hspace{.1cm}  \textbf{DGSAC-G}}   & \includegraphics[width=2.1cm]{qualitative_imgs/york_1_grdy.pdf}&
     \includegraphics[width=2.1cm]{qualitative_imgs/york_2_grdy.pdf}&
     \includegraphics[width=2.1cm]{qualitative_imgs/york_3_grdy.pdf}&
     \includegraphics[width=2.1cm]{qualitative_imgs/york_4_grdy.pdf}&
     \includegraphics[width=2.1cm]{qualitative_imgs/york_5_grdy.pdf} \\
   \rotatebox{90}{\hspace{.1cm}  \textbf{DGSAC-O}}      &\includegraphics[width=2.1cm]{qualitative_imgs/york_1_opt.pdf}&
     \includegraphics[width=2.1cm]{qualitative_imgs/york_2_opt.pdf}&
     \includegraphics[width=2.1cm]{qualitative_imgs/york_3_opt.pdf}&
     \includegraphics[width=2.1cm]{qualitative_imgs/york_4_opt.pdf}&
     \includegraphics[width=2.1cm]{qualitative_imgs/york_5_opt.pdf} \\
\end{tabular}
} \caption{\textbf{Sample Qualitative Results} on York Urban and Toulouse Vanishing Point dataset.}
\label{fig:qual_vp_tvp}
\end{figure*}

%% file: qual_plane.tex
\begin{figure*}
\centering
\setlength{\tabcolsep}{2pt}
\resizebox{\textwidth}{!}{%
\begin{tabular}{ccccccc}
& Data & J-Linkage & T-Linkage & RansaCov & DGSAC-G & DGSAC-O \\
 \rotatebox{90}{\hspace{.2cm}  {CastelVechio}} &  
     \includegraphics[width=2.99cm]{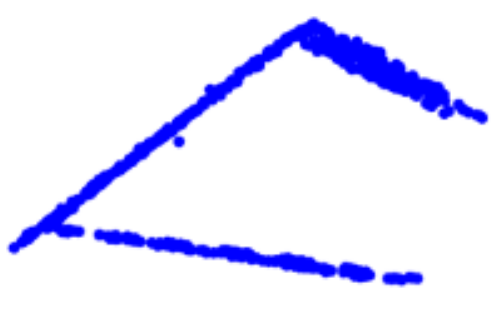}&\hspace{-0.7cm}
     \includegraphics[width=2.99cm]{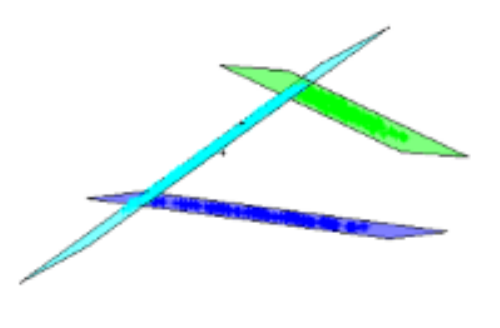}&\hspace{-0.7cm}
     \includegraphics[width=2.99cm]{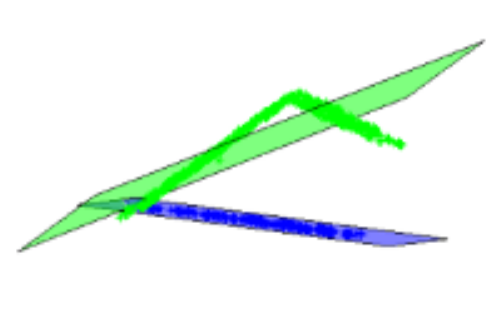}&\hspace{-0.7cm}
     \includegraphics[width=2.99cm]{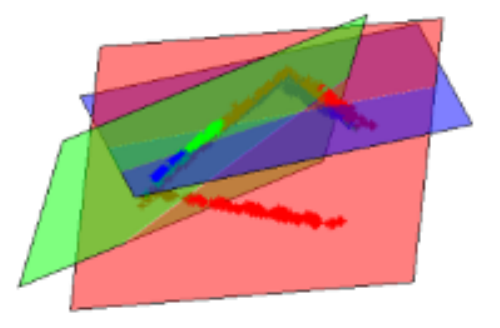}&\hspace{-0.7cm}
     \includegraphics[width=2.99cm]{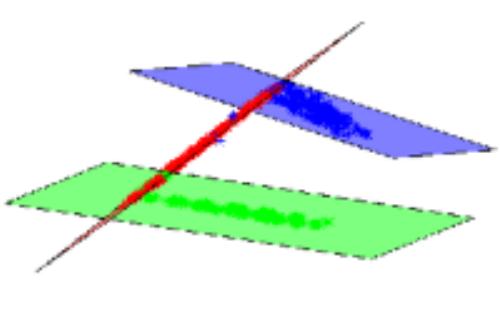}&\hspace{-0.5cm}
     \includegraphics[width=2.99cm]{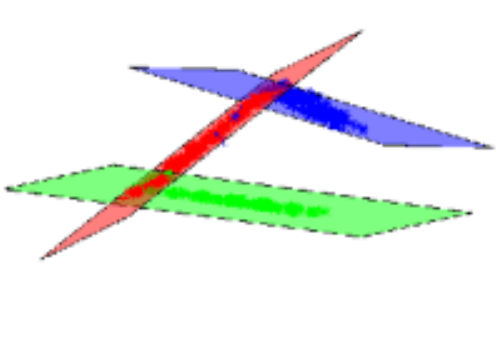}\\
  \rotatebox{90}{\hspace{.2cm}  {PozzoVeggiani}}   &  \includegraphics[width=2.99cm]{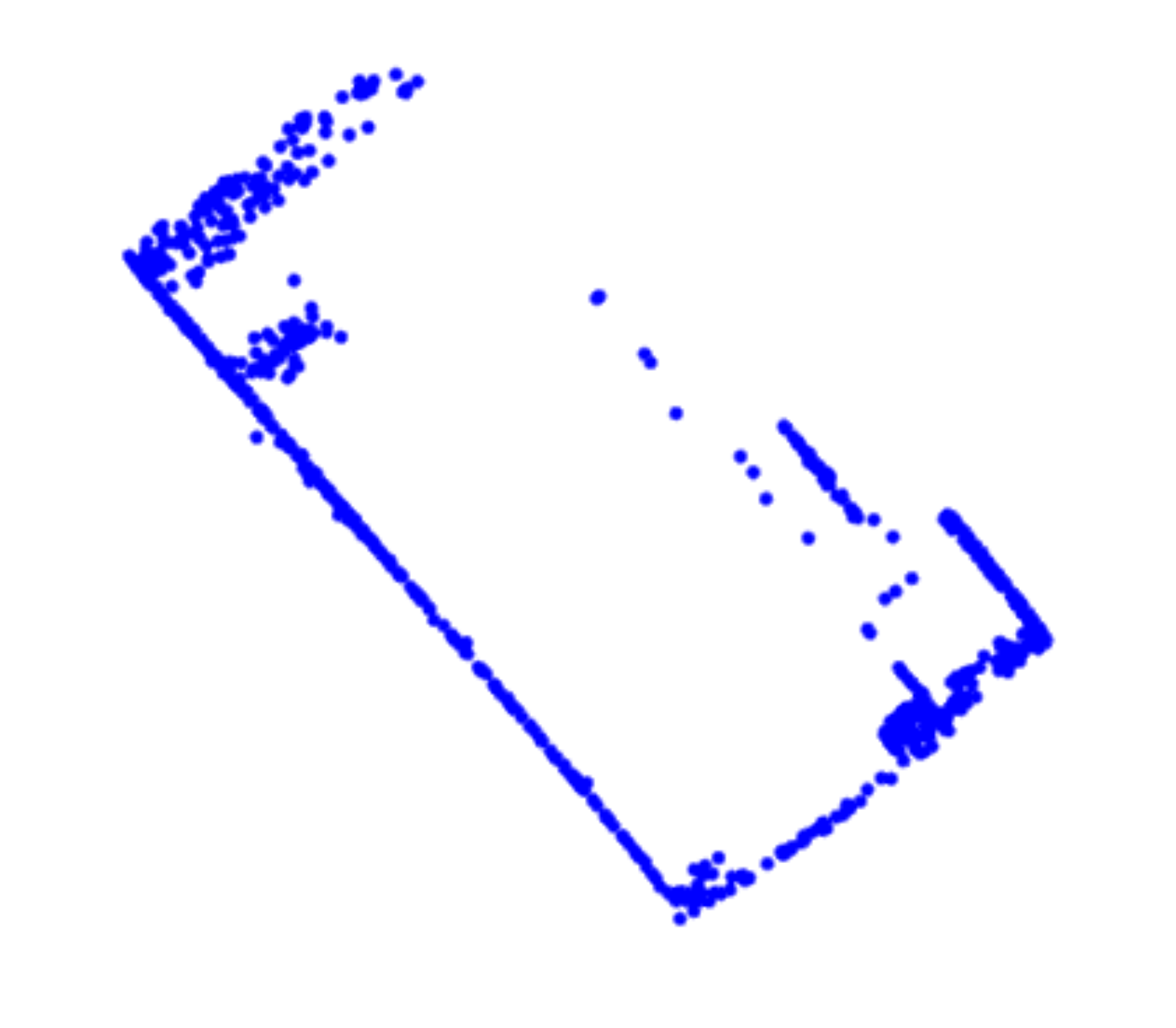}&\hspace{-0.7cm}
     \includegraphics[width=2.99cm]{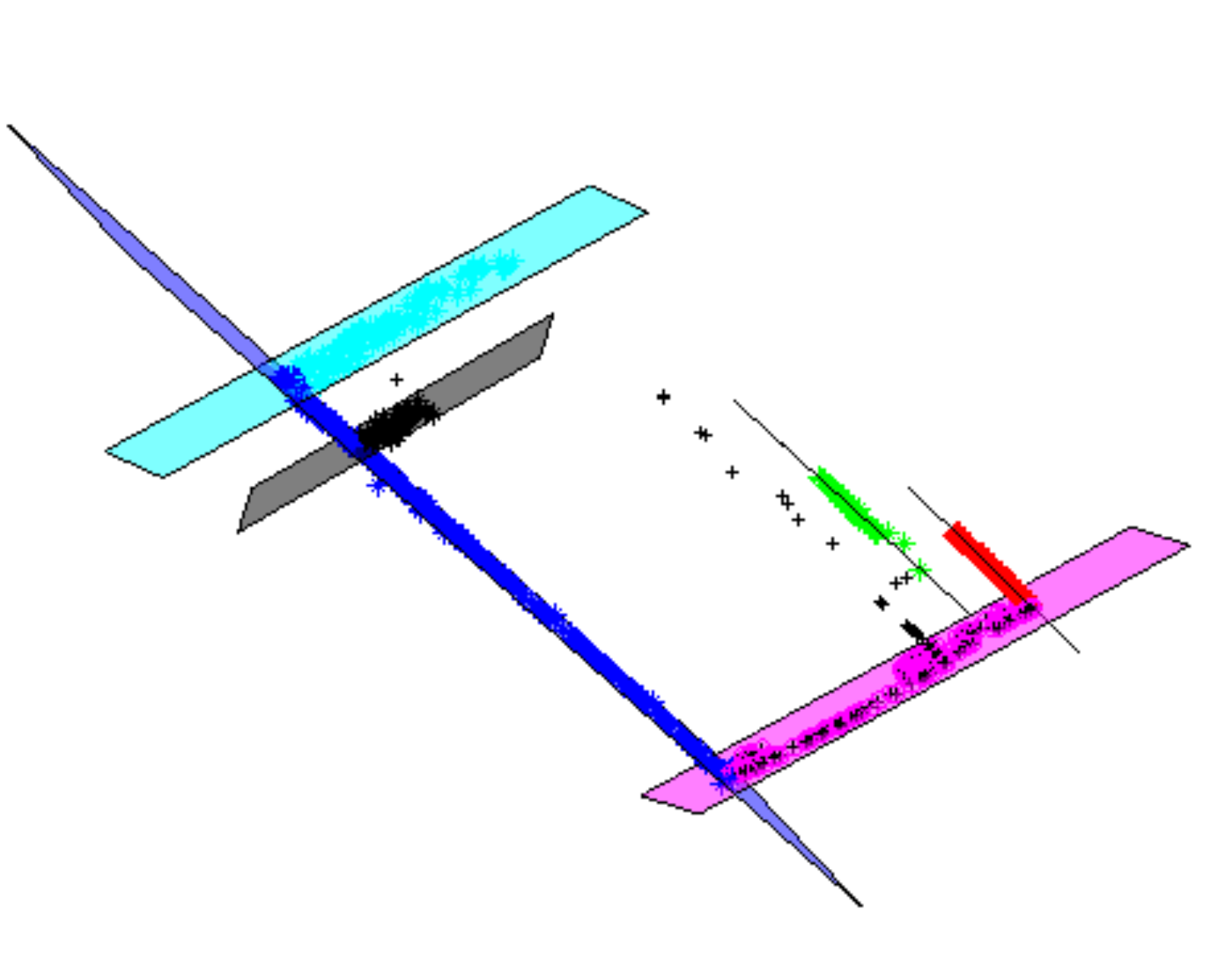}&\hspace{-0.7cm}
     \includegraphics[width=2.99cm]{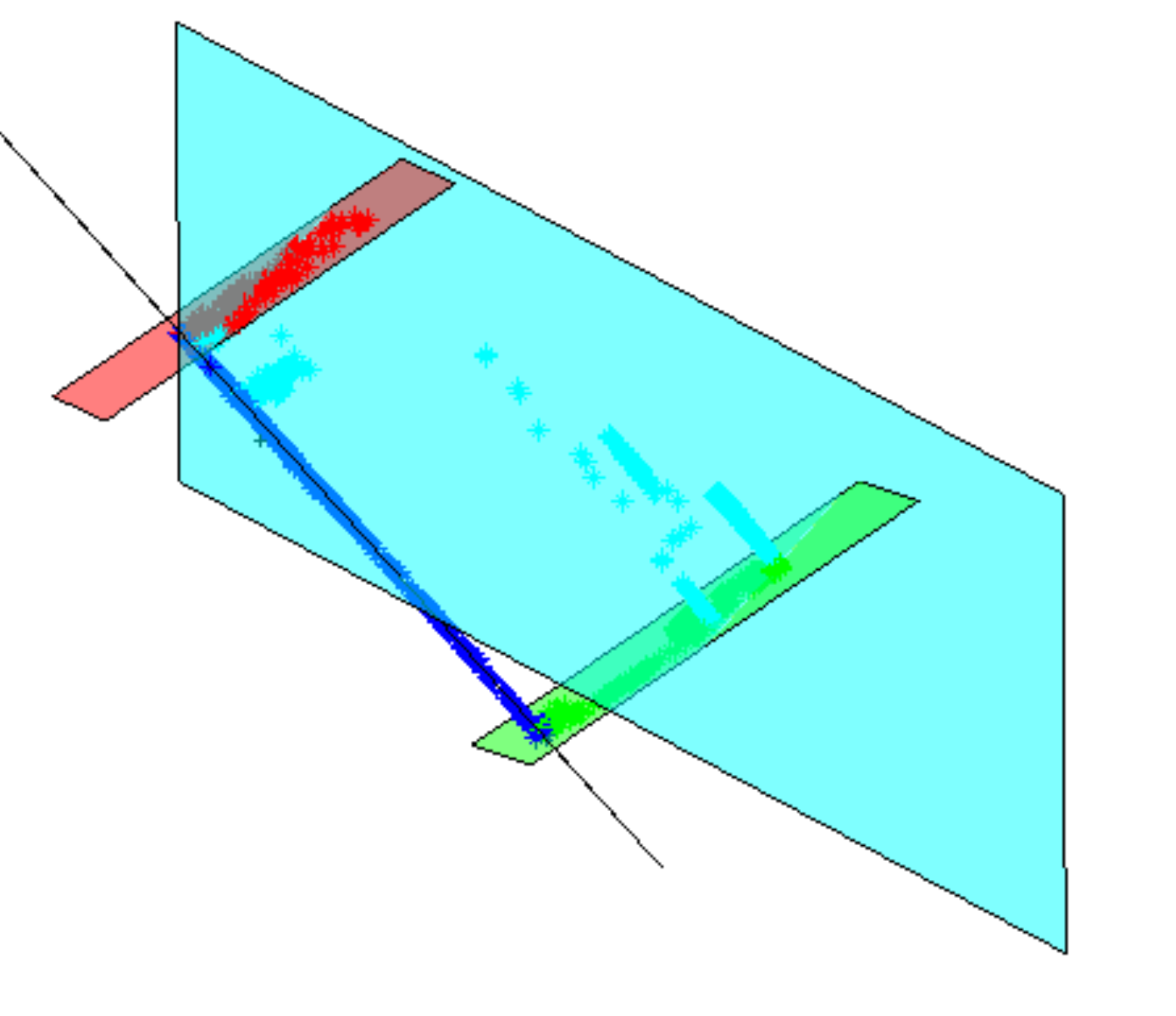}&\hspace{-0.7cm}
     \includegraphics[width=2.99cm]{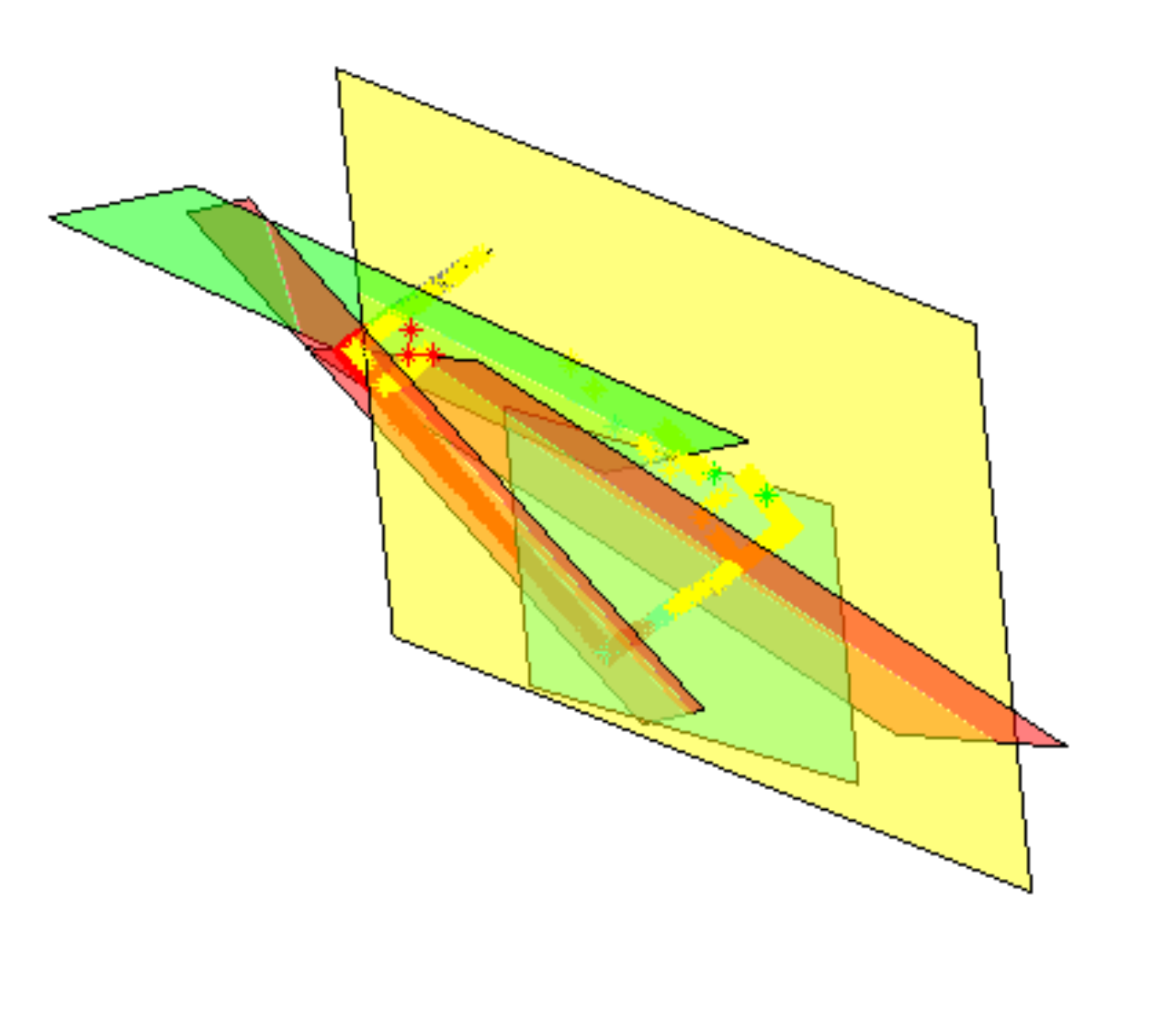}&\hspace{-0.7cm}
     \includegraphics[width=2.99cm]{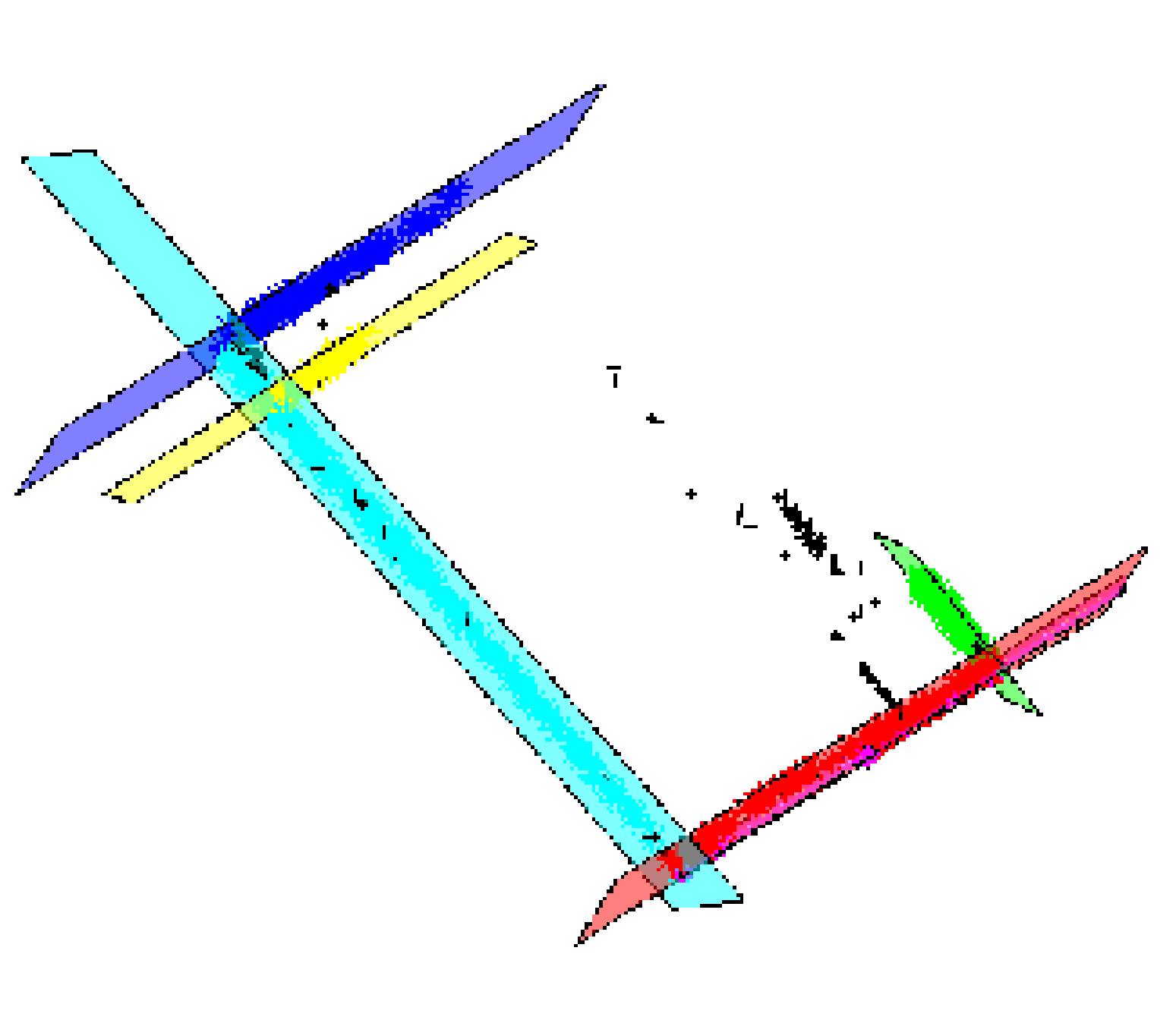}& \hspace{-0.5cm}
     \includegraphics[width=2.99cm]{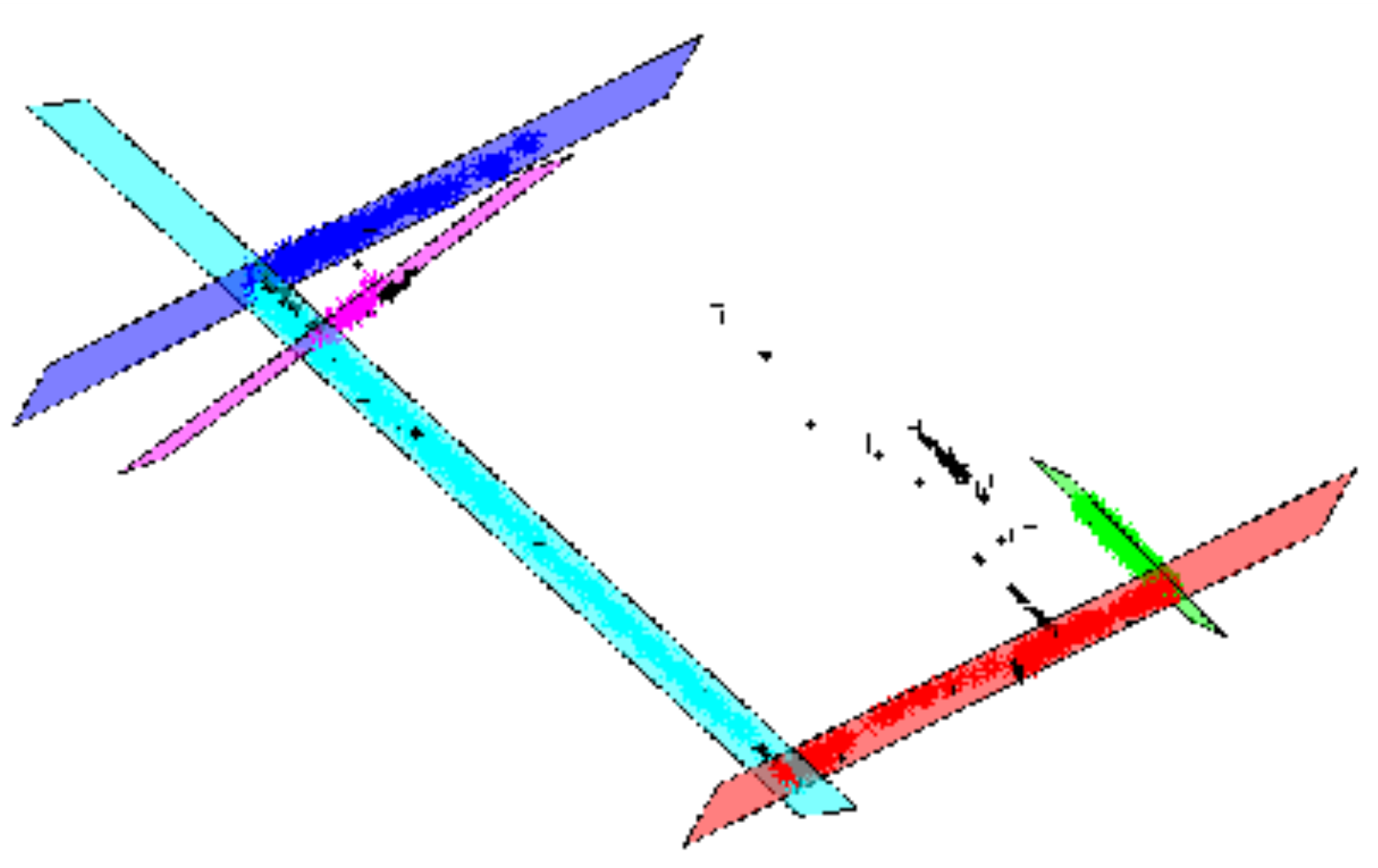}\\
\end{tabular}
}
\caption{\textbf{Multiple Plane Fitting to 3D Point Cloud.} Dataset: \textit{CastelVecchio} and \textit{PozzoVeggiani} examples from SAMANTHA dataset \cite{SAMANTHA}. Point membership is color coded.}
\label{fig:qual_plane}
\end{figure*}

%% file: qual_line.tex
\begin{figure*}
\centering
\setlength{\tabcolsep}{2pt}
\resizebox{\textwidth}{!}{%
\begin{tabular}{ccccccc}
&\hspace{-0.1cm}  Data & J-Linkage &   T-Linkage &   RansaCov &   DGSAC-G & \hspace{-0.1cm}  DGSAC-O \\
 \rotatebox{90}{\hspace{1.2cm}  Star5, O=50\% } &\hspace{-0.1cm}  
     \includegraphics[width=3.2cm,height=3.2cm]{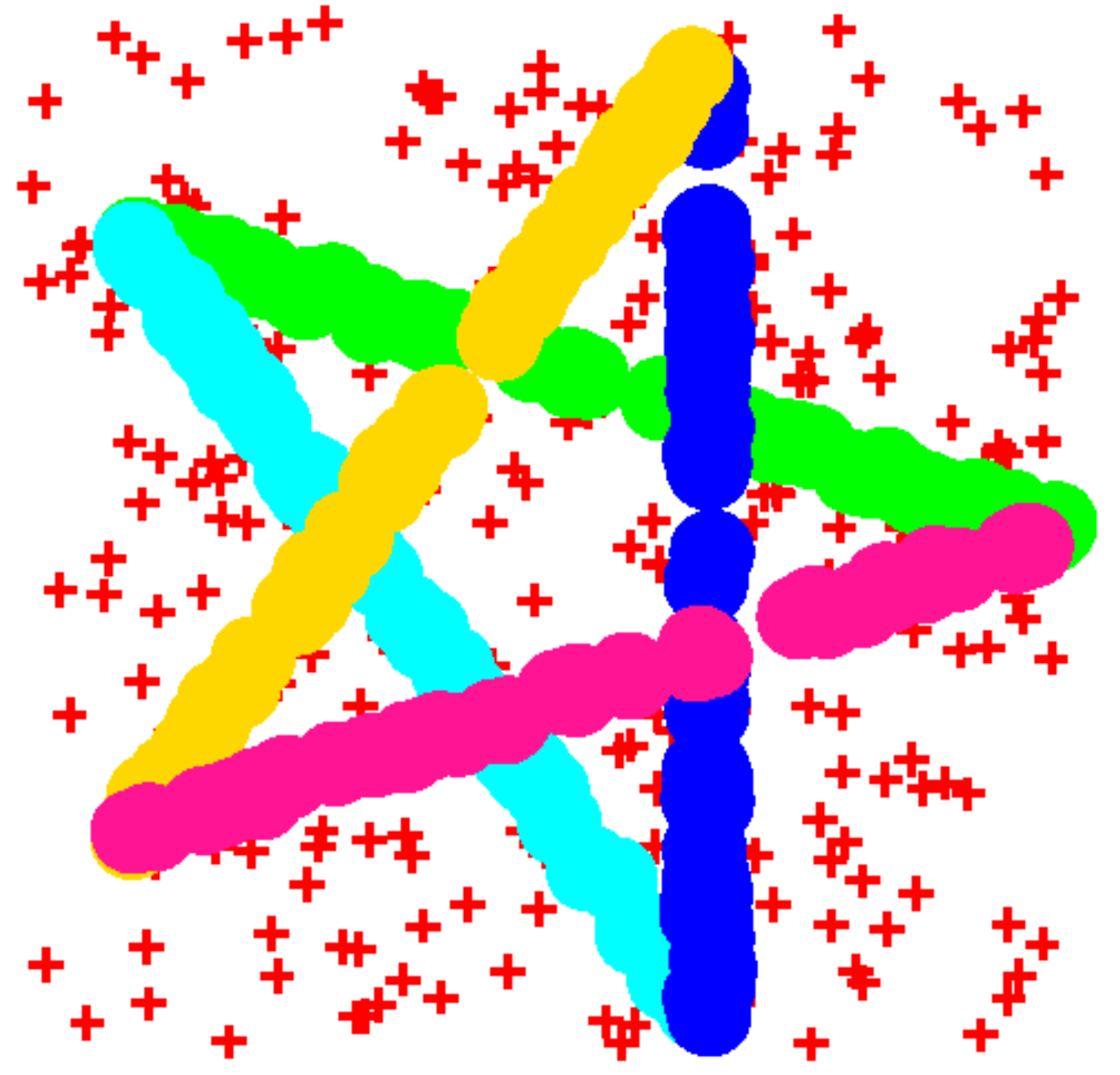}&
     \includegraphics[width=3.2cm,height=3.2cm]{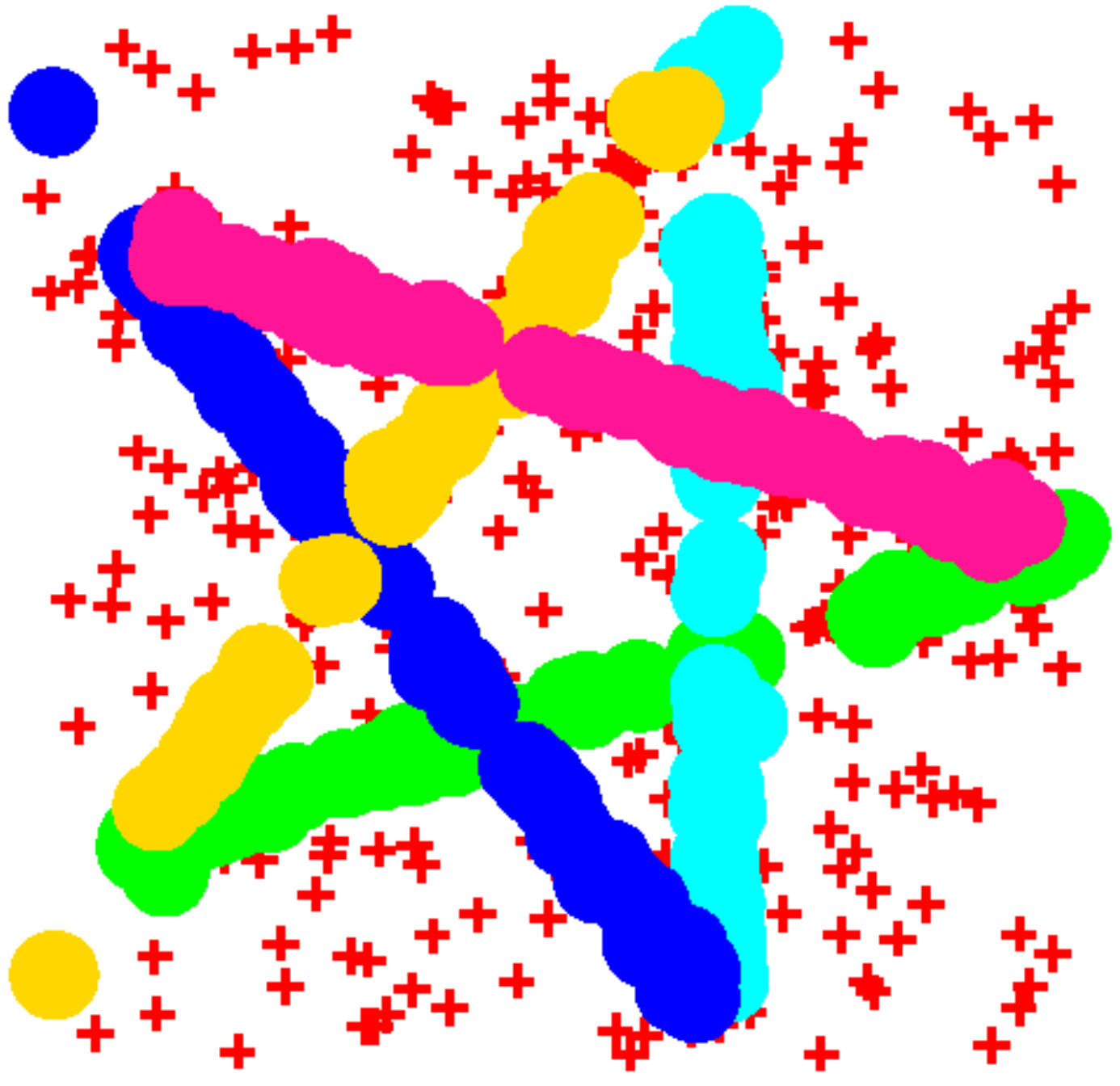}&
     \includegraphics[width=3.2cm,height=3.2cm]{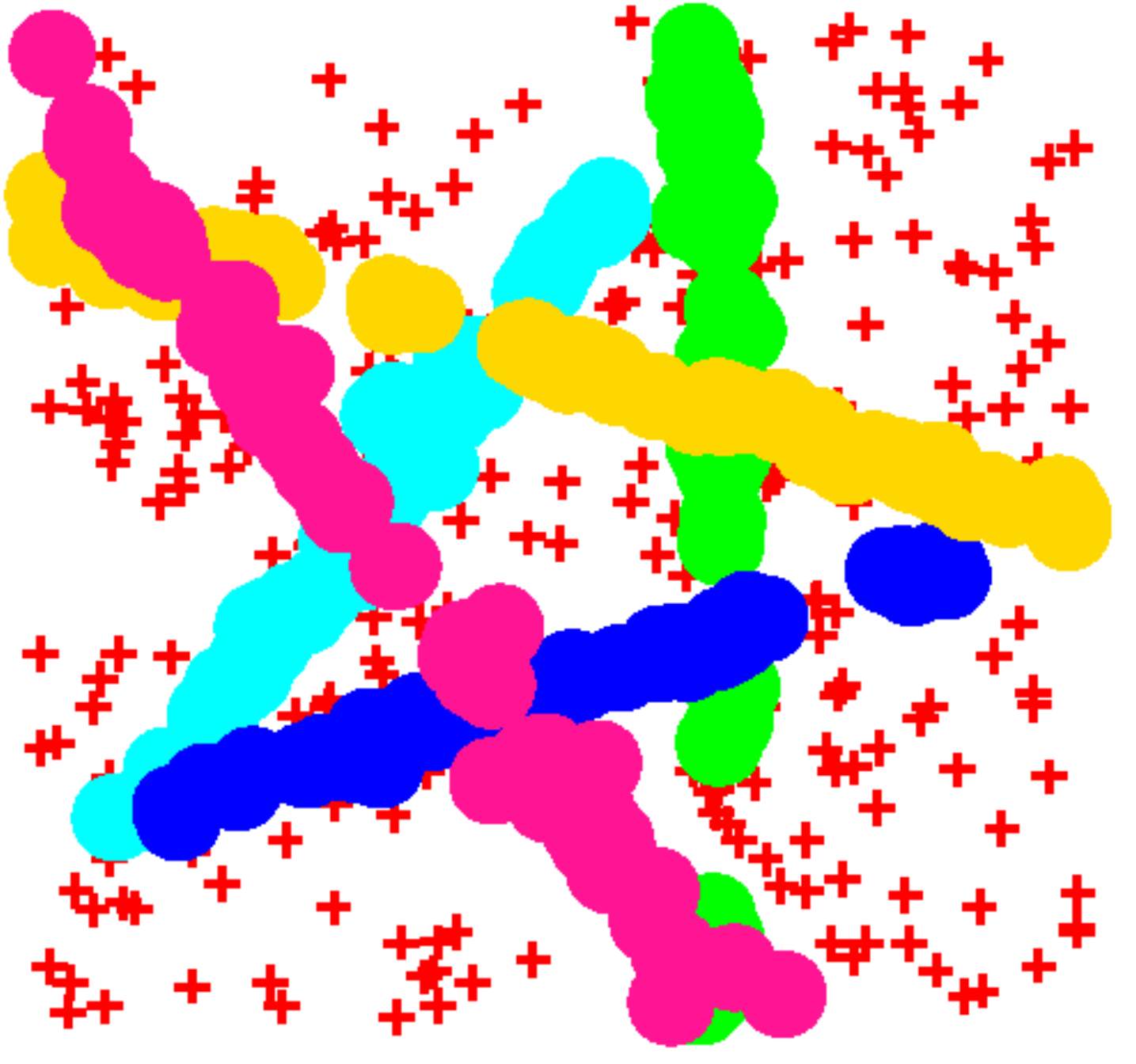}& 
    \includegraphics[width=3.2cm,height=3.2cm]{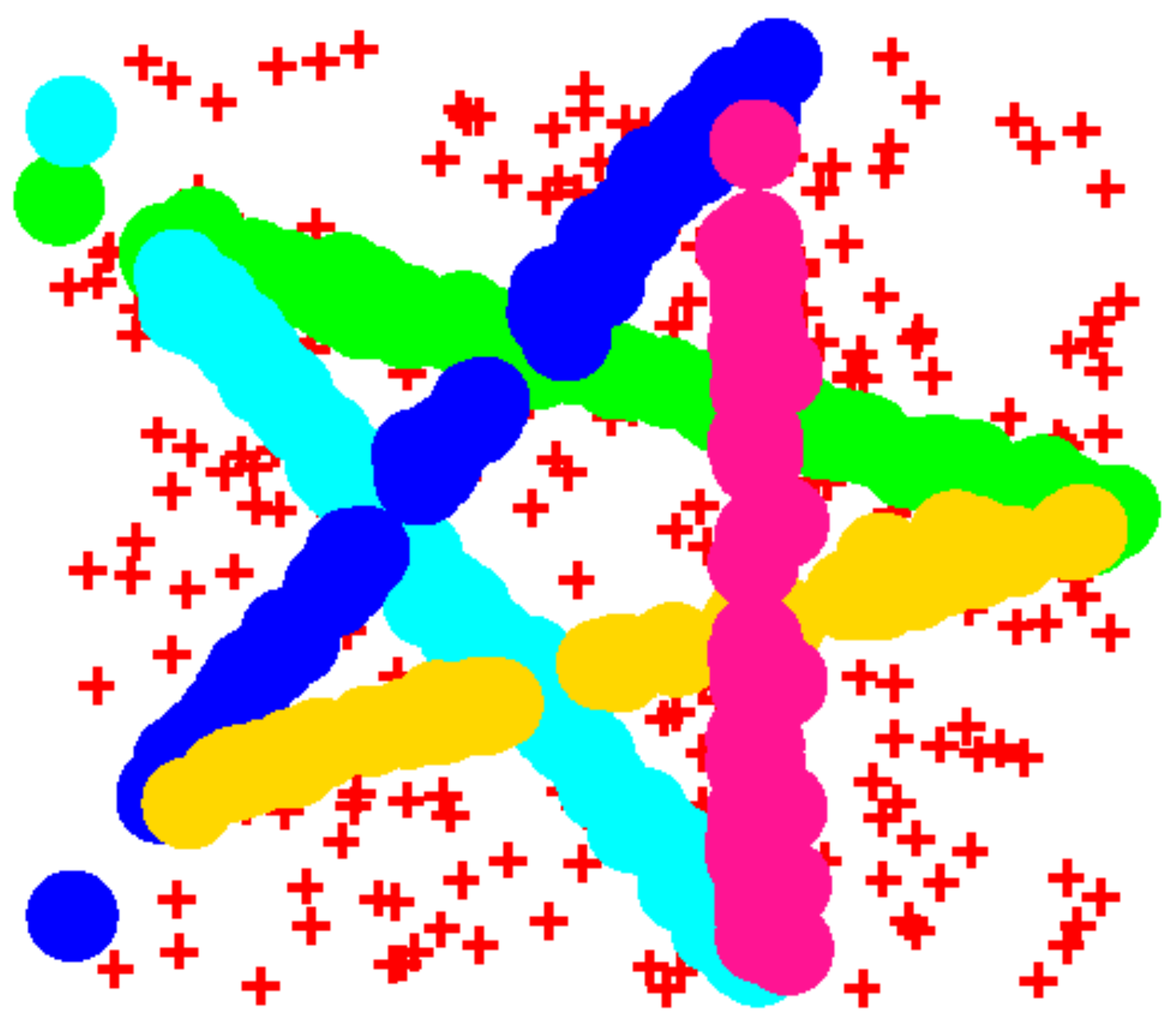}&
     \includegraphics[width=3.2cm,height=3.2cm]{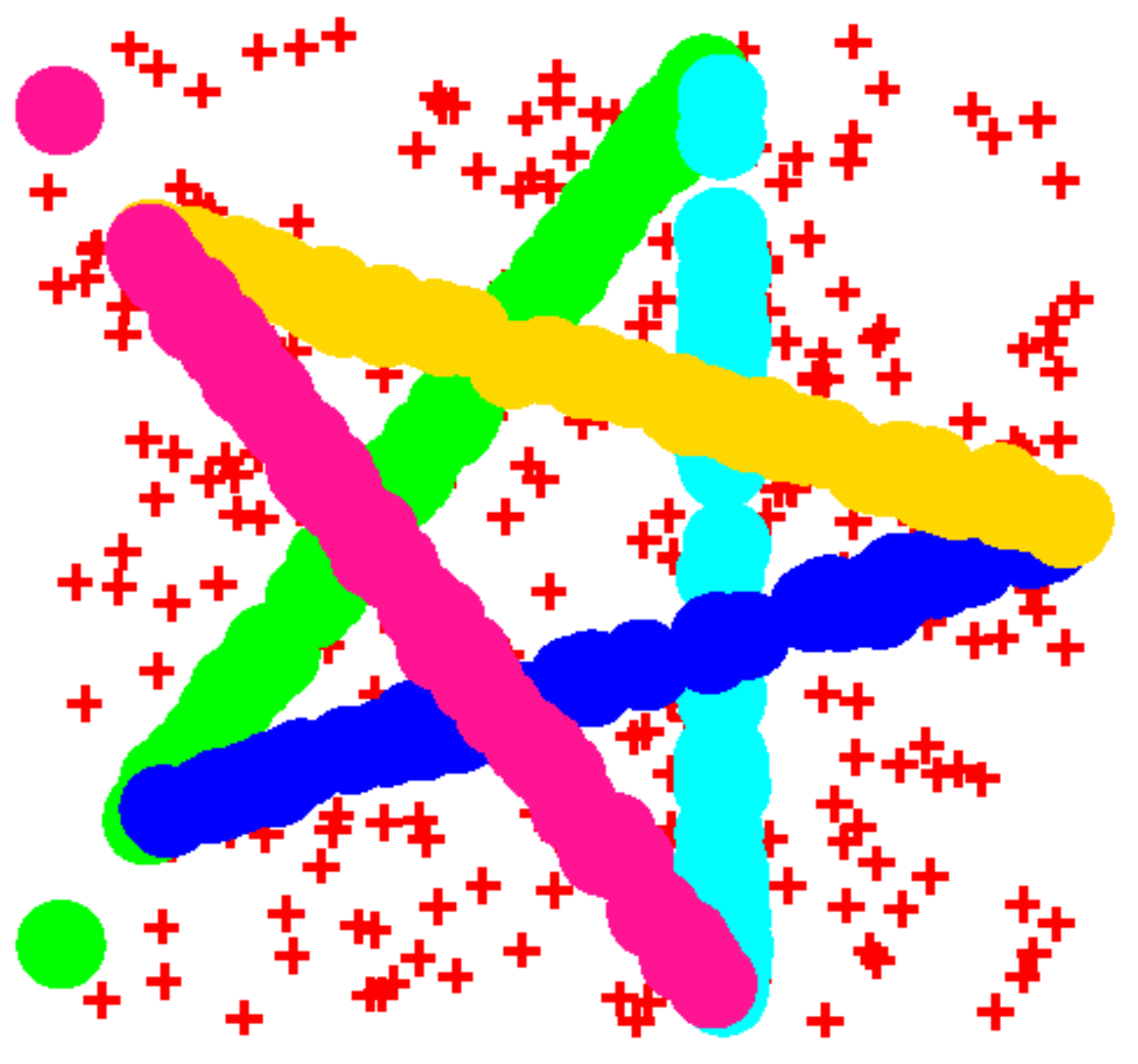}&
     \includegraphics[width=3.2cm,height=3.2cm]{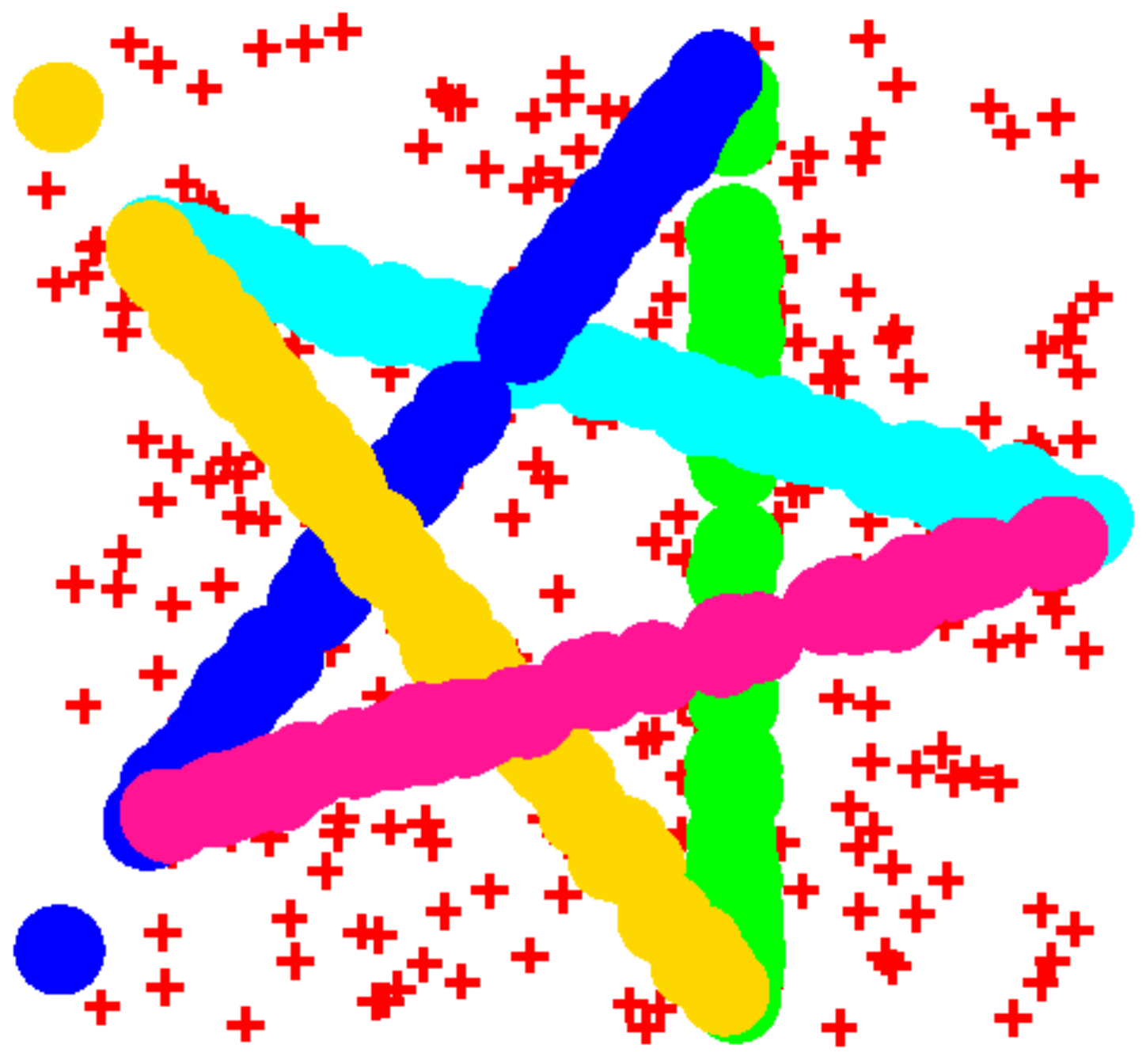} \\
     &\hspace{-0.1cm}   & CA=82.8\% &   CA=78.8\% &   CA=83.6\% &   CA=96.0\% &   CA=96.0\% \\
\end{tabular}
}
\caption{\textbf{Multiple Line Fitting.} Dataset: \textit{Star5} \cite{jlink}. Point membership is color coded. Gross outliers are in red.}
\label{fig:qual_line}
\end{figure*}

%% file: qual_circle.tex
\begin{figure*}
\centering
\setlength{\tabcolsep}{2pt}
\resizebox{\textwidth}{!}{%
\begin{tabular}{ccccccc}
&\hspace{-0.1cm}  Data & J-Linkage &  T-Linkage &  RansaCov &   DGSAC-G &   DGSAC-O \\
 \rotatebox{90}{\hspace{1.5cm}  Circle5, O=50\% } &
     \includegraphics[width=3.2cm,height=4.6cm]{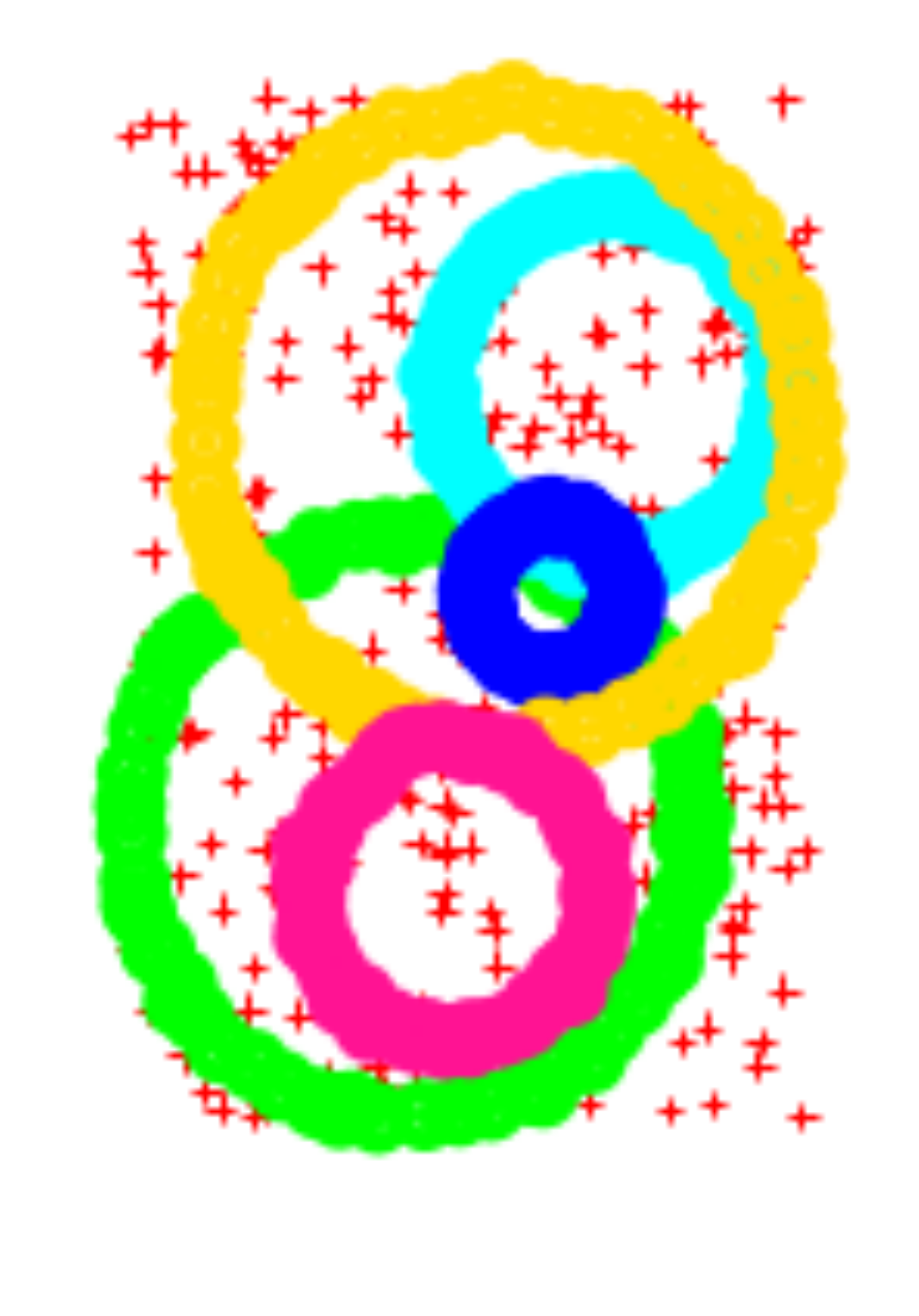}&
     \includegraphics[width=3.2cm,height=4.6cm]{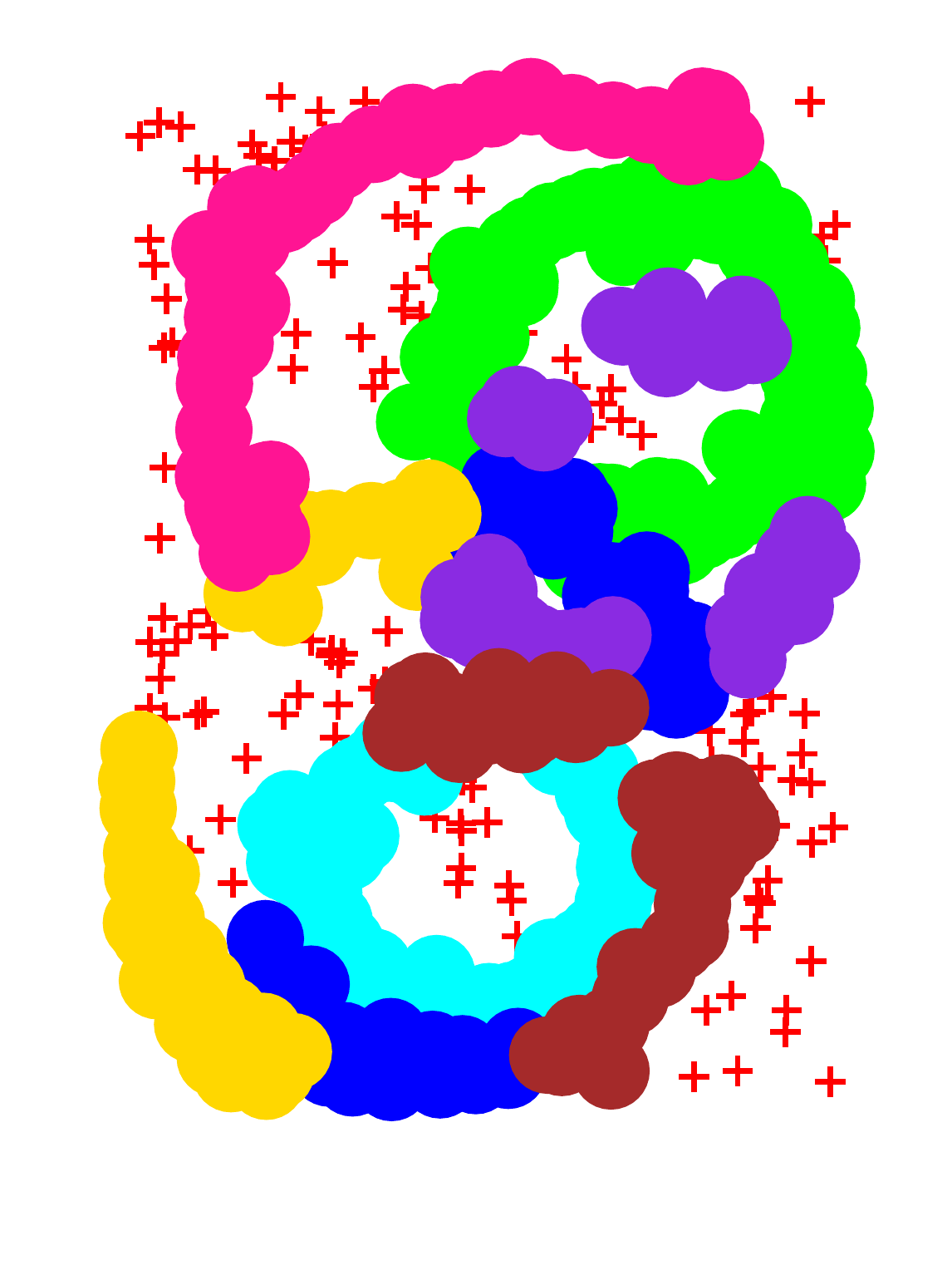}&
     \includegraphics[width=3.2cm,height=4.6cm]{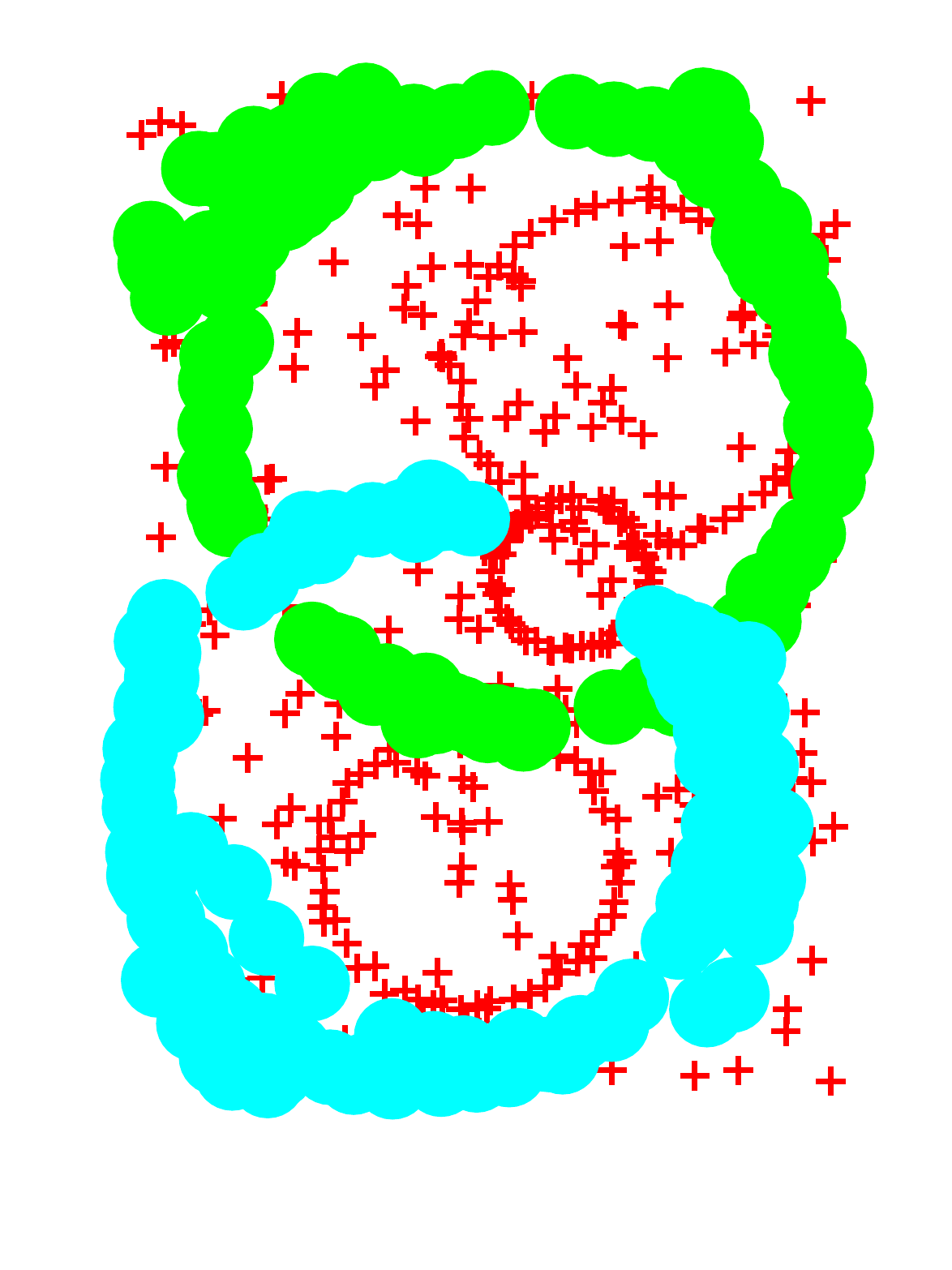}&
    \includegraphics[width=3.2cm,,height=4.6cm]{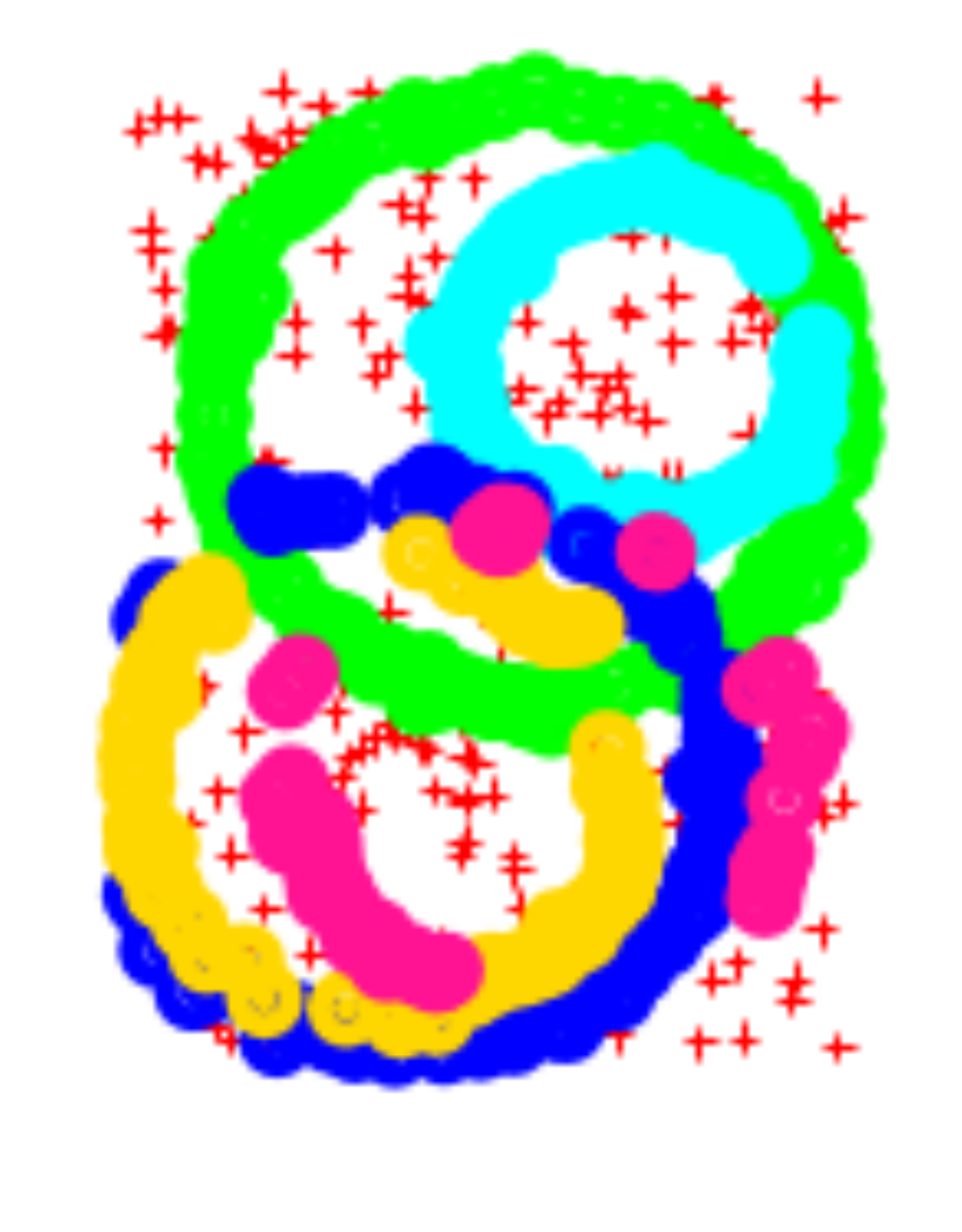}&
     \includegraphics[width=3.2cm,height=4.6cm]{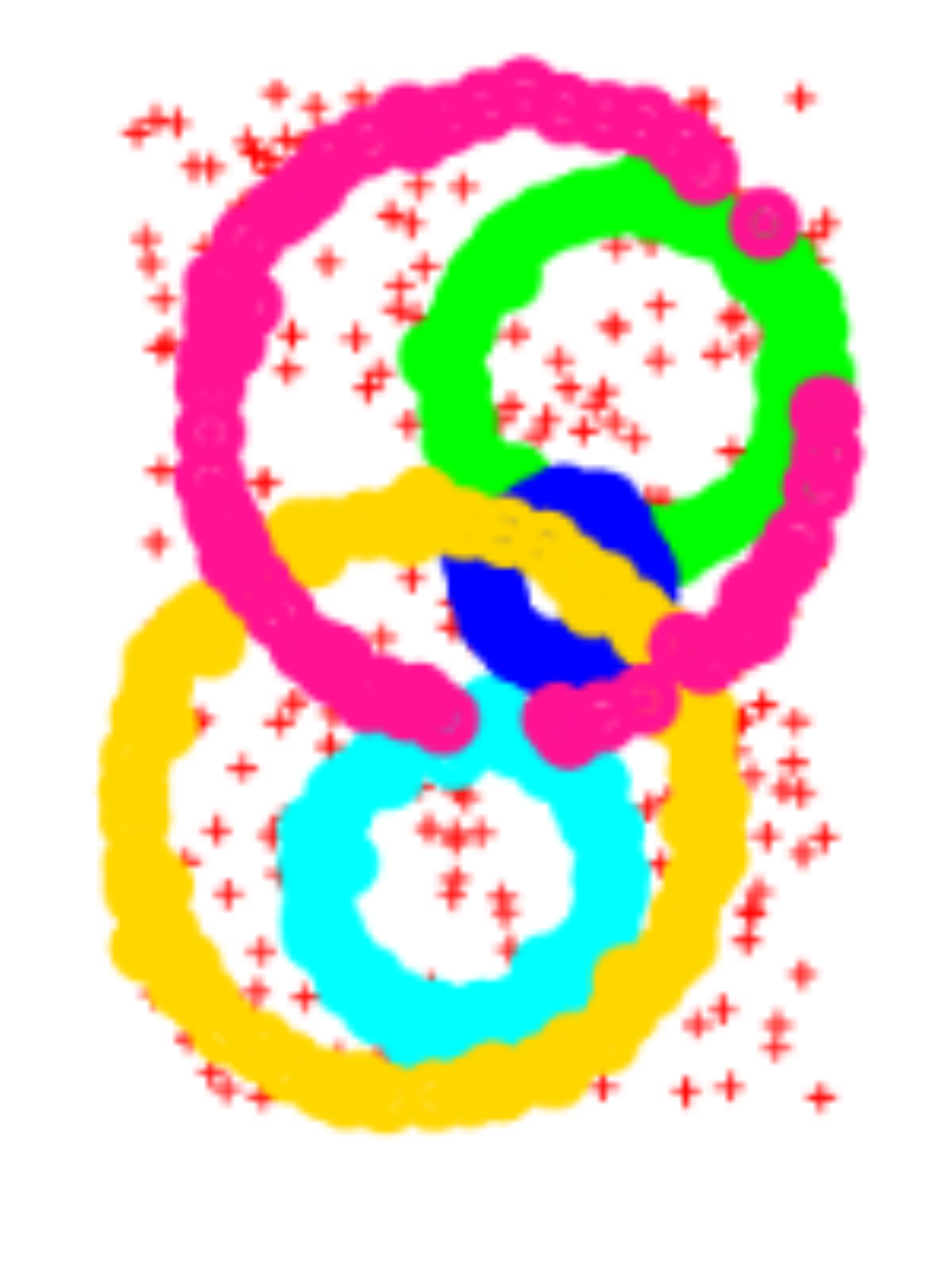}&
     \includegraphics[width=3.2cm,height=4.6cm]{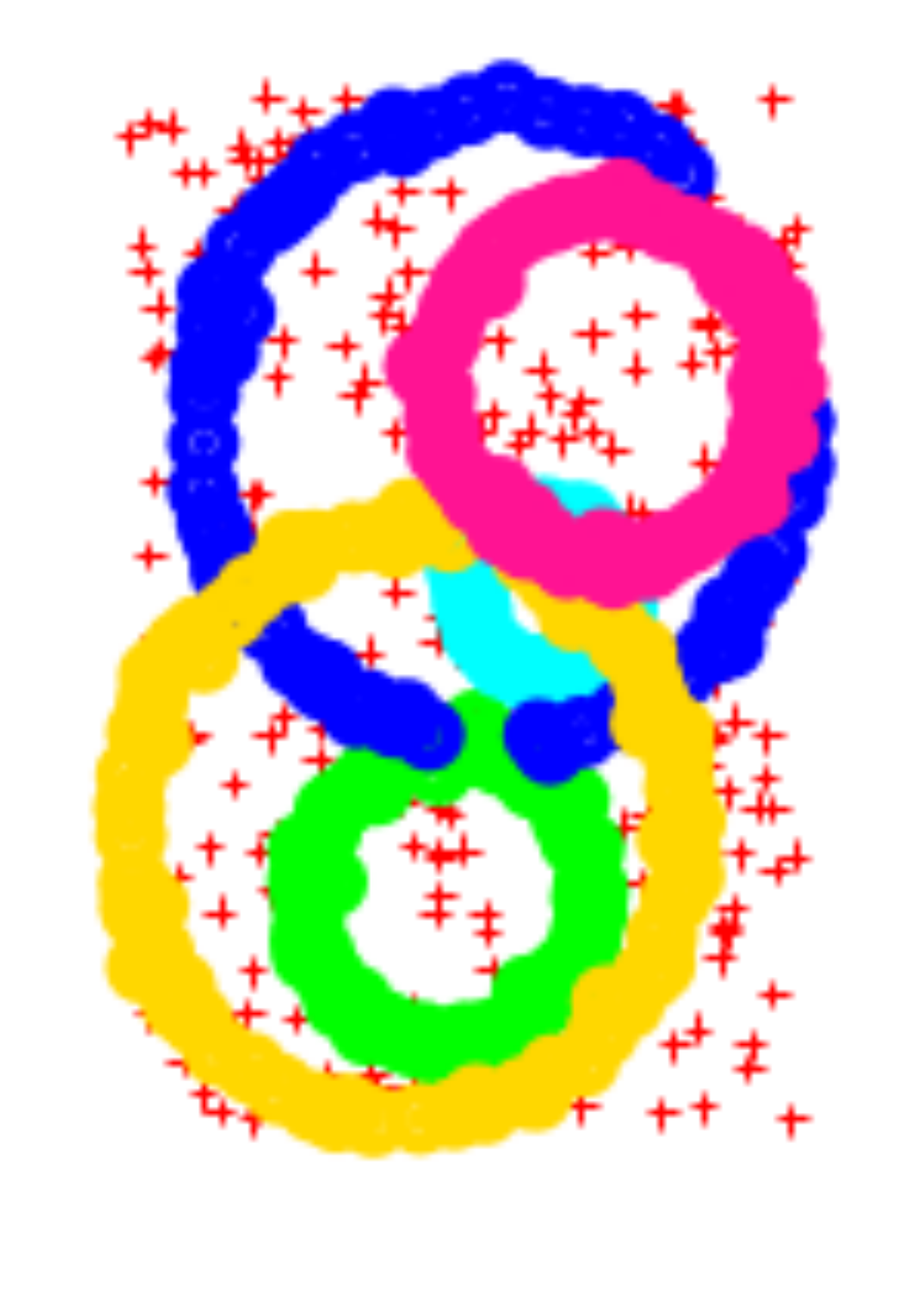} \\
     &   & CA=52.6\% &   CA=40.8\% &  CA=60.5\% &   CA=89.4\% &  CA=89.4\% \\
\end{tabular}
}
\caption{\textbf{Multiple Circle Fitting.} Dataset: \textit{Circle5} \cite{jlink}. Point membership is color coded. Gross outliers are in red.}
\label{fig:qual_circle}
\end{figure*}

%% file: analysis_discussion.tex



\section{Discussion and Conclusion}
\label{sec:ana_discussion}
Inlier noise scale and the ground-truth number of genuine structures present in the data are the two critical parameters of the multi-model fitting process. Most multi-model fitting methods require either or both of the above two parameters to be provided by the user, which introduce user dependency and limit the applicability where automatic execution is required. The best performing method NMU \cite{tepper2016nonnegative} is also susceptible to the user-provided inlier threshold (as presented in the original paper).
In this work, we propose a data-driven unified pipeline for automatic robust multiple structure recovery. The proposed DGSAC utilizes kernel residual density to differentiate inliers and outliers. The KRD is applied to all components of the DGSAC pipeline. Using the KRD based guided sampling, DGSAC generates more relevant hypotheses and performs \textit{greedy} or \textit{optimal} model selection by employing kernel density-based model hypotheses goodness measure. We believe that DGSAC plays a crucial role in the application that requires the automatic extraction of multiple structures. We plan further to improve the running time by parallel and optimized implementation.

%% file: ms.bbl
\begin{thebibliography}{31}
\providecommand{\natexlab}[1]{#1}
\providecommand{\url}[1]{\texttt{#1}}
\expandafter\ifx\csname urlstyle\endcsname\relax
  \providecommand{\doi}[1]{doi: #1}\else
  \providecommand{\doi}{doi: \begingroup \urlstyle{rm}\Url}\fi

\bibitem[Angladon et~al.(2015)Angladon, Gasparini, and Charvillat]{tvp}
V.~Angladon, S.~Gasparini, and V.~Charvillat.
\newblock {The toulouse vanishing points dataset}.
\newblock In \emph{{Proceedings of the 6th ACM Multimedia Systems Conference
  (MMSys '15)}}, Portland, OR, United States, Mar. 2015.
\newblock \doi{10.1145/2713168.2713196}.

\bibitem[Barath and Matas(2019)]{barath2019progressive}
D.~Barath and J.~Matas.
\newblock Progressive-x: Efficient, anytime, multi-model fitting algorithm.
\newblock In \emph{Proceedings of the IEEE International Conference on Computer
  Vision}, pages 3780--3788, 2019.

\bibitem[Barath et~al.(2019)Barath, Matas, and Noskova]{barath2019magsac}
D.~Barath, J.~Matas, and J.~Noskova.
\newblock Magsac: marginalizing sample consensus.
\newblock In \emph{Proceedings of the IEEE Conference on Computer Vision and
  Pattern Recognition}, pages 10197--10205, 2019.

\bibitem[Chin et~al.(2012)Chin, Yu, and Suter]{chin2012accelerated}
T.-J. Chin, J.~Yu, and D.~Suter.
\newblock Accelerated hypothesis generation for multistructure data via
  preference analysis.
\newblock \emph{TPAMI}, 34\penalty0 (4):\penalty0 625--638, 2012.

\bibitem[Chin et~al.(2015)Chin, Purkait, Eriksson, and
  Suter]{chin2015efficient}
T.-J. Chin, P.~Purkait, A.~Eriksson, and D.~Suter.
\newblock Efficient globally optimal consensus maximisation with tree search.
\newblock In \emph{CVPR}, pages 2413--2421, 2015.

\bibitem[Coleman and Li(1996)]{coleman1996interior}
T.~F. Coleman and Y.~Li.
\newblock An interior trust region approach for nonlinear minimization subject
  to bounds.
\newblock \emph{SIAM Journal on optimization}, 6\penalty0 (2):\penalty0
  418--445, 1996.

\bibitem[Conn et~al.(2000)Conn, Gould, and Toint]{conn2000trust}
A.~R. Conn, N.~I. Gould, and P.~L. Toint.
\newblock \emph{Trust region methods}, volume~1.
\newblock Siam, 2000.

\bibitem[Denis et~al.(2008)Denis, Elder, and Estrada]{york}
P.~Denis, J.~H. Elder, and F.~J. Estrada.
\newblock Efficient edge-based methods for estimating manhattan frames in urban
  imagery.
\newblock In \emph{European conference on computer vision}, pages 197--210.
  Springer, 2008.

\bibitem[Fagin et~al.(2003)Fagin, Kumar, and Sivakumar]{topk}
R.~Fagin, R.~Kumar, and D.~Sivakumar.
\newblock Comparing top k lists.
\newblock \emph{SIAM Journal on discrete mathematics}, 17\penalty0
  (1):\penalty0 134--160, 2003.

\bibitem[Farenzena et~al.(2009)Farenzena, Fusiello, and Gherardi]{SAMANTHA}
M.~Farenzena, A.~Fusiello, and R.~Gherardi.
\newblock Structure-and-motion pipeline on a hierarchical cluster tree.
\newblock In \emph{ICCV Workshops}, pages 1489--1496. IEEE, 2009.

\bibitem[Fischler and Bolles(1981)]{fischler81}
M.~A. Fischler and R.~C. Bolles.
\newblock Random sample consensus: A paradigm for model fitting with
  applications to image analysis and automated cartography.
\newblock \emph{Communications of the ACM}, 24\penalty0 (6):\penalty0 381--395,
  1981.

\bibitem[Kuang et~al.(2015)Kuang, Yun, and Park]{symnmf15}
D.~Kuang, S.~Yun, and H.~Park.
\newblock Symnmf: nonnegative low-rank approximation of a similarity matrix for
  graph clustering.
\newblock \emph{Journal of Global Optimization}, 62\penalty0 (3):\penalty0
  545--574, 2015.

\bibitem[Lai et~al.(2016)Lai, Wang, Yan, Xiao, and Suter]{lai2016efficient}
T.~Lai, H.~Wang, Y.~Yan, G.~Xiao, and D.~Suter.
\newblock Efficient guided hypothesis generation for multi-structure epipolar
  geometry estimation.
\newblock \emph{CVIU}, 2016.

\bibitem[Lai et~al.(2017)Lai, Wang, Yan, and Zhang]{lai2017unified}
T.~Lai, H.~Wang, Y.~Yan, and L.~Zhang.
\newblock A unified hypothesis generation framework for multi-structure model
  fitting.
\newblock \emph{Neurocomputing}, 222:\penalty0 144--154, 2017.

\bibitem[Magri and Fusiello(2014)]{magri14tl}
L.~Magri and A.~Fusiello.
\newblock T-linkage: A continuous relaxation of j-linkage for multi-model
  fitting.
\newblock In \emph{CVPR}, pages 3954--3961, 2014.

\bibitem[Magri and Fusiello(2015{\natexlab{a}})]{magri15rpa}
L.~Magri and A.~Fusiello.
\newblock Robust multiple model fitting with preference analysis and low-rank
  approximation.
\newblock In \emph{BMVC}, volume~20, page~12, 2015{\natexlab{a}}.

\bibitem[Magri and Fusiello(2015{\natexlab{b}})]{magri2015scale}
L.~Magri and A.~Fusiello.
\newblock Scale estimation in multiple models fitting via consensus clustering.
\newblock In \emph{International Conference on Computer Analysis of Images and
  Patterns}, pages 13--25. Springer International Publishing,
  2015{\natexlab{b}}.

\bibitem[Magri and Fusiello(2016)]{magri16set}
L.~Magri and A.~Fusiello.
\newblock Multiple model fitting as a set coverage problem.
\newblock In \emph{CVPR}, pages 3318--3326, 2016.

\bibitem[Pham et~al.(2014)Pham, Chin, Yu, and Suter]{pham2014random}
T.~T. Pham, T.-J. Chin, J.~Yu, and D.~Suter.
\newblock The random cluster model for robust geometric fitting.
\newblock \emph{TPAMI}, 36\penalty0 (8):\penalty0 1658--1671, 2014.

\bibitem[Tennakoon et~al.(2015)Tennakoon, Bab-Hadiashar, Cao, Hoseinnezhad, and
  Suter]{tennakoon2015robust}
R.~B. Tennakoon, A.~Bab-Hadiashar, Z.~Cao, R.~Hoseinnezhad, and D.~Suter.
\newblock Robust model fitting using higher than minimal subset sampling.
\newblock \emph{IEEE transactions on pattern analysis and machine
  intelligence}, 38\penalty0 (2):\penalty0 350--362, 2015.

\bibitem[Tepper and Sapiro(2016{\natexlab{a}})]{tepper2016fast}
M.~Tepper and G.~Sapiro.
\newblock Fast l1-nmf for multiple parametric model estimation.
\newblock \emph{arXiv preprint arXiv:1610.05712}, 2016{\natexlab{a}}.

\bibitem[Tepper and Sapiro(2016{\natexlab{b}})]{tepper2016nonnegative}
M.~Tepper and G.~Sapiro.
\newblock Nonnegative matrix underapproximation for robust multiple model
  fitting.
\newblock \emph{arXiv preprint arXiv:1611.01408}, 2016{\natexlab{b}}.

\bibitem[Tiwari and Anand(2018)]{tiwari2018dgsac}
L.~Tiwari and S.~Anand.
\newblock Dgsac: Density guided sampling and consensus.
\newblock In \emph{2018 IEEE Winter Conference on Applications of Computer
  Vision (WACV)}, pages 974--982. IEEE, 2018.

\bibitem[Tiwari et~al.(2016)Tiwari, Anand, and Mittal]{tiwari2016robust}
L.~Tiwari, S.~Anand, and S.~Mittal.
\newblock Robust multi-model fitting using density and preference analysis.
\newblock In \emph{ACCV}, pages 308--323. Springer, 2016.

\bibitem[Toldo and Fusiello(2008)]{jlink}
R.~Toldo and A.~Fusiello.
\newblock Robust multiple structures estimation with j-linkage.
\newblock \emph{ECCV}, pages 537--547, 2008.

\bibitem[Wang and Suter(2004)]{wang2004robust}
H.~Wang and D.~Suter.
\newblock Robust adaptive-scale parametric model estimation for computer
  vision.
\newblock \emph{IEEE transactions on pattern analysis and machine
  intelligence}, 26\penalty0 (11):\penalty0 1459--1474, 2004.

\bibitem[Wang et~al.(2012)Wang, Chin, and Suter]{wang12}
H.~Wang, T.~J. Chin, and D.~Suter.
\newblock Simultaneously fitting and segmenting multiple-structure data with
  outliers.
\newblock \emph{IEEE TPAMI}, 34\penalty0 (6):\penalty0 1177--1192, June 2012.

\bibitem[Wang et~al.(2013)Wang, Cai, and Tang]{wang13}
H.~Wang, J.~Cai, and J.~Tang.
\newblock Amsac: An adaptive robust estimator for model fitting.
\newblock In \emph{2013 IEEE International Conference on Image Processing},
  pages 305--309, 2013.

\bibitem[Wong et~al.(2011)Wong, Chin, Yu, and Suter]{wong2011dynamic}
H.~S. Wong, T.-J. Chin, J.~Yu, and D.~Suter.
\newblock Dynamic and hierarchical multi-structure geometric model fitting.
\newblock In \emph{ICCV}, pages 1044--1051. IEEE, 2011.

\bibitem[Wong et~al.(2013)Wong, Chin, Yu, and Suter]{wong2013simultaneous}
H.~S. Wong, T.-J. Chin, J.~Yu, and D.~Suter.
\newblock A simultaneous sample-and-filter strategy for robust multi-structure
  model fitting.
\newblock \emph{CVIU}, 117\penalty0 (12):\penalty0 1755--1769, 2013.

\bibitem[Yu et~al.(2011)Yu, Chin, and Suter]{yu2011global}
J.~Yu, T.-J. Chin, and D.~Suter.
\newblock A global optimization approach to robust multi-model fitting.
\newblock In \emph{Computer Vision and Pattern Recognition (CVPR), 2011 IEEE
  Conference on}, pages 2041--2048. IEEE, 2011.

\end{thebibliography}
